\documentclass{article} % For LaTeX2e
\usepackage{iclr2026_conference,times}

\usepackage{svg}
\usepackage{caption} % Captions side by side 
\usepackage{multirow}
\usepackage{graphicx} 
\usepackage[most]{tcolorbox}
\usepackage{enumitem}
\usepackage{amssymb}
\usepackage{booktabs}
\usepackage{makecell}

\usepackage{algorithm}
\usepackage{algpseudocode}

\usepackage{subcaption}
\usepackage{caption}
\usepackage{calc}
\usepackage{wrapfig}
\usepackage{placeins}

\usepackage{tikz}
\usepackage{xcolor}
\usepackage{array}

\usepackage{colortbl,booktabs}

\usepackage{booktabs}       % professional-quality tables
\usepackage[table]{xcolor}
\definecolor{vdm_second}{RGB}{235,245,255} % Very light blue
\colorlet{vdm_highlight}{blue!10}          % Blue highlight

\definecolor{color_neutral}{RGB}{230,230,230} 

\definecolor{second}{RGB}{235, 245, 255} % Extremely light blue (lighter than best)

\definecolor{llm_second}{RGB}{255,235,235} % Very light red
\colorlet{llm_highlight}{red!10}           % Red highlight

\definecolor{vdm_intense}{HTML}{00429d}
\definecolor{llm_intense}{HTML}{93003a}

\definecolor{vdm_best}{RGB}{71,113,178}
\definecolor{llm_best}{RGB}{255,165,158}

\colorlet{vdm_best}{vdm_best!35}
\colorlet{llm_best}{llm_best!35}

\newcommand{\vdmbest}[1]{\cellcolor{vdm_best}\textbf{#1}}
\newcommand{\llmbest}[1]{\cellcolor{llm_best}\textbf{#1}}
\newcommand{\vdmbestnt}[1]{\colorbox{vdm_best}{\textbf{#1}}}
\newcommand{\llmbestnt}[1]{\colorbox{llm_best}{\textbf{#1}}}

\definecolor{arc_val0}{RGB}{0,0,0}       % Black
\definecolor{arc_val1}{RGB}{0,116,217}   % Blue
\definecolor{arc_val2}{RGB}{255,65,54}   % Red
\definecolor{arc_val3}{RGB}{46,204,64}   % Green
\definecolor{arc_val4}{RGB}{255,220,0}   % Yellow
\definecolor{arc_val5}{RGB}{170,170,170} % Grey
\definecolor{arc_val6}{RGB}{240,18,190}  % Fuchsia
\definecolor{arc_val7}{RGB}{255,133,27}  % Orange
\definecolor{arc_val8}{RGB}{127,219,255} % Teal
\definecolor{arc_val9}{RGB}{135,12,37}   % Brown
\definecolor{arc_val10}{RGB}{163,73,164} % Purple
\definecolor{arc_val11}{RGB}{255,182,193}% Pink
\definecolor{arc_val12}{RGB}{0,255,255}  % Cyan
\definecolor{arc_val13}{RGB}{128,0,128}  % Dark purple
\definecolor{arc_val14}{RGB}{192,192,192} % Silver
\definecolor{arc_val15}{RGB}{255,255,255} % White

\newcommand{\cnumarc}[1]{\textbf{%
  \ifcase#1
    \textcolor{arc_val0}{0}%
  \or
    \textcolor{arc_val1}{1}%
  \or
    \textcolor{arc_val2}{2}%
  \or
    \textcolor{arc_val3}{3}%
  \or
    \textcolor{arc_val4}{4}%
  \or
    \textcolor{arc_val5}{5}%
  \or
    \textcolor{arc_val6}{6}%
  \or
    \textcolor{arc_val7}{7}%
  \or
    \textcolor{arc_val8}{8}%
  \or
    \textcolor{arc_val9}{9}%
  \or
    \textcolor{arc_val10}{10}%
  \or
    \textcolor{arc_val11}{11}%
  \or
    \textcolor{arc_val12}{12}%
  \or
    \textcolor{arc_val13}{13}%
  \or
    \textcolor{arc_val14}{14}%
  \or
    \textcolor{arc_val15}{15}%
  \fi
}}

\definecolor{maze_val0}{RGB}{71,48,45}   % Wall - dark brown
\definecolor{maze_val1}{RGB}{200,200,200} % Path - light gray (instead of white)
\definecolor{maze_val2}{RGB}{244,96,54}   % End - orange-red
\definecolor{maze_val3}{RGB}{72,191,132}  % Start - green
\definecolor{maze_val5}{RGB}{46,134,171}  % Solution - blue

\newcommand{\cnummazes}[1]{\textbf{%
  \ifcase#1
    \textcolor{maze_val0}{0}%
  \or
    \textcolor{maze_val1}{1}%
  \or
    \textcolor{maze_val2}{2}%
  \or
    \textcolor{maze_val3}{3}%
  \or
    \textcolor{maze_val5}{4}%
  \fi
}}

\usepackage{amsmath,amsfonts,bm}

\def\eqref#1{equation~\ref{#1}}
\def\1{\bm{1}}

\DeclareMathAlphabet{\mathsfit}{\encodingdefault}{\sfdefault}{m}{sl}
\SetMathAlphabet{\mathsfit}{bold}{\encodingdefault}{\sfdefault}{bx}{n}

\usepackage{hyperref}
\usepackage{url}

\title{Rethinking Visual Intelligence: Insights from Video Pretraining}

\arxivversioncopy

\usepackage{authblk}

\author[1]{Pablo Acuaviva}
\author[1]{Aram Davtyan}
\author[2]{Mariam Hassan}
\author[1]{Sebastian Stapf}
\author[2]{Ahmad Rahimi}
\author[2]{Alexandre Alahi}
\author[1]{Paolo Favaro}

\affil[1]{Computer Vision Group, University of Bern, Switzerland}
\affil[2]{VITA Lab, EPFL, Lausanne, Switzerland}

\begin{document}

\maketitle

\renewcommand{\thefootnote}{}
\footnotetext{Code available at \url{https://github.com/PabloAcuaviva/visual-intelligence}}
\renewcommand{\thefootnote}{\arabic{footnote}}

\begin{abstract}
Large language models (LLMs) have demonstrated that large-scale pretraining enables systems to adapt rapidly to new problems with little supervision in the language domain. This success, however, has not translated as effectively to the visual domain, where models, including LLMs, continue to struggle with compositional understanding, sample efficiency, and general-purpose problem-solving. We investigate Video Diffusion Models (VDMs) as a promising direction for bridging this gap. Pretraining on spatiotemporal data endows these models with strong inductive biases for structure and dynamics, which we hypothesize can support broad task adaptability. To test this, we design a controlled evaluation in which both a pretrained LLM and a pretrained VDM are equipped with lightweight adapters and presented with tasks in their natural modalities. Across benchmarks including ARC-AGI, ConceptARC, visual games, route planning, and cellular automata, VDMs demonstrate higher data efficiency than their language counterparts. Taken together, our results indicate that video pretraining offers inductive biases that support progress toward visual foundation models.
\end{abstract}

%%%
% Main Paper
%%%
\section{Introduction}

% \begin{wrapfigure}{r}{0.5\textwidth}
%     \centering
%     \vspace{-10pt} % tighten if needed
%     \includegraphics[width=\linewidth]{figures/main/concept-arc-radar.pdf}
%       \caption{ConceptARC accuracy comparison between \vdmbestnt{CogVideoX1.5-5B} and \llmbestnt{Qwen3-4B-Instruct-2507}. This radar plot highlights the differences between a LLM and a VDM of similar capacity across visual concepts.}
%     \label{fig:concept-arc-radar}
%     \vspace{-5pt}
% \end{wrapfigure}

\begin{figure}[htbp]
    \centering
    \includegraphics[width=\linewidth]{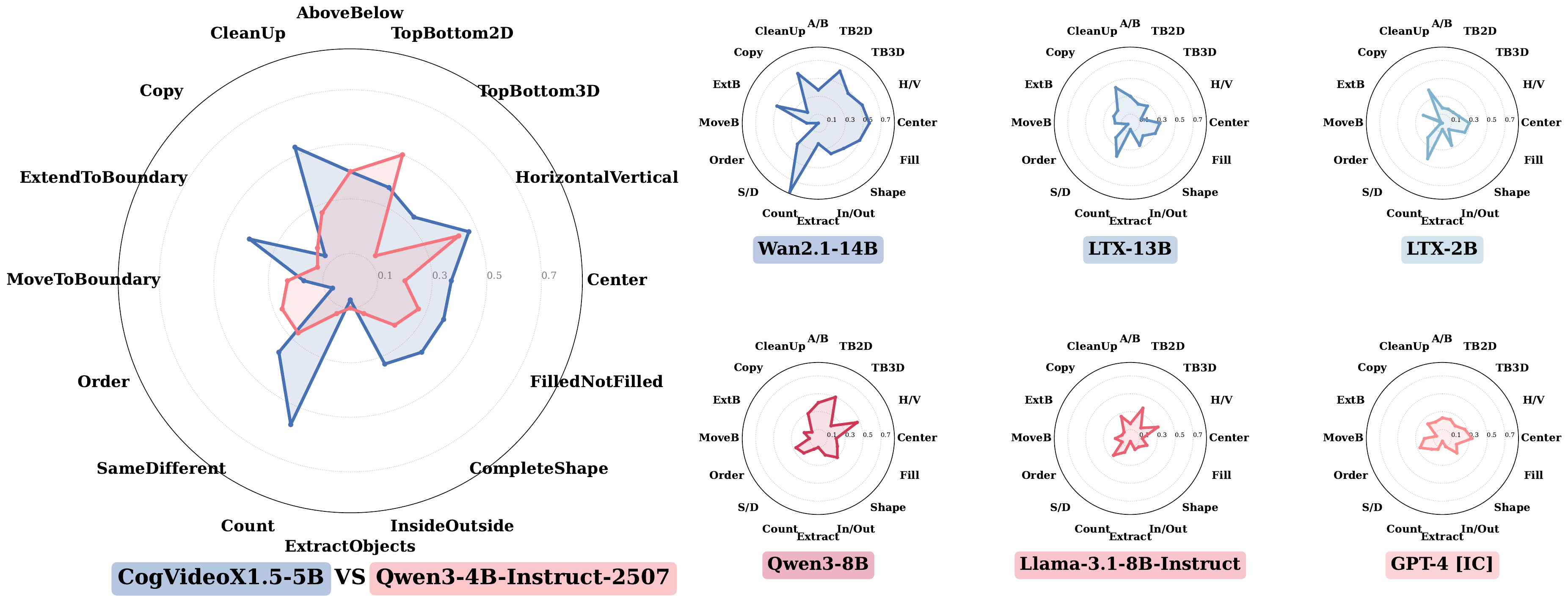}
    \caption{Radar plot showing ConceptARC competencies between \vdmbestnt{\textbf{VDMs}} and \llmbestnt{\textbf{LLMs}}, GPT-4 [IC] is added for additional reference.}
    \label{fig:concept-arc-radar-all}
\end{figure}
 
Foundation models have reshaped natural language processing by showing that large-scale pretraining can equip models with broad knowledge and strong inductive priors. This foundation allows models to adapt quickly and effectively to new tasks through techniques like in-context learning \citep{brown2020language} and parameter-efficient fine-tuning \citep{liu2022fewshotparameterefficientfinetuningbetter}, achieving strong performance with minimal supervision. The success of Large Language Models (LLMs) illustrates how scale and pretraining can create systems that generalize across diverse problems. Achieving a similar level of versatility in vision, however, remains largely unexplored and a major challenge. Despite recent breakthroughs in image and video generation \citep{flux2024, polyak2025moviegencastmedia, brooks2024video}, vision models are not yet on par with LLMs when it comes to compositional skills, sample efficiency, and versatility in problem solving.

Video Diffusion Models (VDMs) represent an exciting direction for narrowing this gap. Pretraining on rich spatiotemporal data endows them with strong inductive biases for spatial structure and temporal dynamics \citep{blattmann2023stablevideodiffusionscaling, deepmind2025veo3, wu2025qwenimage}, which we hypothesize can be harnessed for structured visual understanding. We move beyond treating videos as mere generative artifacts and instead regard them as a natural representation for problem solving, where tasks are expressed as transformations unfolding over time.
Building on this perspective, we introduce a simple and general framework for adapting VDMs to a broad class of visual tasks and evaluate them head-to-head with equally adapted LLMs (see Figure \ref{fig:concept-arc-radar-all}). This setup allows us to test whether large-scale video pretraining offers a complementary foundation for structured visual problem-solving, contrasting the strengths of visually grounded models with those of symbolically trained language models.

Each task is represented consistently but adapted to each model family’s modality: LLMs operate in a text-to-text setting, where inputs and outputs are serialized into structured text, while VDMs receive an image-to-image formulation, where input–output pairs are rendered as short videos to model the task as a temporal transformation. Both model families use identical LoRA-based \cite{hu2022lora} adaptation: adapters are inserted at corresponding layers, pretrained backbones remain frozen, and only lightweight parameters are updated. This symmetry provides a controlled basis for comparison and isolates the impact of video pretraining on structured visual understanding.

Our contributions are as follows:

% Original box

% \newtcolorbox{highlightbox}{
%   colback=vdm_best,
%   colframe=vdm_best,
%   width=\textwidth,
%   boxrule=0pt,
%   arc=3pt,
%   left=6pt,
%   right=6pt,
%   top=6pt,
%   bottom=6pt,
%   enhanced,
%   sharp corners
% }

% \begin{tcolorbox}[colback=second, colframe=second, width=\textwidth, boxrule=0pt, arc=0pt]
% \begin{enumerate}[label=\textbf{\arabic*.}]
%     \item A unified framework for adapting VDMs to image-to-image visual tasks by reframing examples as temporal sequences.

%     \item A controlled evaluation setting where both VDMs and LLMs are fine-tuned with LoRA-based adaptation, enabling direct comparison.
    
%    \item Empirical evidence that VDMs benefit from video pretraining for visual intelligence, hinting at a path toward flexible visual foundation models with both generative and problem-solving strengths.
% \end{enumerate}
% \end{tcolorbox}

% {'vdm_best!40': array([181, 198, 224]), 'vdm_best!80': array([108, 141, 193]), 'llm_best!40': array([255, 219, 216]), 'llm_best!80': array([255, 183, 177])}

% {'vdm_best!10': array([237, 241, 247]), 'llm_best!10': array([255, 246, 245]), 'vdm_best!20': array([218, 227, 240]), 'llm_best!20': array([255, 237, 236]), 'vdm_best!30': array([200, 212, 232]), 'llm_best!30': array([255, 228, 226])}

\newtcolorbox{highlightbox}{
  colback={rgb,255:red,218; green,227; blue,240},
  colframe={rgb,255:red,180; green,195; blue,220},
  width=\textwidth,
  boxrule=0.5pt,
  arc=4pt,
  left=8pt,
  right=8pt,
  top=8pt,
  bottom=8pt,
  enhanced
}

\begin{highlightbox}
\begin{enumerate}[label=\textbf{\arabic*.}]
    \item A unified framework for adapting VDMs to image-to-image visual tasks by reframing examples as temporal sequences.
    \item A controlled evaluation setting where both VDMs and LLMs are fine-tuned with LoRA-based adaptation, enabling direct comparison.
    \item Empirical evidence that VDMs benefit from video pretraining for visual intelligence, hinting at a path toward flexible visual foundation models with both generative and problem-solving strengths.
\end{enumerate}
\end{highlightbox}
\section{Related Work}
\textbf{Language Foundation Models.}
LLMs have demonstrated remarkable generalization and adaptability to new tasks with minimal supervision, mainly due to their large-scale pretraining on diverse text corpora \cite{brown2020language,chowdhery2023palm}. Their extensive pretraining equips LLMs with rich knowledge and strong inductive biases, enabling them to perform few-shot learning \cite{brown2020language} and in-context learning \cite{coda2023meta}, where models learn new tasks only by observing a handful of examples. Parameter-efficient finetuning methods like LoRA \cite{hu2022lora} extend this adaptability allowing LLMs to specialize to new domains while the backbone is completely frozen \cite{liao2025dynamic}. Together, these capabilities make LLMs highly flexible and scalable problem solvers. In this paper, we leverage this adaptability to compare the data efficiency of LLMs and VDMs across diverse visual tasks.

\noindent\textbf{Video Diffusion Models.}  
Diffusion-based generative models have recently achieved remarkable progress in video synthesis. Pioneering approaches such as CogVideo \cite{hong2022cogvideolargescalepretrainingtexttovideo} and \cite{villegas2022phenakivariablelengthvideo} introduced scalable architectures for text-to-video generation. More recent models like Sora \cite{brooks2024video}, MovieGen \cite{polyak2025moviegencastmedia}, Veo~3 \cite{deepmind2025veo3}, and CogVideoX \cite{yang2024cogvideox} set new standards for quality and realism. Recent work has investigated controllable video generation \cite{agarwal2025cosmos, hassan2024gemgeneralizableegovisionmultimodal, kanervisto2025world}, with the goal of producing realistic, high-quality videos while allowing precise control over motion and dynamics. These methods emphasize modeling dynamic environments and predicting plausible future states conditioned on past observations and control inputs.

\noindent\textbf{Visual Foundation Models}
Recent work has investigated the use of generative models as generalist vision models. Methods such as image inpainting for visual prompting \cite{bar2022visual} and image-based in-context learning \cite{wang2023images} demonstrate that structured inputs can enable these models to solve diverse tasks. Diffusion models have further been extended to in-context learning \cite{wang2023context}, instruction following across heterogeneous tasks \cite{geng2024instructdiffusion}, and broader computer vision problem solving \cite{zhao2025diception}. Sequential modeling has been proposed as a unified interface for scaling vision models \cite{bai2024sequential}. Building on this line of work, \cite{lin2025realgeneral} train CogVideoX1.5 with temporal in-context prompts for multi-task learning, but their focus remains on broad computer vision benchmarks rather than visual intelligence, and their method requires extensive training\footnote{We add qualitative results on standard computer vision tasks in the Appendix to show that our framework can also be extended to this setting.}.

Our approach does not attempt to build a foundation model from scratch. Instead, we investigate whether a pretrained VDM, pretrained extensively on next-frame prediction, can begin to exhibit the properties expected of visual foundation models by leveraging inductive biases gained through spatiotemporal pretraining.

\section{Methodology}

\subsection{Setup and Comparison Protocol}

We adopt the definition of intelligence proposed by \cite{chollet2019measure}:

\begin{quote}
\textit{The intelligence of a system is a measure of its skill acquisition efficiency over a scope of tasks with respect to priors, experience, and generalization difficulty.}
\end{quote}

This perspective motivates our evaluation design. We focus not only on absolute accuracy but also on how quickly models acquire new capabilities when exposed to limited supervision.

To evaluate our hypothesis we curate a diverse benchmark of visually grounded tasks that can be specified textually as grid-based problems, including ARC-AGI, Sudoku solving, and route planning. We now describe the evaluation setup in detail.

Let \(\mathcal{T}\) denote a task with dataset \(\mathcal{D}_{\mathcal{T}} = \{(x_i, y_i)\}_{i=1}^{n}\), where each \(x_i\) and \(y_i\) is an input-output pair. Each sample is expressed in two complementary modalities:

\begin{description}[leftmargin=!,labelwidth=\widthof{\bfseries Image}]
    \item[\vdmbestnt{\textbf{Image}}] An \textbf{image pair} \((I(x_i), I(y_i))\), where \(I(\cdot)\) deterministically renders RGB images of size \((3 \times H \times W)\).
    \item[\llmbestnt{\textbf{Text}}] A \textbf{JSON pair} \((J(x_i), J(y_i))\), where \(J(\cdot)\) maps a grid to a compact JSON string. 
\end{description}

We serialize samples in a neutral format that avoids domain-specific priors, requiring both models to infer task rules directly from raw representations. Training and evaluation splits are identical across all models to ensure a fair and controlled comparison. VDMs are trained directly on the image modality using our approach, which we detail in the next section, while LLMs are trained on the text modality.

We define accuracy as the proportion of test instances where the predicted output \textit{exactly matches} the ground truth grid. For tasks where multiple valid solutions may exist (e.g., \textit{Sudoku}, \textit{Sudoku Mini}, \textit{Hitori}), we filter datasets to ensure each instance has an unique solution. When unique solutions cannot easily be guaranteed, as in \textit{Shortest Path}, we introduce complementary metrics to better capture solution quality (see Section~\ref{sec:visual-route-planning}).

To evaluate efficiency of skill acquisition, we consider two complementary settings.  

\textbf{ARC Family.} We evaluate models on ARC-AGI and ConceptARC, where the challenge is to solve diverse tasks from only 2–5 demonstrations. Following prior work \cite{moskvichev2023conceptarc, chollet2019measure, li2025combining}, we measure how many tasks each model can solve under this minimal supervision regime.  

\textbf{Structured Visual Tasks.} We then turn to structured benchmarks. Here we systematically vary \(n\), the number of training examples per task, to trace curves and quantify the rate of skill acquisition rather than focusing solely on endpoint accuracy.

\subsection{Adapting Video Diffusion Models for Image-to -Image}

We adapt pretrained VDMs to image-to-image (I2I) prediction tasks by re-framing each input–output pair \((I_{x_i}, I_{y_i})\) as a short \emph{transition video}. This leverages the generative prior of VDMs, while requiring minimal supervision.

\paragraph{\vdmbestnt{Transition video construction}}  
Each pair \((x_i, y_i)\) is converted into a temporal sequence \(\mbox{$v_i = [v_{i,1}, \dots, v_{i,F}]$}\), where
\[
v_{i,1} = I(x_i), \quad v_{i,F} = I(y_i).
\]
Intermediate frames are generated with an interpolation function \(\phi\). For example, a \textit{convex interpolation} produces a smooth transition
\[
v_{i,f} = \left(1 - \alpha\right) I(x_i) + \alpha I(y_i), \text{ where }\alpha = \tfrac{f-1}{F-1}, \text{ and }
 f = 1, \dots, F,
\]
while a \textit{discrete interpolation} simply holds the input frame for the first half of the sequence and afterwards switches to the output frame:
\[
v_{i,f} = 
\begin{cases}
I(x_i), & f \leq F/2,\\
I(y_i), & f > F/2.
\end{cases}
\]
This yields a dataset $\mathcal{V}_\mathcal{T}$ of input-conditioned video trajectories. 
For our comparisons, we adopt the \textit{discrete interpolation} to avoid introducing any biases.

\paragraph{\vdmbestnt{Fine-tuning}}  
We adapt a pretrained VDM by conditioning on the first frame \(v^0_1\) and a neutral fixed text embedding \(e_{\text{text}}\). Given a noisy video \(v^t\) at step \(t\), the model predicts noise \(\epsilon_\theta\) via
\[
\mathcal{L}_{\text{VDM}} = \mathbb{E}_{v^0 \sim \mathcal{V}_\mathcal{T}, \epsilon \sim \mathcal{N}(0, \mathbf{I}), t}\left[\|\epsilon - \epsilon_\theta(v^t, t, c)\|_2^2\right], \quad c = \{v^0_1, e_{\text{text}}\}.
\]
We use LoRA modules for fine-tuning, updating only these additional weights while keeping the pretrained model frozen.

\paragraph{\vdmbestnt{Inference}}  
At test time, the model generates predictions through reverse diffusion. The procedure is detailed in Algorithm~\ref{alg:vdm-inference}.

% \begin{algorithm}[t]
% \caption{Inference for \vdmbestnt{Video Diffusion Model}}
% \label{alg:vdm-inference}
% \begin{algorithmic}[1]
% \Require Test image $I(x_{\text{test}})$, text embedding $e_{\text{text}}$
% \State Encode input: $c_{\text{test}} \gets \{I(x_{\text{test}}), e_{\text{text}}\}$
% \State Initialize noise: sample $v^T \sim \mathcal{N}(0, \mathbf{I})$
% \State Reverse diffusion: recover $v^0$ conditioned on $c_{\text{test}}$
% \State Output prediction: $\hat{y} \gets v^0_F$ (final frame)
% \end{algorithmic}
% \end{algorithm}

This procedure reframes image-to-image prediction as a conditional video generation problem, enabling efficient adaptation of pretrained VDMs to new tasks.

\subsection{Adapting Large Language Models}

We adapt pretrained LLMs to structured prediction tasks by framing each example as a JSON-to-JSON translation problem.   

\paragraph{\llmbestnt{Fine-tuning}}  
We adapt pretrained LLMs using a standard sequence-to-sequence objective. Given tokenized input–output pairs, the model is trained to maximize the likelihood of the target sequence under teacher forcing:
\[
\mathcal{L}_{\text{LLM}} = \frac{1}{n} \sum_{i=1}^{n} \sum_{t=1}^{|\mathbf{v}_i|} -\log p_\theta(v_{i,t} \mid \mathbf{u}_i, \mathbf{v}_i^{<t}).
\]

We insert LoRA modules into the pretrained backbone, fine-tuning only these lightweight adapters while keeping the majority of parameters frozen.

\paragraph{\llmbestnt{Inference}}  
At test time, predictions are generated autoregressively. The procedure is summarized in Algorithm~\ref{alg:llm-inference}.

% \begin{algorithm}[t]
% \caption{Inference for \llmbestnt{Language Model}}
% \label{alg:llm-inference}
% \begin{algorithmic}[1]
% \State Encode input: $J(x_{\text{test}})$ as JSON string
% \State Tokenize and feed sequence into model
% \State Autoregressively decode output until termination
% \State Return prediction: $\hat{y}$ as JSON string
% \end{algorithmic}
% \end{algorithm}

\begin{figure}[t]
\centering
\begin{minipage}{0.48\linewidth}
\begin{algorithm}[H]
\caption{Inference for \vdmbestnt{VDM}}
\label{alg:vdm-inference}
\begin{algorithmic}[1]
\State Encode input: $c_{\text{test}} \gets \{I(x_{\text{test}}), e_{\text{text}}\}$
\State Initialize noise: sample $v^T \sim \mathcal{N}(0, \mathbf{I})$
\State Reverse diffusion: recover $v^0$ conditioned on $c_{\text{test}}$
\State Output prediction: $\hat{y} \gets v^0_F$ (final frame)
\end{algorithmic}
\end{algorithm}
\end{minipage}
\hfill
\begin{minipage}{0.48\linewidth}
\begin{algorithm}[H]
\caption{Inference for \llmbestnt{LLM}}
\label{alg:llm-inference}
\begin{algorithmic}[1]
\State Encode input: $J(x_{\text{test}})$ as JSON string
\State Tokenize and feed sequence into model
\State Autoregressively decode output until termination
\State Return prediction: $\hat{y}$ as JSON string
\end{algorithmic}
\end{algorithm}
\end{minipage}
\end{figure}

\section{Experiments}

\subsection{ARC Family}

The ARC-AGI benchmark~\cite{chollet2019measure} evaluates an agent's ability to infer and apply abstract patterns through compositional understanding, few-shot learning, and inductive generalization. Each ARC task provides only a handful of input--output examples (typically 2--5), requiring the model to discover the underlying transformation rule and apply it to novel test inputs. This benchmark is widely regarded as a challenging measure of progress in abstraction and generalization.

\begin{figure}[h]
\centering
\begin{minipage}[t]{0.54\textwidth}
    \vspace*{-4.6cm} % adjust this value to push table upward
    \centering
    \captionof{table}{ARC-AGI test performance. Following the official evaluation protocol \cite{chollet2024arc}, models are evaluated with two attempts per test input. We also report single-attempt results for comparability with commercial LLMs, which are only available under this setting.}
    \begin{tabular}{lc}
        \toprule
        \textbf{Model} & \textbf{Accuracy (\%)} \\
        \midrule
        \multicolumn{2}{c}{\textbf{Two-attempts setting}} \\
        \midrule
        \vdmbest{CogVideoX1.5-5B} & \vdmbest{16.75} \\
        \llmbest{Qwen3-4B-Instruct-2507} & \llmbest{8.00} \\
        \midrule
        \multicolumn{2}{c}{\textbf{Single-attempt setting}} \\
        \midrule
        \vdmbest{CogVideoX1.5-5B} & \vdmbest{12.50} \\
        \llmbest{Qwen3-4B-Instruct-2507} & \llmbest{6.75} \\
        OpenAI o1-preview & 21.00 \\
        Anthropic Claude 3.5 Sonnet & 21.00 \\
        OpenAI GPT-4o & \phantom{0}9.00 \\
        Google Gemini 1.5 & \phantom{0}8.00 \\
        \bottomrule
    \end{tabular}
    \label{tab:results-arc-agi}
\end{minipage}\hfill
\begin{minipage}[t]{0.42\textwidth}
    \centering
    \includegraphics[width=\linewidth]{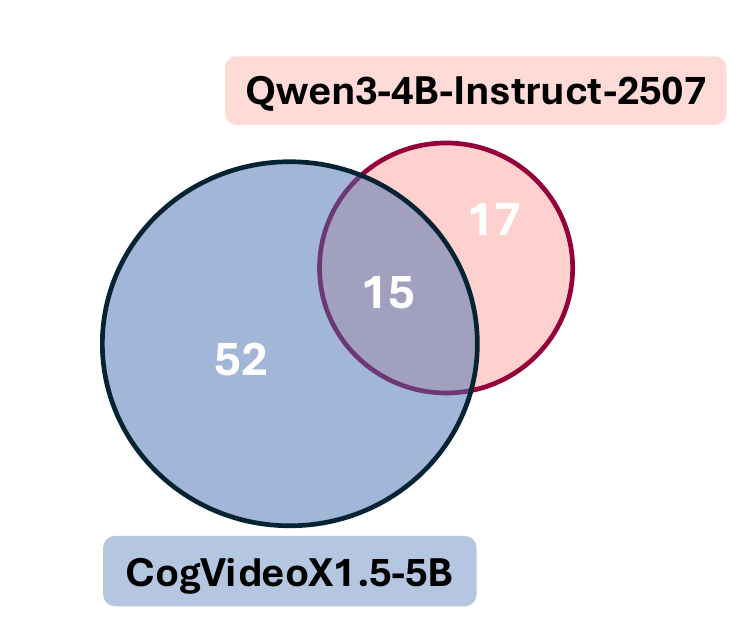}
    \caption{Venn diagram of ARC-AGI tasks showing those solved exclusively by each model and those solved by both.}
    \label{fig:arc-agi-venn}
\end{minipage}
\end{figure}

\begin{figure}[h]
\centering
\renewcommand{\arraystretch}{0.6}
\setlength{\tabcolsep}{2pt}
\begin{tabular}{lcccccc}
& \textbf{Input} & \textbf{Output} & \textbf{Input} & \textbf{Output} & \textbf{Input} & \textbf{Output} \\[0.2em]
\multirow{3}{*}{\rotatebox{90}{\textbf{Training Examples}}} &
\includegraphics[width=1.8cm,height=1.8cm]{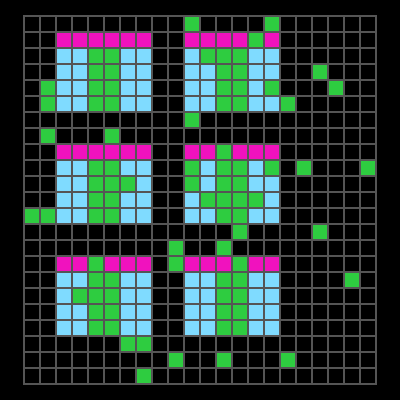} &
\includegraphics[width=1.8cm,height=1.8cm]{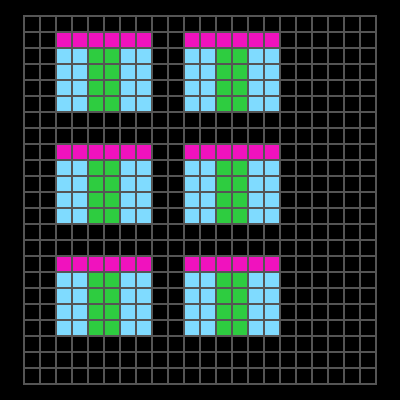} &
\includegraphics[width=1.8cm,height=1.8cm]{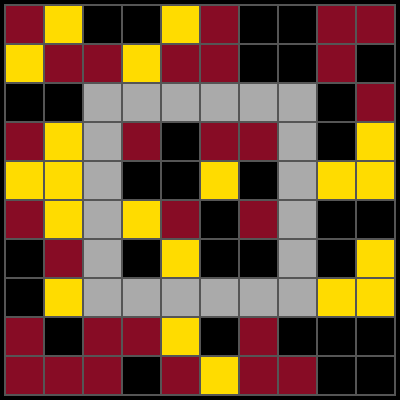} &
\includegraphics[width=1.8cm,height=1.8cm]{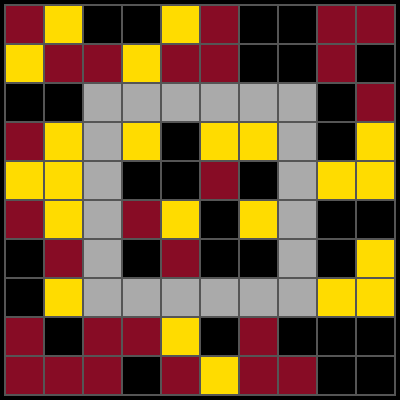} &
\includegraphics[width=1.8cm,height=1.8cm]{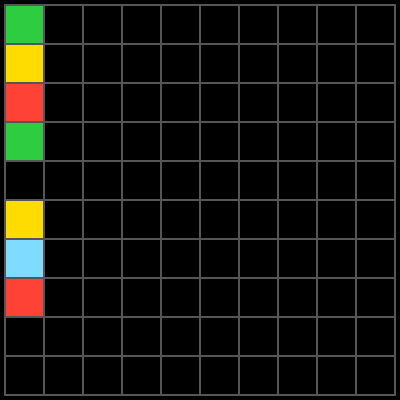} &
\includegraphics[width=1.8cm,height=1.8cm]{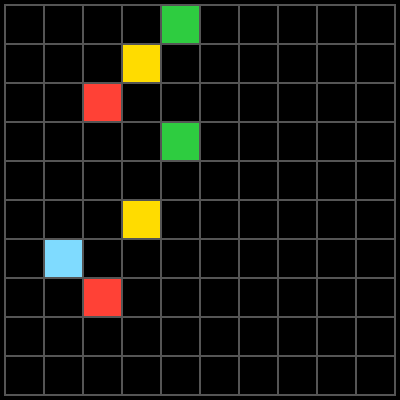} \\
& \includegraphics[width=1.8cm,height=1.8cm]{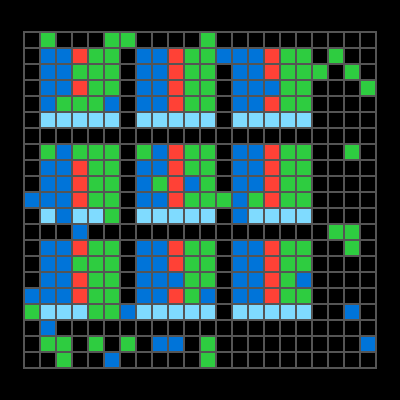} &
\includegraphics[width=1.8cm,height=1.8cm]{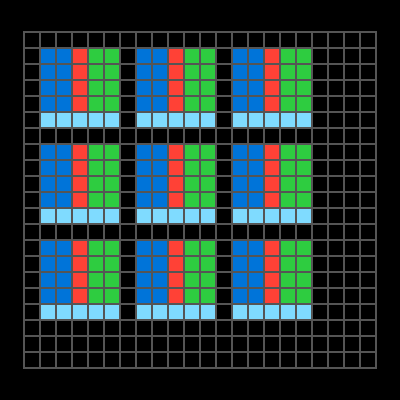} &
\includegraphics[width=1.8cm,height=1.8cm]{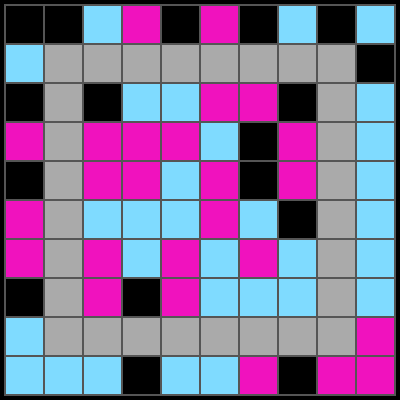} &
\includegraphics[width=1.8cm,height=1.8cm]{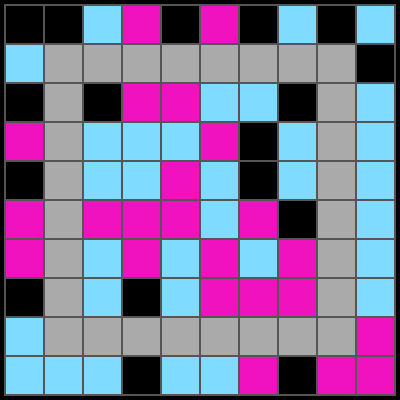} &
\includegraphics[width=1.8cm,height=1.8cm]{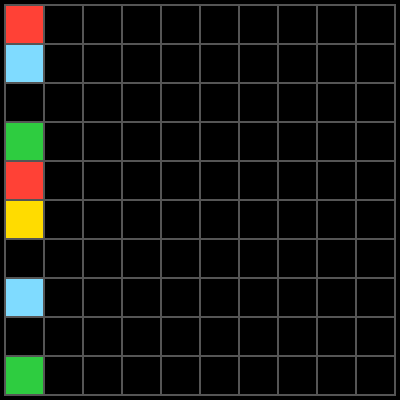} &
\includegraphics[width=1.8cm,height=1.8cm]{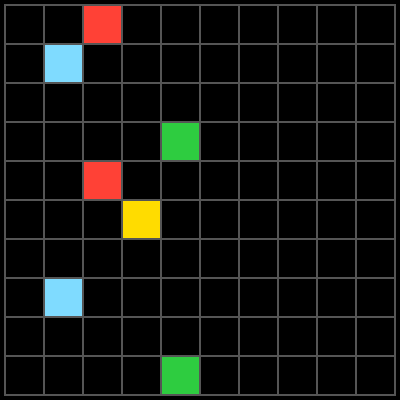} \\
& \includegraphics[width=1.8cm,height=1.8cm]{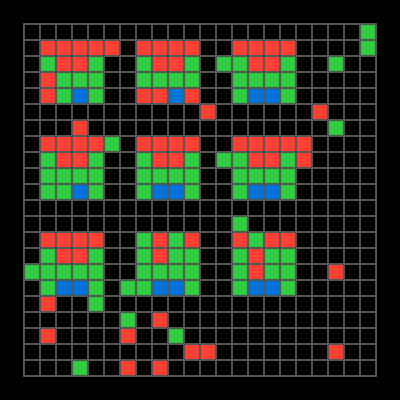} &
\includegraphics[width=1.8cm,height=1.8cm]{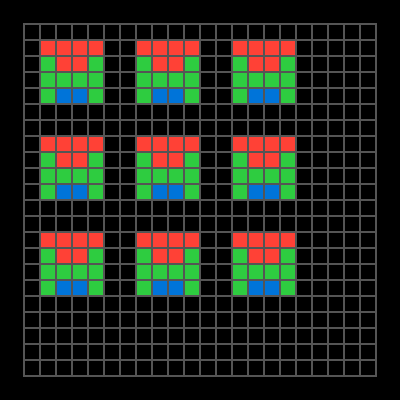} &
\includegraphics[width=1.8cm,height=1.8cm]{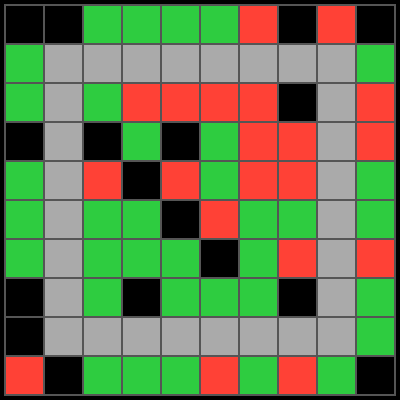} &
\includegraphics[width=1.8cm,height=1.8cm]{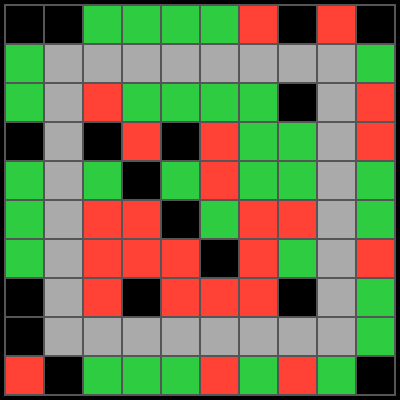} &
\includegraphics[width=1.8cm,height=1.8cm]{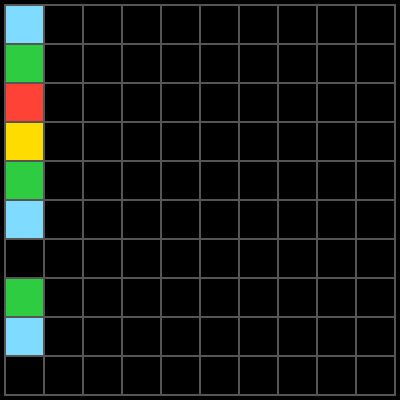} &
\includegraphics[width=1.8cm,height=1.8cm]{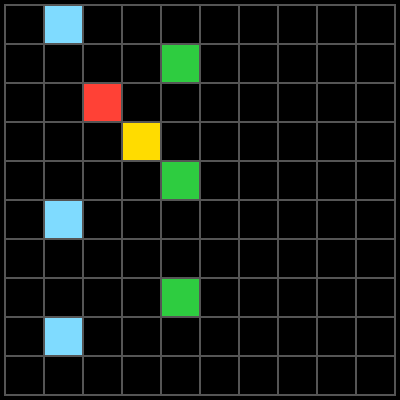} \\[0.4em]

& \textbf{Input} & \textbf{Prediction} & \textbf{Input} & \textbf{Prediction} & \textbf{Input} & \textbf{Prediction} \\
\rotatebox{90}
{\scriptsize\shortstack{\phantom{Q}\\\textbf{\vdmbestnt{CogVideoX1.5-5B}}}} &
\includegraphics[width=1.8cm,height=1.8cm]{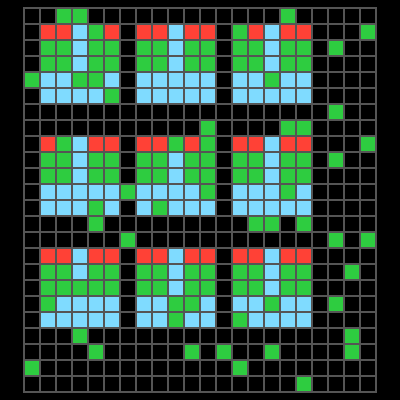} &
\includegraphics[width=1.8cm,height=1.8cm]{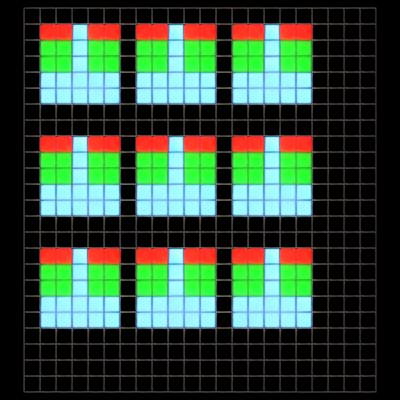} &
\includegraphics[width=1.8cm,height=1.8cm]{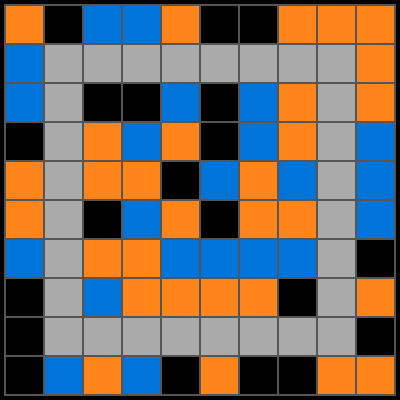} &
\includegraphics[width=1.8cm,height=1.8cm]{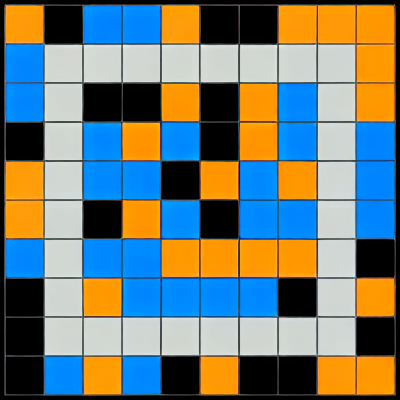} &
\includegraphics[width=1.8cm,height=1.8cm]{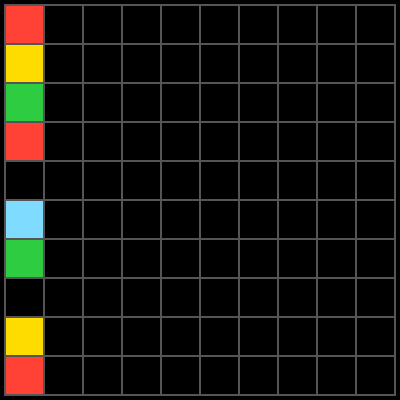} &
\includegraphics[width=1.8cm,height=1.8cm]{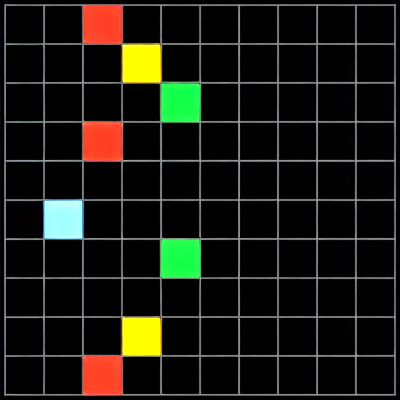} \\
\rotatebox{90}{\llmbestnt{\scriptsize\shortstack{\textbf{Qwen3-4B}\\\textbf{Instruct-2507}}}} &
\includegraphics[width=1.8cm,height=1.8cm]{figures/main/examples/arc_agi_005/arc_agi_gt_image_0_005.png} &
\includegraphics[width=1.8cm,height=1.8cm]{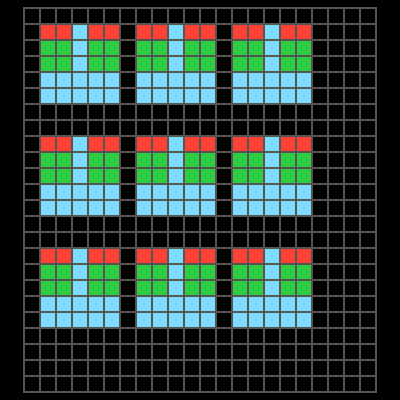} &
\includegraphics[width=1.8cm,height=1.8cm]{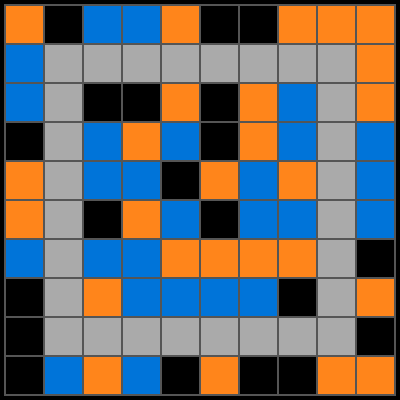} &
\includegraphics[width=1.8cm,height=1.8cm]{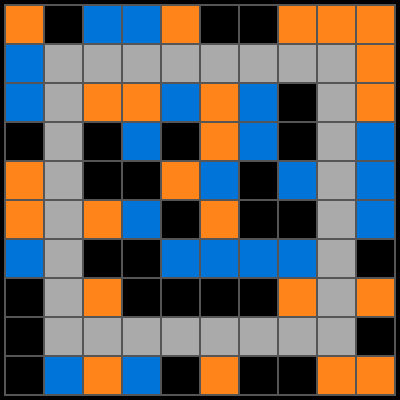} &
\includegraphics[width=1.8cm,height=1.8cm]{figures/main/examples/arc_agi_384/arc_agi_gt_image_0_384.png} &
\includegraphics[width=1.8cm,height=1.8cm]{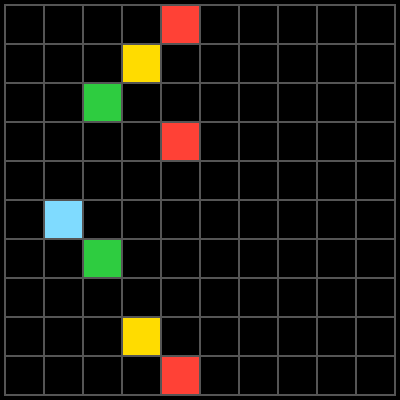} \\
\end{tabular}
\caption{Qualitative results on ARC-AGI for problems \textit{0607ce86}, \textit{7ee1c6ea}, and \textit{f45f5ca7}.}
\label{fig:arc-agi-qualitative}
\end{figure}

We follow the evaluation protocol of \cite{chollet2024arc}, which allows up to two attempts per test input and counts a question as solved only if all predictions match the ground truth. Quantitative results appear in Table \ref{tab:results-arc-agi}, with qualitative examples in Figure \ref{fig:arc-agi-qualitative}. For comparison, we also report single-attempt results of commercial LLMs from \cite{chollet2024arc}. Figure \ref{fig:arc-agi-venn} illustrates the overlap between tasks solved by the VDM and the LLM, underscoring their complementary strengths.

% \begin{table}[t]
%     \centering
%     \caption{ARC-AGI test performance. Following the official evaluation protocol \cite{chollet2024arc}, our models are evaluated with two attempts per test input. We also report single-attempt results for comparability with commercial LLMs, which are only available under this setting.}
%     \begin{tabular}{lc}
%         \toprule
%         \textbf{Model} & \textbf{Accuracy (\%)} \\
%         \midrule
%         \multicolumn{2}{c}{\textbf{Two-attempt setting}} \\
%         \midrule
%         \vdmbest{CogVideoX1.5-5B} & \vdmbest{16.75} \\
%         \llmbest{Qwen3-4B-Instruct-2507} & \llmbest{8.00} \\
%         \midrule
%         \multicolumn{2}{c}{\textbf{Single-attempt setting}} \\
%         \midrule
%         \vdmbest{CogVideoX1.5-5B} & \vdmbest{12.50} \\
%         \llmbest{Qwen3-4B-Instruct-2507} & \llmbest{6.75} \\
%         OpenAI o1-preview & 21.00 \\
%         Anthropic Claude 3.5 Sonnet & 21.00 \\
%         OpenAI GPT-4o & \phantom{0}9.00 \\
%         Google Gemini 1.5 & \phantom{0}8.00 \\
%         \bottomrule
%     \end{tabular}
%     \label{tab:results-arc-agi}
% \end{table}

We evaluate models on ConceptARC \cite{moskvichev2023conceptarc}, a curated variant of ARC designed to systematically measure visual concept understanding and generalization. ConceptARC groups tasks into 16 concept categories (for example, Above and Below, Center, Count), with each category containing 10 tasks. Each task includes 3 distinct test inputs, creating controlled variation in visual patterns and object relationships while maintaining internal consistency within each concept group. Following the protocol of \cite{moskvichev2023conceptarc}, we allow three attempts per test input and mark an input as solved if any attempt is correct. Performance is reported in Figure \ref{fig:concept-arc-radar-all}, where we further include as VDMs: Wan2.1-14B \cite{wan2p1}, LTX-13B, LTX-2B \cite{hacohen2024ltx}, CogVideoX1.5-5B \cite{yang2024cogvideox} and as LLMs: Qwen3-4B-Instruct-2507, Qwen3-8B \cite{qwen3technicalreport}, Llama3.1-8B \cite{meta2024llama31}, and GPT-4 in an IC setting \cite{moskvichev2023conceptarc}. Full table with results is included in the Appendix.

These results highlight the importance of strong visual priors: by leveraging representations that capture spatial structure, compositionality, and low-level visual cues, the VDM is able to approach these abstract tasks in a way that improves upon traditional text-centric approaches.

\noindent

\subsection{Structured Visual Tasks}

From this point onward, we focus on one representative model from each family: \vdmbestnt{\textbf{CogVideoX1.5-5B}} \cite{yang2024cogvideox} for video diffusion models and \llmbestnt{\textbf{Qwen3-4B-Instruct-2507}} \cite{qwen3technicalreport} for language models. This pairing aligns model scale while contrasting pretraining modalities, allowing us to examine how different priors influence adaptability to visually grounded tasks.

\subsubsection{Visual Games}

As part of our broader evaluation, we examine performance on a diverse set of five visual games that span both puzzle-solving and board play. These tasks provide an additional perspective on how the models handle structured visual inputs and varying interaction styles. The puzzle-based tasks, \textit{Hitori 5x5} and two versions of \textit{Sudoku} (standard one and \textit{Mini}), focus on solving constraint-based problems in structured grids, where success depends on extracting spatial patterns and enforcing global consistency from local information. The board games, \textit{Connect 4} and \textit{Chess Mate-in-1}, shift attention to game scenarios where the goal is to identify the winning move in a given configuration. Together, these games cover a range of visual layouts and structured objectives, complementing the other tasks explored in this study. 

\begin{figure}[t]
    \centering
    \includegraphics[width=\linewidth]{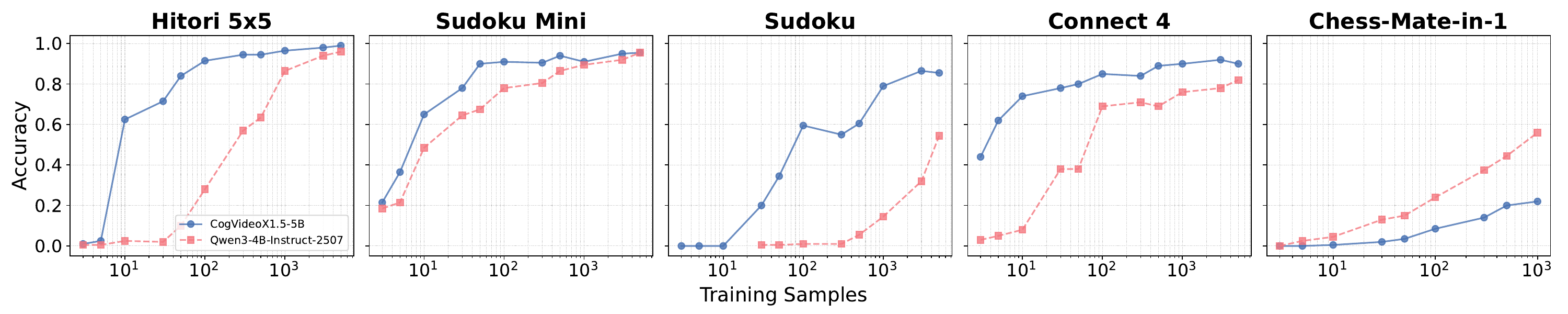}
    \caption{Accuracy as a function of training set size for \vdmbestnt{CogVideoX1.5-5B} and \llmbestnt{Qwen3-4B-Instruct-2507} on five visual games.}
    \label{fig:main-games-accuracies}
\end{figure}

Figure~\ref{fig:main-games-accuracies} presents model performance as a function of the number of training samples. CogVideoX1.5-5B demonstrates strong scaling behavior across most tasks, surpassing Qwen3-4B-Instruct-2507 in four of the five games. Its advantage is particularly clear in \textit{Sudoku} and \textit{Hitori}, which rely on interpreting complex grid layouts and visual compositions. This supports the view that VDMs capture compositional features in visual data more effectively than LLMs, which are primarily optimized for language. The only exception is chess, where Qwen3-4B-Instruct-2507 performs better, likely reflecting the abundance of chess material in textual corpora that LLMs can partially internalize during pretraining \cite{kuo2023chess}.

\subsubsection{Route Planning}\label{sec:visual-route-planning}

We evaluate route planning in \(2\)D grid environments through two tasks: \textit{Maze} and \textit{Shortest Path}. In \textit{Maze}, the model must navigate from the top-left to the bottom-right corner of a grid. In \textit{Shortest Path}, the objective is to connect two arbitrary points with the shortest possible route. For \textit{Shortest Path}, we report two complementary metrics to assess model performance:

\paragraph{Path Success Rate (PSR)} The percentage of evaluation examples where the predicted path forms a continuous connection between the source and target locations.

\paragraph{Relative Path Length (RPL)} For cases \textbf{where a valid path is produced}, we compute
\[
    \text{RPL} = \frac{\text{Predicted Path Length}}{\text{Ground-Truth Shortest Path Length}}.
\]
This value may increase even as overall performance improves, since better models tend to predict good paths for more challenging cases, potentially constructing longer yet valid paths.

\begin{figure}[h]
    \centering
    \includegraphics[width=\linewidth]{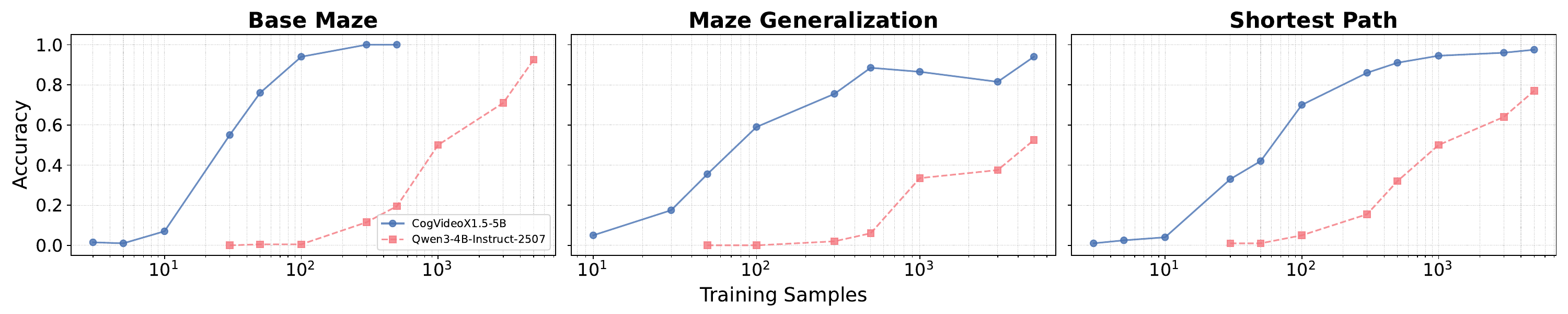}
    \caption{Accuracy as a function of training set size for \vdmbestnt{CogVideoX1.5-5B} and \llmbestnt{Qwen3-4B-Instruct-2507} on \textit{Base Maze}, \textit{Maze Generalization}, and \textit{Shortest Path}.}
    \label{fig:main-navigation-mazes-accuracies}
\end{figure}

For \textit{Maze}, we evaluate in two settings: a \textbf{matched-scale} (\textit{Base Maze}) scenario, where both training and evaluation are conducted on \(21 \times 21\) mazes to study performance as a function of training sample size; and a \textbf{generalization} scenario, where models are trained on smaller \(13 \times 13\) grids and tested on larger \(21 \times 21\) grids to assess cross-scale generalization (\textit{Maze Generalization}).

\begin{figure}[h]
\centering
\begin{minipage}[h]{0.48\textwidth}
    \centering
    \begin{tabular}{lcc}
        & \vdmbestnt{\textbf{CogVideoX1.5-5B}} & \llmbestnt{\shortstack{\textbf{Qwen3-4B}\\\textbf{Instruct-2507}}} \\[0.5em]
        \raisebox{0.15cm}{\rotatebox{90}{\textbf{Base Maze}}} &
        \includegraphics[width=2cm,height=2cm]{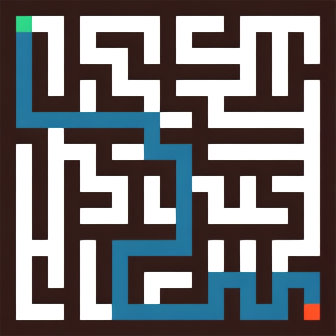} &
        \includegraphics[width=2cm,height=2cm]{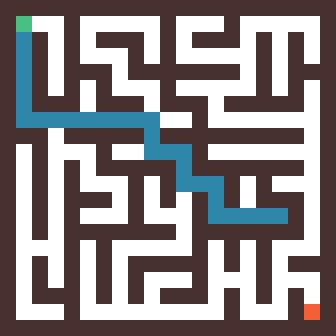} \\
        \raisebox{0.0cm}{\rotatebox{90}{\textbf{Shortest Path}}} &
        \includegraphics[width=2cm,height=2cm]{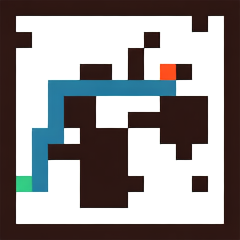} &
        \includegraphics[width=2cm,height=2cm]{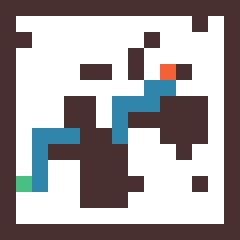} \\
    \end{tabular}
    \caption{Qualitative examples for \textit{Base Maze} and \textit{Shortest Path} tasks, after fine-tuning with \(n=300\) samples.}
\end{minipage}\hfill
\begin{minipage}[h]{0.48\textwidth}
    \vspace*{-1.45cm}
    \centering
    \captionof{table}{Relative Path Length (RPL) and Path Success Rate (PSR) for both models across training sample sizes for \textit{Shortest Path}.}
    \resizebox{\linewidth}{!}{%
    \begin{tabular}{c|cc|cc}
        \toprule
        \multirow{2}{*}{\textbf{Samples}} & \multicolumn{2}{c|}{\textbf{CogVideoX1.5-5B}} & \multicolumn{2}{c}{\textbf{Qwen3-4B-Instruct-2507}} \\
        \cmidrule{2-5}
         & RPL ↓ & PSR ↑ & RPL ↓ & PSR ↑ \\
        \midrule
        3    & 1.005 & 0.115 & --    & --    \\
        5    & 1.089 & 0.160 & --    & --    \\
        10   & 1.060 & 0.245 & --    & --    \\
        30   & 1.028 & 0.670 & 1.020 & 0.015 \\
        50   & 1.013 & 0.645 & 1.038 & 0.060 \\
        100  & 1.017 & 0.870 & 1.025 & 0.205 \\
        300  & 1.007 & 0.940 & 1.040 & 0.530 \\
        500  & 1.005 & 0.985 & 1.019 & 0.605 \\
        1000 & 1.005 & 0.990 & 1.043 & 0.710 \\
        3000 & 1.000 & 0.990 & 1.026 & 0.795 \\
        5000 & 1.001 & 1.000 & 1.016 & 0.870 \\
        \bottomrule
    \end{tabular}
    }
    \label{tab:shortest-path-metrics}
\end{minipage}
\end{figure}

Accuracy results are shown in Figure \ref{fig:main-navigation-mazes-accuracies}. For \textit{Shortest Path}, additional metrics are reported in Table \ref{tab:shortest-path-metrics}. The VDM consistently constructs valid paths with far fewer supervised examples, achieving up to a tenfold reduction in data requirements in low-sample regimes, which underscores its stronger inductive biases relative to the LLM. Moreover, it demonstrates the ability to generalize much quicker from limited training on smaller mazes to larger, more complex ones.

\subsubsection{Cellular Automata}

We evaluate the capacity of both models to capture complex spatial patterns in cellular automata (CA). Our study spans one-dimensional Elementary Cellular Automata (ECA)~\cite{Wolfram1984Universality}, a foundational class of binary-state systems, as well as two-dimensional Life-like Cellular Automata, including Conway's Game of Life~\cite{Gardner1970Life}, defined by various birth and survival (B/S) rules. Additionally, we consider Langton's ant~\cite{Langton1986Studying}, a deterministic agent-based system, where the task is to predict the complete grid state after \(n\) steps of evolution.

\begin{figure}[h]
    \centering
    \includegraphics[width=1.0\linewidth]{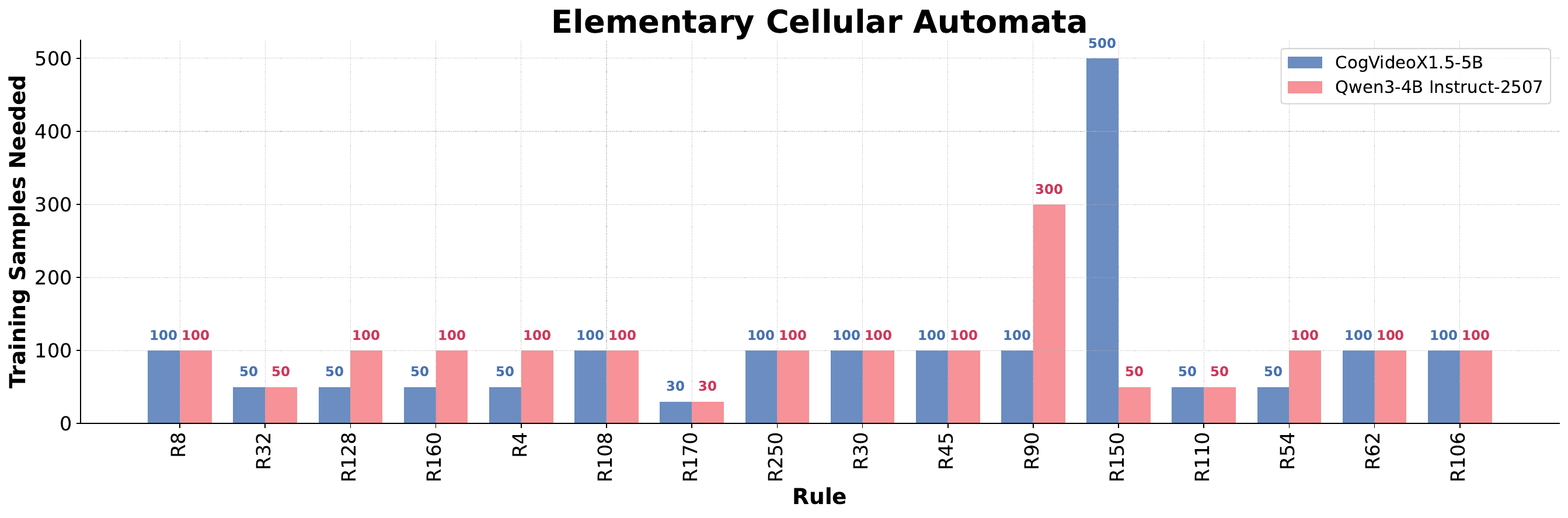}
    
    \caption{Number of training examples required to achieve \(\delta \geq 0.9\) accuracy for selected 1D ECA rules (lower is better).}
    \label{fig:eca-n-to-threshold}
\end{figure}

For the 1D ECA experiments, we evaluate four representative rules from each of Wolfram's four complexity classes. We measure task completion as achieving an accuracy above a fixed threshold \(\delta = 0.9\). Figure~\ref{fig:eca-n-to-threshold} reports the number of training examples required to reach this performance for each rule. Across these rules, both models show broadly similar behavior, with the VDM being better in some cases and worse in others, though overall it remains competitive with the LLM.

\begin{figure}[h]
    \centering
    \begin{minipage}[h]{0.48\linewidth}
        \centering
        \includegraphics[width=\linewidth]{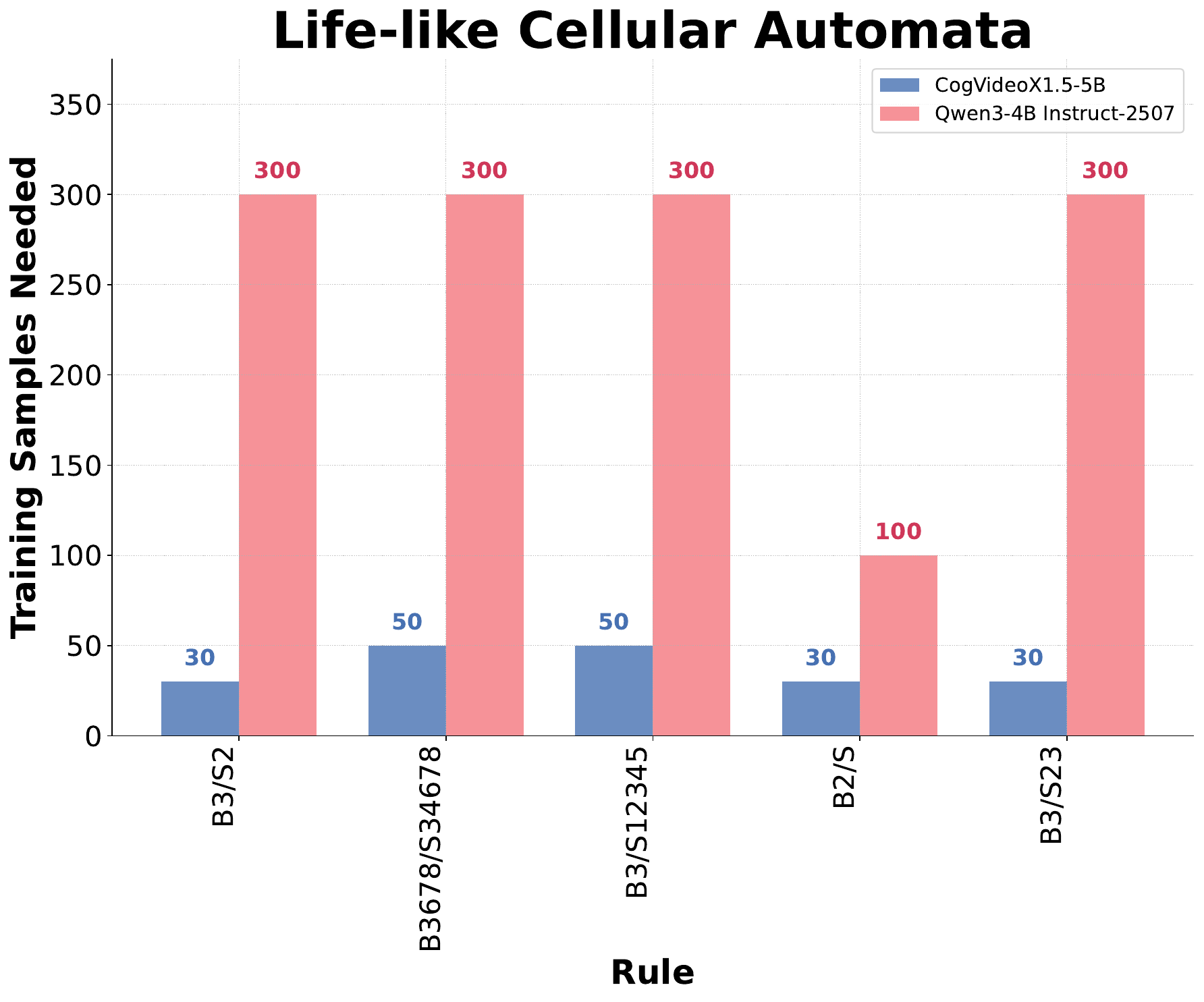}
        \caption{Number of training examples required to achieve \(\delta \geq 0.9\) accuracy for selected Life-like cellular automata rules (lower is better).}
        \label{fig:main:gol-n-to-threshold}
    \end{minipage}\hfill
    \begin{minipage}[h]{0.48\linewidth}
        \centering
        \raisebox{0.55cm}{
            \includegraphics[width=\linewidth]{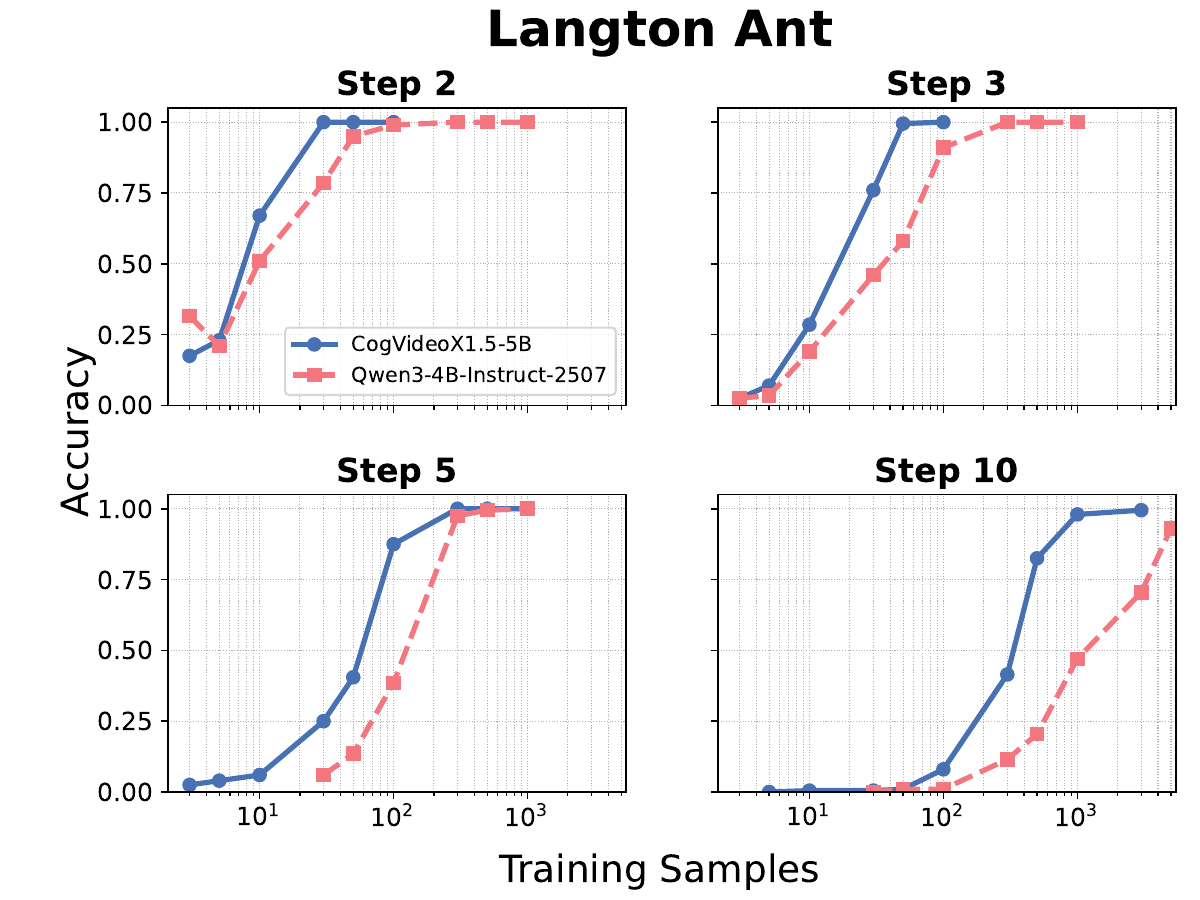}
        }
        \caption{Accuracy as a function of training set size for \vdmbestnt{CogVideoX1.5-5B} and \llmbestnt{Qwen3-4B-Instruct-2507} on \textit{Langton’s Ant} with a prediction horizon of 2,3,5 and 10.}
        \label{fig:main:langton-ant-accuracy}
    \end{minipage}
\end{figure}

In two-dimensional settings, clearer differences emerge (see Figures \ref{fig:main:gol-n-to-threshold}, \ref{fig:main:langton-ant-accuracy}). For Life-like cellular automata, the VDM reaches threshold accuracy with far fewer examples, and a similar advantage is observed in Langton's ant. In the case of Langton's ant, the gap grows larger as the number of steps to be predicted increases, indicating that the VDM scales more effectively on tasks that demand long-range spatial planning.

\begin{figure}[h]
\centering
\begin{tabular}{lcccc}
& \textbf{Input} & \textbf{Output} & \vdmbestnt{\textbf{CogVideoX1.5-5B}} & \llmbestnt{\shortstack{\textbf{Qwen3-4B}\\\textbf{Instruct-2507}}} \\[0.5em]
\raisebox{0.40cm}{\rotatebox{90}{\textbf{B3/S23}}} &
\includegraphics[width=2cm,height=2cm]{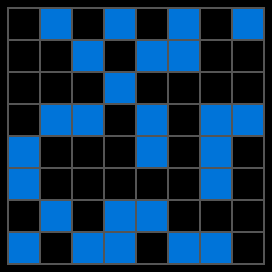} &
\includegraphics[width=2cm,height=2cm]{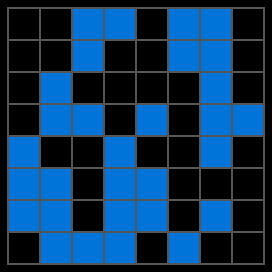} &
\includegraphics[width=2cm,height=2cm]{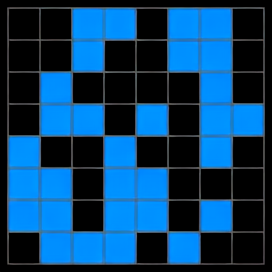} &
\includegraphics[width=2cm,height=2cm]{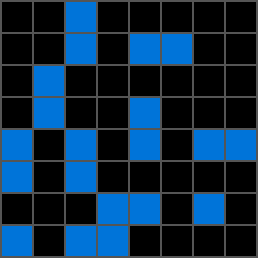} \\[0.5em]
\raisebox{0.45cm}{\rotatebox{90}{\textbf{B2/S}}} &
\includegraphics[width=2cm,height=2cm]{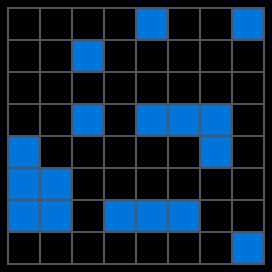} &
\includegraphics[width=2cm,height=2cm]{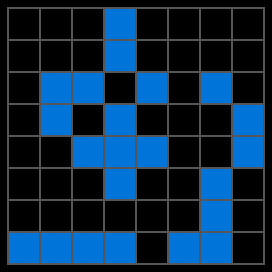} &
\includegraphics[width=2cm,height=2cm]{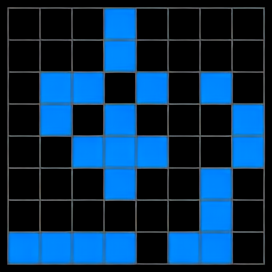} &
\includegraphics[width=2cm,height=2cm]{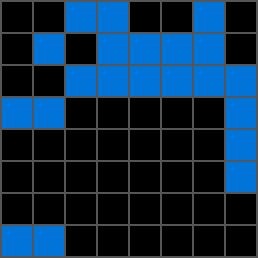} \\
\end{tabular}
\caption{Qualitative examples for Life-like cellular automata with rules \textit{B3/S23} and \textit{B2/S} tasks,  after fine-tuning with \(n = 30\) samples.}
\end{figure}

\section{Conclusions}

Our study shows that VDMs pretrained on spatiotemporal data adapt effectively to structured visual tasks with fewer training examples than comparable LLMs. This demonstrates how modality-aligned pretraining and inductive biases support transfer: VDMs excel in tasks requiring spatial structure and temporal transformation, while LLMs retain strengths in symbol rich domains. Large-scale pretraining on spatiotemporal data with representations aligned to visual structure thus emerges as a promising venue for advancing visual intelligence.

The implications are twofold. For researchers, our benchmarks provide evidence that pretraining pipelines designed around modality-specific structure can unlock new capabilities, offering a path toward more data-efficient models. For practitioners, the inclusion of navigation-style tasks such as mazes and route planning suggests that pretrained VDMs may hold potential for downstream domains like planning, simulation, or robotics. However, validating these capabilities in more realistic, embodied environments remains an important direction for future work.

% Overall, our findings highlight the central role of modality-aligned pretraining in advancing visual intelligence, with alignment to the structure of the data reinforcing its effectiveness.

Overall, these results underline that modality-aligned pretraining plays a central role in advancing visual intelligence.

% Our study shows that VDMs pretrained on spatiotemporal data can adapt effectively to a wide range of structured visual reasoning tasks, requiring fewer training examples than comparable LLMs. This highlights the importance of modality aligned pretraining and inductive biases: VDMs are naturally suited for tasks that involve spatial structure and temporal transformation, while LLMs retain strengths in domains rich in symbolic knowledge.

% These findings suggest that large scale pretraining on spatiotemporal data, combined with a representation aligned to visual structure, supports stronger transfer to tasks that demand visual and spatial understanding. This highlights the potential of domain aligned pretraining as a foundation for advancing visual intelligence.

\section*{Acknowledgements}

This work was supported as part of the Swiss AI Initiative by a grant from the Swiss National Supercomputing Centre (CSCS) under project ID a03 on Alps. Pablo Acuaviva, Aram Davtyan and Sebastian Stapf were supported by SNSF Grant 10001278.

Some of the calculations were performed on UBELIX (https://www.id.unibe.ch/hpc), the HPC cluster at the University of Bern.

\bibliographystyle{iclr2026_conference}
\bibliography{iclr2026_conference}

\begin{thebibliography}{51}
\providecommand{\natexlab}[1]{#1}
\providecommand{\url}[1]{\texttt{#1}}
\expandafter\ifx\csname urlstyle\endcsname\relax
  \providecommand{\doi}[1]{doi: #1}\else
  \providecommand{\doi}{doi: \begingroup \urlstyle{rm}\Url}\fi

\bibitem[Bai et~al.(2024)Bai, Geng, Mangalam, Bar, Yuille, Darrell, Malik, and Efros]{bai2024sequential}
Yutong Bai, Xinyang Geng, Karttikeya Mangalam, Amir Bar, Alan~L Yuille, Trevor Darrell, Jitendra Malik, and Alexei~A Efros.
\newblock Sequential modeling enables scalable learning for large vision models.
\newblock In \emph{Proceedings of the IEEE/CVF Conference on Computer Vision and Pattern Recognition}, pp.\  22861--22872, 2024.

\bibitem[Bar et~al.(2022)Bar, Gandelsman, Darrell, Globerson, and Efros]{bar2022visual}
Amir Bar, Yossi Gandelsman, Trevor Darrell, Amir Globerson, and Alexei~A. Efros.
\newblock Visual prompting via image inpainting.
\newblock \emph{CoRR}, abs/2209.00647, 2022.
\newblock URL \url{https://doi.org/10.48550/arXiv.2209.00647}.

\bibitem[Blattmann et~al.(2023)Blattmann, Dockhorn, Kulal, Mendelevitch, Kilian, Lorenz, Levi, English, Voleti, Letts, Jampani, and Rombach]{blattmann2023stablevideodiffusionscaling}
Andreas Blattmann, Tim Dockhorn, Sumith Kulal, Daniel Mendelevitch, Maciej Kilian, Dominik Lorenz, Yam Levi, Zion English, Vikram Voleti, Adam Letts, Varun Jampani, and Robin Rombach.
\newblock Stable video diffusion: Scaling latent video diffusion models to large datasets.
\newblock \emph{CoRR}, abs/2311.15127, 2023.
\newblock URL \url{https://doi.org/10.48550/arXiv.2311.15127}.

\bibitem[Brown et~al.(2020)Brown, Mann, Ryder, Subbiah, Kaplan, Dhariwal, Neelakantan, Shyam, Sastry, Askell, et~al.]{brown2020language}
Tom Brown, Benjamin Mann, Nick Ryder, Melanie Subbiah, Jared~D Kaplan, Prafulla Dhariwal, Arvind Neelakantan, Pranav Shyam, Girish Sastry, Amanda Askell, et~al.
\newblock Language models are few-shot learners.
\newblock \emph{Advances in neural information processing systems}, 33:\penalty0 1877--1901, 2020.

\bibitem[Chollet(2019)]{chollet2019measure}
Fran{\c{c}}ois Chollet.
\newblock On the measure of intelligence.
\newblock \emph{arXiv preprint arXiv:1911.01547}, 2019.

\bibitem[Chollet et~al.(2024)Chollet, Knoop, Kamradt, and Landers]{chollet2024arc}
Francois Chollet, Mike Knoop, Gregory Kamradt, and Bryan Landers.
\newblock Arc prize 2024: Technical report.
\newblock \emph{arXiv preprint arXiv:2412.04604}, 2024.

\bibitem[Chowdhery et~al.(2023)Chowdhery, Narang, Devlin, Bosma, Mishra, Roberts, Barham, Chung, Sutton, Gehrmann, et~al.]{chowdhery2023palm}
Aakanksha Chowdhery, Sharan Narang, Jacob Devlin, Maarten Bosma, Gaurav Mishra, Adam Roberts, Paul Barham, Hyung~Won Chung, Charles Sutton, Sebastian Gehrmann, et~al.
\newblock Palm: Scaling language modeling with pathways.
\newblock \emph{Journal of Machine Learning Research}, 24\penalty0 (240):\penalty0 1--113, 2023.

\bibitem[Coda-Forno et~al.(2023)Coda-Forno, Binz, Akata, Botvinick, Wang, and Schulz]{coda2023meta}
Julian Coda-Forno, Marcel Binz, Zeynep Akata, Matt Botvinick, Jane Wang, and Eric Schulz.
\newblock Meta-in-context learning in large language models.
\newblock \emph{Advances in Neural Information Processing Systems}, 36:\penalty0 65189--65201, 2023.

\bibitem[Cook(2004)]{Cook2004Universality}
Matthew Cook.
\newblock Universality in elementary cellular automata.
\newblock \emph{Complex Systems}, 15\penalty0 (1):\penalty0 1--40, 2004.

\bibitem[Gardner(1970)]{Gardner1970Life}
Martin Gardner.
\newblock Mathematical games: The fantastic combinations of john conway's new solitaire game "life".
\newblock \emph{Scientific American}, 223\penalty0 (4):\penalty0 120--123, 1970.

\bibitem[{Gemma Team}(2025)]{gemma3}
{Gemma Team}.
\newblock Gemma 3: Technical report.
\newblock \emph{arXiv preprint arXiv:2503.19786}, 2025.
\newblock URL \url{https://arxiv.org/abs/2503.19786}.

\bibitem[Geng et~al.(2024)Geng, Yang, Hang, Li, Gu, Zhang, Bao, Zhang, Li, Hu, et~al.]{geng2024instructdiffusion}
Zigang Geng, Binxin Yang, Tiankai Hang, Chen Li, Shuyang Gu, Ting Zhang, Jianmin Bao, Zheng Zhang, Houqiang Li, Han Hu, et~al.
\newblock Instructdiffusion: A generalist modeling interface for vision tasks.
\newblock In \emph{Proceedings of the IEEE/CVF Conference on computer vision and pattern recognition}, pp.\  12709--12720, 2024.

\bibitem[{Google DeepMind}(2025)]{deepmind2025veo3}
{Google DeepMind}.
\newblock Veo 3.
\newblock \url{https://deepmind.google/models/veo/}, September 2025.
\newblock URL \url{https://deepmind.google/models/veo/}.
\newblock Accessed: 2025-09-23.

\bibitem[HaCohen et~al.(2025)HaCohen, Chiprut, Brazowski, Shalem, Moshe, Richardson, Levin, Shiran, Zabari, Gordon, Panet, Weissbuch, Kulikov, Bitterman, Melumian, and Bibi]{hacohen2024ltx}
Yoav HaCohen, Nisan Chiprut, Benny Brazowski, Daniel Shalem, Dudu Moshe, Eitan Richardson, Eran Levin, Guy Shiran, Nir Zabari, Ori Gordon, Poriya Panet, Sapir Weissbuch, Victor Kulikov, Yaki Bitterman, Zeev Melumian, and Ofir Bibi.
\newblock Ltx-video: Realtime video latent diffusion.
\newblock \emph{CoRR}, abs/2501.00103, January 2025.
\newblock URL \url{https://doi.org/10.48550/arXiv.2501.00103}.

\bibitem[Hassan et~al.(2025)Hassan, Stapf, Rahimi, Rezende, Haghighi, Brüggemann, Katircioglu, Zhang, Chen, Saha, Cannici, Aljalbout, Ye, Wang, Davtyan, Salzmann, Scaramuzza, Pollefeys, Favaro, and Alahi]{hassan2024gemgeneralizableegovisionmultimodal}
Mariam Hassan, Sebastian Stapf, Ahmad Rahimi, Pedro M~B Rezende, Yasaman Haghighi, David Brüggemann, Isinsu Katircioglu, Lin Zhang, Xiaoran Chen, Suman Saha, Marco Cannici, Elie Aljalbout, Botao Ye, Xi~Wang, Aram Davtyan, Mathieu Salzmann, Davide Scaramuzza, Marc Pollefeys, Paolo Favaro, and Alexandre Alahi.
\newblock Gem: A generalizable ego-vision multimodal world model for fine-grained ego-motion, object dynamics, and scene composition control.
\newblock \emph{CVPR}, 2025.

\bibitem[Hong et~al.(2022)Hong, Ding, Zheng, Liu, and Tang]{hong2022cogvideolargescalepretrainingtexttovideo}
Wenyi Hong, Ming Ding, Wendi Zheng, Xinghan Liu, and Jie Tang.
\newblock Cogvideo: Large-scale pretraining for text-to-video generation via transformers.
\newblock \emph{CoRR}, abs/2205.15868, 2022.
\newblock URL \url{https://doi.org/10.48550/arXiv.2205.15868}.

\bibitem[Hu et~al.(2022)Hu, Shen, Wallis, Allen-Zhu, Li, Wang, Wang, Chen, et~al.]{hu2022lora}
Edward~J Hu, Yelong Shen, Phillip Wallis, Zeyuan Allen-Zhu, Yuanzhi Li, Shean Wang, Lu~Wang, Weizhu Chen, et~al.
\newblock Lora: Low-rank adaptation of large language models.
\newblock \emph{ICLR}, 1\penalty0 (2):\penalty0 3, 2022.

\bibitem[Huang et~al.(2024)Huang, Liu, Lin, Pang, Du, and Lin]{huang2024lorahub}
Chengsong Huang, Qian Liu, Bill~Yuchen Lin, Tianyu Pang, Chao Du, and Min Lin.
\newblock Lorahub: Efficient cross-task generalization via dynamic lora composition.
\newblock \emph{arXiv preprint arXiv:2307.13269}, 2024.
\newblock URL \url{https://arxiv.org/abs/2307.13269}.

\bibitem[Jing et~al.(2025)Jing, Chen, Aghazadeh, Wang, and Du]{jing2025a}
Liqiang Jing, Hardy Chen, Ehsan Aghazadeh, Xin~Eric Wang, and Xinya Du.
\newblock A comprehensive analysis for visual object hallucination in large vision-language models.
\newblock In \emph{Knowledgeable Foundation Models at ACL 2025}, 2025.
\newblock URL \url{https://openreview.net/forum?id=Ya4mqbhDP4}.

\bibitem[Kanervisto et~al.(2025)Kanervisto, Bignell, Wen, Grayson, Georgescu, Valcarcel~Macua, Tan, Rashid, Pearce, Cao, et~al.]{kanervisto2025world}
Anssi Kanervisto, Dave Bignell, Linda~Yilin Wen, Martin Grayson, Raluca Georgescu, Sergio Valcarcel~Macua, Shan~Zheng Tan, Tabish Rashid, Tim Pearce, Yuhan Cao, et~al.
\newblock World and human action models towards gameplay ideation.
\newblock \emph{Nature}, 638\penalty0 (8051):\penalty0 656--663, 2025.

\bibitem[Kuo et~al.(2023)Kuo, Hsueh, and Tsai]{kuo2023chess}
Mu-Tien Kuo, Chih-Chung Hsueh, and Richard Tzong-Han Tsai.
\newblock Large language models on the chessboard: A study on chatgpt’s formal language comprehension and complex reasoning skills.
\newblock 2023.
\newblock Preprint, arXiv.

\bibitem[Labs(2025)]{flux2024}
Black~Forest Labs.
\newblock Flux.1-dev.
\newblock \url{https://huggingface.co/black-forest-labs/FLUX.1-dev}, 2025.

\bibitem[Langton(1986)]{Langton1986Studying}
Christopher~G. Langton.
\newblock Studying artificial life with cellular automata.
\newblock \emph{Physica D: Nonlinear Phenomena}, 22\penalty0 (1-3):\penalty0 120--149, 1986.

\bibitem[Li et~al.(2025)Li, Hu, Larsen, Wu, Alford, Woo, Dunn, Tang, Zheng, Pu, and Ellis]{li2025combining}
Wen-Ding Li, Keya Hu, Carter Larsen, Yuqing Wu, Simon Alford, Caleb Woo, Spencer~M. Dunn, Hao Tang, Wei-Long Zheng, Yewen Pu, and Kevin Ellis.
\newblock Combining induction and transduction for abstract reasoning.
\newblock In \emph{The Thirteenth International Conference on Learning Representations}, 2025.
\newblock URL \url{https://openreview.net/forum?id=UmdotAAVDe}.

\bibitem[Liao et~al.(2025)Liao, Wang, Zhou, Hu, Zheng, and Gao]{liao2025dynamic}
Xiaoxuan Liao, Chihang Wang, Shicheng Zhou, Jiacheng Hu, Hongye Zheng, and Jia Gao.
\newblock Dynamic adaptation of lora fine-tuning for efficient and task-specific optimization of large language models.
\newblock In \emph{Proceedings of the 2025 International Conference on Artificial Intelligence and Computational Intelligence}, pp.\  120--125, 2025.

\bibitem[Lin et~al.(2014)Lin, Maire, Belongie, Bourdev, Girshick, Hays, Perona, Ramanan, Dollár, and Zitnick]{lin2014microsoft}
Tsung-Yi Lin, Michael Maire, Serge~J. Belongie, Lubomir~D. Bourdev, Ross~B. Girshick, James Hays, Pietro Perona, Deva Ramanan, Piotr Dollár, and C.~Lawrence Zitnick.
\newblock Microsoft coco: Common objects in context.
\newblock \emph{CoRR}, abs/1405.0312, 2014.
\newblock URL \url{http://arxiv.org/abs/1405.0312}.

\bibitem[Lin et~al.(2025)Lin, Huang, Zhuang, and Mao]{lin2025realgeneral}
Yijing Lin, Mengqi Huang, Shuhan Zhuang, and Zhendong Mao.
\newblock Realgeneral: Unifying visual generation via temporal in-context learning with video models.
\newblock \emph{arXiv preprint arXiv:2503.10406}, 2025.

\bibitem[Liu et~al.(2022)Liu, Tam, Muqeeth, Mohta, Huang, Bansal, and Raffel]{liu2022fewshotparameterefficientfinetuningbetter}
Haokun Liu, Derek Tam, Mohammed Muqeeth, Jay Mohta, Tenghao Huang, Mohit Bansal, and Colin Raffel.
\newblock Few-shot parameter-efficient fine-tuning is better and cheaper than in-context learning.
\newblock \emph{CoRR}, abs/2205.05638, 2022.
\newblock URL \url{https://doi.org/10.48550/arXiv.2205.05638}.

\bibitem[Meta-AI(2024)]{meta2024llama31}
Meta-AI.
\newblock Llama 3.1 models.
\newblock \url{https://ai.meta.com/blog/meta-llama-3-1} and \url{https://huggingface.co/meta-llama/Llama-3.1-8B}, 2024.

\bibitem[Moskvichev et~al.(2023)Moskvichev, Odouard, and Mitchell]{moskvichev2023conceptarc}
Arsenii Moskvichev, Victor~Vikram Odouard, and Melanie Mitchell.
\newblock The conceptarc benchmark: Evaluating understanding and generalization in the arc domain.
\newblock \emph{Trans. Mach. Learn. Res.}, 2023, 2023.
\newblock URL \url{https://openreview.net/forum?id=8ykyGbtt2q}.

\bibitem[Nathan~Silberman \& Fergus(2012)Nathan~Silberman and Fergus]{Silberman:ECCV12}
Pushmeet~Kohli Nathan~Silberman, Derek~Hoiem and Rob Fergus.
\newblock Indoor segmentation and support inference from rgbd images.
\newblock In \emph{ECCV}, 2012.

\bibitem[NVIDIA et~al.(2025)NVIDIA, :, Agarwal, Ali, Bala, Balaji, Barker, Cai, Chattopadhyay, Chen, Cui, Ding, Dworakowski, Fan, Fenzi, Ferroni, Fidler, Fox, Ge, Ge, Gu, Gururani, He, Huang, Huffman, Jannaty, Jin, Kim, Klár, Lam, Lan, Leal-Taixe, Li, Li, Lin, Lin, Ling, Liu, Liu, Luo, Ma, Mao, Mo, Mousavian, Nah, Niverty, Page, Paschalidou, Patel, Pavao, Ramezanali, Reda, Ren, Sabavat, Schmerling, Shi, Stefaniak, Tang, Tchapmi, Tredak, Tseng, Varghese, Wang, Wang, Wang, Wang, Wei, Wei, Wu, Xu, Yang, Yen-Chen, Zeng, Zeng, Zhang, Zhang, Zhang, Zhao, and Zolkowski]{agarwal2025cosmos}
NVIDIA, :, Niket Agarwal, Arslan Ali, Maciej Bala, Yogesh Balaji, Erik Barker, Tiffany Cai, Prithvijit Chattopadhyay, Yongxin Chen, Yin Cui, Yifan Ding, Daniel Dworakowski, Jiaojiao Fan, Michele Fenzi, Francesco Ferroni, Sanja Fidler, Dieter Fox, Songwei Ge, Yunhao Ge, Jinwei Gu, Siddharth Gururani, Ethan He, Jiahui Huang, Jacob Huffman, Pooya Jannaty, Jingyi Jin, Seung~Wook Kim, Gergely Klár, Grace Lam, Shiyi Lan, Laura Leal-Taixe, Anqi Li, Zhaoshuo Li, Chen-Hsuan Lin, Tsung-Yi Lin, Huan Ling, Ming-Yu Liu, Xian Liu, Alice Luo, Qianli Ma, Hanzi Mao, Kaichun Mo, Arsalan Mousavian, Seungjun Nah, Sriharsha Niverty, David Page, Despoina Paschalidou, Zeeshan Patel, Lindsey Pavao, Morteza Ramezanali, Fitsum Reda, Xiaowei Ren, Vasanth Rao~Naik Sabavat, Ed~Schmerling, Stella Shi, Bartosz Stefaniak, Shitao Tang, Lyne Tchapmi, Przemek Tredak, Wei-Cheng Tseng, Jibin Varghese, Hao Wang, Haoxiang Wang, Heng Wang, Ting-Chun Wang, Fangyin Wei, Xinyue Wei, Jay~Zhangjie Wu, Jiashu Xu, Wei Yang, Lin Yen-Chen, Xiaohui Zeng,
  Yu~Zeng, Jing Zhang, Qinsheng Zhang, Yuxuan Zhang, Qingqing Zhao, and Artur Zolkowski.
\newblock Cosmos world foundation model platform for physical ai, 2025.
\newblock URL \url{https://arxiv.org/abs/2501.03575}.

\bibitem[Polyak et~al.(2024)]{polyak2025moviegencastmedia}
Adam Polyak et~al.
\newblock Movie gen: A cast of media foundation models.
\newblock \emph{CoRR}, abs/2410.13720, 2024.
\newblock URL \url{https://doi.org/10.48550/arXiv.2410.13720}.

\bibitem[Prabhakar et~al.(2024)Prabhakar, Li, Narasimhan, Kakade, Malach, and Jelassi]{prabhakar2024lorasoups}
Akshara Prabhakar, Yuanzhi Li, Karthik Narasimhan, Sham~M. Kakade, Eran Malach, and Samy Jelassi.
\newblock Lora soups: Merging loras for practical skill composition tasks.
\newblock \emph{CoRR}, abs/2410.13025, 2024.
\newblock URL \url{https://doi.org/10.48550/arXiv.2410.13025}.

\bibitem[Qin et~al.(2024)Qin, Shi, Yu, Wang, Zhou, Li, Yin, Liu, Sheng, Shao, Bai, Ouyang, and Zhang]{brooks2024video}
Yiran Qin, Zhelun Shi, Jiwen Yu, Xijun Wang, Enshen Zhou, Lijun Li, Zhenfei Yin, Xihui Liu, Lu~Sheng, Jing Shao, Lei Bai, Wanli Ouyang, and Ruimao Zhang.
\newblock Worldsimbench: Towards video generation models as world simulators.
\newblock \emph{CoRR}, abs/2410.18072, 2024.
\newblock URL \url{https://doi.org/10.48550/arXiv.2410.18072}.

\bibitem[quantum24(2023)]{quantum24chess}
quantum24.
\newblock Chess puzzles 10k in pgn san.
\newblock \url{https://huggingface.co/datasets/quantum24/chess_puzzles_10k_in_pgn_san}, 2023.
\newblock Curated collection of checkmate-in-1, -2, and -3 puzzles derived from the Lichess community puzzle database. Licensed under CC0 1.0.

\bibitem[{Qwen3-4B-Instruct-2507 Team}(2025)]{qwen3technicalreport}
{Qwen3-4B-Instruct-2507 Team}.
\newblock Qwen3 technical report.
\newblock \emph{arXiv preprint arXiv:2505.09388}, 2025.

\bibitem[Roberts et~al.(2021)Roberts, Ramapuram, Ranjan, Kumar, Bautista, Paczan, Webb, and Susskind]{roberts:2021}
Mike Roberts, Jason Ramapuram, Anurag Ranjan, Atulit Kumar, Miguel~Angel Bautista, Nathan Paczan, Russ Webb, and Joshua~M. Susskind.
\newblock {Hypersim}: {A} photorealistic synthetic dataset for holistic indoor scene understanding.
\newblock In \emph{International Conference on Computer Vision (ICCV) 2021}, 2021.

\bibitem[Ruiz et~al.(2022)Ruiz, Li, Jampani, Pritch, Rubinstein, and Aberman]{ruiz2023dreambooth}
Nataniel Ruiz, Yuanzhen Li, Varun Jampani, Yael Pritch, Michael Rubinstein, and Kfir Aberman.
\newblock Dreambooth: Fine tuning text-to-image diffusion models for subject-driven generation.
\newblock \emph{CoRR}, abs/2208.12242, 2022.
\newblock URL \url{https://doi.org/10.48550/arXiv.2208.12242}.

\bibitem[Sim et~al.(2025)Sim, Zhang, Dai, and Fang]{sim2025can}
Mong~Yuan Sim, Wei~Emma Zhang, Xiang Dai, and Biaoyan Fang.
\newblock Can vlms actually see and read? a survey on modality collapse in vision-language models.
\newblock In \emph{Findings of the Association for Computational Linguistics: ACL 2025}, pp.\  24452--24470, 2025.

\bibitem[Villegas et~al.(2022)]{villegas2022phenakivariablelengthvideo}
Ruben Villegas et~al.
\newblock Phenaki: Variable length video generation from open domain textual description.
\newblock \emph{CoRR}, abs/2210.02399, 2022.
\newblock URL \url{https://doi.org/10.48550/arXiv.2210.02399}.

\bibitem[Wang et~al.(2025)Wang, Ai, Wen, Mao, Xie, Chen, Yu, Zhao, Yang, Zeng, Wang, Zhang, Zhou, Wang, Chen, Zhu, Zhao, Yan, Huang, Meng, Zhang, Li, Wu, Chu, Feng, Zhang, Sun, Fang, Wang, Gui, Weng, Shen, Lin, Wang, Wang, Zhou, Wang, Shen, Yu, Shi, Huang, Xu, Kou, Lv, Li, Liu, Wang, Zhang, Huang, Li, Wu, Liu, Pan, Zheng, Hong, Shi, Feng, Jiang, Han, Wu, and Liu]{wan2p1}
Ang Wang, Baole Ai, Bin Wen, Chaojie Mao, Chen-Wei Xie, Di~Chen, Feiwu Yu, Haiming Zhao, Jianxiao Yang, Jianyuan Zeng, Jiayu Wang, Jingfeng Zhang, Jingren Zhou, Jinkai Wang, Jixuan Chen, Kai Zhu, Kang Zhao, Keyu Yan, Lianghua Huang, Xiaofeng Meng, Ningyi Zhang, Pandeng Li, Pingyu Wu, Ruihang Chu, Ruili Feng, Shiwei Zhang, Siyang Sun, Tao Fang, Tianxing Wang, Tianyi Gui, Tingyu Weng, Tong Shen, Wei Lin, Wei Wang, Wei Wang, Wenmeng Zhou, Wente Wang, Wenting Shen, Wenyuan Yu, Xianzhong Shi, Xiaoming Huang, Xin Xu, Yan Kou, Yangyu Lv, Yifei Li, Yijing Liu, Yiming Wang, Yingya Zhang, Yitong Huang, Yong Li, You Wu, Yu~Liu, Yulin Pan, Yun Zheng, Yuntao Hong, Yupeng Shi, Yutong Feng, Zeyinzi Jiang, Zhen Han, Zhi-Fan Wu, and Ziyu Liu.
\newblock Wan: Open and advanced large-scale video generative models.
\newblock \emph{CoRR}, abs/2503.20314, March 2025.
\newblock URL \url{https://doi.org/10.48550/arXiv.2503.20314}.

\bibitem[Wang et~al.(2023{\natexlab{a}})Wang, Wang, Cao, Shen, and Huang]{wang2023images}
Xinlong Wang, Wen Wang, Yue Cao, Chunhua Shen, and Tiejun Huang.
\newblock Images speak in images: A generalist painter for in-context visual learning.
\newblock In \emph{Proceedings of the IEEE/CVF Conference on Computer Vision and Pattern Recognition}, pp.\  6830--6839, 2023{\natexlab{a}}.

\bibitem[Wang et~al.(2023{\natexlab{b}})Wang, Jiang, Lu, He, Chen, Wang, Zhou, et~al.]{wang2023context}
Zhendong Wang, Yifan Jiang, Yadong Lu, Pengcheng He, Weizhu Chen, Zhangyang Wang, Mingyuan Zhou, et~al.
\newblock In-context learning unlocked for diffusion models.
\newblock \emph{Advances in Neural Information Processing Systems}, 36:\penalty0 8542--8562, 2023{\natexlab{b}}.

\bibitem[Wolfram(1984)]{Wolfram1984Universality}
Stephen Wolfram.
\newblock Universality and complexity in cellular automata.
\newblock \emph{Physica D: Nonlinear Phenomena}, 10\penalty0 (1):\penalty0 1--35, 1984.
\newblock ISSN 0167-2789.
\newblock \doi{https://doi.org/10.1016/0167-2789(84)90245-8}.
\newblock URL \url{https://www.sciencedirect.com/science/article/pii/0167278984902458}.

\bibitem[Wu et~al.(2025)Wu, Li, Zhou, Lin, Gao, Yan, Yin, Bai, Xu, Chen, Chen, Tang, Zhang, Wang, Yang, Yu, Cheng, Liu, Li, Zhang, Meng, Wei, Ni, Chen, Cao, Peng, Qu, Wu, Wang, Yu, Wen, Feng, Xu, Wang, Zhang, Zhu, Wu, Cai, and Liu]{wu2025qwenimage}
Chenfei Wu, Jiahao Li, Jingren Zhou, Junyang Lin, Kaiyuan Gao, Kun Yan, Sheng-ming Yin, Shuai Bai, Xiao Xu, Yilei Chen, Yuxiang Chen, Zecheng Tang, Zekai Zhang, Zhengyi Wang, An~Yang, Bowen Yu, Chen Cheng, Dayiheng Liu, Deqing Li, Hang Zhang, Hao Meng, Hu~Wei, Jingyuan Ni, Kai Chen, Kuan Cao, Liang Peng, Lin Qu, Minggang Wu, Peng Wang, Shuting Yu, Tingkun Wen, Wensen Feng, Xiaoxiao Xu, Yi~Wang, Yichang Zhang, Yongqiang Zhu, Yujia Wu, Yuxuan Cai, and Zenan Liu.
\newblock Qwen-image technical report.
\newblock Technical report, Qwen Team, August 2025.
\newblock URL \url{https://arxiv.org/abs/2508.02324}.
\newblock Accessed: 2025-09-23.

\bibitem[Xu et~al.(2025)Xu, Ge, Liu, Fan, Xie, Zhao, Chen, and Shen]{xu2024matters}
Guangkai Xu, Yongtao Ge, Mingyu Liu, Chengxiang Fan, Kangyang Xie, Zhiyue Zhao, Hao Chen, and Chunhua Shen.
\newblock What matters when repurposing diffusion models for general dense perception tasks?
\newblock \emph{The Thirteenth International Conference on Learning Representations (ICLR)}, 2025.
\newblock URL \url{https://openreview.net/forum?id=BgYbk6ZmeX}.

\bibitem[Yang et~al.(2024)Yang, Wang, Li, Zhang, Chen, Wang, Liu, Li, Du, Zhou, et~al.]{yang2024cogvideox}
Zhuoyi Yang, Shuhong Wang, Jing Li, Haoran Zhang, Junpeng Chen, Zeyu Wang, Qian Liu, Jinzhe Li, Yifan Du, Kun Zhou, et~al.
\newblock Cogvideox: Text-to-video diffusion models with an expert transformer.
\newblock \emph{arXiv preprint arXiv:2408.06072}, 2024.

\bibitem[Zhao et~al.(2025)Zhao, Liu, Zheng, Zhu, Zhao, Chen, He, and Shen]{zhao2025diception}
Canyu Zhao, Mingyu Liu, Huanyi Zheng, Muzhi Zhu, Zhiyue Zhao, Hao Chen, Tong He, and Chunhua Shen.
\newblock Diception: A generalist diffusion model for visual perceptual tasks.
\newblock \emph{arXiv preprint arXiv:2502.17157}, 2025.
\newblock URL \url{https://arxiv.org/abs/2502.17157}.

\bibitem[Zhou et~al.(2017)Zhou, Zhao, Puig, Fidler, Barriuso, and Torralba]{zhou2017scene}
Bolei Zhou, Hang Zhao, Xavier Puig, Sanja Fidler, Adela Barriuso, and Antonio Torralba.
\newblock Scene parsing through ade20k dataset.
\newblock In \emph{Proceedings of the IEEE Conference on Computer Vision and Pattern Recognition}, 2017.

\bibitem[Zhou et~al.(2019)Zhou, Zhao, Puig, Xiao, Fidler, Barriuso, and Torralba]{zhou2019semantic}
Bolei Zhou, Hang Zhao, Xavier Puig, Tete Xiao, Sanja Fidler, Adela Barriuso, and Antonio Torralba.
\newblock Semantic understanding of scenes through the ade20k dataset.
\newblock \emph{International Journal of Computer Vision}, 127\penalty0 (3):\penalty0 302--321, 2019.

\end{thebibliography}

\appendix

%%%
% Appendix
%%%
\appendix
\section*{Appendix}
\section*{Limitations and future work}
Our study focuses on grid-based benchmarks such as ARC-AGI, ConceptARC, and synthetic puzzles. This controlled setup provides a systematic framework for comparing models under equivalent conditions, offering a clear interface through which LLMs can demonstrate visual understanding. While these benchmarks do not capture the full diversity of real-world challenges, they remain valuable for isolating and analyzing the role of modality-aligned pretraining in visual intelligence. 

The iterative nature of diffusion sampling also adds significant computational overhead. While some tasks can perform well with only a few sampling steps \cite{xu2024matters}, complex domains such as ARC-AGI often require longer sampling schedules to maintain structural consistency and coherence.

Fine-tuning VDMs remains computationally demanding even when using parameter-efficient methods such as LoRAs. Future work could explore modular and composable LoRA strategies \cite{huang2024lorahub, prabhakar2024lorasoups}, potentially reducing retraining costs while enhancing cross-task generalization. Another promising direction is to extend these models toward in-context task adaptation. Just as LLMs have evolved from next-token prediction in their pretraining phase to in-context question answering, VDMs could benefit from fine-tuning approaches that enable flexible, context-dependent adaptation.

Beyond improving adaptability, understanding the mechanisms that give rise to visual intelligence in these models is an equally important research direction. Inspired by ongoing advances in mechanistic interpretability for LLMs, future work could aim to uncover how VDMs internally represent and manipulate concepts.
\section{Experimental Details} 

We report here the detailed computational costs and hyperparameter settings used in our experiments. 
Tables~\ref{tab:conceptarc-gpuhours} and \ref{tab:arcagi-structured-gpuhours} summarize the GPU hours required across different tasks, 
while Tables~\ref{appdx:tab:hyperparameters} and \ref{tab:model-configs} provide the LoRA fine-tuning configurations for both VDMs and LLMs.

\begin{table}[ht]
\centering
\caption{GPU hours required for ConceptARC across VDMs and LLMs. 
Reported hours are wall-clock time and depend on hardware.}
\label{tab:conceptarc-gpuhours}
\resizebox{0.9\linewidth}{!}{%
    \begin{tabular}{l r l r}
    \toprule
    \textbf{VDM Model (GPU)} & \textbf{Hours} & \textbf{LLM Model (GPU)} & \textbf{Hours} \\
    \midrule
    Wan2.1-14B (H100) & 100 & Llama3.1-8B (H100) & 80 \\
    LTX-13B (H100) & 95 & Qwen3-8B (2$\times$RTX4090) & 100 \\
    CogVideoX1.5-5B (RTX4090) & 130 & Qwen3-4B-Instruct-2507 (RTX4090) & 135 \\
    LTX-2B (H100) & 40 &  &  \\
    \bottomrule
    \end{tabular}
}
\end{table}

\begin{table}[ht]
\centering
\caption{GPU hours required for ARC-AGI and Structured Visual Tasks. 
Reported hours are wall-clock time and depend on hardware.}
\label{tab:arcagi-structured-gpuhours}
\resizebox{0.9\linewidth}{!}{%
    \begin{tabular}{l r l r}
    \toprule
    \textbf{ARCAGI Model (GPU)} & \textbf{Hours} & \textbf{Structured Task Model (GPU)} & \textbf{Hours} \\
    \midrule
    CogVideoX1.5-5B (RTX4090) & 450 & CogVideoX1.5-5B (RTX4090) & 1650 \\
    Qwen3-4B-Instruct-2507 (RTX4090) & 475 & Qwen3-4B-Instruct-2507 (RTX4090) & 2000 \\
    \bottomrule
    \end{tabular}
}
\end{table}

To ensure reproducibility, we also include the fine-tuning hyperparameters for each model. 
The following two tables detail the LoRA, training, and optimizer configurations used for VDMs (Table~\ref{appdx:tab:hyperparameters}) 
and LLMs (Table~\ref{tab:model-configs}).

\begin{table}[h]
    \centering
    \caption{LoRA finetuning configuration for VDM experiments.}
    \label{appdx:tab:hyperparameters}
    \resizebox{0.95\linewidth}{!}{%
    \begin{tabular}{l p{2.6cm} p{2.6cm} p{2.6cm} p{2.6cm}}
    \toprule
    \textbf{Parameter} & \textbf{LTX-13B} & \textbf{LTX-2B} & \textbf{CogVideoX1.5-5B} & \textbf{Wan2.1-14B} \\
    \midrule
    \multicolumn{5}{l}{\textit{LoRA Configuration}} \\
    Rank & 64 & 64 & 64 & 64 \\
    Alpha & 64 & 64 & 32 & 32 \\
    Target modules & to\_q, to\_k, to\_v, to\_out.0, ff.net.0.proj, ff.net.2 & to\_q, to\_k, to\_v, to\_out.0, ff.net.0.proj, ff.net.2 & QKVO & – \\
    \midrule
    \multicolumn{5}{l}{\textit{Training Configuration}} \\
    Seed & 42 & 42 & 42 & 42 \\
    Batch size & 2 & 4 & 2 & 1 \\
    Gradient accumulation steps & 2 & 1 & 1 & 1 \\
    \midrule
    \multicolumn{5}{l}{\textit{Optimizer Configuration}} \\
    Optimizer & AdamW & AdamW & AdamW & AdamW \\
    Learning rate & 2e-4 & 2e-4 & 1e-4 & 1e-4 \\
    Scheduler & Linear & Linear & Constant & Constant \\
    Max grad norm & 1.0 & 1.0 & 1.0 & 0.05 \\
    \bottomrule
    \end{tabular}%
    }
\end{table}

\begin{table}[h]
\centering
\caption{LoRA finetuning configuration for LLMs used.}
\label{tab:model-configs}
\resizebox{0.95\linewidth}{!}{%
\begin{tabular}{l p{4.2cm} p{4.2cm} p{4.2cm}}
\toprule
\textbf{Parameter} & \textbf{Qwen3-4B-Instruct-2507} & \textbf{Qwen3-8B} & \textbf{LLaMA-3.1-8B} \\
\midrule
\multicolumn{4}{l}{\textit{LoRA Configuration}} \\
Rank & 32 & 32 & 32 \\
Alpha & 32 & 32 & 64 \\
Dropout & 0 & 0 & 0.05 \\
Target modules & q\_proj, k\_proj, v\_proj, o\_proj, gate\_proj, up\_proj, down\_proj & q\_proj, k\_proj, v\_proj, o\_proj, gate\_proj, up\_proj, down\_proj & q\_proj, k\_proj, v\_proj, o\_proj, gate\_proj, up\_proj, down\_proj, lm\_head \\
\midrule
\multicolumn{4}{l}{\textit{Model Setup}} \\
Max sequence length & 8192 & 8192 & 4096 \\
Random seed & 3407 & 3407 & 3407 \\
\midrule
\multicolumn{4}{l}{\textit{Training Configuration}} \\
Batch size per device & 2 & 1 & 1 \\
Effective batch size & 8 & 8 & 8 \\
Gradient accumulation steps & 4 & 8 & 8 \\
Learning rate & 2e-4 & 2e-4 & 2e-4 \\
Scheduler & Linear & Linear & Linear \\
Warmup steps & 5 & 5 & 5 \\
Weight decay & 0.01 & 0.01 & 0.01 \\
\midrule
\multicolumn{4}{l}{\textit{Generation Configuration}} \\
Max new tokens & 4096 & 4096 & 4096 \\
Temperature & 0.7 & 0.7 & 0.7 \\
Top-$p$ & 0.8 & 0.8 & 0.8 \\
Top-$k$ & 20 & 20 & 20 \\
\bottomrule
\end{tabular}%
}
\end{table}

\noindent\textit{Note.} 
LoRA ranks differ slightly across model families (VDMs use rank 64, whereas LLMs use rank 32). We verified that performance is largely insensitive to this setting: Qwen3 models with rank 64 perform comparably to rank 32, and CogVideoX1.5-5B models with rank 32 match the reported rank 64 results. In both cases, we report the configuration that yielded stronger results in our initial trials. All reported results in the paper correspond to the configurations shown in the tables.

\section{Task Details}

For completeness, we provide additional explanations of the tasks considered in our evaluation. Each subsection introduces a task family and highlights the key rules and objectives, we further provide examples on how the task is encoded into image and text.

\subsection{Visual Games}

% We include a diverse set of visual games to complement our evaluation. These tasks test the models' ability to operate under structured constraints and goal-directed objectives, ranging from puzzle solving to board play. Puzzle-based tasks such as \textit{Hitori} and \textit{Sudoku} emphasize spatial consistency and constraint satisfaction on grid-based layouts. In contrast, competitive board games such as \textit{Connect 4} and \textit{Chess Mate-in-1} highlight the need to identify decisive actions in discrete states. Together, these games provide a complementary perspective on how models handle structured, rule-based environments.

\subsubsection{Hitori 5x5}
\textbf{Objective:} Eliminate cells so that each number appears at most once per row and column.  

\textbf{Rules:}
\begin{enumerate}[leftmargin=*]
    \item A number must not be repeated in any row or column.
    \item Shaded cells cannot be orthogonally adjacent.
    \item All unshaded cells must form a single connected component.
\end{enumerate}

We add an example of the task in Figure \ref{tab:task-example-hitori}.

%%%%%%%%%%%%%%%
% HITORI 5 EASY
%%%%%%%%%%%%%%%

\begin{table}[ht]
\centering
\renewcommand{\arraystretch}{3}
\begin{tabular}{c >{\centering\arraybackslash}m{4cm} >{\centering\arraybackslash}m{4cm}}
& \textbf{Input} & \textbf{Output} \\[1ex]
\rotatebox{90}{\textbf{\small Image Representation}} & 
\vspace{-2cm} % ADJUST TO ALIGN
\includegraphics[width=3cm]{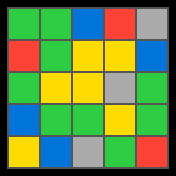} & 
\vspace{-2cm} % ADJUST TO ALIGN
\includegraphics[width=3cm]
{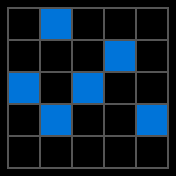} \\[2ex]
\rotatebox{90}{\textbf{\small Text Representation}} & 
\begin{minipage}[c]{\linewidth}
\vspace{-2.5cm} % ADJUST TO ALIGN
\noindent
\texttt{[} \\
\phantom{X}\texttt{[\cnumarc{3}, \cnumarc{3}, \cnumarc{1}, \cnumarc{2}, \cnumarc{5}],} \\
\phantom{X}\texttt{[\cnumarc{2}, \cnumarc{3}, \cnumarc{4}, \cnumarc{4}, \cnumarc{1}],} \\
\phantom{X}\texttt{[\cnumarc{3}, \cnumarc{4}, \cnumarc{4}, \cnumarc{5}, \cnumarc{3}],} \\
\phantom{X}\texttt{[\cnumarc{1}, \cnumarc{3}, \cnumarc{3}, \cnumarc{4}, \cnumarc{3}],} \\
\phantom{X}\texttt{[\cnumarc{4}, \cnumarc{1}, \cnumarc{5}, \cnumarc{3}, \cnumarc{2}],} \\
\texttt{]}
\end{minipage} & 
\begin{minipage}[c]{\linewidth}
\vspace{-2.5cm} % ADJUST TO ALIGN
\noindent
\texttt{[} \\
\phantom{X}\texttt{[\cnumarc{0}, \cnumarc{1}, \cnumarc{0}, \cnumarc{0}, \cnumarc{0}],} \\
\phantom{X}\texttt{[\cnumarc{0}, \cnumarc{0}, \cnumarc{0}, \cnumarc{1}, \cnumarc{0}],} \\
\phantom{X}\texttt{[\cnumarc{1}, \cnumarc{0}, \cnumarc{1}, \cnumarc{0}, \cnumarc{0}],} \\
\phantom{X}\texttt{[\cnumarc{0}, \cnumarc{1}, \cnumarc{0}, \cnumarc{0}, \cnumarc{1}],} \\
\phantom{X}\texttt{[\cnumarc{0}, \cnumarc{0}, \cnumarc{0}, \cnumarc{0}, \cnumarc{0}],} \\
\texttt{]}
\end{minipage} \\
\end{tabular}
\captionof{figure}{Example input-output pair for task \textit{Hitori}.}

\label{tab:task-example-hitori}
\end{table}

\subsubsection{Sudoku}
\textbf{Objective:} Fill the grid so that all constraints are satisfied. 

\textbf{Rules:}
\begin{enumerate}[leftmargin=*]
    \item Each row must contain all required digits without repetition.
    \item Each column must contain all required digits without repetition.
    \item Each subgrid must contain all required digits without repetition.
\end{enumerate}
We evaluate two variants: \textit{Mini Sudoku} (4x4 with 2x2 subgrids, see Figure \ref{tab:task-example-sudoku-mini}) and \textit{Sudoku} (9x9 with 3x3 subgrids, see Figure \ref{tab:task-example-sudoku}).

%%%%%%%%%%%%%%%
% SUDOKU MINI EASY
%%%%%%%%%%%%%%%

\begin{table}[ht]
\centering
\renewcommand{\arraystretch}{3}
\begin{tabular}{c >{\centering\arraybackslash}m{4cm} >{\centering\arraybackslash}m{4cm}}
& \textbf{Input} & \textbf{Output} \\[1ex]
\rotatebox{90}{\textbf{\small Image Representation}} & 
\vspace{-2cm} % ADJUST TO ALIGN
\includegraphics[width=3cm]{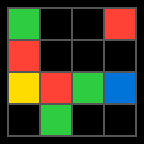} & 
\vspace{-2cm} % ADJUST TO ALIGN
\includegraphics[width=3cm]
{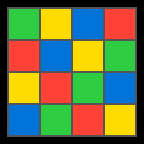} \\[2ex]
\rotatebox{90}{\textbf{\small Text Representation}} & 
\begin{minipage}[c]{\linewidth}
\vspace{-2.5cm} % ADJUST TO ALIGN

\noindent
\texttt{[} \\
\phantom{X}\texttt{[\cnumarc{3}, \cnumarc{0}, \cnumarc{0}, \cnumarc{2}],} \\
\phantom{X}\texttt{[\cnumarc{2}, \cnumarc{0}, \cnumarc{0}, \cnumarc{0}],} \\
\phantom{X}\texttt{[\cnumarc{4}, \cnumarc{2}, \cnumarc{3}, \cnumarc{1}],} \\
\phantom{X}\texttt{[\cnumarc{0}, \cnumarc{3}, \cnumarc{0}, \cnumarc{0}],} \\
\texttt{]}
\end{minipage} & 
\begin{minipage}[c]{\linewidth}
\vspace{-2.5cm} % ADJUST TO ALIGN

\noindent
\texttt{[} \\
\phantom{X}\texttt{[\cnumarc{3}, \cnumarc{4}, \cnumarc{1}, \cnumarc{2}],} \\
\phantom{X}\texttt{[\cnumarc{2}, \cnumarc{1}, \cnumarc{4}, \cnumarc{3}],} \\
\phantom{X}\texttt{[\cnumarc{4}, \cnumarc{2}, \cnumarc{3}, \cnumarc{1}],} \\
\phantom{X}\texttt{[\cnumarc{1}, \cnumarc{3}, \cnumarc{2}, \cnumarc{4}],} \\
\texttt{]}
\end{minipage} \\
\end{tabular}
\captionof{figure}{Example input-output pair for task \textit{Sudoku Mini}.}
\label{tab:task-example-sudoku-mini}
\end{table}

%%%%%%%%%%%%%%%
% SUDOKU STANDARD EASY
%%%%%%%%%%%%%%%

\begin{table}[ht]
\centering
\renewcommand{\arraystretch}{3}
\begin{tabular}{c >{\centering\arraybackslash}m{4cm} >{\centering\arraybackslash}m{4cm}}
& \textbf{Input} & \textbf{Output} \\[1ex]
\rotatebox{90}{\textbf{\small Image Representation}} & 
\vspace{-2cm} % ADJUST TO ALIGN
\includegraphics[width=3cm]{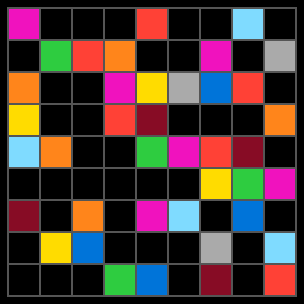} & 
\vspace{-2cm} % ADJUST TO ALIGN
\includegraphics[width=3cm]
{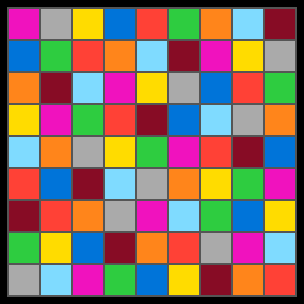} \\[2ex]
\rotatebox{90}{\textbf{\small Text Representation}} & 
\begin{minipage}[c]{\linewidth}
\vspace{-2.5cm} % ADJUST TO ALIGN
\tiny
\noindent
\texttt{[} \\
\phantom{X}\texttt{[\cnumarc{6}, \cnumarc{0}, \cnumarc{0}, \cnumarc{0}, \cnumarc{2}, \cnumarc{0}, \cnumarc{0}, \cnumarc{8}, \cnumarc{0}],} \\
\phantom{X}\texttt{[\cnumarc{0}, \cnumarc{3}, \cnumarc{2}, \cnumarc{7}, \cnumarc{0}, \cnumarc{0}, \cnumarc{6}, \cnumarc{0}, \cnumarc{5}],} \\
\phantom{X}\texttt{[\cnumarc{7}, \cnumarc{0}, \cnumarc{0}, \cnumarc{6}, \cnumarc{4}, \cnumarc{5}, \cnumarc{1}, \cnumarc{2}, \cnumarc{0}],} \\
\phantom{X}\texttt{[\cnumarc{4}, \cnumarc{0}, \cnumarc{0}, \cnumarc{2}, \cnumarc{9}, \cnumarc{0}, \cnumarc{0}, \cnumarc{0}, \cnumarc{7}],} \\
\phantom{X}\texttt{[\cnumarc{8}, \cnumarc{7}, \cnumarc{0}, \cnumarc{0}, \cnumarc{3}, \cnumarc{6}, \cnumarc{2}, \cnumarc{9}, \cnumarc{0}],} \\
\phantom{X}\texttt{[\cnumarc{0}, \cnumarc{0}, \cnumarc{0}, \cnumarc{0}, \cnumarc{0}, \cnumarc{0}, \cnumarc{4}, \cnumarc{3}, \cnumarc{6}],} \\
\phantom{X}\texttt{[\cnumarc{9}, \cnumarc{0}, \cnumarc{7}, \cnumarc{0}, \cnumarc{6}, \cnumarc{8}, \cnumarc{0}, \cnumarc{1}, \cnumarc{0}],} \\
\phantom{X}\texttt{[\cnumarc{0}, \cnumarc{4}, \cnumarc{1}, \cnumarc{0}, \cnumarc{0}, \cnumarc{0}, \cnumarc{5}, \cnumarc{0}, \cnumarc{8}],} \\
\phantom{X}\texttt{[\cnumarc{0}, \cnumarc{0}, \cnumarc{0}, \cnumarc{3}, \cnumarc{1}, \cnumarc{0}, \cnumarc{9}, \cnumarc{0}, \cnumarc{2}],} \\
\texttt{]}
\end{minipage} & 
\begin{minipage}[c]{\linewidth}
\vspace{-2.5cm} % ADJUST TO ALIGN
\tiny
\noindent
\texttt{[} \\
\phantom{X}\texttt{[\cnumarc{6}, \cnumarc{5}, \cnumarc{4}, \cnumarc{1}, \cnumarc{2}, \cnumarc{3}, \cnumarc{7}, \cnumarc{8}, \cnumarc{9}],} \\
\phantom{X}\texttt{[\cnumarc{1}, \cnumarc{3}, \cnumarc{2}, \cnumarc{7}, \cnumarc{8}, \cnumarc{9}, \cnumarc{6}, \cnumarc{4}, \cnumarc{5}],} \\
\phantom{X}\texttt{[\cnumarc{7}, \cnumarc{9}, \cnumarc{8}, \cnumarc{6}, \cnumarc{4}, \cnumarc{5}, \cnumarc{1}, \cnumarc{2}, \cnumarc{3}],} \\
\phantom{X}\texttt{[\cnumarc{4}, \cnumarc{6}, \cnumarc{3}, \cnumarc{2}, \cnumarc{9}, \cnumarc{1}, \cnumarc{8}, \cnumarc{5}, \cnumarc{7}],} \\
\phantom{X}\texttt{[\cnumarc{8}, \cnumarc{7}, \cnumarc{5}, \cnumarc{4}, \cnumarc{3}, \cnumarc{6}, \cnumarc{2}, \cnumarc{9}, \cnumarc{1}],} \\
\phantom{X}\texttt{[\cnumarc{2}, \cnumarc{1}, \cnumarc{9}, \cnumarc{8}, \cnumarc{5}, \cnumarc{7}, \cnumarc{4}, \cnumarc{3}, \cnumarc{6}],} \\
\phantom{X}\texttt{[\cnumarc{9}, \cnumarc{2}, \cnumarc{7}, \cnumarc{5}, \cnumarc{6}, \cnumarc{8}, \cnumarc{3}, \cnumarc{1}, \cnumarc{4}],} \\
\phantom{X}\texttt{[\cnumarc{3}, \cnumarc{4}, \cnumarc{1}, \cnumarc{9}, \cnumarc{7}, \cnumarc{2}, \cnumarc{5}, \cnumarc{6}, \cnumarc{8}],} \\
\phantom{X}\texttt{[\cnumarc{5}, \cnumarc{8}, \cnumarc{6}, \cnumarc{3}, \cnumarc{1}, \cnumarc{4}, \cnumarc{9}, \cnumarc{7}, \cnumarc{2}],} \\
\texttt{]}
\end{minipage} \\
\end{tabular}
\captionof{figure}{Example input-output pair for task \textit{Sudoku}.}
\label{tab:task-example-sudoku}
\end{table}

\subsubsection{Connect 4}
\textbf{Objective:} Place tokens to align four in a row. 

\textbf{Rules:}
\begin{enumerate}[leftmargin=*]
    \item Players alternate dropping tokens into one of the seven columns.
    \item A token occupies the lowest available cell in the chosen column.
    \item A player wins by forming a horizontal, vertical, or diagonal line of four tokens.
\end{enumerate}
We restrict evaluation to single-move winning scenarios, see Figure \ref{tab:task-example-connect4}.

%%%%%%%%%%%%%%%
% CONNECT 4
%%%%%%%%%%%%%%%

\begin{table}[ht]
\centering
\renewcommand{\arraystretch}{3}
\begin{tabular}{c >{\centering\arraybackslash}m{4cm} >{\centering\arraybackslash}m{4cm}}
& \textbf{Input} & \textbf{Output} \\[1ex]
\rotatebox{90}{\textbf{\small Image Representation}} & 
\vspace{-2cm} % ADJUST TO ALIGN
\includegraphics[width=3cm]{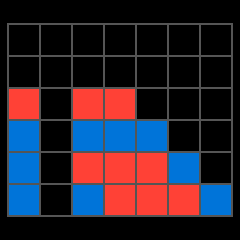} & 
\vspace{-2cm} % ADJUST TO ALIGN
\includegraphics[width=3cm]
{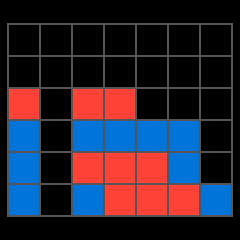} \\[2ex]
\rotatebox{90}{\textbf{\small Text Representation}} & 
\begin{minipage}[c]{\linewidth}
\vspace{-2.5cm} % ADJUST TO ALIGN
\scriptsize
\noindent
\texttt{[} \\
\phantom{X}\texttt{[\cnumarc{0}, \cnumarc{0}, \cnumarc{0}, \cnumarc{0}, \cnumarc{0}, \cnumarc{0}, \cnumarc{0}],} \\
\phantom{X}\texttt{[\cnumarc{0}, \cnumarc{0}, \cnumarc{0}, \cnumarc{0}, \cnumarc{0}, \cnumarc{0}, \cnumarc{0}],} \\
\phantom{X}\texttt{[\cnumarc{2}, \cnumarc{0}, \cnumarc{2}, \cnumarc{2}, \cnumarc{0}, \cnumarc{0}, \cnumarc{0}],} \\
\phantom{X}\texttt{[\cnumarc{1}, \cnumarc{0}, \cnumarc{1}, \cnumarc{1}, \cnumarc{1}, \cnumarc{0}, \cnumarc{0}],} \\
\phantom{X}\texttt{[\cnumarc{1}, \cnumarc{0}, \cnumarc{2}, \cnumarc{2}, \cnumarc{2}, \cnumarc{1}, \cnumarc{0}],} \\
\phantom{X}\texttt{[\cnumarc{1}, \cnumarc{0}, \cnumarc{1}, \cnumarc{2}, \cnumarc{2}, \cnumarc{2}, \cnumarc{1}],} \\
\texttt{]}

\end{minipage} & 
\begin{minipage}[c]{\linewidth}
\vspace{-2.5cm} % ADJUST TO ALIGN
\scriptsize
\noindent
\texttt{[} \\
\phantom{X}\texttt{[\cnumarc{0}, \cnumarc{0}, \cnumarc{0}, \cnumarc{0}, \cnumarc{0}, \cnumarc{0}, \cnumarc{0}],} \\
\phantom{X}\texttt{[\cnumarc{0}, \cnumarc{0}, \cnumarc{0}, \cnumarc{0}, \cnumarc{0}, \cnumarc{0}, \cnumarc{0}],} \\
\phantom{X}\texttt{[\cnumarc{2}, \cnumarc{0}, \cnumarc{2}, \cnumarc{2}, \cnumarc{0}, \cnumarc{0}, \cnumarc{0}],} \\
\phantom{X}\texttt{[\cnumarc{1}, \cnumarc{0}, \cnumarc{1}, \cnumarc{1}, \cnumarc{1}, \cnumarc{1}, \cnumarc{0}],} \\
\phantom{X}\texttt{[\cnumarc{1}, \cnumarc{0}, \cnumarc{2}, \cnumarc{2}, \cnumarc{2}, \cnumarc{1}, \cnumarc{0}],} \\
\phantom{X}\texttt{[\cnumarc{1}, \cnumarc{0}, \cnumarc{1}, \cnumarc{2}, \cnumarc{2}, \cnumarc{2}, \cnumarc{1}],} \\
\texttt{]}
\end{minipage} \\
\end{tabular}
\captionof{figure}{Example input-output pair for task \textit{Connect4}.}
\label{tab:task-example-connect4}
\end{table}

\subsubsection{Chess Mate-in-1}
\textbf{Objective:} Deliver checkmate in a single move.  
\textbf{Rules:}
\begin{enumerate}[leftmargin=*]
    \item All standard chess movement rules apply.
    \item A move is correct only if it results in an immediate checkmate of the opposing king.
\end{enumerate}
To ensure the task is well defined, we filter scenarios so that they always correspond to white moves. The original dataset is extracted from \cite{quantum24chess}, and an illustrative example is shown in Figure~\ref{tab:task-example-chess-mate-in-1-w}.

%%%%%%%%%%%%%%%
% CHESS MATE IN 1
%%%%%%%%%%%%%%%

\begin{table}[ht]
\centering
\renewcommand{\arraystretch}{3}
\begin{tabular}{c >{\centering\arraybackslash}m{4cm} >{\centering\arraybackslash}m{4cm}}
& \textbf{Input} & \textbf{Output} \\[1ex]
\raisebox{-1.0em}{\rotatebox{90}{\textbf{\small Chess state}}} & 
\vspace{-0.1cm} % ADJUST TO ALIGN
\includegraphics[width=3cm]{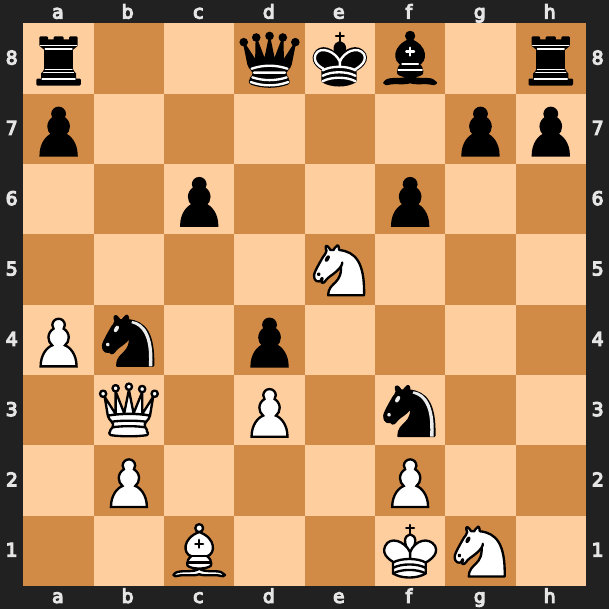} & 
%\includesvg[width=3cm]{figures/appendix/examples_tasks/human_repr_chess_mate_in_1_w_0} & 
\vspace{-0.1cm} % ADJUST TO ALIGN
\includegraphics[width=3cm]{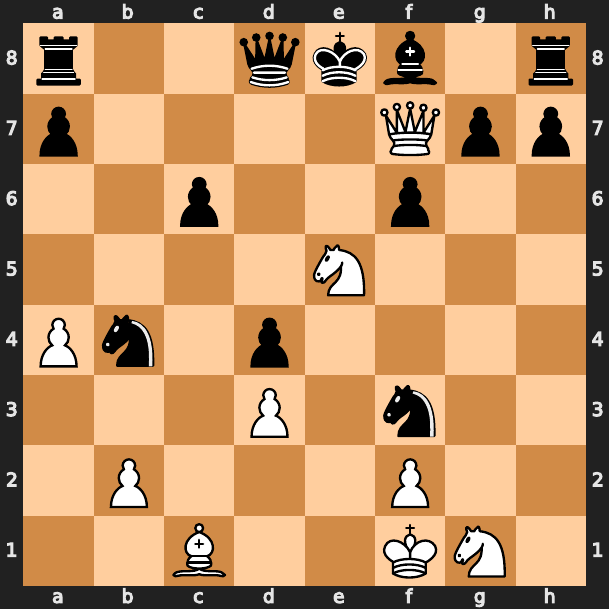} \\[2ex]
% \includesvg[width=3cm]{figures/appendix/examples_tasks/human_repr_chess_mate_in_1_w_1} \\[2ex]
\rotatebox{90}{\textbf{\small Image Representation}} & 
\vspace{-2cm} % ADJUST TO ALIGN
\includegraphics[width=3cm]{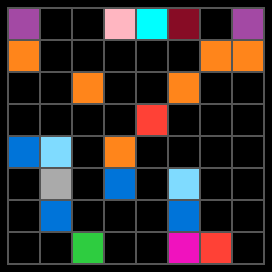} & 
\vspace{-2cm} % ADJUST TO ALIGN
\includegraphics[width=3cm]{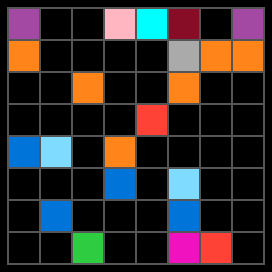} \\[2ex]
\rotatebox{90}{\textbf{\small Text Representation}} & 
\begin{minipage}[c]{\linewidth}
\vspace{-2.5cm} % ADJUST TO ALIGN
\tiny
\noindent
\texttt{[} \vspace{-1.0pt}\\
\phantom{X}\texttt{[\cnumarc{10}, \cnumarc{0}, \cnumarc{0}, \cnumarc{11}, \cnumarc{12}, \cnumarc{9}, \cnumarc{0}, \cnumarc{10}],} \vspace{-1.0pt} \\
\phantom{X}\texttt{[\cnumarc{7}, \cnumarc{0}, \cnumarc{0}, \cnumarc{0}, \cnumarc{0}, \cnumarc{0}, \cnumarc{7}, \cnumarc{7}],} \vspace{-1.0pt} \\
\phantom{X}\texttt{[\cnumarc{0}, \cnumarc{0}, \cnumarc{7}, \cnumarc{0}, \cnumarc{0}, \cnumarc{7}, \cnumarc{0}, \cnumarc{0}],} \vspace{-1.0pt} \\
\phantom{X}\texttt{[\cnumarc{0}, \cnumarc{0}, \cnumarc{0}, \cnumarc{0}, \cnumarc{2}, \cnumarc{0}, \cnumarc{0}, \cnumarc{0}],} \vspace{-1.0pt} \\
\phantom{X}\texttt{[\cnumarc{1}, \cnumarc{8}, \cnumarc{0}, \cnumarc{7}, \cnumarc{0}, \cnumarc{0}, \cnumarc{0}, \cnumarc{0}],} \vspace{-1.0pt} \\
\phantom{X}\texttt{[\cnumarc{0}, \cnumarc{5}, \cnumarc{0}, \cnumarc{1}, \cnumarc{0}, \cnumarc{8}, \cnumarc{0}, \cnumarc{0}],} \vspace{-1.0pt} \\
\phantom{X}\texttt{[\cnumarc{0}, \cnumarc{1}, \cnumarc{0}, \cnumarc{0}, \cnumarc{0}, \cnumarc{1}, \cnumarc{0}, \cnumarc{0}],} \vspace{-1.0pt} \\
\phantom{X}\texttt{[\cnumarc{0}, \cnumarc{0}, \cnumarc{3}, \cnumarc{0}, \cnumarc{0}, \cnumarc{6}, \cnumarc{2}, \cnumarc{0}],} \vspace{-1.0pt} \\
\texttt{]}
\end{minipage} & 
\begin{minipage}[c]{\linewidth}
\vspace{-2.5cm} % ADJUST TO ALIGN
\tiny
\noindent
\texttt{[} \vspace{-1.0pt}\\
\phantom{X}\texttt{[\cnumarc{10}, \cnumarc{0}, \cnumarc{0}, \cnumarc{11}, \cnumarc{12}, \cnumarc{9}, \cnumarc{0}, \cnumarc{10}],} \vspace{-1.0pt} \\
\phantom{X}\texttt{[\cnumarc{7}, \cnumarc{0}, \cnumarc{0}, \cnumarc{0}, \cnumarc{0}, \cnumarc{5}, \cnumarc{7}, \cnumarc{7}],} \vspace{-1.0pt} \\
\phantom{X}\texttt{[\cnumarc{0}, \cnumarc{0}, \cnumarc{7}, \cnumarc{0}, \cnumarc{0}, \cnumarc{7}, \cnumarc{0}, \cnumarc{0}],} \vspace{-1.0pt} \\
\phantom{X}\texttt{[\cnumarc{0}, \cnumarc{0}, \cnumarc{0}, \cnumarc{0}, \cnumarc{2}, \cnumarc{0}, \cnumarc{0}, \cnumarc{0}],} \vspace{-1.0pt} \\
\phantom{X}\texttt{[\cnumarc{1}, \cnumarc{8}, \cnumarc{0}, \cnumarc{7}, \cnumarc{0}, \cnumarc{0}, \cnumarc{0}, \cnumarc{0}],} \vspace{-1.0pt} \\
\phantom{X}\texttt{[\cnumarc{0}, \cnumarc{0}, \cnumarc{0}, \cnumarc{1}, \cnumarc{0}, \cnumarc{8}, \cnumarc{0}, \cnumarc{0}],} \vspace{-1.0pt} \\
\phantom{X}\texttt{[\cnumarc{0}, \cnumarc{1}, \cnumarc{0}, \cnumarc{0}, \cnumarc{0}, \cnumarc{1}, \cnumarc{0}, \cnumarc{0}],} \vspace{-1.0pt} \\
\phantom{X}\texttt{[\cnumarc{0}, \cnumarc{0}, \cnumarc{3}, \cnumarc{0}, \cnumarc{0}, \cnumarc{6}, \cnumarc{2}, \cnumarc{0}],} \vspace{-1.0pt} \\
\texttt{]}
\end{minipage} \\
\end{tabular}
\captionof{figure}{Example input-output pair for task \textit{Chess Mate in 1}.}
\label{tab:task-example-chess-mate-in-1-w}
\end{table}

\subsection{Route Planning}\label{sec:appdx-route-planning}

We evaluate route planning in two-dimensional grid environments. The objective across tasks is to construct valid paths that connect designated start and goal locations under different structural constraints. We consider two tasks: \textit{Maze} and \textit{Shortest Path}.

\subsubsection{Maze}
\textbf{Objective:} Navigate from the start cell to the goal cell through a grid containing blocked and open positions.  

\textbf{Rules:}
\begin{enumerate}[leftmargin=*]
    \item The agent starts at the top-left cell and must reach the bottom-right cell.  
    \item Movement is allowed only through open cells.  
    \item Allowed moves are up, down, left, and right (no diagonal moves).  
    \item A valid solution is a continuous sequence of moves from start to goal.  
\end{enumerate}

%%%%%%%%%%%%%%%
% MAZE MINI
%%%%%%%%%%%%%%%

\begin{table}[ht]
\centering
\renewcommand{\arraystretch}{3}
\begin{tabular}{c >{\centering\arraybackslash}m{4cm} >{\centering\arraybackslash}m{4cm}}
& \textbf{Input} & \textbf{Output} \\[1ex]
\rotatebox{90}{\textbf{\small Image Representation}} & 
\vspace{-2cm} % ADJUST TO ALIGN
\includegraphics[width=3cm]{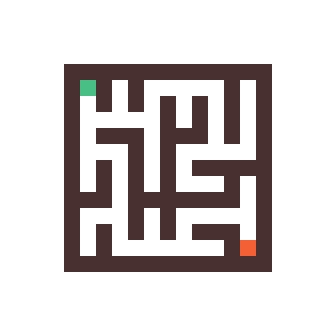} & 
\vspace{-2cm} % ADJUST TO ALIGN
\includegraphics[width=3cm]
{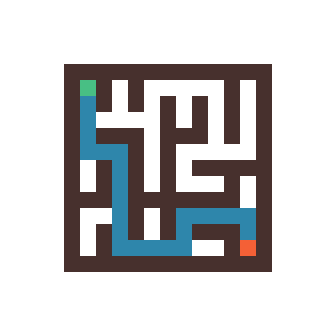} \\[2ex]
\rotatebox{90}{\textbf{\small Text Representation}} & 
\begin{minipage}[c]{\linewidth}
\vspace{-2.5cm} % ADJUST TO ALIGN
\fontsize{2.7}{2}
\noindent
\texttt{[} \vspace{-7.0pt}\\
\phantom{X}\texttt{[\cnummazes{1}, \cnummazes{1}, \cnummazes{1}, \cnummazes{1}, \cnummazes{1}, \cnummazes{1}, \cnummazes{1}, \cnummazes{1}, \cnummazes{1}, \cnummazes{1}, \cnummazes{1}, \cnummazes{1}, \cnummazes{1}, \cnummazes{1}, \cnummazes{1}, \cnummazes{1}, \cnummazes{1}, \cnummazes{1}, \cnummazes{1}, \cnummazes{1}, \cnummazes{1}],} \vspace{-7.0pt} \\
\phantom{X}\texttt{[\cnummazes{1}, \cnummazes{1}, \cnummazes{1}, \cnummazes{1}, \cnummazes{1}, \cnummazes{1}, \cnummazes{1}, \cnummazes{1}, \cnummazes{1}, \cnummazes{1}, \cnummazes{1}, \cnummazes{1}, \cnummazes{1}, \cnummazes{1}, \cnummazes{1}, \cnummazes{1}, \cnummazes{1}, \cnummazes{1}, \cnummazes{1}, \cnummazes{1}, \cnummazes{1}],} \vspace{-7.0pt} \\
\phantom{X}\texttt{[\cnummazes{1}, \cnummazes{1}, \cnummazes{1}, \cnummazes{1}, \cnummazes{1}, \cnummazes{1}, \cnummazes{1}, \cnummazes{1}, \cnummazes{1}, \cnummazes{1}, \cnummazes{1}, \cnummazes{1}, \cnummazes{1}, \cnummazes{1}, \cnummazes{1}, \cnummazes{1}, \cnummazes{1}, \cnummazes{1}, \cnummazes{1}, \cnummazes{1}, \cnummazes{1}],} \vspace{-7.0pt} \\
\phantom{X}\texttt{[\cnummazes{1}, \cnummazes{1}, \cnummazes{1}, \cnummazes{1}, \cnummazes{1}, \cnummazes{1}, \cnummazes{1}, \cnummazes{1}, \cnummazes{1}, \cnummazes{1}, \cnummazes{1}, \cnummazes{1}, \cnummazes{1}, \cnummazes{1}, \cnummazes{1}, \cnummazes{1}, \cnummazes{1}, \cnummazes{1}, \cnummazes{1}, \cnummazes{1}, \cnummazes{1}],} \vspace{-7.0pt} \\
\phantom{X}\texttt{[\cnummazes{1}, \cnummazes{1}, \cnummazes{1}, \cnummazes{1}, \cnummazes{0}, \cnummazes{0}, \cnummazes{0}, \cnummazes{0}, \cnummazes{0}, \cnummazes{0}, \cnummazes{0}, \cnummazes{0}, \cnummazes{0}, \cnummazes{0}, \cnummazes{0}, \cnummazes{0}, \cnummazes{0}, \cnummazes{1}, \cnummazes{1}, \cnummazes{1}, \cnummazes{1}],} \vspace{-7.0pt} \\
\phantom{X}\texttt{[\cnummazes{1}, \cnummazes{1}, \cnummazes{1}, \cnummazes{1}, \cnummazes{0}, \cnummazes{3}, \cnummazes{0}, \cnummazes{1}, \cnummazes{0}, \cnummazes{1}, \cnummazes{1}, \cnummazes{1}, \cnummazes{1}, \cnummazes{1}, \cnummazes{0}, \cnummazes{1}, \cnummazes{0}, \cnummazes{1}, \cnummazes{1}, \cnummazes{1}, \cnummazes{1}],} \vspace{-7.0pt} \\
\phantom{X}\texttt{[\cnummazes{1}, \cnummazes{1}, \cnummazes{1}, \cnummazes{1}, \cnummazes{0}, \cnummazes{1}, \cnummazes{0}, \cnummazes{1}, \cnummazes{0}, \cnummazes{1}, \cnummazes{0}, \cnummazes{1}, \cnummazes{0}, \cnummazes{1}, \cnummazes{0}, \cnummazes{1}, \cnummazes{0}, \cnummazes{1}, \cnummazes{1}, \cnummazes{1}, \cnummazes{1}],} \vspace{-7.0pt} \\
\phantom{X}\texttt{[\cnummazes{1}, \cnummazes{1}, \cnummazes{1}, \cnummazes{1}, \cnummazes{0}, \cnummazes{1}, \cnummazes{1}, \cnummazes{1}, \cnummazes{1}, \cnummazes{1}, \cnummazes{0}, \cnummazes{1}, \cnummazes{0}, \cnummazes{1}, \cnummazes{0}, \cnummazes{1}, \cnummazes{0}, \cnummazes{1}, \cnummazes{1}, \cnummazes{1}, \cnummazes{1}],} \vspace{-7.0pt} \\
\phantom{X}\texttt{[\cnummazes{1}, \cnummazes{1}, \cnummazes{1}, \cnummazes{1}, \cnummazes{0}, \cnummazes{1}, \cnummazes{0}, \cnummazes{0}, \cnummazes{0}, \cnummazes{1}, \cnummazes{0}, \cnummazes{0}, \cnummazes{0}, \cnummazes{1}, \cnummazes{0}, \cnummazes{1}, \cnummazes{0}, \cnummazes{1}, \cnummazes{1}, \cnummazes{1}, \cnummazes{1}],} \vspace{-7.0pt} \\
\phantom{X}\texttt{[\cnummazes{1}, \cnummazes{1}, \cnummazes{1}, \cnummazes{1}, \cnummazes{0}, \cnummazes{1}, \cnummazes{1}, \cnummazes{1}, \cnummazes{0}, \cnummazes{1}, \cnummazes{0}, \cnummazes{1}, \cnummazes{1}, \cnummazes{1}, \cnummazes{1}, \cnummazes{1}, \cnummazes{0}, \cnummazes{1}, \cnummazes{1}, \cnummazes{1}, \cnummazes{1}],} \vspace{-7.0pt} \\
\phantom{X}\texttt{[\cnummazes{1}, \cnummazes{1}, \cnummazes{1}, \cnummazes{1}, \cnummazes{0}, \cnummazes{1}, \cnummazes{0}, \cnummazes{1}, \cnummazes{0}, \cnummazes{1}, \cnummazes{0}, \cnummazes{1}, \cnummazes{0}, \cnummazes{0}, \cnummazes{0}, \cnummazes{0}, \cnummazes{0}, \cnummazes{1}, \cnummazes{1}, \cnummazes{1}, \cnummazes{1}],} \vspace{-7.0pt} \\
\phantom{X}\texttt{[\cnummazes{1}, \cnummazes{1}, \cnummazes{1}, \cnummazes{1}, \cnummazes{0}, \cnummazes{1}, \cnummazes{0}, \cnummazes{1}, \cnummazes{0}, \cnummazes{1}, \cnummazes{0}, \cnummazes{1}, \cnummazes{1}, \cnummazes{1}, \cnummazes{0}, \cnummazes{1}, \cnummazes{0}, \cnummazes{1}, \cnummazes{1}, \cnummazes{1}, \cnummazes{1}],} \vspace{-7.0pt} \\
\phantom{X}\texttt{[\cnummazes{1}, \cnummazes{1}, \cnummazes{1}, \cnummazes{1}, \cnummazes{0}, \cnummazes{0}, \cnummazes{0}, \cnummazes{1}, \cnummazes{0}, \cnummazes{0}, \cnummazes{0}, \cnummazes{0}, \cnummazes{0}, \cnummazes{0}, \cnummazes{0}, \cnummazes{1}, \cnummazes{0}, \cnummazes{1}, \cnummazes{1}, \cnummazes{1}, \cnummazes{1}],} \vspace{-7.0pt} \\
\phantom{X}\texttt{[\cnummazes{1}, \cnummazes{1}, \cnummazes{1}, \cnummazes{1}, \cnummazes{0}, \cnummazes{1}, \cnummazes{1}, \cnummazes{1}, \cnummazes{0}, \cnummazes{1}, \cnummazes{0}, \cnummazes{1}, \cnummazes{1}, \cnummazes{1}, \cnummazes{1}, \cnummazes{1}, \cnummazes{0}, \cnummazes{1}, \cnummazes{1}, \cnummazes{1}, \cnummazes{1}],} \vspace{-7.0pt} \\
\phantom{X}\texttt{[\cnummazes{1}, \cnummazes{1}, \cnummazes{1}, \cnummazes{1}, \cnummazes{0}, \cnummazes{1}, \cnummazes{0}, \cnummazes{1}, \cnummazes{0}, \cnummazes{1}, \cnummazes{0}, \cnummazes{1}, \cnummazes{0}, \cnummazes{0}, \cnummazes{0}, \cnummazes{1}, \cnummazes{0}, \cnummazes{1}, \cnummazes{1}, \cnummazes{1}, \cnummazes{1}],} \vspace{-7.0pt} \\
\phantom{X}\texttt{[\cnummazes{1}, \cnummazes{1}, \cnummazes{1}, \cnummazes{1}, \cnummazes{0}, \cnummazes{1}, \cnummazes{0}, \cnummazes{1}, \cnummazes{1}, \cnummazes{1}, \cnummazes{1}, \cnummazes{1}, \cnummazes{1}, \cnummazes{1}, \cnummazes{0}, \cnummazes{2}, \cnummazes{0}, \cnummazes{1}, \cnummazes{1}, \cnummazes{1}, \cnummazes{1}],} \vspace{-7.0pt} \\
\phantom{X}\texttt{[\cnummazes{1}, \cnummazes{1}, \cnummazes{1}, \cnummazes{1}, \cnummazes{0}, \cnummazes{0}, \cnummazes{0}, \cnummazes{0}, \cnummazes{0}, \cnummazes{0}, \cnummazes{0}, \cnummazes{0}, \cnummazes{0}, \cnummazes{0}, \cnummazes{0}, \cnummazes{0}, \cnummazes{0}, \cnummazes{1}, \cnummazes{1}, \cnummazes{1}, \cnummazes{1}],} \vspace{-7.0pt} \\
\phantom{X}\texttt{[\cnummazes{1}, \cnummazes{1}, \cnummazes{1}, \cnummazes{1}, \cnummazes{1}, \cnummazes{1}, \cnummazes{1}, \cnummazes{1}, \cnummazes{1}, \cnummazes{1}, \cnummazes{1}, \cnummazes{1}, \cnummazes{1}, \cnummazes{1}, \cnummazes{1}, \cnummazes{1}, \cnummazes{1}, \cnummazes{1}, \cnummazes{1}, \cnummazes{1}, \cnummazes{1}],} \vspace{-7.0pt} \\
\phantom{X}\texttt{[\cnummazes{1}, \cnummazes{1}, \cnummazes{1}, \cnummazes{1}, \cnummazes{1}, \cnummazes{1}, \cnummazes{1}, \cnummazes{1}, \cnummazes{1}, \cnummazes{1}, \cnummazes{1}, \cnummazes{1}, \cnummazes{1}, \cnummazes{1}, \cnummazes{1}, \cnummazes{1}, \cnummazes{1}, \cnummazes{1}, \cnummazes{1}, \cnummazes{1}, \cnummazes{1}],} \vspace{-7.0pt} \\
\phantom{X}\texttt{[\cnummazes{1}, \cnummazes{1}, \cnummazes{1}, \cnummazes{1}, \cnummazes{1}, \cnummazes{1}, \cnummazes{1}, \cnummazes{1}, \cnummazes{1}, \cnummazes{1}, \cnummazes{1}, \cnummazes{1}, \cnummazes{1}, \cnummazes{1}, \cnummazes{1}, \cnummazes{1}, \cnummazes{1}, \cnummazes{1}, \cnummazes{1}, \cnummazes{1}, \cnummazes{1}],} \vspace{-7.0pt} \\
\phantom{X}\texttt{[\cnummazes{1}, \cnummazes{1}, \cnummazes{1}, \cnummazes{1}, \cnummazes{1}, \cnummazes{1}, \cnummazes{1}, \cnummazes{1}, \cnummazes{1}, \cnummazes{1}, \cnummazes{1}, \cnummazes{1}, \cnummazes{1}, \cnummazes{1}, \cnummazes{1}, \cnummazes{1}, \cnummazes{1}, \cnummazes{1}, \cnummazes{1}, \cnummazes{1}, \cnummazes{1}],} \vspace{-7.0pt} \\
\texttt{]}
\end{minipage} & 
\begin{minipage}[c]{\linewidth}
\vspace{-2.5cm} % ADJUST TO ALIGN
\fontsize{2.7}{2}
\noindent
\texttt{[} \vspace{-7.0pt}\\
\phantom{X}\texttt{[\cnummazes{1}, \cnummazes{1}, \cnummazes{1}, \cnummazes{1}, \cnummazes{1}, \cnummazes{1}, \cnummazes{1}, \cnummazes{1}, \cnummazes{1}, \cnummazes{1}, \cnummazes{1}, \cnummazes{1}, \cnummazes{1}, \cnummazes{1}, \cnummazes{1}, \cnummazes{1}, \cnummazes{1}, \cnummazes{1}, \cnummazes{1}, \cnummazes{1}, \cnummazes{1}],} \vspace{-7.0pt} \\
\phantom{X}\texttt{[\cnummazes{1}, \cnummazes{1}, \cnummazes{1}, \cnummazes{1}, \cnummazes{1}, \cnummazes{1}, \cnummazes{1}, \cnummazes{1}, \cnummazes{1}, \cnummazes{1}, \cnummazes{1}, \cnummazes{1}, \cnummazes{1}, \cnummazes{1}, \cnummazes{1}, \cnummazes{1}, \cnummazes{1}, \cnummazes{1}, \cnummazes{1}, \cnummazes{1}, \cnummazes{1}],} \vspace{-7.0pt} \\
\phantom{X}\texttt{[\cnummazes{1}, \cnummazes{1}, \cnummazes{1}, \cnummazes{1}, \cnummazes{1}, \cnummazes{1}, \cnummazes{1}, \cnummazes{1}, \cnummazes{1}, \cnummazes{1}, \cnummazes{1}, \cnummazes{1}, \cnummazes{1}, \cnummazes{1}, \cnummazes{1}, \cnummazes{1}, \cnummazes{1}, \cnummazes{1}, \cnummazes{1}, \cnummazes{1}, \cnummazes{1}],} \vspace{-7.0pt} \\
\phantom{X}\texttt{[\cnummazes{1}, \cnummazes{1}, \cnummazes{1}, \cnummazes{1}, \cnummazes{1}, \cnummazes{1}, \cnummazes{1}, \cnummazes{1}, \cnummazes{1}, \cnummazes{1}, \cnummazes{1}, \cnummazes{1}, \cnummazes{1}, \cnummazes{1}, \cnummazes{1}, \cnummazes{1}, \cnummazes{1}, \cnummazes{1}, \cnummazes{1}, \cnummazes{1}, \cnummazes{1}],} \vspace{-7.0pt} \\
\phantom{X}\texttt{[\cnummazes{1}, \cnummazes{1}, \cnummazes{1}, \cnummazes{1}, \cnummazes{0}, \cnummazes{0}, \cnummazes{0}, \cnummazes{0}, \cnummazes{0}, \cnummazes{0}, \cnummazes{0}, \cnummazes{0}, \cnummazes{0}, \cnummazes{0}, \cnummazes{0}, \cnummazes{0}, \cnummazes{0}, \cnummazes{1}, \cnummazes{1}, \cnummazes{1}, \cnummazes{1}],} \vspace{-7.0pt} \\
\phantom{X}\texttt{[\cnummazes{1}, \cnummazes{1}, \cnummazes{1}, \cnummazes{1}, \cnummazes{0}, \cnummazes{3}, \cnummazes{0}, \cnummazes{1}, \cnummazes{0}, \cnummazes{1}, \cnummazes{1}, \cnummazes{1}, \cnummazes{1}, \cnummazes{1}, \cnummazes{0}, \cnummazes{1}, \cnummazes{0}, \cnummazes{1}, \cnummazes{1}, \cnummazes{1}, \cnummazes{1}],} \vspace{-7.0pt} \\
\phantom{X}\texttt{[\cnummazes{1}, \cnummazes{1}, \cnummazes{1}, \cnummazes{1}, \cnummazes{0}, \cnummazes{4}, \cnummazes{0}, \cnummazes{1}, \cnummazes{0}, \cnummazes{1}, \cnummazes{0}, \cnummazes{1}, \cnummazes{0}, \cnummazes{1}, \cnummazes{0}, \cnummazes{1}, \cnummazes{0}, \cnummazes{1}, \cnummazes{1}, \cnummazes{1}, \cnummazes{1}],} \vspace{-7.0pt} \\
\phantom{X}\texttt{[\cnummazes{1}, \cnummazes{1}, \cnummazes{1}, \cnummazes{1}, \cnummazes{0}, \cnummazes{4}, \cnummazes{1}, \cnummazes{1}, \cnummazes{1}, \cnummazes{1}, \cnummazes{0}, \cnummazes{1}, \cnummazes{0}, \cnummazes{1}, \cnummazes{0}, \cnummazes{1}, \cnummazes{0}, \cnummazes{1}, \cnummazes{1}, \cnummazes{1}, \cnummazes{1}],} \vspace{-7.0pt} \\
\phantom{X}\texttt{[\cnummazes{1}, \cnummazes{1}, \cnummazes{1}, \cnummazes{1}, \cnummazes{0}, \cnummazes{4}, \cnummazes{0}, \cnummazes{0}, \cnummazes{0}, \cnummazes{1}, \cnummazes{0}, \cnummazes{0}, \cnummazes{0}, \cnummazes{1}, \cnummazes{0}, \cnummazes{1}, \cnummazes{0}, \cnummazes{1}, \cnummazes{1}, \cnummazes{1}, \cnummazes{1}],} \vspace{-7.0pt} \\
\phantom{X}\texttt{[\cnummazes{1}, \cnummazes{1}, \cnummazes{1}, \cnummazes{1}, \cnummazes{0}, \cnummazes{4}, \cnummazes{4}, \cnummazes{4}, \cnummazes{0}, \cnummazes{1}, \cnummazes{0}, \cnummazes{1}, \cnummazes{1}, \cnummazes{1}, \cnummazes{1}, \cnummazes{1}, \cnummazes{0}, \cnummazes{1}, \cnummazes{1}, \cnummazes{1}, \cnummazes{1}],} \vspace{-7.0pt} \\
\phantom{X}\texttt{[\cnummazes{1}, \cnummazes{1}, \cnummazes{1}, \cnummazes{1}, \cnummazes{0}, \cnummazes{1}, \cnummazes{0}, \cnummazes{4}, \cnummazes{0}, \cnummazes{1}, \cnummazes{0}, \cnummazes{1}, \cnummazes{0}, \cnummazes{0}, \cnummazes{0}, \cnummazes{0}, \cnummazes{0}, \cnummazes{1}, \cnummazes{1}, \cnummazes{1}, \cnummazes{1}],} \vspace{-7.0pt} \\
\phantom{X}\texttt{[\cnummazes{1}, \cnummazes{1}, \cnummazes{1}, \cnummazes{1}, \cnummazes{0}, \cnummazes{1}, \cnummazes{0}, \cnummazes{4}, \cnummazes{0}, \cnummazes{1}, \cnummazes{0}, \cnummazes{1}, \cnummazes{1}, \cnummazes{1}, \cnummazes{0}, \cnummazes{1}, \cnummazes{0}, \cnummazes{1}, \cnummazes{1}, \cnummazes{1}, \cnummazes{1}],} \vspace{-7.0pt} \\
\phantom{X}\texttt{[\cnummazes{1}, \cnummazes{1}, \cnummazes{1}, \cnummazes{1}, \cnummazes{0}, \cnummazes{0}, \cnummazes{0}, \cnummazes{4}, \cnummazes{0}, \cnummazes{0}, \cnummazes{0}, \cnummazes{0}, \cnummazes{0}, \cnummazes{0}, \cnummazes{0}, \cnummazes{1}, \cnummazes{0}, \cnummazes{1}, \cnummazes{1}, \cnummazes{1}, \cnummazes{1}],} \vspace{-7.0pt} \\
\phantom{X}\texttt{[\cnummazes{1}, \cnummazes{1}, \cnummazes{1}, \cnummazes{1}, \cnummazes{0}, \cnummazes{1}, \cnummazes{1}, \cnummazes{4}, \cnummazes{0}, \cnummazes{1}, \cnummazes{0}, \cnummazes{4}, \cnummazes{4}, \cnummazes{4}, \cnummazes{4}, \cnummazes{4}, \cnummazes{0}, \cnummazes{1}, \cnummazes{1}, \cnummazes{1}, \cnummazes{1}],} \vspace{-7.0pt} \\
\phantom{X}\texttt{[\cnummazes{1}, \cnummazes{1}, \cnummazes{1}, \cnummazes{1}, \cnummazes{0}, \cnummazes{1}, \cnummazes{0}, \cnummazes{4}, \cnummazes{0}, \cnummazes{1}, \cnummazes{0}, \cnummazes{4}, \cnummazes{0}, \cnummazes{0}, \cnummazes{0}, \cnummazes{4}, \cnummazes{0}, \cnummazes{1}, \cnummazes{1}, \cnummazes{1}, \cnummazes{1}],} \vspace{-7.0pt} \\
\phantom{X}\texttt{[\cnummazes{1}, \cnummazes{1}, \cnummazes{1}, \cnummazes{1}, \cnummazes{0}, \cnummazes{1}, \cnummazes{0}, \cnummazes{4}, \cnummazes{4}, \cnummazes{4}, \cnummazes{4}, \cnummazes{4}, \cnummazes{1}, \cnummazes{1}, \cnummazes{0}, \cnummazes{2}, \cnummazes{0}, \cnummazes{1}, \cnummazes{1}, \cnummazes{1}, \cnummazes{1}],} \vspace{-7.0pt} \\
\phantom{X}\texttt{[\cnummazes{1}, \cnummazes{1}, \cnummazes{1}, \cnummazes{1}, \cnummazes{0}, \cnummazes{0}, \cnummazes{0}, \cnummazes{0}, \cnummazes{0}, \cnummazes{0}, \cnummazes{0}, \cnummazes{0}, \cnummazes{0}, \cnummazes{0}, \cnummazes{0}, \cnummazes{0}, \cnummazes{0}, \cnummazes{1}, \cnummazes{1}, \cnummazes{1}, \cnummazes{1}],} \vspace{-7.0pt} \\
\phantom{X}\texttt{[\cnummazes{1}, \cnummazes{1}, \cnummazes{1}, \cnummazes{1}, \cnummazes{1}, \cnummazes{1}, \cnummazes{1}, \cnummazes{1}, \cnummazes{1}, \cnummazes{1}, \cnummazes{1}, \cnummazes{1}, \cnummazes{1}, \cnummazes{1}, \cnummazes{1}, \cnummazes{1}, \cnummazes{1}, \cnummazes{1}, \cnummazes{1}, \cnummazes{1}, \cnummazes{1}],} \vspace{-7.0pt} \\
\phantom{X}\texttt{[\cnummazes{1}, \cnummazes{1}, \cnummazes{1}, \cnummazes{1}, \cnummazes{1}, \cnummazes{1}, \cnummazes{1}, \cnummazes{1}, \cnummazes{1}, \cnummazes{1}, \cnummazes{1}, \cnummazes{1}, \cnummazes{1}, \cnummazes{1}, \cnummazes{1}, \cnummazes{1}, \cnummazes{1}, \cnummazes{1}, \cnummazes{1}, \cnummazes{1}, \cnummazes{1}],} \vspace{-7.0pt} \\
\phantom{X}\texttt{[\cnummazes{1}, \cnummazes{1}, \cnummazes{1}, \cnummazes{1}, \cnummazes{1}, \cnummazes{1}, \cnummazes{1}, \cnummazes{1}, \cnummazes{1}, \cnummazes{1}, \cnummazes{1}, \cnummazes{1}, \cnummazes{1}, \cnummazes{1}, \cnummazes{1}, \cnummazes{1}, \cnummazes{1}, \cnummazes{1}, \cnummazes{1}, \cnummazes{1}, \cnummazes{1}],} \vspace{-7.0pt} \\
\phantom{X}\texttt{[\cnummazes{1}, \cnummazes{1}, \cnummazes{1}, \cnummazes{1}, \cnummazes{1}, \cnummazes{1}, \cnummazes{1}, \cnummazes{1}, \cnummazes{1}, \cnummazes{1}, \cnummazes{1}, \cnummazes{1}, \cnummazes{1}, \cnummazes{1}, \cnummazes{1}, \cnummazes{1}, \cnummazes{1}, \cnummazes{1}, \cnummazes{1}, \cnummazes{1}, \cnummazes{1}],} \vspace{-7.0pt} \\
\texttt{]}
\end{minipage} \\
\end{tabular}
\captionof{figure}{Example input-output pair for task \textit{Maze Small}.}
\label{tab:task-example-maze-small}
\end{table}

%%%%%%%%%%%%%%%
% MAZE
%%%%%%%%%%%%%%%

\begin{table}[ht]
\centering
\renewcommand{\arraystretch}{3}
\begin{tabular}{c >{\centering\arraybackslash}m{4cm} >{\centering\arraybackslash}m{4cm}}
& \textbf{Input} & \textbf{Output} \\[1ex]
\rotatebox{90}{\textbf{\small Image Representation}} & 
\vspace{-2cm} % ADJUST TO ALIGN
\includegraphics[width=3cm]{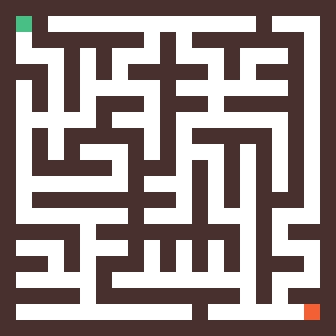} & 
\vspace{-2cm} % ADJUST TO ALIGN
\includegraphics[width=3cm]
{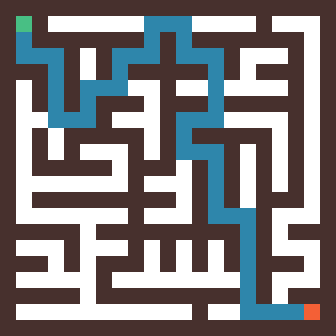} \\[2ex]
\rotatebox{90}{\textbf{\small Text Representation}} & 
\begin{minipage}[c]{\linewidth}
\vspace{-2.5cm} % ADJUST TO ALIGN
\fontsize{2.7}{2}
\noindent
\texttt{[} \vspace{-7.0pt}\\
\phantom{X}\texttt{[\cnummazes{0}, \cnummazes{0}, \cnummazes{0}, \cnummazes{0}, \cnummazes{0}, \cnummazes{0}, \cnummazes{0}, \cnummazes{0}, \cnummazes{0}, \cnummazes{0}, \cnummazes{0}, \cnummazes{0}, \cnummazes{0}, \cnummazes{0}, \cnummazes{0}, \cnummazes{0}, \cnummazes{0}, \cnummazes{0}, \cnummazes{0}, \cnummazes{0}, \cnummazes{0}],} \vspace{-7.0pt} \\
\phantom{X}\texttt{[\cnummazes{0}, \cnummazes{3}, \cnummazes{0}, \cnummazes{1}, \cnummazes{1}, \cnummazes{1}, \cnummazes{1}, \cnummazes{1}, \cnummazes{1}, \cnummazes{1}, \cnummazes{1}, \cnummazes{1}, \cnummazes{1}, \cnummazes{1}, \cnummazes{1}, \cnummazes{1}, \cnummazes{0}, \cnummazes{1}, \cnummazes{1}, \cnummazes{1}, \cnummazes{0}],} \vspace{-7.0pt} \\
\phantom{X}\texttt{[\cnummazes{0}, \cnummazes{1}, \cnummazes{0}, \cnummazes{0}, \cnummazes{0}, \cnummazes{0}, \cnummazes{0}, \cnummazes{0}, \cnummazes{0}, \cnummazes{1}, \cnummazes{0}, \cnummazes{1}, \cnummazes{0}, \cnummazes{0}, \cnummazes{0}, \cnummazes{0}, \cnummazes{0}, \cnummazes{0}, \cnummazes{0}, \cnummazes{1}, \cnummazes{0}],} \vspace{-7.0pt} \\
\phantom{X}\texttt{[\cnummazes{0}, \cnummazes{1}, \cnummazes{1}, \cnummazes{1}, \cnummazes{0}, \cnummazes{1}, \cnummazes{0}, \cnummazes{1}, \cnummazes{1}, \cnummazes{1}, \cnummazes{0}, \cnummazes{1}, \cnummazes{1}, \cnummazes{1}, \cnummazes{0}, \cnummazes{1}, \cnummazes{1}, \cnummazes{1}, \cnummazes{0}, \cnummazes{1}, \cnummazes{0}],} \vspace{-7.0pt} \\
\phantom{X}\texttt{[\cnummazes{0}, \cnummazes{0}, \cnummazes{0}, \cnummazes{1}, \cnummazes{0}, \cnummazes{1}, \cnummazes{0}, \cnummazes{1}, \cnummazes{0}, \cnummazes{0}, \cnummazes{0}, \cnummazes{0}, \cnummazes{0}, \cnummazes{1}, \cnummazes{0}, \cnummazes{1}, \cnummazes{0}, \cnummazes{0}, \cnummazes{0}, \cnummazes{1}, \cnummazes{0}],} \vspace{-7.0pt} \\
\phantom{X}\texttt{[\cnummazes{0}, \cnummazes{1}, \cnummazes{0}, \cnummazes{1}, \cnummazes{0}, \cnummazes{1}, \cnummazes{1}, \cnummazes{1}, \cnummazes{1}, \cnummazes{1}, \cnummazes{0}, \cnummazes{1}, \cnummazes{1}, \cnummazes{1}, \cnummazes{1}, \cnummazes{1}, \cnummazes{1}, \cnummazes{1}, \cnummazes{0}, \cnummazes{1}, \cnummazes{0}],} \vspace{-7.0pt} \\
\phantom{X}\texttt{[\cnummazes{0}, \cnummazes{1}, \cnummazes{0}, \cnummazes{1}, \cnummazes{0}, \cnummazes{1}, \cnummazes{0}, \cnummazes{0}, \cnummazes{0}, \cnummazes{1}, \cnummazes{0}, \cnummazes{0}, \cnummazes{0}, \cnummazes{1}, \cnummazes{0}, \cnummazes{0}, \cnummazes{0}, \cnummazes{0}, \cnummazes{0}, \cnummazes{1}, \cnummazes{0}],} \vspace{-7.0pt} \\
\phantom{X}\texttt{[\cnummazes{0}, \cnummazes{1}, \cnummazes{1}, \cnummazes{1}, \cnummazes{1}, \cnummazes{1}, \cnummazes{0}, \cnummazes{1}, \cnummazes{1}, \cnummazes{1}, \cnummazes{0}, \cnummazes{1}, \cnummazes{1}, \cnummazes{1}, \cnummazes{1}, \cnummazes{1}, \cnummazes{1}, \cnummazes{1}, \cnummazes{0}, \cnummazes{1}, \cnummazes{0}],} \vspace{-7.0pt} \\
\phantom{X}\texttt{[\cnummazes{0}, \cnummazes{1}, \cnummazes{0}, \cnummazes{1}, \cnummazes{0}, \cnummazes{0}, \cnummazes{0}, \cnummazes{0}, \cnummazes{0}, \cnummazes{1}, \cnummazes{0}, \cnummazes{1}, \cnummazes{0}, \cnummazes{0}, \cnummazes{0}, \cnummazes{0}, \cnummazes{0}, \cnummazes{1}, \cnummazes{0}, \cnummazes{1}, \cnummazes{0}],} \vspace{-7.0pt} \\
\phantom{X}\texttt{[\cnummazes{0}, \cnummazes{1}, \cnummazes{0}, \cnummazes{1}, \cnummazes{0}, \cnummazes{1}, \cnummazes{1}, \cnummazes{1}, \cnummazes{0}, \cnummazes{1}, \cnummazes{0}, \cnummazes{1}, \cnummazes{1}, \cnummazes{1}, \cnummazes{0}, \cnummazes{1}, \cnummazes{0}, \cnummazes{1}, \cnummazes{0}, \cnummazes{1}, \cnummazes{0}],} \vspace{-7.0pt} \\
\phantom{X}\texttt{[\cnummazes{0}, \cnummazes{1}, \cnummazes{0}, \cnummazes{0}, \cnummazes{0}, \cnummazes{0}, \cnummazes{0}, \cnummazes{1}, \cnummazes{0}, \cnummazes{0}, \cnummazes{0}, \cnummazes{1}, \cnummazes{0}, \cnummazes{1}, \cnummazes{0}, \cnummazes{1}, \cnummazes{0}, \cnummazes{1}, \cnummazes{0}, \cnummazes{1}, \cnummazes{0}],} \vspace{-7.0pt} \\
\phantom{X}\texttt{[\cnummazes{0}, \cnummazes{1}, \cnummazes{1}, \cnummazes{1}, \cnummazes{1}, \cnummazes{1}, \cnummazes{1}, \cnummazes{1}, \cnummazes{0}, \cnummazes{1}, \cnummazes{1}, \cnummazes{1}, \cnummazes{0}, \cnummazes{1}, \cnummazes{0}, \cnummazes{1}, \cnummazes{0}, \cnummazes{1}, \cnummazes{0}, \cnummazes{1}, \cnummazes{0}],} \vspace{-7.0pt} \\
\phantom{X}\texttt{[\cnummazes{0}, \cnummazes{1}, \cnummazes{0}, \cnummazes{0}, \cnummazes{0}, \cnummazes{0}, \cnummazes{0}, \cnummazes{0}, \cnummazes{0}, \cnummazes{0}, \cnummazes{0}, \cnummazes{1}, \cnummazes{0}, \cnummazes{1}, \cnummazes{0}, \cnummazes{1}, \cnummazes{0}, \cnummazes{0}, \cnummazes{0}, \cnummazes{1}, \cnummazes{0}],} \vspace{-7.0pt} \\
\phantom{X}\texttt{[\cnummazes{0}, \cnummazes{1}, \cnummazes{1}, \cnummazes{1}, \cnummazes{1}, \cnummazes{1}, \cnummazes{1}, \cnummazes{1}, \cnummazes{0}, \cnummazes{1}, \cnummazes{1}, \cnummazes{1}, \cnummazes{0}, \cnummazes{1}, \cnummazes{1}, \cnummazes{1}, \cnummazes{0}, \cnummazes{1}, \cnummazes{1}, \cnummazes{1}, \cnummazes{0}],} \vspace{-7.0pt} \\
\phantom{X}\texttt{[\cnummazes{0}, \cnummazes{0}, \cnummazes{0}, \cnummazes{0}, \cnummazes{0}, \cnummazes{1}, \cnummazes{0}, \cnummazes{0}, \cnummazes{0}, \cnummazes{0}, \cnummazes{0}, \cnummazes{0}, \cnummazes{0}, \cnummazes{0}, \cnummazes{0}, \cnummazes{1}, \cnummazes{0}, \cnummazes{1}, \cnummazes{0}, \cnummazes{0}, \cnummazes{0}],} \vspace{-7.0pt} \\
\phantom{X}\texttt{[\cnummazes{0}, \cnummazes{1}, \cnummazes{1}, \cnummazes{1}, \cnummazes{0}, \cnummazes{1}, \cnummazes{1}, \cnummazes{1}, \cnummazes{0}, \cnummazes{1}, \cnummazes{0}, \cnummazes{1}, \cnummazes{0}, \cnummazes{1}, \cnummazes{0}, \cnummazes{1}, \cnummazes{0}, \cnummazes{1}, \cnummazes{1}, \cnummazes{1}, \cnummazes{0}],} \vspace{-7.0pt} \\
\phantom{X}\texttt{[\cnummazes{0}, \cnummazes{0}, \cnummazes{0}, \cnummazes{1}, \cnummazes{0}, \cnummazes{1}, \cnummazes{0}, \cnummazes{0}, \cnummazes{0}, \cnummazes{1}, \cnummazes{0}, \cnummazes{1}, \cnummazes{0}, \cnummazes{1}, \cnummazes{0}, \cnummazes{1}, \cnummazes{0}, \cnummazes{0}, \cnummazes{0}, \cnummazes{1}, \cnummazes{0}],} \vspace{-7.0pt} \\
\phantom{X}\texttt{[\cnummazes{0}, \cnummazes{1}, \cnummazes{1}, \cnummazes{1}, \cnummazes{1}, \cnummazes{1}, \cnummazes{0}, \cnummazes{1}, \cnummazes{1}, \cnummazes{1}, \cnummazes{1}, \cnummazes{1}, \cnummazes{1}, \cnummazes{1}, \cnummazes{1}, \cnummazes{1}, \cnummazes{0}, \cnummazes{1}, \cnummazes{1}, \cnummazes{1}, \cnummazes{0}],} \vspace{-7.0pt} \\
\phantom{X}\texttt{[\cnummazes{0}, \cnummazes{0}, \cnummazes{0}, \cnummazes{0}, \cnummazes{0}, \cnummazes{1}, \cnummazes{0}, \cnummazes{0}, \cnummazes{0}, \cnummazes{0}, \cnummazes{0}, \cnummazes{0}, \cnummazes{0}, \cnummazes{0}, \cnummazes{0}, \cnummazes{1}, \cnummazes{0}, \cnummazes{1}, \cnummazes{0}, \cnummazes{0}, \cnummazes{0}],} \vspace{-7.0pt} \\
\phantom{X}\texttt{[\cnummazes{0}, \cnummazes{1}, \cnummazes{1}, \cnummazes{1}, \cnummazes{1}, \cnummazes{1}, \cnummazes{1}, \cnummazes{1}, \cnummazes{1}, \cnummazes{1}, \cnummazes{1}, \cnummazes{1}, \cnummazes{0}, \cnummazes{1}, \cnummazes{1}, \cnummazes{1}, \cnummazes{1}, \cnummazes{1}, \cnummazes{1}, \cnummazes{2}, \cnummazes{0}],} \vspace{-7.0pt} \\
\phantom{X}\texttt{[\cnummazes{0}, \cnummazes{0}, \cnummazes{0}, \cnummazes{0}, \cnummazes{0}, \cnummazes{0}, \cnummazes{0}, \cnummazes{0}, \cnummazes{0}, \cnummazes{0}, \cnummazes{0}, \cnummazes{0}, \cnummazes{0}, \cnummazes{0}, \cnummazes{0}, \cnummazes{0}, \cnummazes{0}, \cnummazes{0}, \cnummazes{0}, \cnummazes{0}, \cnummazes{0}],} \vspace{-7.0pt} \\
\texttt{]}
\end{minipage} & 
\begin{minipage}[c]{\linewidth}
\vspace{-2.5cm} % ADJUST TO ALIGN
\fontsize{2.7}{2}
\noindent
\texttt{[} \vspace{-7.0pt}\\
\phantom{X}\texttt{[\cnummazes{0}, \cnummazes{0}, \cnummazes{0}, \cnummazes{0}, \cnummazes{0}, \cnummazes{0}, \cnummazes{0}, \cnummazes{0}, \cnummazes{0}, \cnummazes{0}, \cnummazes{0}, \cnummazes{0}, \cnummazes{0}, \cnummazes{0}, \cnummazes{0}, \cnummazes{0}, \cnummazes{0}, \cnummazes{0}, \cnummazes{0}, \cnummazes{0}, \cnummazes{0}],} \vspace{-7.0pt} \\
\phantom{X}\texttt{[\cnummazes{0}, \cnummazes{3}, \cnummazes{0}, \cnummazes{1}, \cnummazes{1}, \cnummazes{1}, \cnummazes{1}, \cnummazes{1}, \cnummazes{1}, \cnummazes{4}, \cnummazes{4}, \cnummazes{4}, \cnummazes{1}, \cnummazes{1}, \cnummazes{1}, \cnummazes{1}, \cnummazes{0}, \cnummazes{1}, \cnummazes{1}, \cnummazes{1}, \cnummazes{0}],} \vspace{-7.0pt} \\
\phantom{X}\texttt{[\cnummazes{0}, \cnummazes{4}, \cnummazes{0}, \cnummazes{0}, \cnummazes{0}, \cnummazes{0}, \cnummazes{0}, \cnummazes{0}, \cnummazes{0}, \cnummazes{4}, \cnummazes{0}, \cnummazes{4}, \cnummazes{0}, \cnummazes{0}, \cnummazes{0}, \cnummazes{0}, \cnummazes{0}, \cnummazes{0}, \cnummazes{0}, \cnummazes{1}, \cnummazes{0}],} \vspace{-7.0pt} \\
\phantom{X}\texttt{[\cnummazes{0}, \cnummazes{4}, \cnummazes{4}, \cnummazes{4}, \cnummazes{0}, \cnummazes{1}, \cnummazes{0}, \cnummazes{4}, \cnummazes{4}, \cnummazes{4}, \cnummazes{0}, \cnummazes{4}, \cnummazes{4}, \cnummazes{4}, \cnummazes{0}, \cnummazes{1}, \cnummazes{1}, \cnummazes{1}, \cnummazes{0}, \cnummazes{1}, \cnummazes{0}],} \vspace{-7.0pt} \\
\phantom{X}\texttt{[\cnummazes{0}, \cnummazes{0}, \cnummazes{0}, \cnummazes{4}, \cnummazes{0}, \cnummazes{1}, \cnummazes{0}, \cnummazes{4}, \cnummazes{0}, \cnummazes{0}, \cnummazes{0}, \cnummazes{0}, \cnummazes{0}, \cnummazes{4}, \cnummazes{0}, \cnummazes{1}, \cnummazes{0}, \cnummazes{0}, \cnummazes{0}, \cnummazes{1}, \cnummazes{0}],} \vspace{-7.0pt} \\
\phantom{X}\texttt{[\cnummazes{0}, \cnummazes{1}, \cnummazes{0}, \cnummazes{4}, \cnummazes{0}, \cnummazes{4}, \cnummazes{4}, \cnummazes{4}, \cnummazes{1}, \cnummazes{1}, \cnummazes{0}, \cnummazes{1}, \cnummazes{1}, \cnummazes{4}, \cnummazes{1}, \cnummazes{1}, \cnummazes{1}, \cnummazes{1}, \cnummazes{0}, \cnummazes{1}, \cnummazes{0}],} \vspace{-7.0pt} \\
\phantom{X}\texttt{[\cnummazes{0}, \cnummazes{1}, \cnummazes{0}, \cnummazes{4}, \cnummazes{0}, \cnummazes{4}, \cnummazes{0}, \cnummazes{0}, \cnummazes{0}, \cnummazes{1}, \cnummazes{0}, \cnummazes{0}, \cnummazes{0}, \cnummazes{4}, \cnummazes{0}, \cnummazes{0}, \cnummazes{0}, \cnummazes{0}, \cnummazes{0}, \cnummazes{1}, \cnummazes{0}],} \vspace{-7.0pt} \\
\phantom{X}\texttt{[\cnummazes{0}, \cnummazes{1}, \cnummazes{1}, \cnummazes{4}, \cnummazes{4}, \cnummazes{4}, \cnummazes{0}, \cnummazes{1}, \cnummazes{1}, \cnummazes{1}, \cnummazes{0}, \cnummazes{4}, \cnummazes{4}, \cnummazes{4}, \cnummazes{1}, \cnummazes{1}, \cnummazes{1}, \cnummazes{1}, \cnummazes{0}, \cnummazes{1}, \cnummazes{0}],} \vspace{-7.0pt} \\
\phantom{X}\texttt{[\cnummazes{0}, \cnummazes{1}, \cnummazes{0}, \cnummazes{1}, \cnummazes{0}, \cnummazes{0}, \cnummazes{0}, \cnummazes{0}, \cnummazes{0}, \cnummazes{1}, \cnummazes{0}, \cnummazes{4}, \cnummazes{0}, \cnummazes{0}, \cnummazes{0}, \cnummazes{0}, \cnummazes{0}, \cnummazes{1}, \cnummazes{0}, \cnummazes{1}, \cnummazes{0}],} \vspace{-7.0pt} \\
\phantom{X}\texttt{[\cnummazes{0}, \cnummazes{1}, \cnummazes{0}, \cnummazes{1}, \cnummazes{0}, \cnummazes{1}, \cnummazes{1}, \cnummazes{1}, \cnummazes{0}, \cnummazes{1}, \cnummazes{0}, \cnummazes{4}, \cnummazes{4}, \cnummazes{4}, \cnummazes{0}, \cnummazes{1}, \cnummazes{0}, \cnummazes{1}, \cnummazes{0}, \cnummazes{1}, \cnummazes{0}],} \vspace{-7.0pt} \\
\phantom{X}\texttt{[\cnummazes{0}, \cnummazes{1}, \cnummazes{0}, \cnummazes{0}, \cnummazes{0}, \cnummazes{0}, \cnummazes{0}, \cnummazes{1}, \cnummazes{0}, \cnummazes{0}, \cnummazes{0}, \cnummazes{1}, \cnummazes{0}, \cnummazes{4}, \cnummazes{0}, \cnummazes{1}, \cnummazes{0}, \cnummazes{1}, \cnummazes{0}, \cnummazes{1}, \cnummazes{0}],} \vspace{-7.0pt} \\
\phantom{X}\texttt{[\cnummazes{0}, \cnummazes{1}, \cnummazes{1}, \cnummazes{1}, \cnummazes{1}, \cnummazes{1}, \cnummazes{1}, \cnummazes{1}, \cnummazes{0}, \cnummazes{1}, \cnummazes{1}, \cnummazes{1}, \cnummazes{0}, \cnummazes{4}, \cnummazes{0}, \cnummazes{1}, \cnummazes{0}, \cnummazes{1}, \cnummazes{0}, \cnummazes{1}, \cnummazes{0}],} \vspace{-7.0pt} \\
\phantom{X}\texttt{[\cnummazes{0}, \cnummazes{1}, \cnummazes{0}, \cnummazes{0}, \cnummazes{0}, \cnummazes{0}, \cnummazes{0}, \cnummazes{0}, \cnummazes{0}, \cnummazes{0}, \cnummazes{0}, \cnummazes{1}, \cnummazes{0}, \cnummazes{4}, \cnummazes{0}, \cnummazes{1}, \cnummazes{0}, \cnummazes{0}, \cnummazes{0}, \cnummazes{1}, \cnummazes{0}],} \vspace{-7.0pt} \\
\phantom{X}\texttt{[\cnummazes{0}, \cnummazes{1}, \cnummazes{1}, \cnummazes{1}, \cnummazes{1}, \cnummazes{1}, \cnummazes{1}, \cnummazes{1}, \cnummazes{0}, \cnummazes{1}, \cnummazes{1}, \cnummazes{1}, \cnummazes{0}, \cnummazes{4}, \cnummazes{4}, \cnummazes{4}, \cnummazes{0}, \cnummazes{1}, \cnummazes{1}, \cnummazes{1}, \cnummazes{0}],} \vspace{-7.0pt} \\
\phantom{X}\texttt{[\cnummazes{0}, \cnummazes{0}, \cnummazes{0}, \cnummazes{0}, \cnummazes{0}, \cnummazes{1}, \cnummazes{0}, \cnummazes{0}, \cnummazes{0}, \cnummazes{0}, \cnummazes{0}, \cnummazes{0}, \cnummazes{0}, \cnummazes{0}, \cnummazes{0}, \cnummazes{4}, \cnummazes{0}, \cnummazes{1}, \cnummazes{0}, \cnummazes{0}, \cnummazes{0}],} \vspace{-7.0pt} \\
\phantom{X}\texttt{[\cnummazes{0}, \cnummazes{1}, \cnummazes{1}, \cnummazes{1}, \cnummazes{0}, \cnummazes{1}, \cnummazes{1}, \cnummazes{1}, \cnummazes{0}, \cnummazes{1}, \cnummazes{0}, \cnummazes{1}, \cnummazes{0}, \cnummazes{1}, \cnummazes{0}, \cnummazes{4}, \cnummazes{0}, \cnummazes{1}, \cnummazes{1}, \cnummazes{1}, \cnummazes{0}],} \vspace{-7.0pt} \\
\phantom{X}\texttt{[\cnummazes{0}, \cnummazes{0}, \cnummazes{0}, \cnummazes{1}, \cnummazes{0}, \cnummazes{1}, \cnummazes{0}, \cnummazes{0}, \cnummazes{0}, \cnummazes{1}, \cnummazes{0}, \cnummazes{1}, \cnummazes{0}, \cnummazes{1}, \cnummazes{0}, \cnummazes{4}, \cnummazes{0}, \cnummazes{0}, \cnummazes{0}, \cnummazes{1}, \cnummazes{0}],} \vspace{-7.0pt} \\
\phantom{X}\texttt{[\cnummazes{0}, \cnummazes{1}, \cnummazes{1}, \cnummazes{1}, \cnummazes{1}, \cnummazes{1}, \cnummazes{0}, \cnummazes{1}, \cnummazes{1}, \cnummazes{1}, \cnummazes{1}, \cnummazes{1}, \cnummazes{1}, \cnummazes{1}, \cnummazes{1}, \cnummazes{4}, \cnummazes{0}, \cnummazes{1}, \cnummazes{1}, \cnummazes{1}, \cnummazes{0}],} \vspace{-7.0pt} \\
\phantom{X}\texttt{[\cnummazes{0}, \cnummazes{0}, \cnummazes{0}, \cnummazes{0}, \cnummazes{0}, \cnummazes{1}, \cnummazes{0}, \cnummazes{0}, \cnummazes{0}, \cnummazes{0}, \cnummazes{0}, \cnummazes{0}, \cnummazes{0}, \cnummazes{0}, \cnummazes{0}, \cnummazes{4}, \cnummazes{0}, \cnummazes{1}, \cnummazes{0}, \cnummazes{0}, \cnummazes{0}],} \vspace{-7.0pt} \\
\phantom{X}\texttt{[\cnummazes{0}, \cnummazes{1}, \cnummazes{1}, \cnummazes{1}, \cnummazes{1}, \cnummazes{1}, \cnummazes{1}, \cnummazes{1}, \cnummazes{1}, \cnummazes{1}, \cnummazes{1}, \cnummazes{1}, \cnummazes{0}, \cnummazes{1}, \cnummazes{1}, \cnummazes{4}, \cnummazes{4}, \cnummazes{4}, \cnummazes{4}, \cnummazes{2}, \cnummazes{0}],} \vspace{-7.0pt} \\
\phantom{X}\texttt{[\cnummazes{0}, \cnummazes{0}, \cnummazes{0}, \cnummazes{0}, \cnummazes{0}, \cnummazes{0}, \cnummazes{0}, \cnummazes{0}, \cnummazes{0}, \cnummazes{0}, \cnummazes{0}, \cnummazes{0}, \cnummazes{0}, \cnummazes{0}, \cnummazes{0}, \cnummazes{0}, \cnummazes{0}, \cnummazes{0}, \cnummazes{0}, \cnummazes{0}, \cnummazes{0}],} \vspace{-7.0pt} \\
\texttt{]}
\end{minipage} \\
\end{tabular}
\captionof{figure}{Example input-output pair for task \textit{Maze}.}
\label{tab:task-example-maze}
\end{table}

We evaluate two scenarios:
\begin{itemize}[leftmargin=*]
    \item \textbf{Base Maze:} Training and evaluation on \(21 \times 21\) grids.  
    \item \textbf{Maze Generalization:} Training on smaller \(13 \times 13\) grids and testing on larger \(21 \times 21\) grids.  
\end{itemize}

We illustrate a sample \(21 \times 21\) maze in Figure~\ref{tab:task-example-maze}, which serves as training and evaluation data in the \textit{Base Maze} setting and as evaluation data in the \textit{Maze Generalization} setting. Figure~\ref{tab:task-example-maze-small} shows a sample \(13 \times 13\) maze, which is used as training data in the \textit{Maze Generalization} setting.

\subsubsection{Shortest Path}
\textbf{Objective:} Connect two arbitrary points with the shortest possible route.  

\textbf{Rules:}
\begin{enumerate}[leftmargin=*]
    \item Start and goal cells are specified anywhere on the grid.  
    \item Movement is allowed only through open cells.  
    \item Allowed moves are up, down, left, and right (no diagonal moves).  
    \item A valid solution is a continuous path from start to goal with minimal length among all possible paths.  
\end{enumerate}

We provide an example in Figure \ref{tab:task-example-navigation2d-any-to-any}.

%%%%%%%%%%%%%%%
% NAVIGATION 2D ANY TO ANY / SHORTEST PATH
%%%%%%%%%%%%%%%

\begin{table}[ht]
\centering
\renewcommand{\arraystretch}{3}
\begin{tabular}{c >{\centering\arraybackslash}m{4cm} >{\centering\arraybackslash}m{4cm}}
& \textbf{Input} & \textbf{Output} \\[1ex]
\rotatebox{90}{\textbf{\small Image Representation}} & 
\vspace{-2cm} % ADJUST TO ALIGN
\includegraphics[width=3cm]{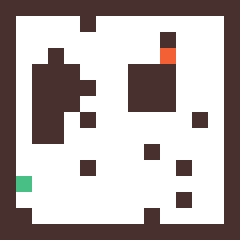} & 
\vspace{-2cm} % ADJUST TO ALIGN
\includegraphics[width=3cm]{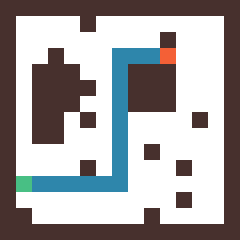} \\[2ex]
\rotatebox{90}{\textbf{\small Text Representation}} & 
\begin{minipage}[c]{\linewidth}
\vspace{-2.5cm} % ADJUST TO ALIGN
\fontsize{3.75}{2}
\noindent
\texttt{[} \vspace{-7.0pt}\\
\phantom{X}\texttt{[\cnummazes{0}, \cnummazes{0}, \cnummazes{0}, \cnummazes{0}, \cnummazes{0}, \cnummazes{0}, \cnummazes{0}, \cnummazes{0}, \cnummazes{0}, \cnummazes{0}, \cnummazes{0}, \cnummazes{0}, \cnummazes{0}, \cnummazes{0}, \cnummazes{0}],} \vspace{-7.0pt} \\
\phantom{X}\texttt{[\cnummazes{0}, \cnummazes{1}, \cnummazes{1}, \cnummazes{1}, \cnummazes{1}, \cnummazes{0}, \cnummazes{1}, \cnummazes{1}, \cnummazes{1}, \cnummazes{1}, \cnummazes{1}, \cnummazes{1}, \cnummazes{1}, \cnummazes{1}, \cnummazes{0}],} \vspace{-7.0pt} \\
\phantom{X}\texttt{[\cnummazes{0}, \cnummazes{1}, \cnummazes{1}, \cnummazes{1}, \cnummazes{1}, \cnummazes{1}, \cnummazes{1}, \cnummazes{1}, \cnummazes{1}, \cnummazes{1}, \cnummazes{0}, \cnummazes{1}, \cnummazes{1}, \cnummazes{1}, \cnummazes{0}],} \vspace{-7.0pt} \\
\phantom{X}\texttt{[\cnummazes{0}, \cnummazes{1}, \cnummazes{1}, \cnummazes{0}, \cnummazes{1}, \cnummazes{1}, \cnummazes{1}, \cnummazes{1}, \cnummazes{1}, \cnummazes{1}, \cnummazes{2}, \cnummazes{1}, \cnummazes{1}, \cnummazes{1}, \cnummazes{0}],} \vspace{-7.0pt} \\
\phantom{X}\texttt{[\cnummazes{0}, \cnummazes{1}, \cnummazes{0}, \cnummazes{0}, \cnummazes{0}, \cnummazes{1}, \cnummazes{1}, \cnummazes{1}, \cnummazes{0}, \cnummazes{0}, \cnummazes{0}, \cnummazes{1}, \cnummazes{1}, \cnummazes{1}, \cnummazes{0}],} \vspace{-7.0pt} \\
\phantom{X}\texttt{[\cnummazes{0}, \cnummazes{1}, \cnummazes{0}, \cnummazes{0}, \cnummazes{0}, \cnummazes{0}, \cnummazes{1}, \cnummazes{1}, \cnummazes{0}, \cnummazes{0}, \cnummazes{0}, \cnummazes{1}, \cnummazes{1}, \cnummazes{1}, \cnummazes{0}],} \vspace{-7.0pt} \\
\phantom{X}\texttt{[\cnummazes{0}, \cnummazes{1}, \cnummazes{0}, \cnummazes{0}, \cnummazes{0}, \cnummazes{1}, \cnummazes{1}, \cnummazes{1}, \cnummazes{0}, \cnummazes{0}, \cnummazes{0}, \cnummazes{1}, \cnummazes{1}, \cnummazes{1}, \cnummazes{0}],} \vspace{-7.0pt} \\
\phantom{X}\texttt{[\cnummazes{0}, \cnummazes{1}, \cnummazes{0}, \cnummazes{0}, \cnummazes{1}, \cnummazes{0}, \cnummazes{1}, \cnummazes{1}, \cnummazes{1}, \cnummazes{1}, \cnummazes{1}, \cnummazes{1}, \cnummazes{0}, \cnummazes{1}, \cnummazes{0}],} \vspace{-7.0pt} \\
\phantom{X}\texttt{[\cnummazes{0}, \cnummazes{1}, \cnummazes{0}, \cnummazes{0}, \cnummazes{1}, \cnummazes{1}, \cnummazes{1}, \cnummazes{1}, \cnummazes{1}, \cnummazes{1}, \cnummazes{1}, \cnummazes{1}, \cnummazes{1}, \cnummazes{1}, \cnummazes{0}],} \vspace{-7.0pt} \\
\phantom{X}\texttt{[\cnummazes{0}, \cnummazes{1}, \cnummazes{1}, \cnummazes{1}, \cnummazes{1}, \cnummazes{1}, \cnummazes{1}, \cnummazes{1}, \cnummazes{1}, \cnummazes{0}, \cnummazes{1}, \cnummazes{1}, \cnummazes{1}, \cnummazes{1}, \cnummazes{0}],} \vspace{-7.0pt} \\
\phantom{X}\texttt{[\cnummazes{0}, \cnummazes{1}, \cnummazes{1}, \cnummazes{1}, \cnummazes{1}, \cnummazes{0}, \cnummazes{1}, \cnummazes{1}, \cnummazes{1}, \cnummazes{1}, \cnummazes{1}, \cnummazes{0}, \cnummazes{1}, \cnummazes{1}, \cnummazes{0}],} \vspace{-7.0pt} \\
\phantom{X}\texttt{[\cnummazes{0}, \cnummazes{3}, \cnummazes{1}, \cnummazes{1}, \cnummazes{1}, \cnummazes{1}, \cnummazes{1}, \cnummazes{1}, \cnummazes{1}, \cnummazes{1}, \cnummazes{1}, \cnummazes{1}, \cnummazes{1}, \cnummazes{1}, \cnummazes{0}],} \vspace{-7.0pt} \\
\phantom{X}\texttt{[\cnummazes{0}, \cnummazes{1}, \cnummazes{1}, \cnummazes{1}, \cnummazes{1}, \cnummazes{1}, \cnummazes{1}, \cnummazes{1}, \cnummazes{1}, \cnummazes{1}, \cnummazes{1}, \cnummazes{0}, \cnummazes{1}, \cnummazes{1}, \cnummazes{0}],} \vspace{-7.0pt} \\
\phantom{X}\texttt{[\cnummazes{0}, \cnummazes{0}, \cnummazes{1}, \cnummazes{1}, \cnummazes{1}, \cnummazes{1}, \cnummazes{1}, \cnummazes{1}, \cnummazes{1}, \cnummazes{0}, \cnummazes{1}, \cnummazes{1}, \cnummazes{1}, \cnummazes{1}, \cnummazes{0}],} \vspace{-7.0pt} \\
\phantom{X}\texttt{[\cnummazes{0}, \cnummazes{0}, \cnummazes{0}, \cnummazes{0}, \cnummazes{0}, \cnummazes{0}, \cnummazes{0}, \cnummazes{0}, \cnummazes{0}, \cnummazes{0}, \cnummazes{0}, \cnummazes{0}, \cnummazes{0}, \cnummazes{0}, \cnummazes{0}],} \vspace{-7.0pt} \\
\texttt{]}
\end{minipage} & 
\begin{minipage}[c]{\linewidth}
\vspace{-2.5cm} % ADJUST TO ALIGN
\fontsize{3.75}{2}
\noindent
\texttt{[} \vspace{-7.0pt}\\
\phantom{X}\texttt{[\cnummazes{0}, \cnummazes{0}, \cnummazes{0}, \cnummazes{0}, \cnummazes{0}, \cnummazes{0}, \cnummazes{0}, \cnummazes{0}, \cnummazes{0}, \cnummazes{0}, \cnummazes{0}, \cnummazes{0}, \cnummazes{0}, \cnummazes{0}, \cnummazes{0}],} \vspace{-7.0pt} \\
\phantom{X}\texttt{[\cnummazes{0}, \cnummazes{1}, \cnummazes{1}, \cnummazes{1}, \cnummazes{1}, \cnummazes{0}, \cnummazes{1}, \cnummazes{1}, \cnummazes{1}, \cnummazes{1}, \cnummazes{1}, \cnummazes{1}, \cnummazes{1}, \cnummazes{1}, \cnummazes{0}],} \vspace{-7.0pt} \\
\phantom{X}\texttt{[\cnummazes{0}, \cnummazes{1}, \cnummazes{1}, \cnummazes{1}, \cnummazes{1}, \cnummazes{1}, \cnummazes{1}, \cnummazes{1}, \cnummazes{1}, \cnummazes{1}, \cnummazes{0}, \cnummazes{1}, \cnummazes{1}, \cnummazes{1}, \cnummazes{0}],} \vspace{-7.0pt} \\
\phantom{X}\texttt{[\cnummazes{0}, \cnummazes{1}, \cnummazes{1}, \cnummazes{0}, \cnummazes{1}, \cnummazes{1}, \cnummazes{1}, \cnummazes{4}, \cnummazes{4}, \cnummazes{4}, \cnummazes{2}, \cnummazes{1}, \cnummazes{1}, \cnummazes{1}, \cnummazes{0}],} \vspace{-7.0pt} \\
\phantom{X}\texttt{[\cnummazes{0}, \cnummazes{1}, \cnummazes{0}, \cnummazes{0}, \cnummazes{0}, \cnummazes{1}, \cnummazes{1}, \cnummazes{4}, \cnummazes{0}, \cnummazes{0}, \cnummazes{0}, \cnummazes{1}, \cnummazes{1}, \cnummazes{1}, \cnummazes{0}],} \vspace{-7.0pt} \\
\phantom{X}\texttt{[\cnummazes{0}, \cnummazes{1}, \cnummazes{0}, \cnummazes{0}, \cnummazes{0}, \cnummazes{0}, \cnummazes{1}, \cnummazes{4}, \cnummazes{0}, \cnummazes{0}, \cnummazes{0}, \cnummazes{1}, \cnummazes{1}, \cnummazes{1}, \cnummazes{0}],} \vspace{-7.0pt} \\
\phantom{X}\texttt{[\cnummazes{0}, \cnummazes{1}, \cnummazes{0}, \cnummazes{0}, \cnummazes{0}, \cnummazes{1}, \cnummazes{1}, \cnummazes{4}, \cnummazes{0}, \cnummazes{0}, \cnummazes{0}, \cnummazes{1}, \cnummazes{1}, \cnummazes{1}, \cnummazes{0}],} \vspace{-7.0pt} \\
\phantom{X}\texttt{[\cnummazes{0}, \cnummazes{1}, \cnummazes{0}, \cnummazes{0}, \cnummazes{1}, \cnummazes{0}, \cnummazes{1}, \cnummazes{4}, \cnummazes{1}, \cnummazes{1}, \cnummazes{1}, \cnummazes{1}, \cnummazes{0}, \cnummazes{1}, \cnummazes{0}],} \vspace{-7.0pt} \\
\phantom{X}\texttt{[\cnummazes{0}, \cnummazes{1}, \cnummazes{0}, \cnummazes{0}, \cnummazes{1}, \cnummazes{1}, \cnummazes{1}, \cnummazes{4}, \cnummazes{1}, \cnummazes{1}, \cnummazes{1}, \cnummazes{1}, \cnummazes{1}, \cnummazes{1}, \cnummazes{0}],} \vspace{-7.0pt} \\
\phantom{X}\texttt{[\cnummazes{0}, \cnummazes{1}, \cnummazes{1}, \cnummazes{1}, \cnummazes{1}, \cnummazes{1}, \cnummazes{1}, \cnummazes{4}, \cnummazes{1}, \cnummazes{0}, \cnummazes{1}, \cnummazes{1}, \cnummazes{1}, \cnummazes{1}, \cnummazes{0}],} \vspace{-7.0pt} \\
\phantom{X}\texttt{[\cnummazes{0}, \cnummazes{1}, \cnummazes{1}, \cnummazes{1}, \cnummazes{1}, \cnummazes{0}, \cnummazes{1}, \cnummazes{4}, \cnummazes{1}, \cnummazes{1}, \cnummazes{1}, \cnummazes{0}, \cnummazes{1}, \cnummazes{1}, \cnummazes{0}],} \vspace{-7.0pt} \\
\phantom{X}\texttt{[\cnummazes{0}, \cnummazes{3}, \cnummazes{4}, \cnummazes{4}, \cnummazes{4}, \cnummazes{4}, \cnummazes{4}, \cnummazes{4}, \cnummazes{1}, \cnummazes{1}, \cnummazes{1}, \cnummazes{1}, \cnummazes{1}, \cnummazes{1}, \cnummazes{0}],} \vspace{-7.0pt} \\
\phantom{X}\texttt{[\cnummazes{0}, \cnummazes{1}, \cnummazes{1}, \cnummazes{1}, \cnummazes{1}, \cnummazes{1}, \cnummazes{1}, \cnummazes{1}, \cnummazes{1}, \cnummazes{1}, \cnummazes{1}, \cnummazes{0}, \cnummazes{1}, \cnummazes{1}, \cnummazes{0}],} \vspace{-7.0pt} \\
\phantom{X}\texttt{[\cnummazes{0}, \cnummazes{0}, \cnummazes{1}, \cnummazes{1}, \cnummazes{1}, \cnummazes{1}, \cnummazes{1}, \cnummazes{1}, \cnummazes{1}, \cnummazes{0}, \cnummazes{1}, \cnummazes{1}, \cnummazes{1}, \cnummazes{1}, \cnummazes{0}],} \vspace{-7.0pt} \\
\phantom{X}\texttt{[\cnummazes{0}, \cnummazes{0}, \cnummazes{0}, \cnummazes{0}, \cnummazes{0}, \cnummazes{0}, \cnummazes{0}, \cnummazes{0}, \cnummazes{0}, \cnummazes{0}, \cnummazes{0}, \cnummazes{0}, \cnummazes{0}, \cnummazes{0}, \cnummazes{0}],} \vspace{-7.0pt} \\
\texttt{]}
\end{minipage} \\
\end{tabular}
\captionof{figure}{Example input-output pair for task \textit{Shortest Path}.}
\label{tab:task-example-navigation2d-any-to-any}
\end{table}

\subsection{Cellular Automata}

\subsubsection{Elementary Cellular Automata (ECA)}
Elementary Cellular Automata (ECA) are one-dimensional binary-state automata defined on a line of cells. Each cell \(c_i^t \in \{0,1\}\) at time \(t\) updates based on itself and its two neighbors:
\[
c_i^{t+1} = f(c_{i-1}^t, c_i^t, c_{i+1}^t),
\]
where \(f\) is specified by a rule number between 0 and 255.  

For example, Rule 110 is encoded by the binary string \texttt{01101110}, which maps the eight possible neighborhoods \((c_{i-1}^t, c_i^t, c_{i+1}^t)\) to the next state:
\[
\begin{array}{c|cccccccc}
\text{Neighborhood} & 111 & 110 & 101 & 100 & 011 & 010 & 001 & 000 \\
\hline
\text{Next state}   & 0   & 1   & 1   & 0   & 1   & 1   & 1   & 0
\end{array}
\]

We evaluate four representative rules from each of Wolfram's classes~\cite{Wolfram1984Universality}, summarized in Table~\ref{tab:eca-rules}.

\begin{table}[t]
\centering
\caption{Representative Elementary Cellular Automata rules by Wolfram class.}
\label{tab:eca-rules}
\begin{tabular}{ll}
\toprule
\textbf{Class} & \textbf{Rules} \\
\midrule
Class 1 & 8, 32, 128, 160 \\
Class 2 & 4, 108, 170, 250 \\
Class 3 & 30, 45, 90, 150 \\
Class 4 & 110, 54, 62, 106 \\
\bottomrule
\end{tabular}
\end{table}

Rule 110 is well known for its complex localized structures and universality~\cite{Cook2004Universality}. We show an example in Figure \ref{tab:task-example-eca-rule110}.

%%%%%%%%%%%%%%%
% ECA RULE 110
%%%%%%%%%%%%%%%
\begin{table}[ht]
\centering
\renewcommand{\arraystretch}{3}
\begin{tabular}{c >{\centering\arraybackslash}m{4cm} >{\centering\arraybackslash}m{4cm}}
& \textbf{Input} & \textbf{Output} \\[1ex]
\rotatebox{90}{\textbf{\small Image Representation}} & 
\vspace{-3cm} % ADJUST TO ALIGN
\includegraphics[width=3cm]{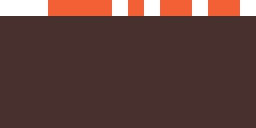} & 
\vspace{-3cm} % ADJUST TO ALIGN
\includegraphics[width=3cm]{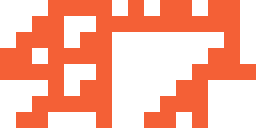} \\[2ex]
\rotatebox{90}{\textbf{\small Text Representation}} & 
\begin{minipage}[c]{\linewidth}
\vspace{-2.5cm} % ADJUST TO ALIGN
\fontsize{3.25}{2}
\noindent
\texttt{[} \vspace{-7.0pt}\\
\phantom{X}\texttt{[\cnummazes{1}, \cnummazes{1}, \cnummazes{1}, \cnummazes{2}, \cnummazes{2}, \cnummazes{2}, \cnummazes{2}, \cnummazes{1}, \cnummazes{2}, \cnummazes{1}, \cnummazes{2}, \cnummazes{2}, \cnummazes{1}, \cnummazes{2}, \cnummazes{2}, \cnummazes{1}],} \vspace{-7.0pt} \\
\phantom{X}\texttt{[\cnummazes{0}, \cnummazes{0}, \cnummazes{0}, \cnummazes{0}, \cnummazes{0}, \cnummazes{0}, \cnummazes{0}, \cnummazes{0}, \cnummazes{0}, \cnummazes{0}, \cnummazes{0}, \cnummazes{0}, \cnummazes{0}, \cnummazes{0}, \cnummazes{0}, \cnummazes{0}],} \vspace{-7.0pt} \\
\phantom{X}\texttt{[\cnummazes{0}, \cnummazes{0}, \cnummazes{0}, \cnummazes{0}, \cnummazes{0}, \cnummazes{0}, \cnummazes{0}, \cnummazes{0}, \cnummazes{0}, \cnummazes{0}, \cnummazes{0}, \cnummazes{0}, \cnummazes{0}, \cnummazes{0}, \cnummazes{0}, \cnummazes{0}],} \vspace{-7.0pt} \\
\phantom{X}\texttt{[\cnummazes{0}, \cnummazes{0}, \cnummazes{0}, \cnummazes{0}, \cnummazes{0}, \cnummazes{0}, \cnummazes{0}, \cnummazes{0}, \cnummazes{0}, \cnummazes{0}, \cnummazes{0}, \cnummazes{0}, \cnummazes{0}, \cnummazes{0}, \cnummazes{0}, \cnummazes{0}],} \vspace{-7.0pt} \\
\phantom{X}\texttt{[\cnummazes{0}, \cnummazes{0}, \cnummazes{0}, \cnummazes{0}, \cnummazes{0}, \cnummazes{0}, \cnummazes{0}, \cnummazes{0}, \cnummazes{0}, \cnummazes{0}, \cnummazes{0}, \cnummazes{0}, \cnummazes{0}, \cnummazes{0}, \cnummazes{0}, \cnummazes{0}],} \vspace{-7.0pt} \\
\phantom{X}\texttt{[\cnummazes{0}, \cnummazes{0}, \cnummazes{0}, \cnummazes{0}, \cnummazes{0}, \cnummazes{0}, \cnummazes{0}, \cnummazes{0}, \cnummazes{0}, \cnummazes{0}, \cnummazes{0}, \cnummazes{0}, \cnummazes{0}, \cnummazes{0}, \cnummazes{0}, \cnummazes{0}],} \vspace{-7.0pt} \\
\phantom{X}\texttt{[\cnummazes{0}, \cnummazes{0}, \cnummazes{0}, \cnummazes{0}, \cnummazes{0}, \cnummazes{0}, \cnummazes{0}, \cnummazes{0}, \cnummazes{0}, \cnummazes{0}, \cnummazes{0}, \cnummazes{0}, \cnummazes{0}, \cnummazes{0}, \cnummazes{0}, \cnummazes{0}],} \vspace{-7.0pt} \\
\phantom{X}\texttt{[\cnummazes{0}, \cnummazes{0}, \cnummazes{0}, \cnummazes{0}, \cnummazes{0}, \cnummazes{0}, \cnummazes{0}, \cnummazes{0}, \cnummazes{0}, \cnummazes{0}, \cnummazes{0}, \cnummazes{0}, \cnummazes{0}, \cnummazes{0}, \cnummazes{0}, \cnummazes{0}],} \vspace{-7.0pt} \\
\texttt{]}
\end{minipage} & 
\begin{minipage}[c]{\linewidth}
\vspace{-2.5cm} % ADJUST TO ALIGN
\fontsize{3.25}{2}
\noindent
\texttt{[} \vspace{-7.0pt}\\
\phantom{X}\texttt{[\cnummazes{1}, \cnummazes{1}, \cnummazes{1}, \cnummazes{2}, \cnummazes{2}, \cnummazes{2}, \cnummazes{2}, \cnummazes{1}, \cnummazes{2}, \cnummazes{1}, \cnummazes{2}, \cnummazes{2}, \cnummazes{1}, \cnummazes{2}, \cnummazes{2}, \cnummazes{1}],} \vspace{-7.0pt} \\
\phantom{X}\texttt{[\cnummazes{1}, \cnummazes{1}, \cnummazes{2}, \cnummazes{2}, \cnummazes{1}, \cnummazes{1}, \cnummazes{2}, \cnummazes{2}, \cnummazes{2}, \cnummazes{2}, \cnummazes{2}, \cnummazes{2}, \cnummazes{2}, \cnummazes{2}, \cnummazes{2}, \cnummazes{1}],} \vspace{-7.0pt} \\
\phantom{X}\texttt{[\cnummazes{1}, \cnummazes{2}, \cnummazes{2}, \cnummazes{2}, \cnummazes{1}, \cnummazes{2}, \cnummazes{2}, \cnummazes{1}, \cnummazes{1}, \cnummazes{1}, \cnummazes{1}, \cnummazes{1}, \cnummazes{1}, \cnummazes{1}, \cnummazes{2}, \cnummazes{1}],} \vspace{-7.0pt} \\
\phantom{X}\texttt{[\cnummazes{2}, \cnummazes{2}, \cnummazes{1}, \cnummazes{2}, \cnummazes{2}, \cnummazes{2}, \cnummazes{2}, \cnummazes{1}, \cnummazes{1}, \cnummazes{1}, \cnummazes{1}, \cnummazes{1}, \cnummazes{1}, \cnummazes{2}, \cnummazes{2}, \cnummazes{1}],} \vspace{-7.0pt} \\
\phantom{X}\texttt{[\cnummazes{2}, \cnummazes{2}, \cnummazes{2}, \cnummazes{2}, \cnummazes{1}, \cnummazes{1}, \cnummazes{2}, \cnummazes{1}, \cnummazes{1}, \cnummazes{1}, \cnummazes{1}, \cnummazes{1}, \cnummazes{2}, \cnummazes{2}, \cnummazes{2}, \cnummazes{2}],} \vspace{-7.0pt} \\
\phantom{X}\texttt{[\cnummazes{1}, \cnummazes{1}, \cnummazes{1}, \cnummazes{2}, \cnummazes{1}, \cnummazes{2}, \cnummazes{2}, \cnummazes{1}, \cnummazes{1}, \cnummazes{1}, \cnummazes{1}, \cnummazes{2}, \cnummazes{2}, \cnummazes{1}, \cnummazes{1}, \cnummazes{1}],} \vspace{-7.0pt} \\
\phantom{X}\texttt{[\cnummazes{1}, \cnummazes{1}, \cnummazes{2}, \cnummazes{2}, \cnummazes{2}, \cnummazes{2}, \cnummazes{2}, \cnummazes{1}, \cnummazes{1}, \cnummazes{1}, \cnummazes{2}, \cnummazes{2}, \cnummazes{2}, \cnummazes{1}, \cnummazes{1}, \cnummazes{1}],} \vspace{-7.0pt} \\
\phantom{X}\texttt{[\cnummazes{1}, \cnummazes{2}, \cnummazes{2}, \cnummazes{1}, \cnummazes{1}, \cnummazes{1}, \cnummazes{2}, \cnummazes{1}, \cnummazes{1}, \cnummazes{2}, \cnummazes{2}, \cnummazes{1}, \cnummazes{2}, \cnummazes{1}, \cnummazes{1}, \cnummazes{1}],} \vspace{-7.0pt} \\
\texttt{]}
\end{minipage} \\
\end{tabular}
\captionof{figure}{Example input-output pair for task \textit{ECA rule 101}.}
\label{tab:task-example-eca-rule110}
\end{table}

\subsubsection{Life-like Cellular Automata}
Life-like CA generalize Conway's Game of Life~\cite{Gardner1970Life}, using binary cells on a two-dimensional grid. Each cell updates according to the number of live neighbors in the Moore neighborhood (eight adjacent cells). In standard Game of Life (\(B3/S23\)):
\[
c_{i,j}^{t+1} =
\begin{cases}
1 & \text{if cell is dead and has exactly 3 live neighbors (birth)},\\
1 & \text{if cell is alive and has 2 or 3 live neighbors (survival)},\\
0 & \text{otherwise (death).}
\end{cases}
\]

We consider several well-known Life-like variants. These rules, summarized in Table~\ref{tab:lifelike-rules}, capture diverse behaviors ranging from explosive growth to symmetry under inversion. We shown an example in Figure \ref{tab:task-example-game-of-life-step1} of the basic Game of Life.

%%%%%%%%%%%%%%%
% GAME OF LIFE STEP 1
%%%%%%%%%%%%%%%

\begin{table}[ht]
\centering
\renewcommand{\arraystretch}{3}
\begin{tabular}{c >{\centering\arraybackslash}m{4cm} >{\centering\arraybackslash}m{4cm}}
& \textbf{Input} & \textbf{Output} \\[1ex]
\rotatebox{90}{\textbf{\small Image Representation}} & 
\vspace{-2cm} % ADJUST TO ALIGN
\includegraphics[width=3cm]{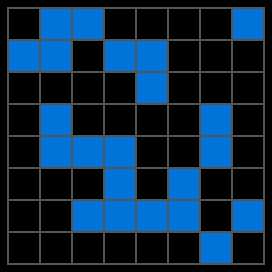} & 
\vspace{-2cm} % ADJUST TO ALIGN
\includegraphics[width=3cm]
{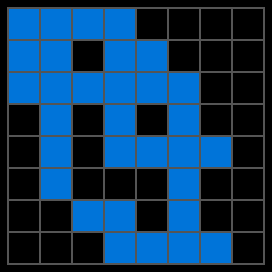} \\[2ex]
\rotatebox{90}{\textbf{\small Text Representation}} & 
\begin{minipage}[c]{\linewidth}
\vspace{-2.5cm} % ADJUST TO ALIGN
\tiny
\noindent
\texttt{[} \\
\phantom{X}\texttt{[\cnumarc{0}, \cnumarc{1}, \cnumarc{1}, \cnumarc{0}, \cnumarc{0}, \cnumarc{0}, \cnumarc{0}, \cnumarc{1}],} \\
\phantom{X}\texttt{[\cnumarc{1}, \cnumarc{1}, \cnumarc{0}, \cnumarc{1}, \cnumarc{1}, \cnumarc{0}, \cnumarc{0}, \cnumarc{0}],} \\
\phantom{X}\texttt{[\cnumarc{0}, \cnumarc{0}, \cnumarc{0}, \cnumarc{0}, \cnumarc{1}, \cnumarc{0}, \cnumarc{0}, \cnumarc{0}],} \\
\phantom{X}\texttt{[\cnumarc{0}, \cnumarc{1}, \cnumarc{0}, \cnumarc{0}, \cnumarc{0}, \cnumarc{0}, \cnumarc{1}, \cnumarc{0}],} \\
\phantom{X}\texttt{[\cnumarc{0}, \cnumarc{1}, \cnumarc{1}, \cnumarc{1}, \cnumarc{0}, \cnumarc{0}, \cnumarc{1}, \cnumarc{0}],} \\
\phantom{X}\texttt{[\cnumarc{0}, \cnumarc{0}, \cnumarc{0}, \cnumarc{1}, \cnumarc{0}, \cnumarc{1}, \cnumarc{0}, \cnumarc{0}],} \\
\phantom{X}\texttt{[\cnumarc{0}, \cnumarc{0}, \cnumarc{1}, \cnumarc{1}, \cnumarc{1}, \cnumarc{1}, \cnumarc{0}, \cnumarc{1}],} \\
\phantom{X}\texttt{[\cnumarc{0}, \cnumarc{0}, \cnumarc{0}, \cnumarc{0}, \cnumarc{0}, \cnumarc{0}, \cnumarc{1}, \cnumarc{0}],} \\
\texttt{]}
\end{minipage} & 
\begin{minipage}[c]{\linewidth}
\vspace{-2.5cm} % ADJUST TO ALIGN
\tiny
\noindent
\texttt{[} \\
\phantom{X}\texttt{[\cnumarc{1}, \cnumarc{1}, \cnumarc{1}, \cnumarc{1}, \cnumarc{0}, \cnumarc{0}, \cnumarc{0}, \cnumarc{0}],} \\
\phantom{X}\texttt{[\cnumarc{1}, \cnumarc{1}, \cnumarc{0}, \cnumarc{1}, \cnumarc{1}, \cnumarc{0}, \cnumarc{0}, \cnumarc{0}],} \\
\phantom{X}\texttt{[\cnumarc{1}, \cnumarc{1}, \cnumarc{1}, \cnumarc{1}, \cnumarc{1}, \cnumarc{1}, \cnumarc{0}, \cnumarc{0}],} \\
\phantom{X}\texttt{[\cnumarc{0}, \cnumarc{1}, \cnumarc{0}, \cnumarc{1}, \cnumarc{0}, \cnumarc{1}, \cnumarc{0}, \cnumarc{0}],} \\
\phantom{X}\texttt{[\cnumarc{0}, \cnumarc{1}, \cnumarc{0}, \cnumarc{1}, \cnumarc{1}, \cnumarc{1}, \cnumarc{1}, \cnumarc{0}],} \\
\phantom{X}\texttt{[\cnumarc{0}, \cnumarc{1}, \cnumarc{0}, \cnumarc{0}, \cnumarc{0}, \cnumarc{1}, \cnumarc{0}, \cnumarc{0}],} \\
\phantom{X}\texttt{[\cnumarc{0}, \cnumarc{0}, \cnumarc{1}, \cnumarc{1}, \cnumarc{0}, \cnumarc{1}, \cnumarc{0}, \cnumarc{0}],} \\
\phantom{X}\texttt{[\cnumarc{0}, \cnumarc{0}, \cnumarc{0}, \cnumarc{1}, \cnumarc{1}, \cnumarc{1}, \cnumarc{1}, \cnumarc{0}],} \\
\texttt{]}
\end{minipage} \\
\end{tabular}
\captionof{figure}{Example input-output pair for task \textit{Game of Life step 1}.}
\label{tab:task-example-game-of-life-step1}
\end{table}

\begin{table}[t]
\centering
\caption{Life-like cellular automata variants evaluated.}
\label{tab:lifelike-rules}
\begin{tabular}{lll}
\toprule
\textbf{Name} & \textbf{Rule (B/S)} & \textbf{Description} \\
\midrule
Day \& Night & B3678/S34678 & Symmetric under inversion; complex dynamics \\
Maze & B3/S12345 & Generates labyrinth-like, maze-like growth \\
Seeds & B2/S$\varnothing$ & All live cells die each step; explosive expansion \\
Life & B3/S2 & Sparse survival; promotes small, mobile clusters \\
\bottomrule
\end{tabular}
\end{table}

\subsubsection{Langton's Ant}
Langton's ant~\cite{Langton1986Studying} is an agent-based CA where a single agent moves on a binary grid. At each step:
\[
(x,y), d, g(x,y) \rightarrow (x',y'), d', g'(x,y),
\]
where \((x,y)\) is the current cell, \(d\) is direction, and \(g(x,y)\in\{0,1\}\) is the cell state.
\begin{enumerate}[leftmargin=*]
    \item If \(g(x,y)=0\), turn right; if \(g(x,y)=1\), turn left.
    \item Flip the cell color: \(g'(x,y)=1-g(x,y)\).
    \item Move forward one step.
\end{enumerate}
After many steps, chaotic behavior gives way to a repeating “highway” structure. To make the task predictable, \textbf{we always start with the ant facing on the same initial direction and being on top of a 0 cell.} For an example see Figure \ref{tab:task-example-langton_ant_step2}

%%%%%%%%%%%%%%%
% LANGTON ANT STEP 2
%%%%%%%%%%%%%%%
\begin{table}[ht]
\centering
\renewcommand{\arraystretch}{3}
\begin{tabular}{c >{\centering\arraybackslash}m{4cm} >{\centering\arraybackslash}m{4cm}}
& \textbf{Input} & \textbf{Output} \\[1ex]
\rotatebox{90}{\textbf{\small Image Representation}} & 
\vspace{-2cm} % ADJUST TO ALIGN
\includegraphics[width=3cm]{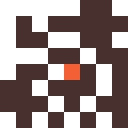} & 
\vspace{-2cm} % ADJUST TO ALIGN
\includegraphics[width=3cm]
{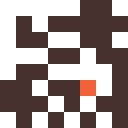} \\[2ex]
\rotatebox{90}{\textbf{\small Text Representation}} & 
\begin{minipage}[c]{\linewidth}
\vspace{-2.5cm} % ADJUST TO ALIGN
\tiny
\texttt{[} \\
\phantom{X}\texttt{[\cnummazes{1}, \cnummazes{1}, \cnummazes{1}, \cnummazes{1}, \cnummazes{1}, \cnummazes{0}, \cnummazes{0}, \cnummazes{1}],} \\
\phantom{X}\texttt{[\cnummazes{1}, \cnummazes{0}, \cnummazes{1}, \cnummazes{0}, \cnummazes{0}, \cnummazes{0}, \cnummazes{0}, \cnummazes{0}],} \\
\phantom{X}\texttt{[\cnummazes{1}, \cnummazes{0}, \cnummazes{0}, \cnummazes{1}, \cnummazes{1}, \cnummazes{0}, \cnummazes{0}, \cnummazes{1}],} \\
\phantom{X}\texttt{[\cnummazes{1}, \cnummazes{1}, \cnummazes{1}, \cnummazes{0}, \cnummazes{0}, \cnummazes{1}, \cnummazes{0}, \cnummazes{0}],} \\
\phantom{X}\texttt{[\cnummazes{1}, \cnummazes{0}, \cnummazes{0}, \cnummazes{1}, \cnummazes{2}, \cnummazes{0}, \cnummazes{1}, \cnummazes{0}],} \\
\phantom{X}\texttt{[\cnummazes{0}, \cnummazes{0}, \cnummazes{1}, \cnummazes{0}, \cnummazes{0}, \cnummazes{0}, \cnummazes{1}, \cnummazes{0}],} \\
\phantom{X}\texttt{[\cnummazes{0}, \cnummazes{0}, \cnummazes{0}, \cnummazes{1}, \cnummazes{0}, \cnummazes{1}, \cnummazes{0}, \cnummazes{0}],} \\
\phantom{X}\texttt{[\cnummazes{1}, \cnummazes{0}, \cnummazes{1}, \cnummazes{0}, \cnummazes{1}, \cnummazes{0}, \cnummazes{1}, \cnummazes{0}],} \\
\texttt{]}
\end{minipage} & 
\begin{minipage}[c]{\linewidth}
\vspace{-2.5cm} % ADJUST TO ALIGN
\tiny
\texttt{[} \\
\phantom{X}\texttt{[\cnummazes{1}, \cnummazes{1}, \cnummazes{1}, \cnummazes{1}, \cnummazes{1}, \cnummazes{0}, \cnummazes{0}, \cnummazes{1}],} \\
\phantom{X}\texttt{[\cnummazes{1}, \cnummazes{0}, \cnummazes{1}, \cnummazes{0}, \cnummazes{0}, \cnummazes{0}, \cnummazes{0}, \cnummazes{0}],} \\
\phantom{X}\texttt{[\cnummazes{1}, \cnummazes{0}, \cnummazes{0}, \cnummazes{1}, \cnummazes{1}, \cnummazes{0}, \cnummazes{0}, \cnummazes{1}],} \\
\phantom{X}\texttt{[\cnummazes{1}, \cnummazes{1}, \cnummazes{1}, \cnummazes{0}, \cnummazes{0}, \cnummazes{1}, \cnummazes{0}, \cnummazes{0}],} \\
\phantom{X}\texttt{[\cnummazes{1}, \cnummazes{0}, \cnummazes{0}, \cnummazes{1}, \cnummazes{1}, \cnummazes{1}, \cnummazes{1}, \cnummazes{0}],} \\
\phantom{X}\texttt{[\cnummazes{0}, \cnummazes{0}, \cnummazes{1}, \cnummazes{0}, \cnummazes{0}, \cnummazes{2}, \cnummazes{1}, \cnummazes{0}],} \\
\phantom{X}\texttt{[\cnummazes{0}, \cnummazes{0}, \cnummazes{0}, \cnummazes{1}, \cnummazes{0}, \cnummazes{1}, \cnummazes{0}, \cnummazes{0}],} \\
\phantom{X}\texttt{[\cnummazes{1}, \cnummazes{0}, \cnummazes{1}, \cnummazes{0}, \cnummazes{1}, \cnummazes{0}, \cnummazes{1}, \cnummazes{0}],} \\
\texttt{]}
\end{minipage} \\
\end{tabular}
\captionof{figure}{Example input-output pair for task \textit{Langton ant step 2}.}
\label{tab:task-example-langton_ant_step2}
\end{table}

\section{Additional Qualitative Results}

\subsection{ARC-AGI}

\begin{figure}[htbp]
\centering
\renewcommand{\arraystretch}{0.6}
\setlength{\tabcolsep}{2pt}
\begin{tabular}{lcccccc}
& \textbf{Input} & \textbf{Output} & \textbf{Input} & \textbf{Output} & \textbf{Input} & \textbf{Output} \\[0.2em]
\multirow{3}{*}{\rotatebox{90}{\textbf{Training Examples}}} & \includegraphics[width=1.8cm,height=1.8cm]{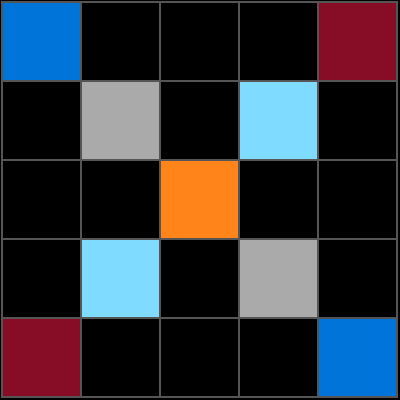} & \includegraphics[width=1.8cm,height=1.8cm]{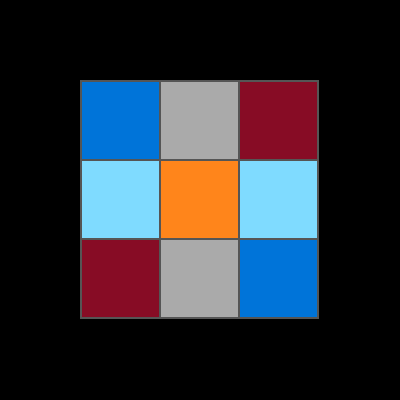} & \includegraphics[width=1.8cm,height=1.8cm]{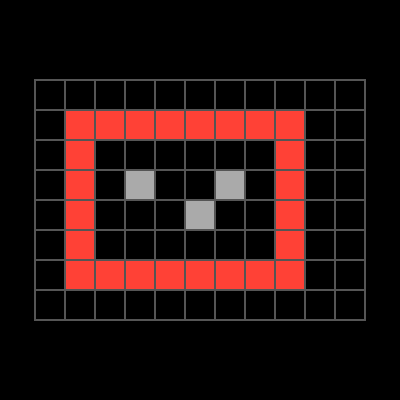} & \includegraphics[width=1.8cm,height=1.8cm]{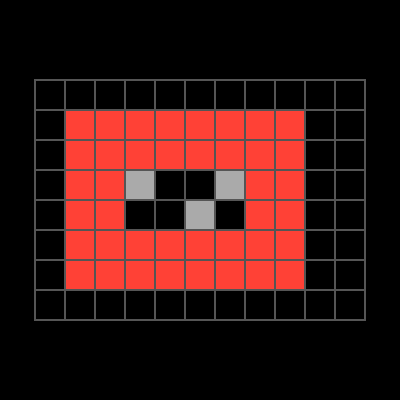} & \includegraphics[width=1.8cm,height=1.8cm]{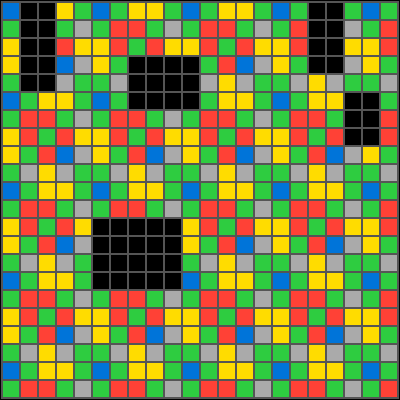} & \includegraphics[width=1.8cm,height=1.8cm]{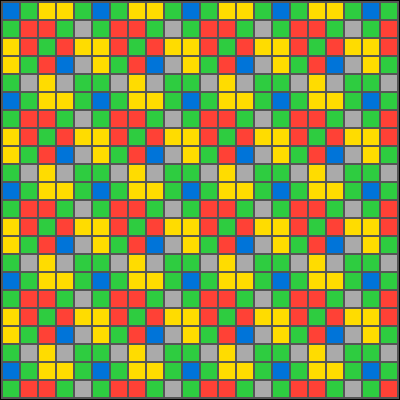} \\
& \includegraphics[width=1.8cm,height=1.8cm]{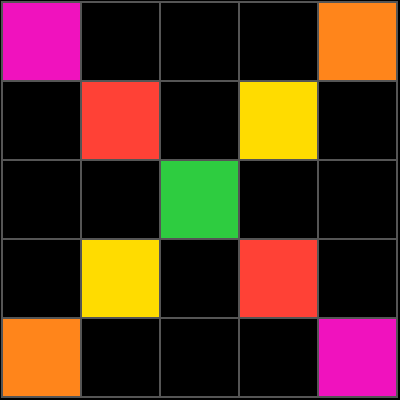} & \includegraphics[width=1.8cm,height=1.8cm]{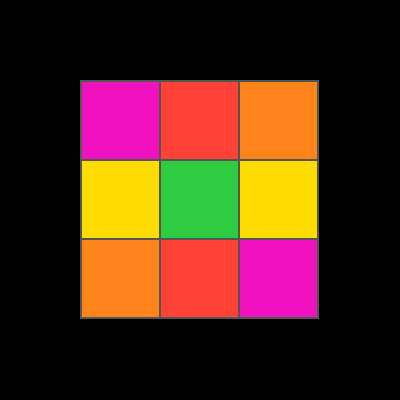} & \includegraphics[width=1.8cm,height=1.8cm]{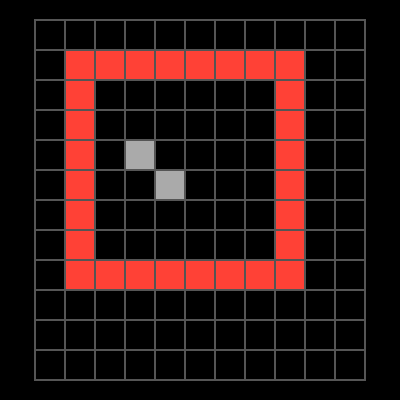} & \includegraphics[width=1.8cm,height=1.8cm]{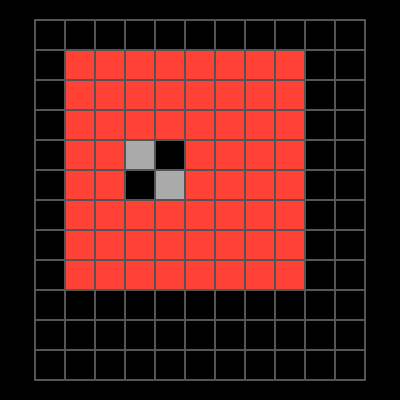} & \includegraphics[width=1.8cm,height=1.8cm]{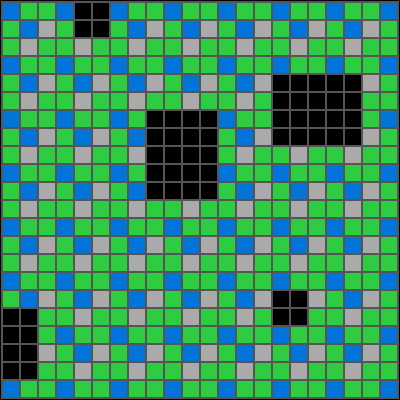} & \includegraphics[width=1.8cm,height=1.8cm]{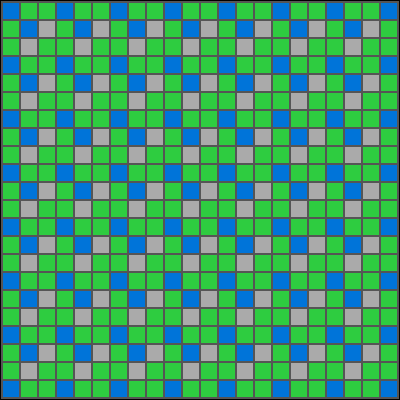} \\
& \includegraphics[width=1.8cm,height=1.8cm]{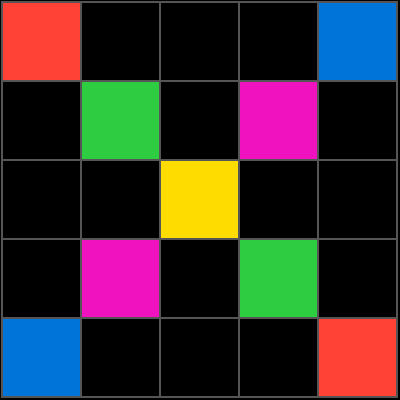} & \includegraphics[width=1.8cm,height=1.8cm]{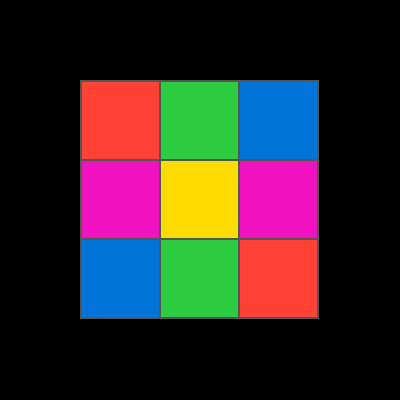} & \includegraphics[width=1.8cm,height=1.8cm]{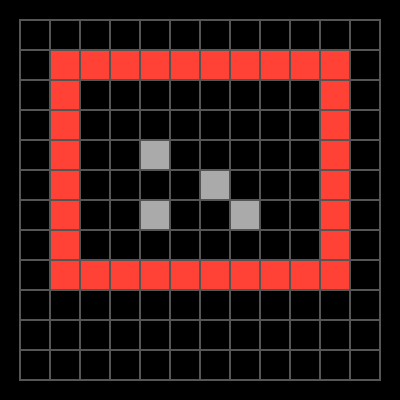} & \includegraphics[width=1.8cm,height=1.8cm]{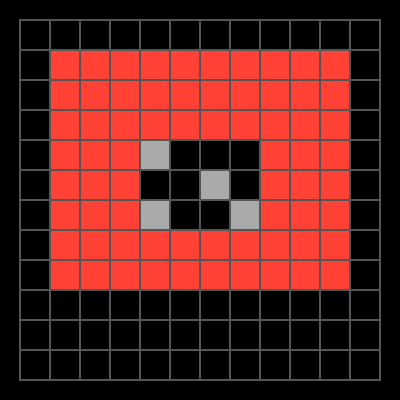} & \includegraphics[width=1.8cm,height=1.8cm]{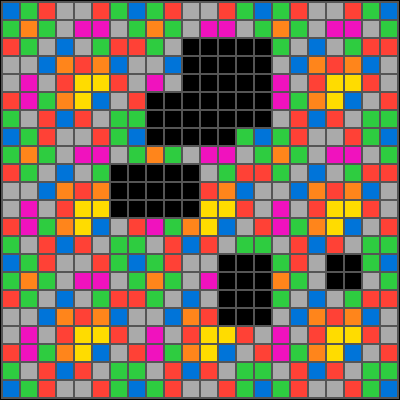} & \includegraphics[width=1.8cm,height=1.8cm]{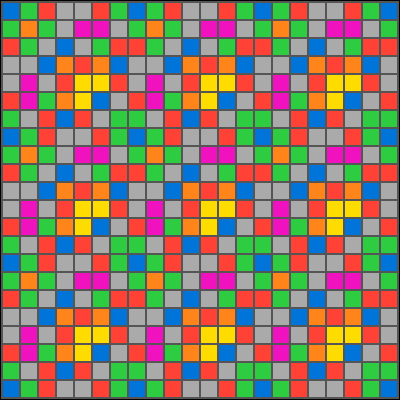} \\
\\[0.4em]
& \textbf{Input} & \textbf{Prediction} & \textbf{Input} & \textbf{Prediction} & \textbf{Input} & \textbf{Prediction} \\
\rotatebox{90}{\scriptsize\shortstack{\phantom{Q}\\\textbf{\vdmbestnt{CogVideoX1.5-5B}}}} & \includegraphics[width=1.8cm,height=1.8cm]{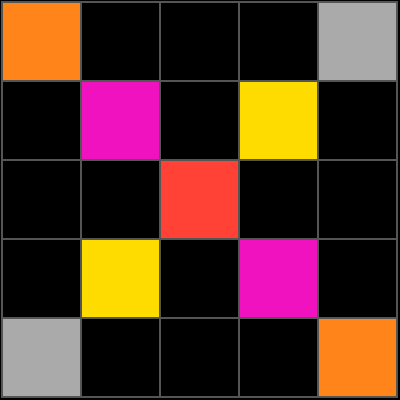} & \includegraphics[width=1.8cm,height=1.8cm]{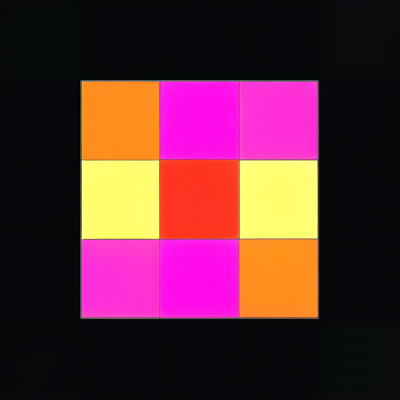} & \includegraphics[width=1.8cm,height=1.8cm]{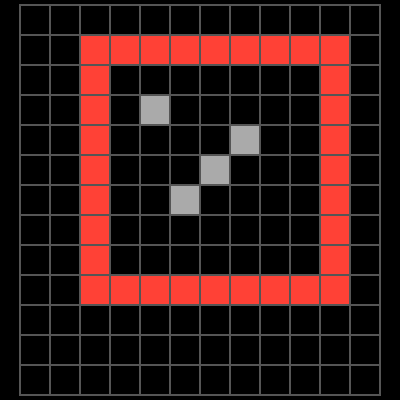} & \includegraphics[width=1.8cm,height=1.8cm]{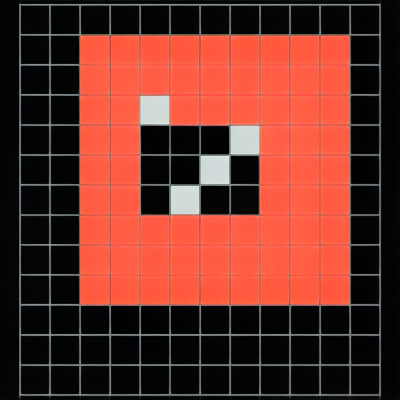} & \includegraphics[width=1.8cm,height=1.8cm]{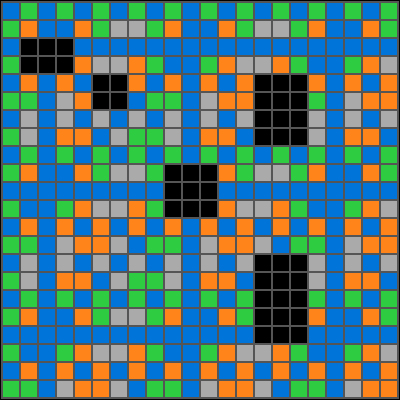} & \includegraphics[width=1.8cm,height=1.8cm]{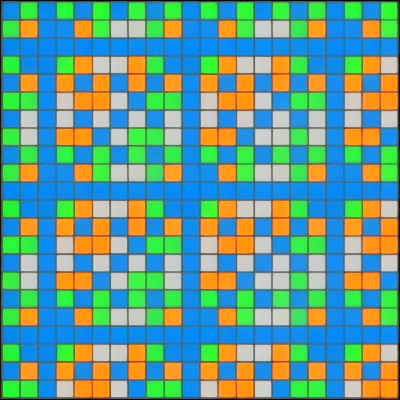} \\
\rotatebox{90}{\llmbestnt{\scriptsize\shortstack{\textbf{Qwen3-4B}\\\textbf{Instruct-2507}}}} & \includegraphics[width=1.8cm,height=1.8cm]{figures/appendix/arc-agi/llm-but-not-vdm/ca8de6ea/test_000_gt_in.png} & \includegraphics[width=1.8cm,height=1.8cm]{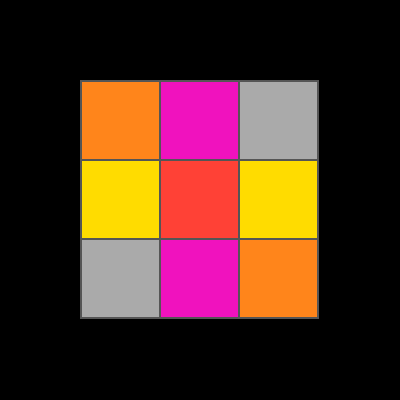} & \includegraphics[width=1.8cm,height=1.8cm]{figures/appendix/arc-agi/llm-but-not-vdm/d37a1ef5/test_000_gt_in.png} & \includegraphics[width=1.8cm,height=1.8cm]{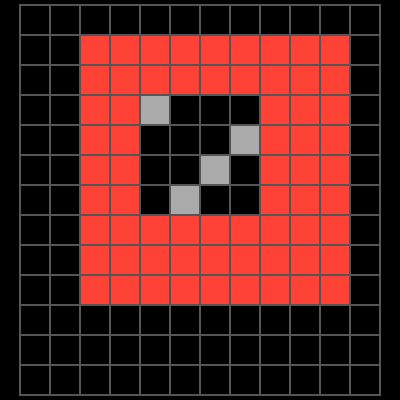} & \includegraphics[width=1.8cm,height=1.8cm]{figures/appendix/arc-agi/llm-but-not-vdm/e95e3d8e/test_000_gt_in.png} & \includegraphics[width=1.8cm,height=1.8cm]{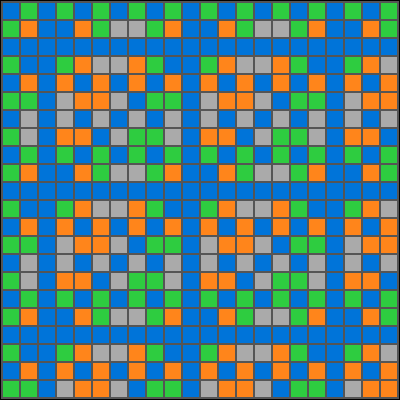} \\
\end{tabular}
\caption{Qualitative results on ARC-AGI for problems \textit{ca8de6ea}, \textit{d37a1ef5}, \textit{e95e3d8e}.}
\label{fig:arc-agi-ca8de6ea-d37a1ef5-e95e3d8e}
\end{figure}

To further illustrate the complementary strengths of VDMs and LLMs, we include qualitative examples of ARC-AGI tasks.  
In some cases, the LLM enables it to find the correct solution, while the VDM fails.  
Examples of this behavior is shown in Figure~\ref{fig:arc-agi-ca8de6ea-d37a1ef5-e95e3d8e}.  

\begin{figure}[htbp]
\centering
\renewcommand{\arraystretch}{0.6}
\setlength{\tabcolsep}{2pt}
\begin{tabular}{lcccccc}
& \textbf{Input} & \textbf{Output} & \textbf{Input} & \textbf{Output} & \textbf{Input} & \textbf{Output} \\[0.2em]
\multirow{3}{*}{\rotatebox{90}{\textbf{Training Examples}}} & \includegraphics[width=1.8cm,height=1.8cm]{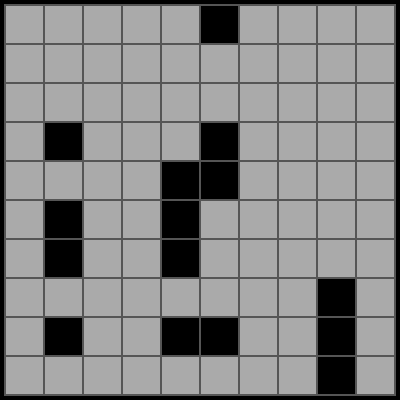} & \includegraphics[width=1.8cm,height=1.8cm]{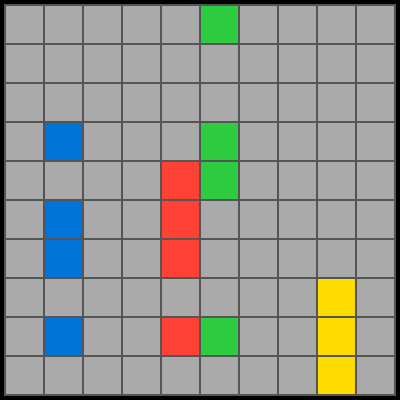} & \includegraphics[width=1.8cm,height=1.8cm]{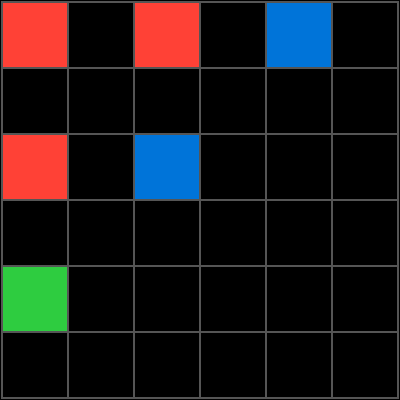} & \includegraphics[width=1.8cm,height=1.8cm]{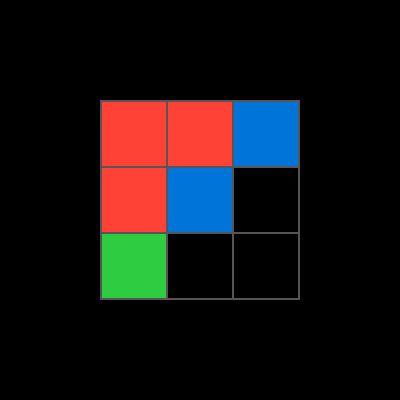} & \includegraphics[width=1.8cm,height=1.8cm]{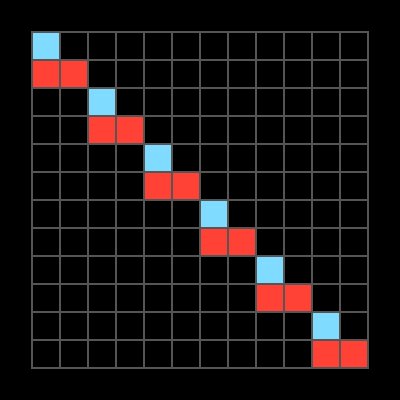} & \includegraphics[width=1.8cm,height=1.8cm]{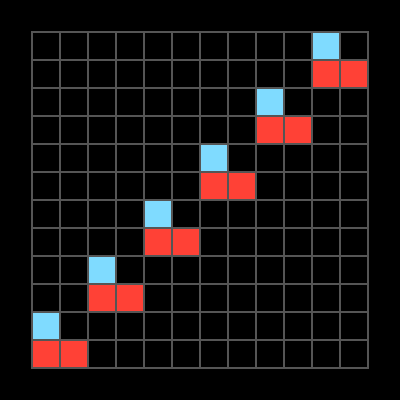} \\
& \includegraphics[width=1.8cm,height=1.8cm]{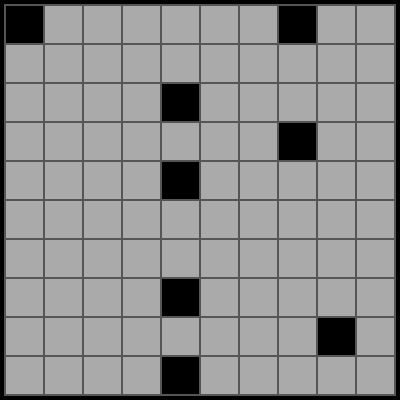} & \includegraphics[width=1.8cm,height=1.8cm]{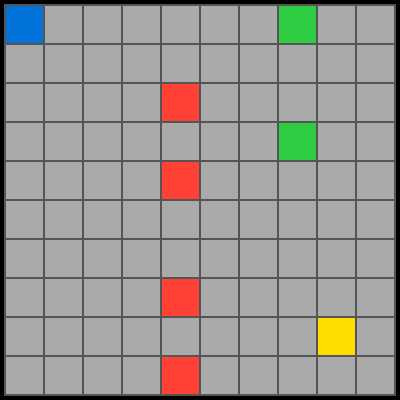} & \includegraphics[width=1.8cm,height=1.8cm]{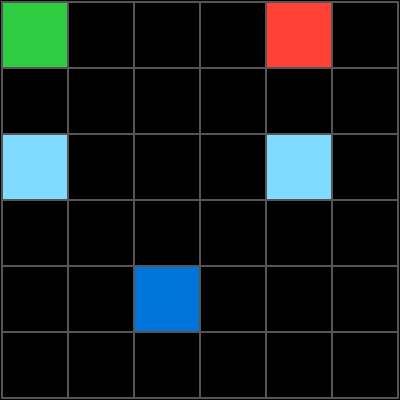} & \includegraphics[width=1.8cm,height=1.8cm]{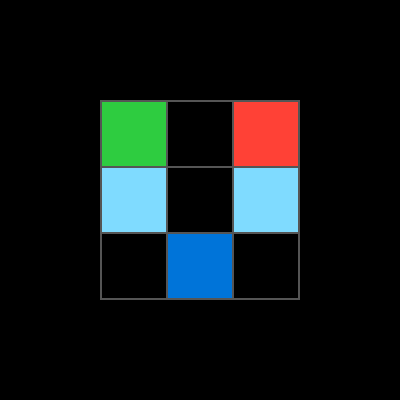} & \includegraphics[width=1.8cm,height=1.8cm]{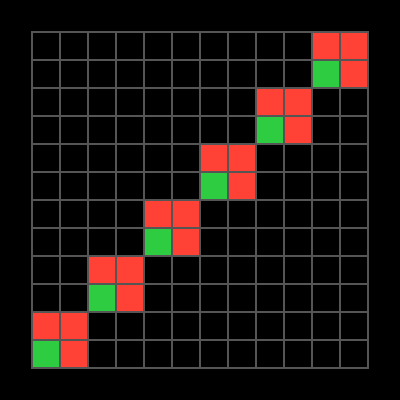} & \includegraphics[width=1.8cm,height=1.8cm]{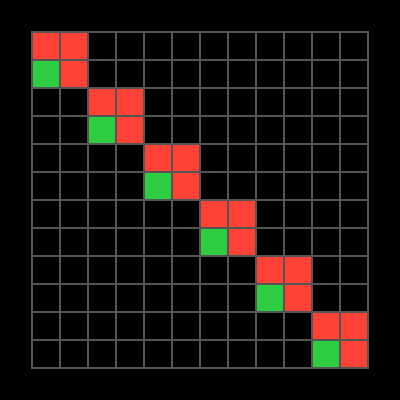} \\
& \includegraphics[width=1.8cm,height=1.8cm]{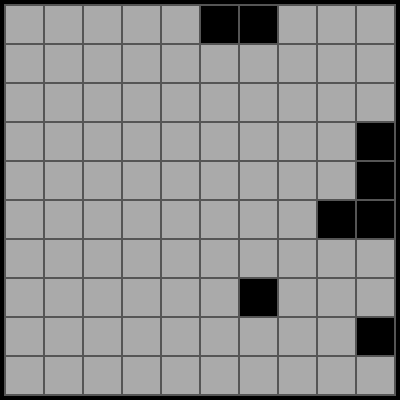} & \includegraphics[width=1.8cm,height=1.8cm]{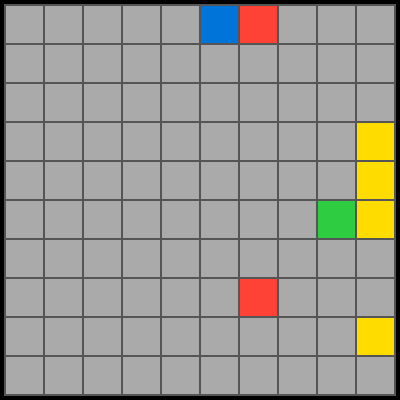} & \includegraphics[width=1.8cm,height=1.8cm]{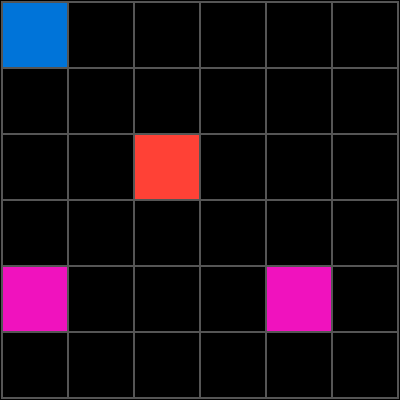} & \includegraphics[width=1.8cm,height=1.8cm]{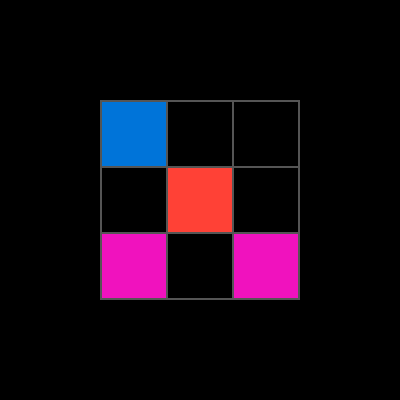} & \includegraphics[width=1.8cm,height=1.8cm]{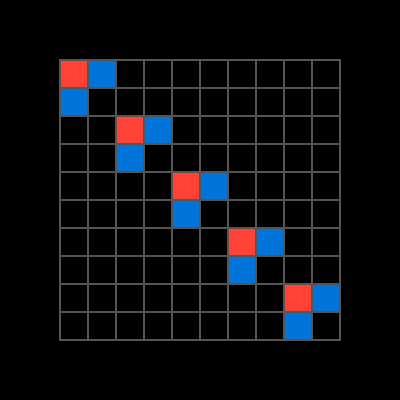} & \includegraphics[width=1.8cm,height=1.8cm]{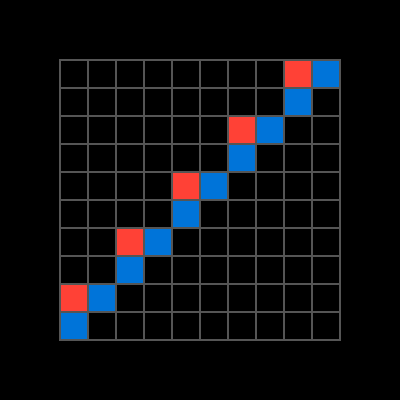} \\
\\[0.4em]
& \textbf{Input} & \textbf{Prediction} & \textbf{Input} & \textbf{Prediction} & \textbf{Input} & \textbf{Prediction} \\
\rotatebox{90}{\scriptsize\shortstack{\phantom{Q}\\\textbf{\vdmbestnt{CogVideoX1.5-5B}}}} & \includegraphics[width=1.8cm,height=1.8cm]{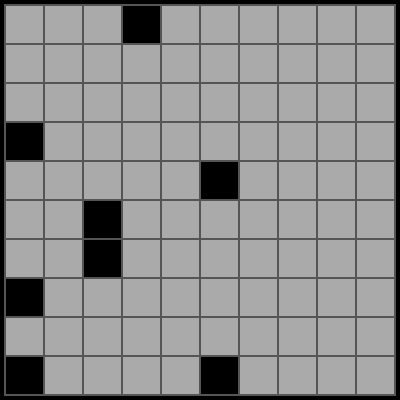} & \includegraphics[width=1.8cm,height=1.8cm]{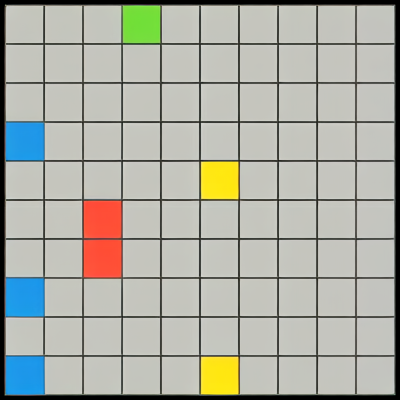} & \includegraphics[width=1.8cm,height=1.8cm]{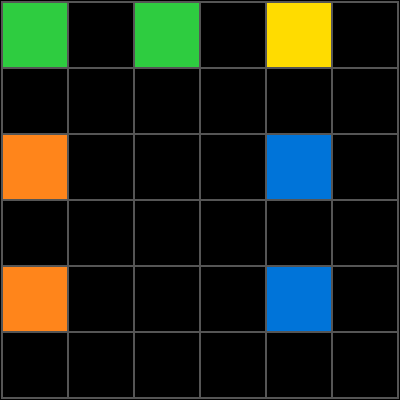} & \includegraphics[width=1.8cm,height=1.8cm]{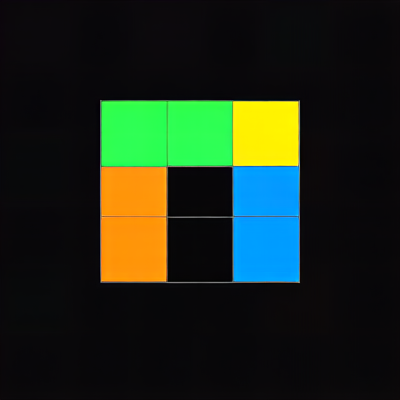} & \includegraphics[width=1.8cm,height=1.8cm]{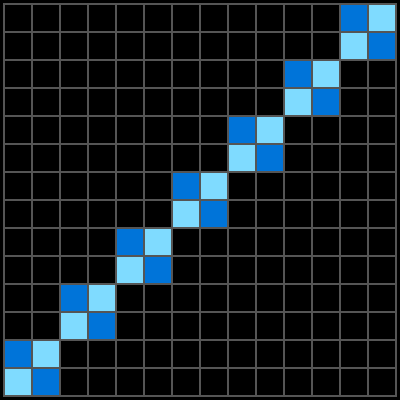} & \includegraphics[width=1.8cm,height=1.8cm]{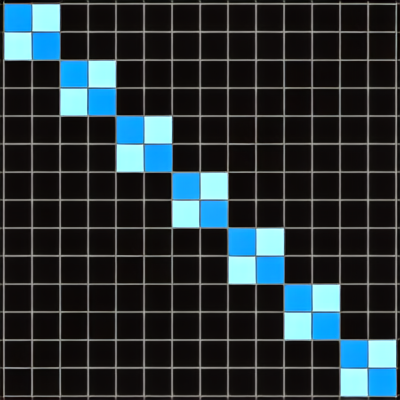} \\
\rotatebox{90}{\llmbestnt{\scriptsize\shortstack{\textbf{Qwen3-4B}\\\textbf{Instruct-2507}}}} & \includegraphics[width=1.8cm,height=1.8cm]{figures/appendix/arc-agi/both/575b1a71/test_000_gt_in.png} & \includegraphics[width=1.8cm,height=1.8cm]{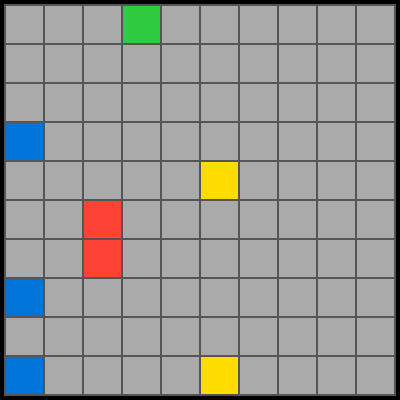} & \includegraphics[width=1.8cm,height=1.8cm]{figures/appendix/arc-agi/both/68b67ca3/test_000_gt_in.png} & \includegraphics[width=1.8cm,height=1.8cm]{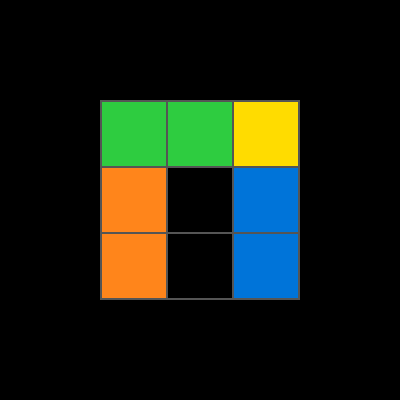} & \includegraphics[width=1.8cm,height=1.8cm]{figures/appendix/arc-agi/both/8ee62060/test_000_gt_in.png} & \includegraphics[width=1.8cm,height=1.8cm]{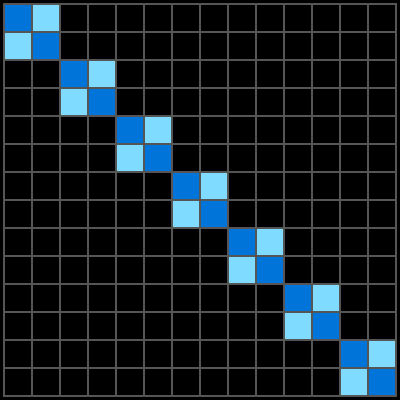} \\
\end{tabular}
\caption{Qualitative results on ARC-AGI for problems \textit{575b1a71}, \textit{68b67ca3}, \textit{8ee62060}.}
\label{fig:arc-agi-575b1a71-68b67ca3-8ee62060-both}
\end{figure}

In contrast, there are tasks where both models succeed, suggesting that the underlying structure can be captured through either symbolic reasoning or visual pattern learning.  
One such case is given in Figure~\ref{fig:arc-agi-575b1a71-68b67ca3-8ee62060-both}.  

\begin{figure}[htbp]
\centering
\renewcommand{\arraystretch}{0.6}
\setlength{\tabcolsep}{2pt}
\begin{tabular}{lcccccc}
& \textbf{Input} & \textbf{Output} & \textbf{Input} & \textbf{Output} & \textbf{Input} & \textbf{Output} \\[0.2em]
\multirow{3}{*}{\rotatebox{90}{\textbf{Training Examples}}} & \includegraphics[width=1.8cm,height=1.8cm]{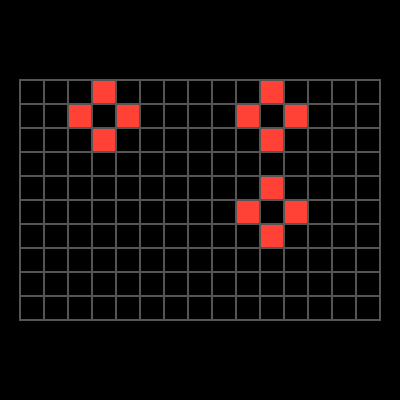} & \includegraphics[width=1.8cm,height=1.8cm]{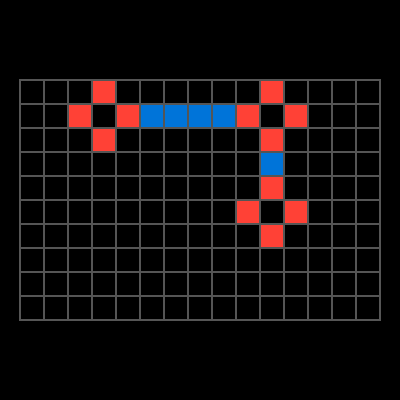} & \includegraphics[width=1.8cm,height=1.8cm]{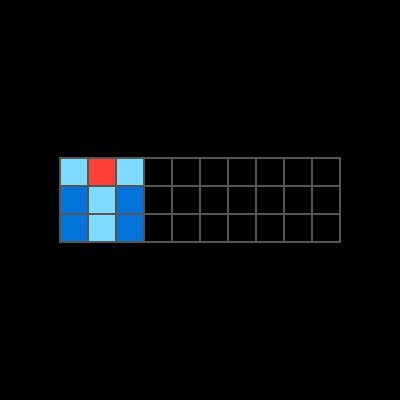} & \includegraphics[width=1.8cm,height=1.8cm]{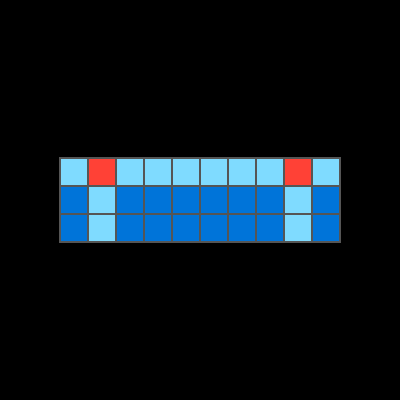} & \includegraphics[width=1.8cm,height=1.8cm]{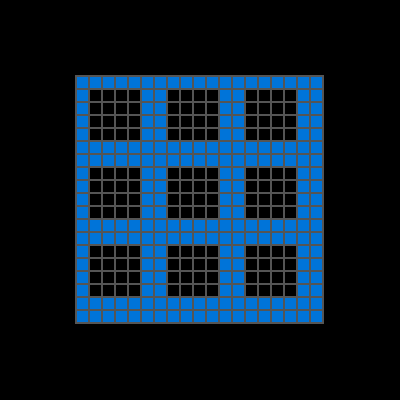} & \includegraphics[width=1.8cm,height=1.8cm]{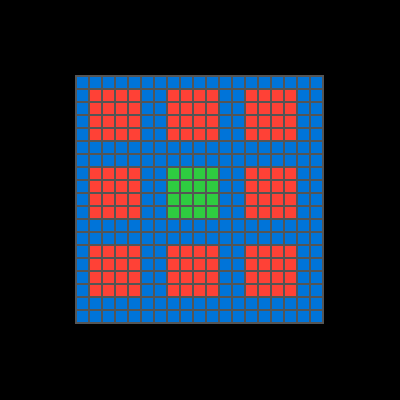} \\
& \includegraphics[width=1.8cm,height=1.8cm]{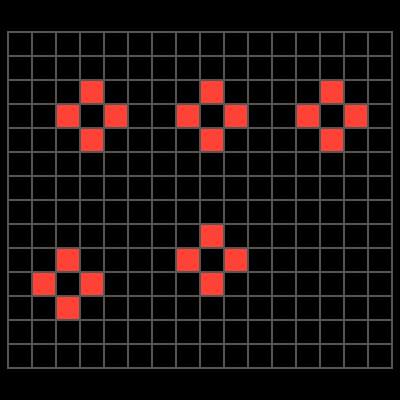} & \includegraphics[width=1.8cm,height=1.8cm]{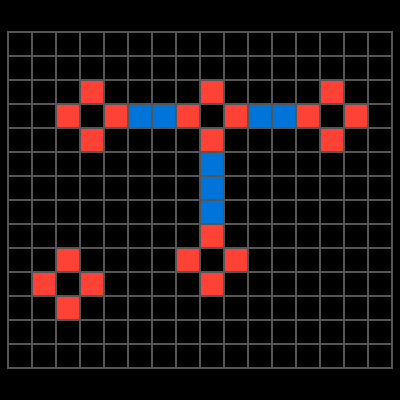} & \includegraphics[width=1.8cm,height=1.8cm]{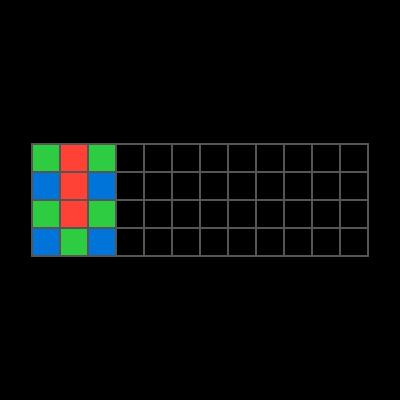} & \includegraphics[width=1.8cm,height=1.8cm]{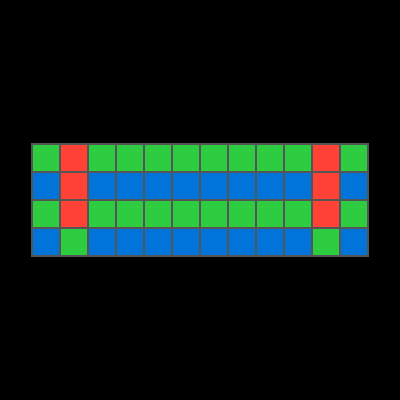} & \includegraphics[width=1.8cm,height=1.8cm]{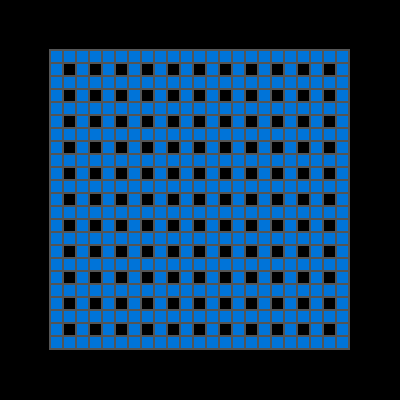} & \includegraphics[width=1.8cm,height=1.8cm]{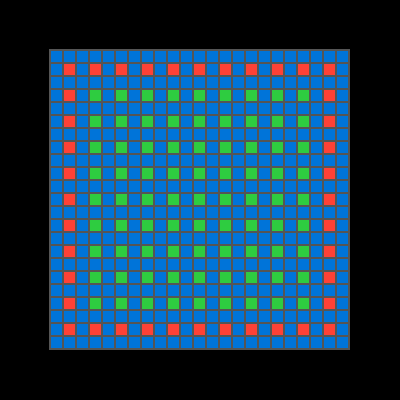} \\
& \includegraphics[width=1.8cm,height=1.8cm]{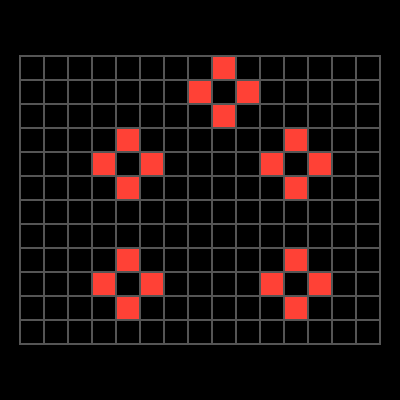} & \includegraphics[width=1.8cm,height=1.8cm]{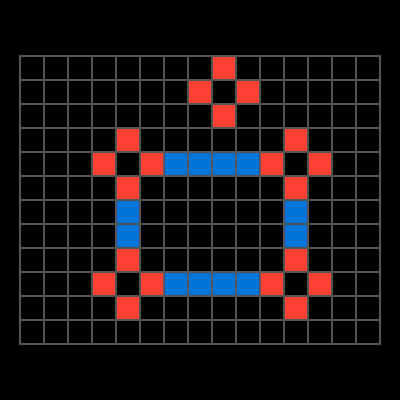} & \includegraphics[width=1.8cm,height=1.8cm]{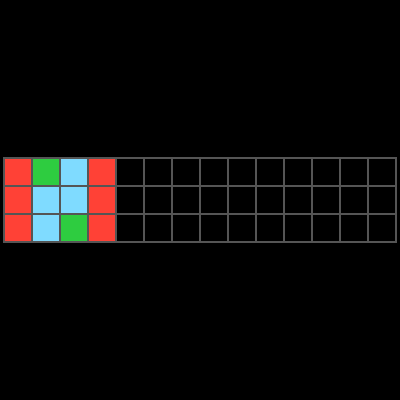} & \includegraphics[width=1.8cm,height=1.8cm]{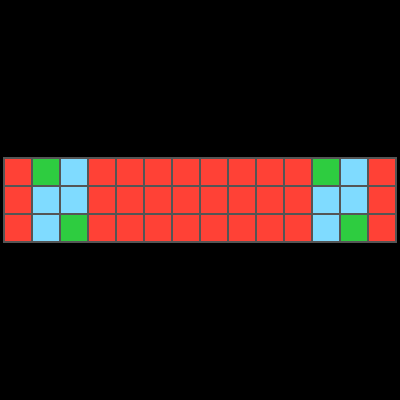} & \includegraphics[width=1.8cm,height=1.8cm]{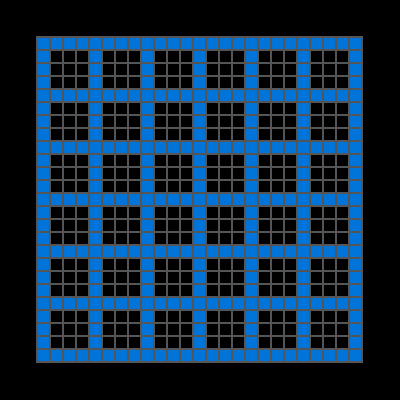} & \includegraphics[width=1.8cm,height=1.8cm]{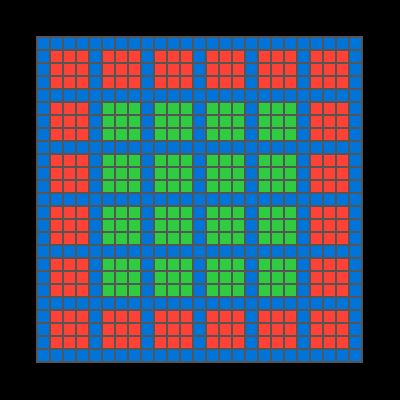} \\
\\[0.4em]
& \textbf{Input} & \textbf{Prediction} & \textbf{Input} & \textbf{Prediction} & \textbf{Input} & \textbf{Prediction} \\
\rotatebox{90}{\scriptsize\shortstack{\phantom{Q}\\\textbf{\vdmbestnt{CogVideoX1.5-5B}}}} & \includegraphics[width=1.8cm,height=1.8cm]{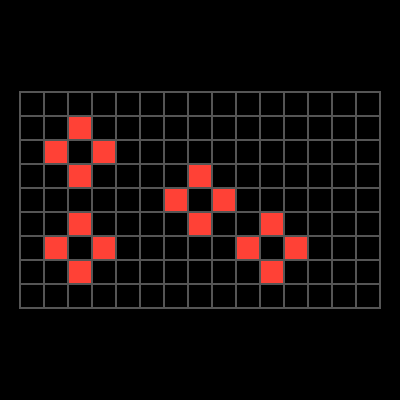} & \includegraphics[width=1.8cm,height=1.8cm]{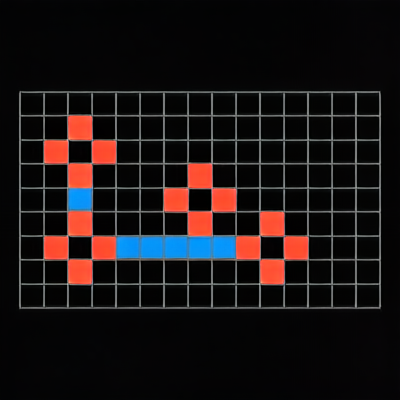} & \includegraphics[width=1.8cm,height=1.8cm]{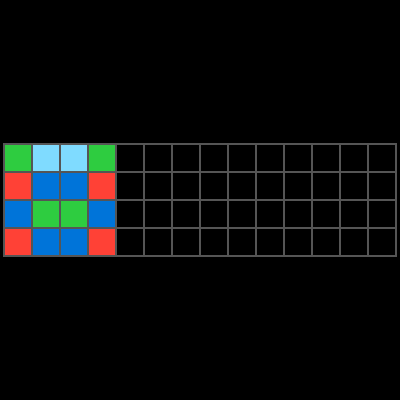} & \includegraphics[width=1.8cm,height=1.8cm]{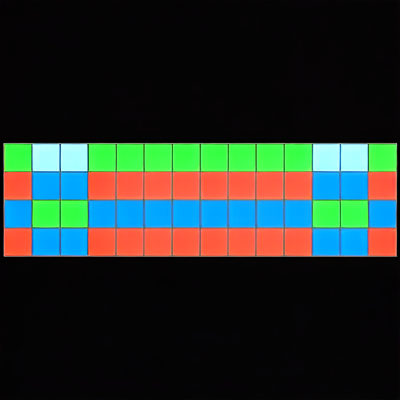} & \includegraphics[width=1.8cm,height=1.8cm]{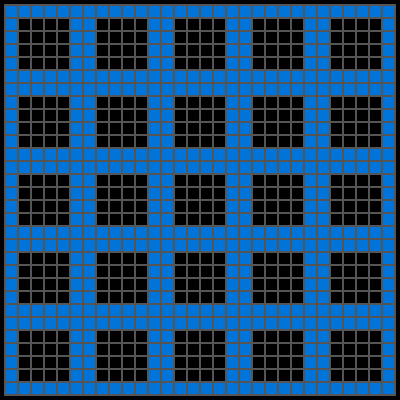} & \includegraphics[width=1.8cm,height=1.8cm]{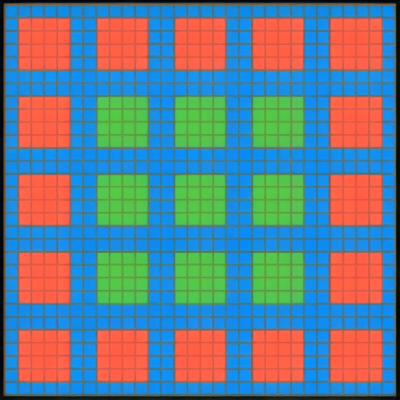} \\
\rotatebox{90}{\llmbestnt{\scriptsize\shortstack{\textbf{Qwen3-4B}\\\textbf{Instruct-2507}}}} & \includegraphics[width=1.8cm,height=1.8cm]{figures/appendix/arc-agi/vdm-but-not-llm/60a26a3e/test_000_gt_in.png} & \includegraphics[width=1.8cm,height=1.8cm]{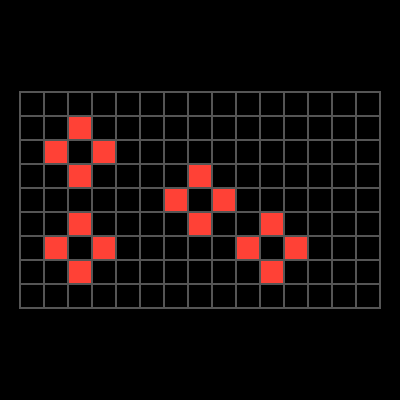} & \includegraphics[width=1.8cm,height=1.8cm]{figures/appendix/arc-agi/vdm-but-not-llm/62b74c02/test_000_gt_in.png} & \includegraphics[width=1.8cm,height=1.8cm]{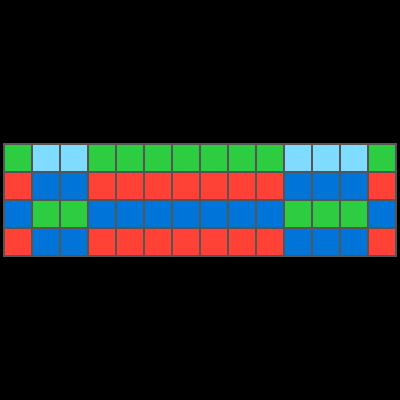} & \includegraphics[width=1.8cm,height=1.8cm]{figures/appendix/arc-agi/vdm-but-not-llm/8a371977/test_000_gt_in.png} & \includegraphics[width=1.8cm,height=1.8cm]{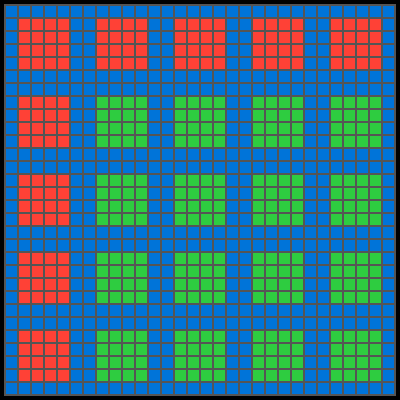} \\
\end{tabular}
\caption{Qualitative results on ARC-AGI for problems \textit{60a26a3e}, \textit{62b74c02}, \textit{8a371977}.}
\label{fig:arc-agi-60a26a3e-62b74c02-8a371977-only-vdm}
\end{figure}

\begin{figure}[htbp]
\centering
\renewcommand{\arraystretch}{0.6}
\setlength{\tabcolsep}{2pt}
\begin{tabular}{lcccccc}
& \textbf{Input} & \textbf{Output} & \textbf{Input} & \textbf{Output} & \textbf{Input} & \textbf{Output} \\[0.2em]
\multirow{3}{*}{\rotatebox{90}{\textbf{Training Examples}}} & \includegraphics[width=1.8cm,height=1.8cm]{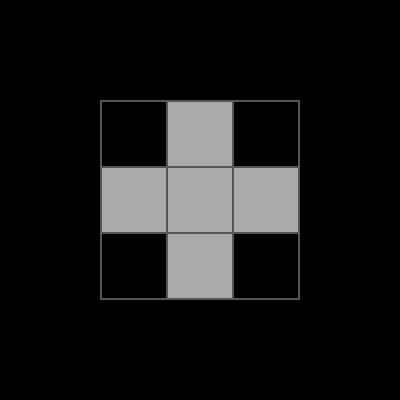} & \includegraphics[width=1.8cm,height=1.8cm]{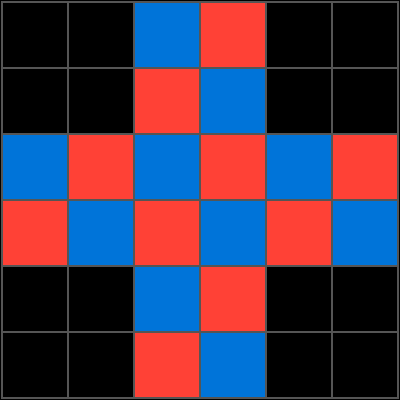} & \includegraphics[width=1.8cm,height=1.8cm]{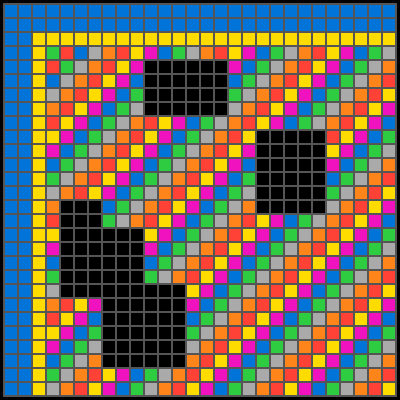} & \includegraphics[width=1.8cm,height=1.8cm]{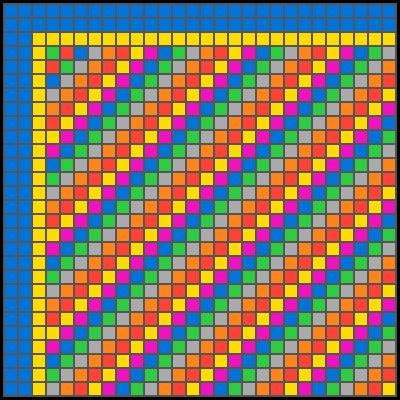} & \includegraphics[width=1.8cm,height=1.8cm]{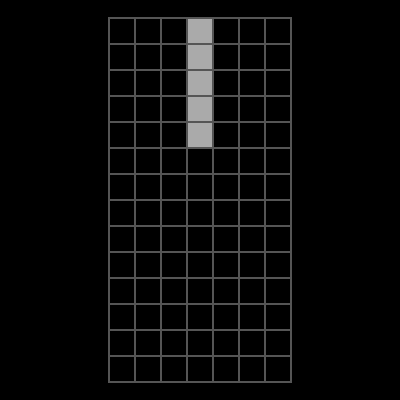} & \includegraphics[width=1.8cm,height=1.8cm]{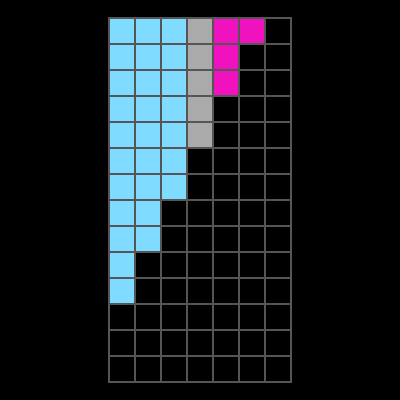} \\
& \includegraphics[width=1.8cm,height=1.8cm]{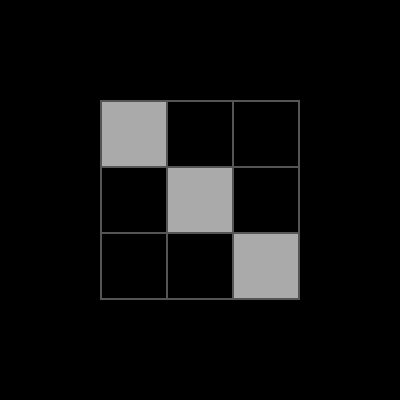} & \includegraphics[width=1.8cm,height=1.8cm]{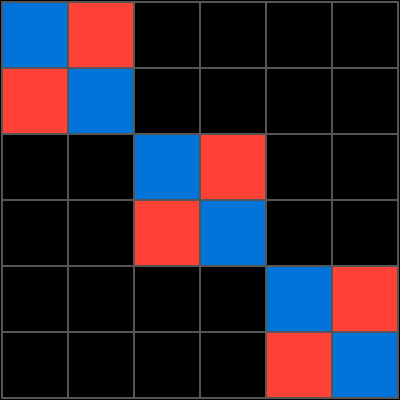} & \includegraphics[width=1.8cm,height=1.8cm]{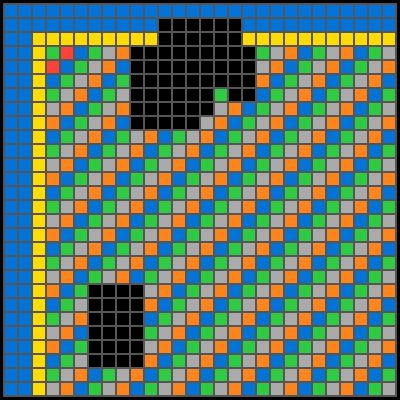} & \includegraphics[width=1.8cm,height=1.8cm]{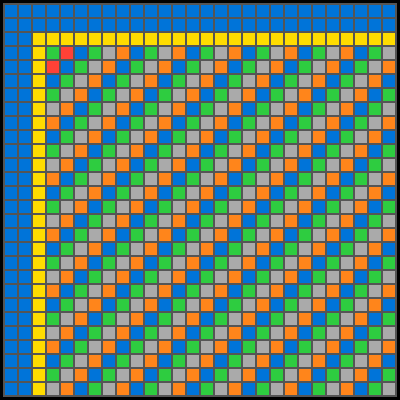} & \includegraphics[width=1.8cm,height=1.8cm]{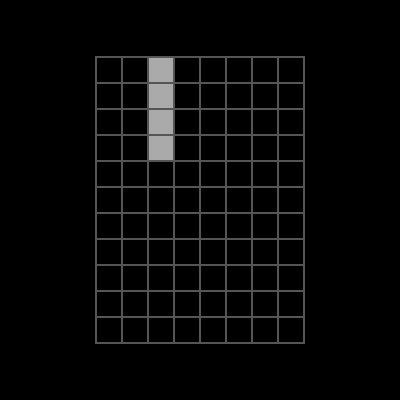} & \includegraphics[width=1.8cm,height=1.8cm]{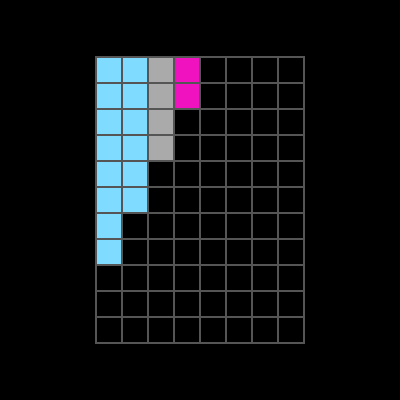} \\
& \includegraphics[width=1.8cm,height=1.8cm]{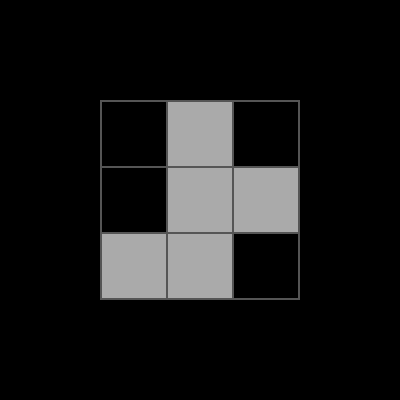} & \includegraphics[width=1.8cm,height=1.8cm]{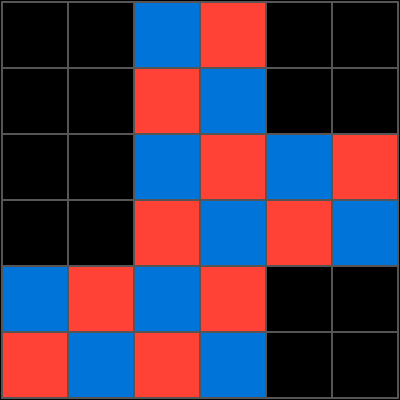} & \includegraphics[width=1.8cm,height=1.8cm]{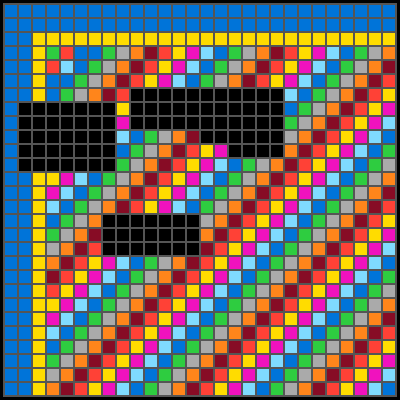} & \includegraphics[width=1.8cm,height=1.8cm]{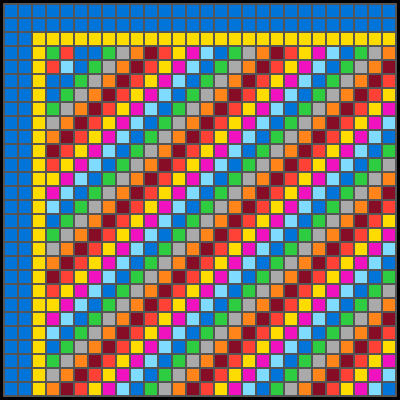} & \includegraphics[width=1.8cm,height=1.8cm]{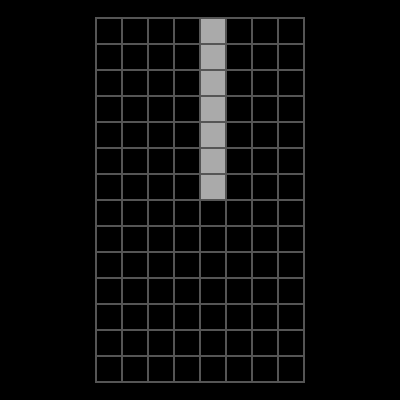} & \includegraphics[width=1.8cm,height=1.8cm]{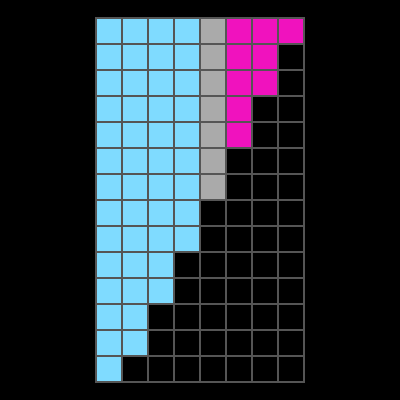} \\
\\[0.4em]
& \textbf{Input} & \textbf{Prediction} & \textbf{Input} & \textbf{Prediction} & \textbf{Input} & \textbf{Prediction} \\
\rotatebox{90}{\scriptsize\shortstack{\phantom{Q}\\\textbf{\vdmbestnt{CogVideoX1.5-5B}}}} & \includegraphics[width=1.8cm,height=1.8cm]{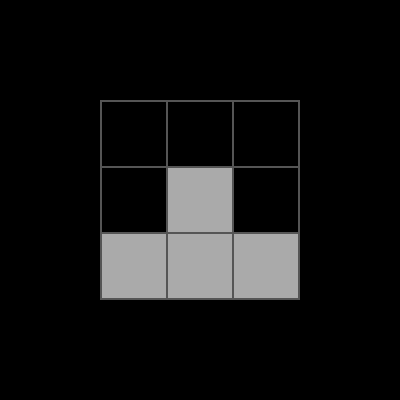} & \includegraphics[width=1.8cm,height=1.8cm]{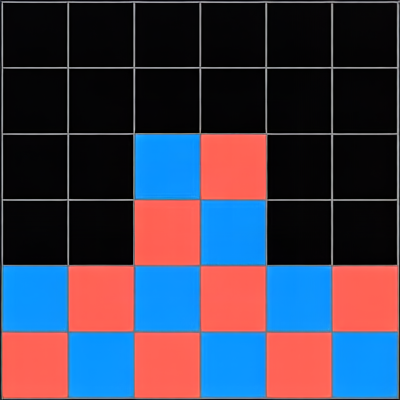} & \includegraphics[width=1.8cm,height=1.8cm]{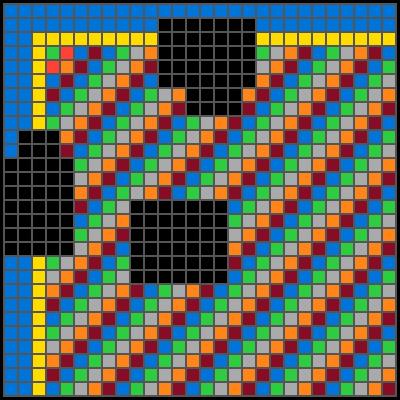} & \includegraphics[width=1.8cm,height=1.8cm]{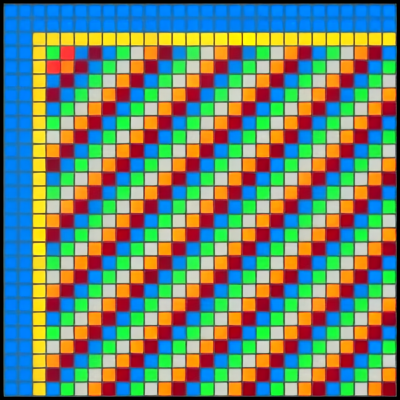} & \includegraphics[width=1.8cm,height=1.8cm]{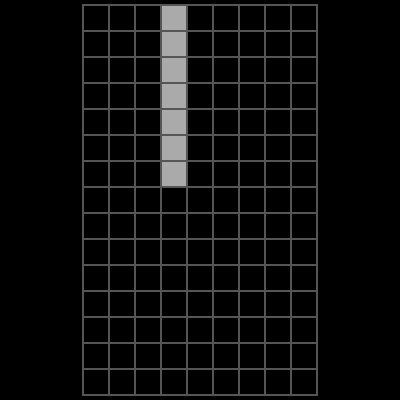} & \includegraphics[width=1.8cm,height=1.8cm]{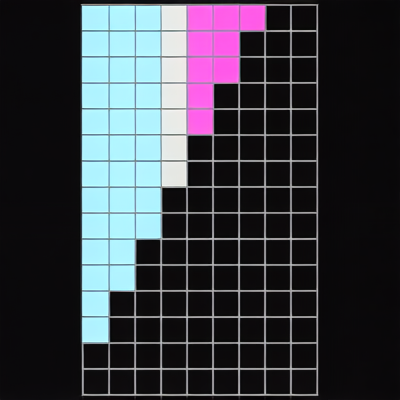} \\
\rotatebox{90}{\llmbestnt{\scriptsize\shortstack{\textbf{Qwen3-4B}\\\textbf{Instruct-2507}}}} & \includegraphics[width=1.8cm,height=1.8cm]{figures/appendix/arc-agi/vdm-but-not-llm/2072aba6/test_000_gt_in.png} & \includegraphics[width=1.8cm,height=1.8cm]{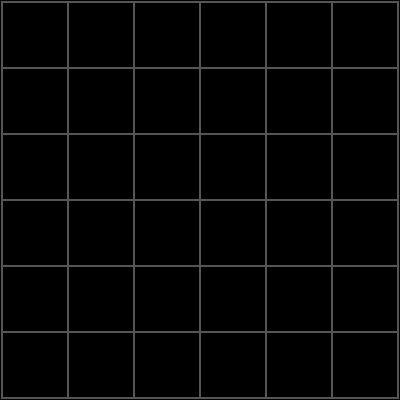} & \includegraphics[width=1.8cm,height=1.8cm]{figures/appendix/arc-agi/vdm-but-not-llm/4aab4007/test_000_gt_in.png} & \includegraphics[width=1.8cm,height=1.8cm]{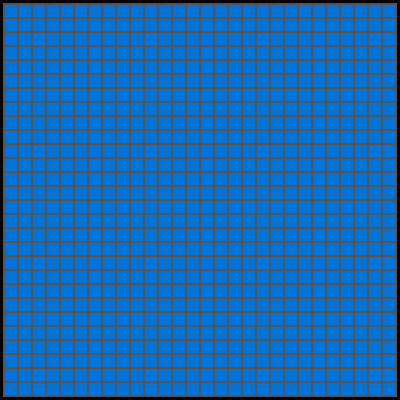} & \includegraphics[width=1.8cm,height=1.8cm]{figures/appendix/arc-agi/vdm-but-not-llm/5207a7b5/test_000_gt_in.png} & \includegraphics[width=1.8cm,height=1.8cm]{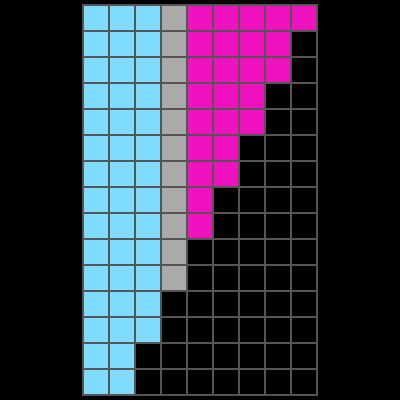} \\
\end{tabular}
\caption{Qualitative results on ARC-AGI for problems \textit{2072aba6}, \textit{4aab4007}, \textit{5207a7b5}.}
\label{fig:arc-agi-2072aba6-4aab4007-5207a7b5-only-vdm}
\end{figure}

Finally, we highlight situations where only the VDM solves the task correctly (Figures~\ref{fig:arc-agi-60a26a3e-62b74c02-8a371977-only-vdm} and \ref{fig:arc-agi-2072aba6-4aab4007-5207a7b5-only-vdm}).  
These examples emphasize how visual inductive biases allow the VDM to generalize in settings where symbolic reasoning alone appears insufficient.

\subsection{Structured Visual Tasks}
We include additional qualitative examples from structured visual tasks such as mazes, route planning, and cellular automata, complementing the quantitative results in the main text.

\begin{figure}[htbp]
\centering
\begin{tabular}{ccc}
    \textbf{Input} & \vdmbestnt{\textbf{CogVideoX1.5-5B}} & \llmbestnt{\shortstack{\textbf{Qwen3-4B}\\\textbf{Instruct-2507}}} \\[0.3em]
    
    \includegraphics[width=2.5cm,height=2.5cm]{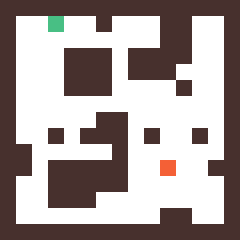} &
    \includegraphics[width=2.5cm,height=2.5cm]{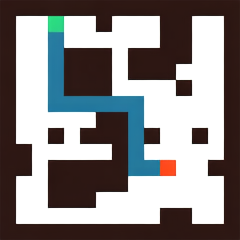} &
    \includegraphics[width=2.5cm,height=2.5cm]{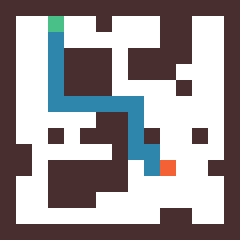} \\[0.1em]
    
    \includegraphics[width=2.5cm,height=2.5cm]{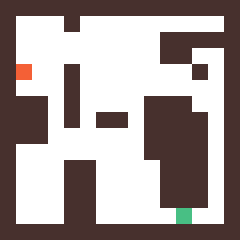} &
    \includegraphics[width=2.5cm,height=2.5cm]{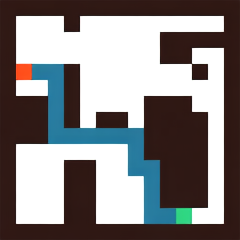} &
    \includegraphics[width=2.5cm,height=2.5cm]{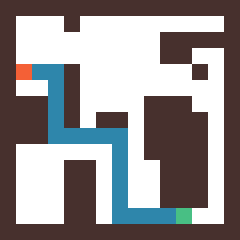} \\[0.1em]
    
    \includegraphics[width=2.5cm,height=2.5cm]{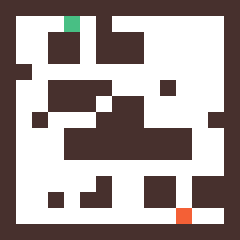} &
    \includegraphics[width=2.5cm,height=2.5cm]{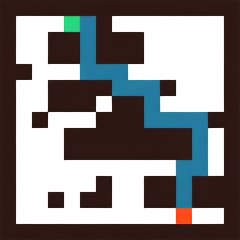} &
    \includegraphics[width=2.5cm,height=2.5cm]{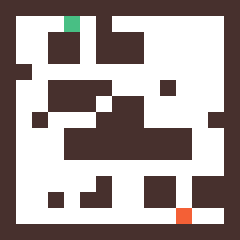} \\[0.1em]
    
    \includegraphics[width=2.5cm,height=2.5cm]{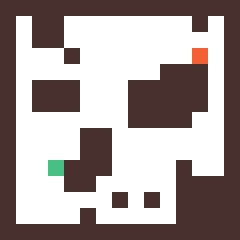} &
    \includegraphics[width=2.5cm,height=2.5cm]{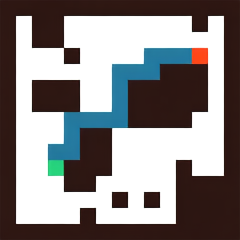} &
    \includegraphics[width=2.5cm,height=2.5cm]{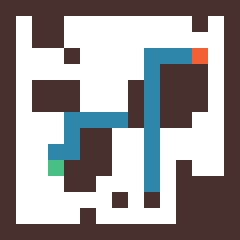} \\[0.1em]
    
    \includegraphics[width=2.5cm,height=2.5cm]{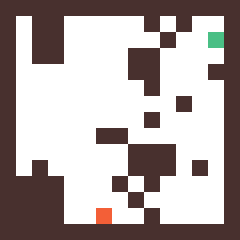} &
    \includegraphics[width=2.5cm,height=2.5cm]{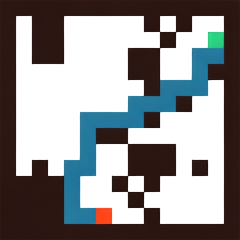} &
    \includegraphics[width=2.5cm,height=2.5cm]{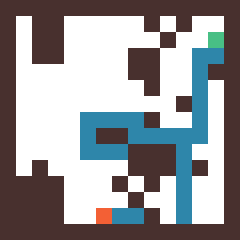} \\[0.1em]
    
    \includegraphics[width=2.5cm,height=2.5cm]{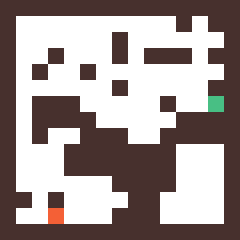} &
    \includegraphics[width=2.5cm,height=2.5cm]{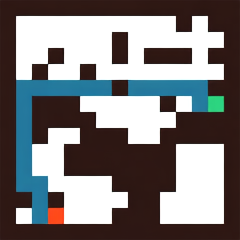} &
    \includegraphics[width=2.5cm,height=2.5cm]{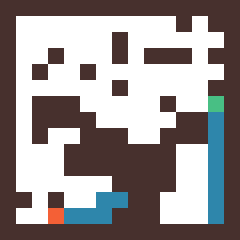} \\[0.1em]
    
    \includegraphics[width=2.5cm,height=2.5cm]{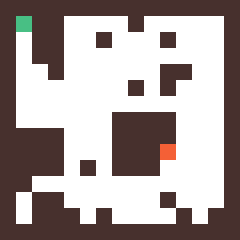} &
    \includegraphics[width=2.5cm,height=2.5cm]{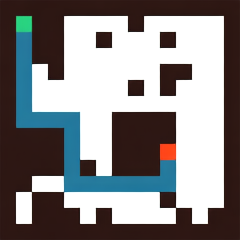} &
    \includegraphics[width=2.5cm,height=2.5cm]{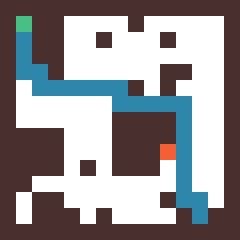} \\[0.1em]
    
    \includegraphics[width=2.5cm,height=2.5cm]{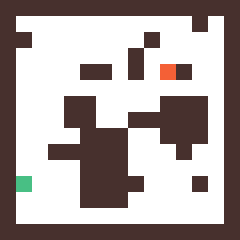} &
    \includegraphics[width=2.5cm,height=2.5cm]{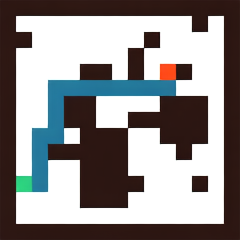} &
    \includegraphics[width=2.5cm,height=2.5cm]{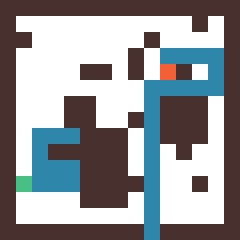} \\
\end{tabular}

\caption{Representative examples for the \textit{Shortest Path} task, showing ground truth inputs (left) and model predictions (center and right) after finetuning with \(n=300\) samples.}
\end{figure}

\begin{figure}[htbp]
\centering
\begin{tabular}{cccc}
    \textbf{Input} & \textbf{Output} & \vdmbestnt{\textbf{CogVideoX1.5-5B}} & \llmbestnt{\shortstack{\textbf{Qwen3-4B}\\\textbf{Instruct-2507}}} \\[0.3em]
    
    \includegraphics[width=2.5cm,height=2.5cm]{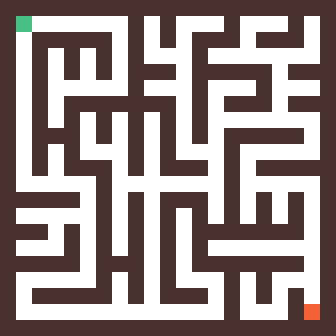} &
    \includegraphics[width=2.5cm,height=2.5cm]{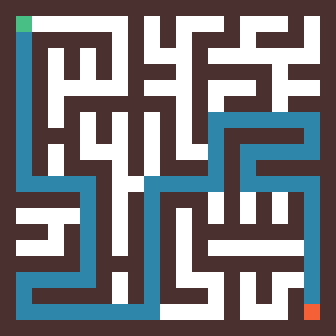} &
    \includegraphics[width=2.5cm,height=2.5cm]{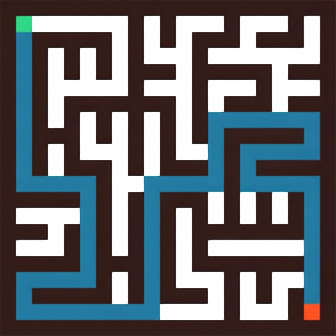} &
    \includegraphics[width=2.5cm,height=2.5cm]{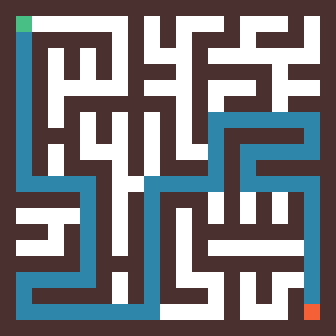} \\[0.1em]
    
    \includegraphics[width=2.5cm,height=2.5cm]{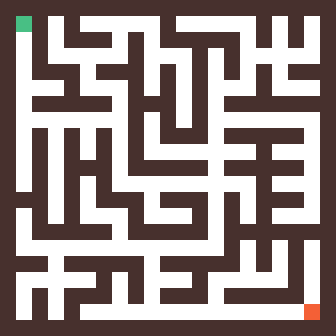} &
    \includegraphics[width=2.5cm,height=2.5cm]{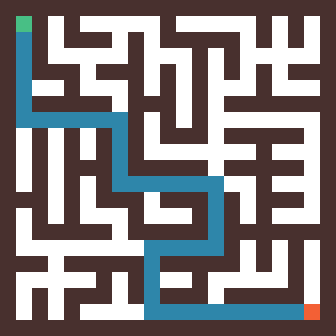} &
    \includegraphics[width=2.5cm,height=2.5cm]{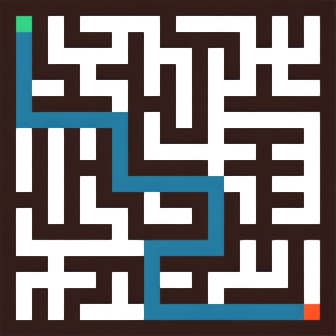} &
    \includegraphics[width=2.5cm,height=2.5cm]{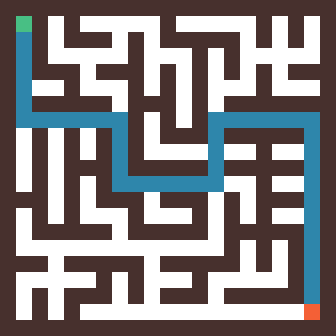} \\[0.1em]
    
    \includegraphics[width=2.5cm,height=2.5cm]{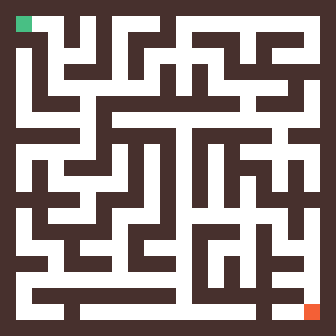} &
    \includegraphics[width=2.5cm,height=2.5cm]{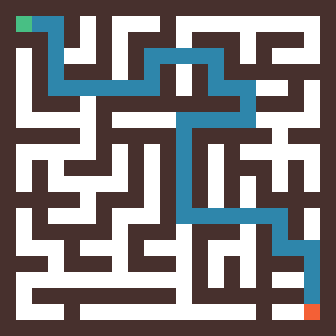} &
    \includegraphics[width=2.5cm,height=2.5cm]{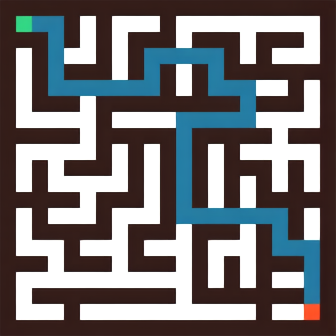} &
    \includegraphics[width=2.5cm,height=2.5cm]{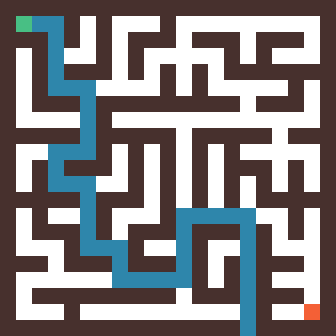} \\[0.1em]
    
    \includegraphics[width=2.5cm,height=2.5cm]{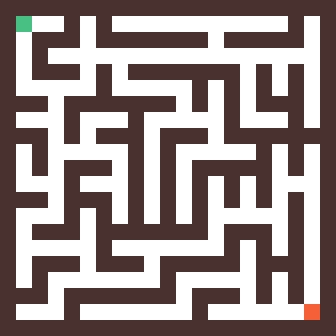} &
    \includegraphics[width=2.5cm,height=2.5cm]{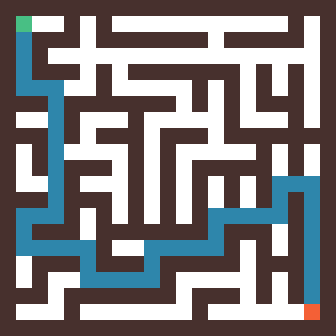} &
    \includegraphics[width=2.5cm,height=2.5cm]{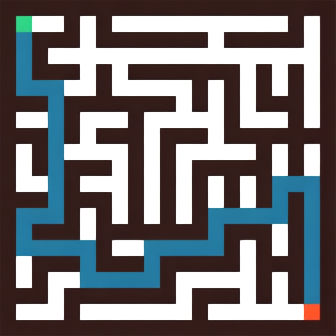} &
    \includegraphics[width=2.5cm,height=2.5cm]{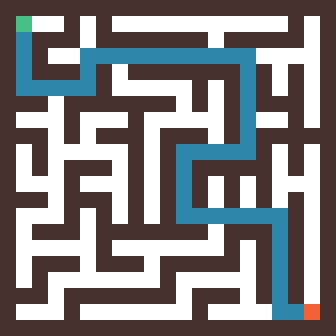} \\[0.1em]
    
    \includegraphics[width=2.5cm,height=2.5cm]{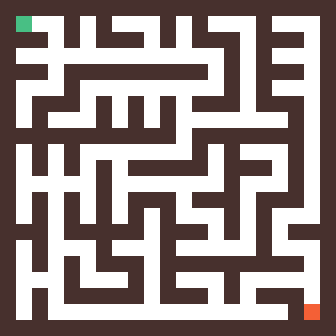} &
    \includegraphics[width=2.5cm,height=2.5cm]{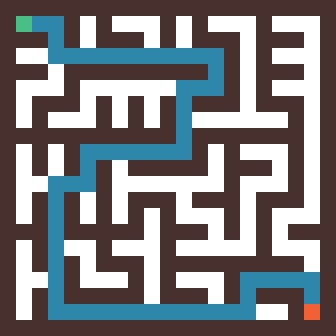} &
    \includegraphics[width=2.5cm,height=2.5cm]{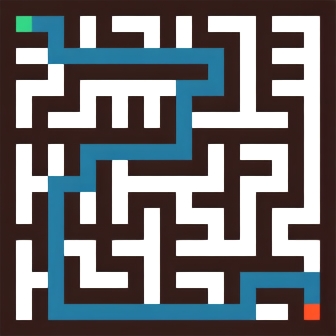} &
    \includegraphics[width=2.5cm,height=2.5cm]{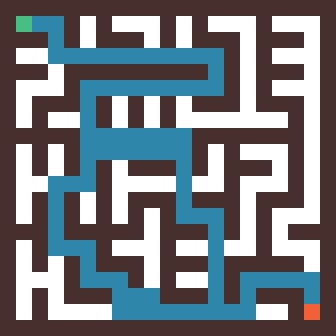} \\[0.1em]
    
    \includegraphics[width=2.5cm,height=2.5cm]{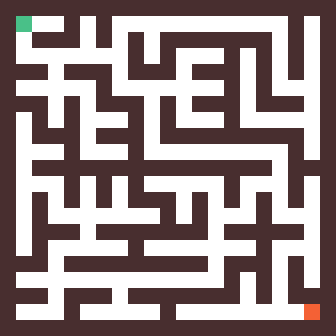} &
    \includegraphics[width=2.5cm,height=2.5cm]{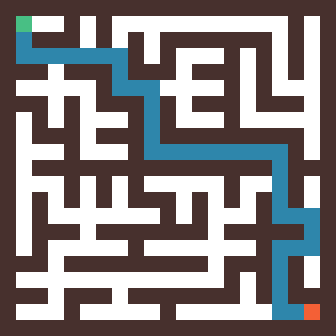} &
    \includegraphics[width=2.5cm,height=2.5cm]{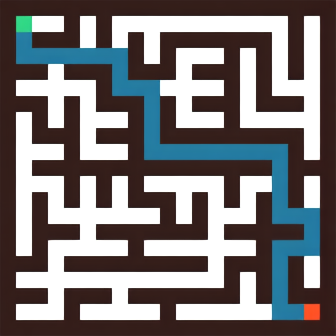} &
    \includegraphics[width=2.5cm,height=2.5cm]{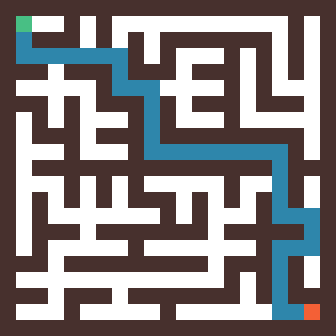} \\[0.1em]
    
    \includegraphics[width=2.5cm,height=2.5cm]{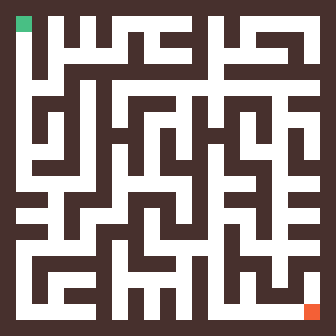} &
    \includegraphics[width=2.5cm,height=2.5cm]{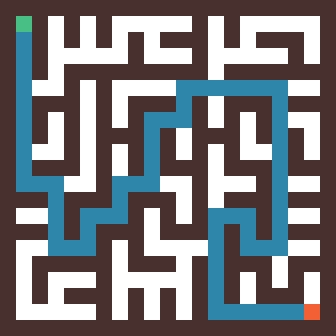} &
    \includegraphics[width=2.5cm,height=2.5cm]{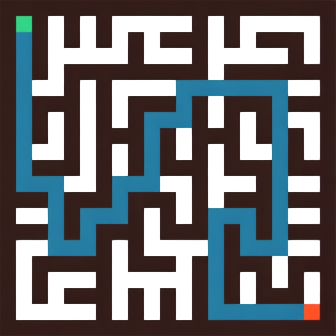} &
    \includegraphics[width=2.5cm,height=2.5cm]{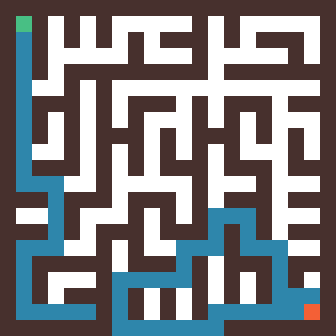} \\[0.1em]
    
    \includegraphics[width=2.5cm,height=2.5cm]{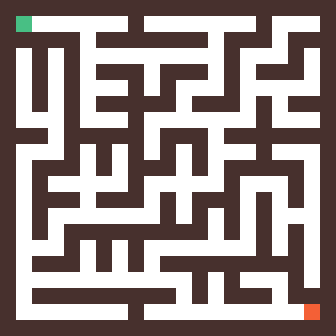} &
    \includegraphics[width=2.5cm,height=2.5cm]{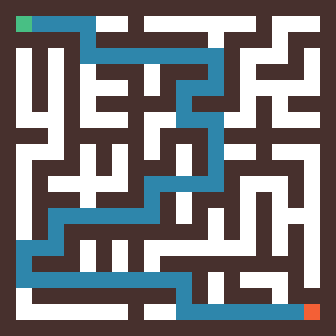} &
    \includegraphics[width=2.5cm,height=2.5cm]{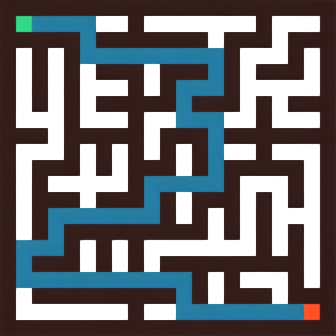} &
    \includegraphics[width=2.5cm,height=2.5cm]{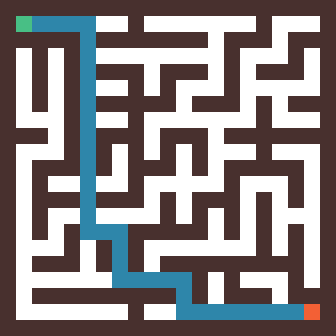} \\
\end{tabular}

\caption{Additional qualitative examples for the \textit{Maze} task, showing inputs, ground truth outputs, and model predictions after finetuning with \(n=300\) samples.}
\end{figure}

\begin{figure}[htbp]
\centering
\begin{tabular}{cccc}
    \textbf{Input} & \textbf{Output} & \vdmbestnt{\textbf{CogVideoX1.5-5B}} & \llmbestnt{\shortstack{\textbf{Qwen3-4B}\\\textbf{Instruct-2507}}} \\[0.3em]
    
    \includegraphics[width=2.5cm,height=2.5cm]{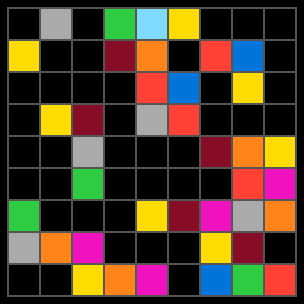} &
    \includegraphics[width=2.5cm,height=2.5cm]{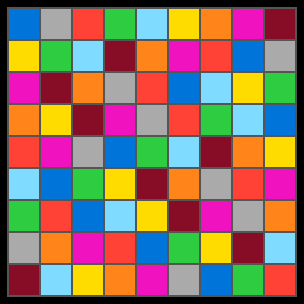} &
    \includegraphics[width=2.5cm,height=2.5cm]{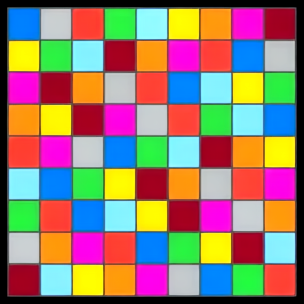} &
    \includegraphics[width=2.5cm,height=2.5cm]{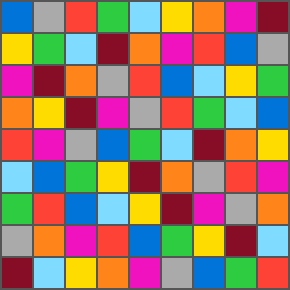} \\[0.1em]
    
    \includegraphics[width=2.5cm,height=2.5cm]{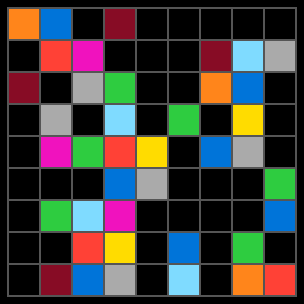} &
    \includegraphics[width=2.5cm,height=2.5cm]{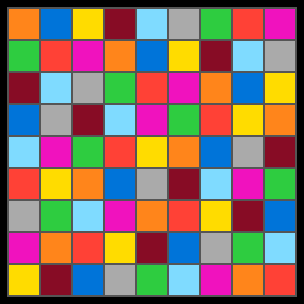} &
    \includegraphics[width=2.5cm,height=2.5cm]{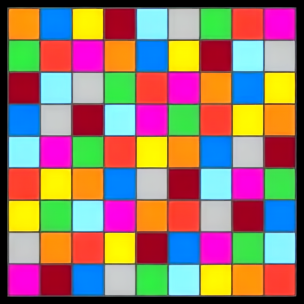} &
    \includegraphics[width=2.5cm,height=2.5cm]{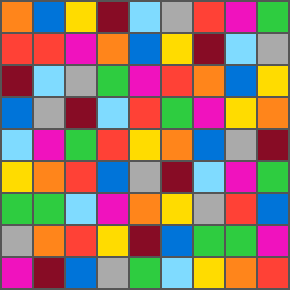} \\[0.1em]
    
    \includegraphics[width=2.5cm,height=2.5cm]{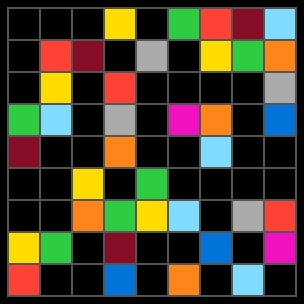} &
    \includegraphics[width=2.5cm,height=2.5cm]{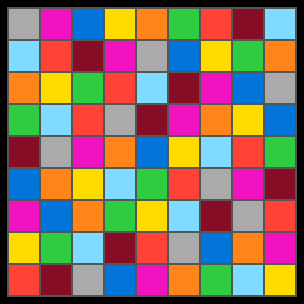} &
    \includegraphics[width=2.5cm,height=2.5cm]{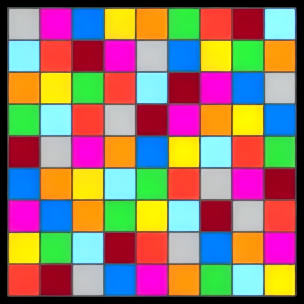} &
    \includegraphics[width=2.5cm,height=2.5cm]{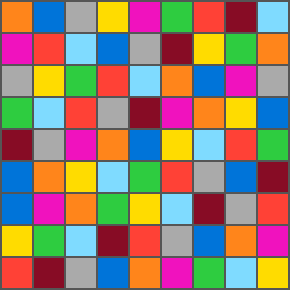} \\[0.1em]
    
    \includegraphics[width=2.5cm,height=2.5cm]{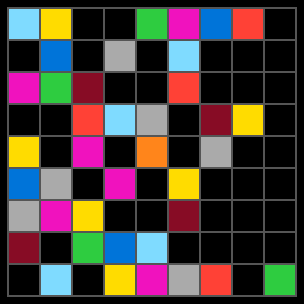} &
    \includegraphics[width=2.5cm,height=2.5cm]{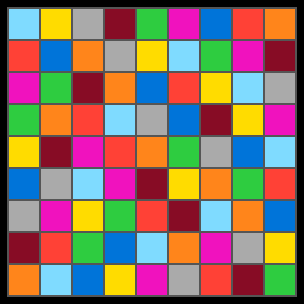} &
    \includegraphics[width=2.5cm,height=2.5cm]{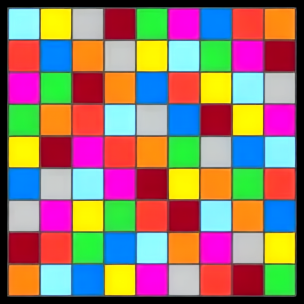} &
    \includegraphics[width=2.5cm,height=2.5cm]{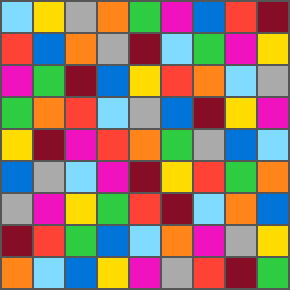} \\[0.1em]
    
    \includegraphics[width=2.5cm,height=2.5cm]{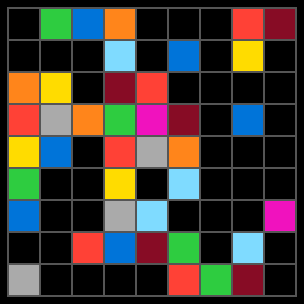} &
    \includegraphics[width=2.5cm,height=2.5cm]{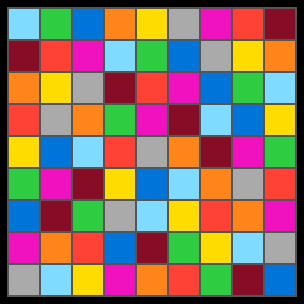} &
    \includegraphics[width=2.5cm,height=2.5cm]{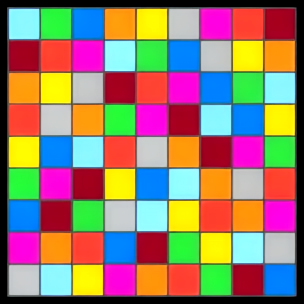} &
    \includegraphics[width=2.5cm,height=2.5cm]{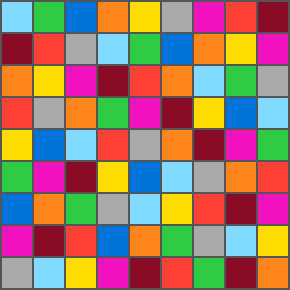} \\[0.1em]
    
    \includegraphics[width=2.5cm,height=2.5cm]{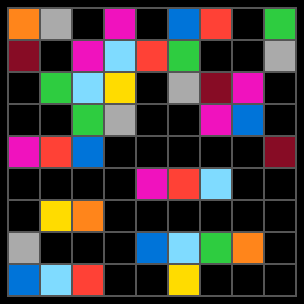} &
    \includegraphics[width=2.5cm,height=2.5cm]{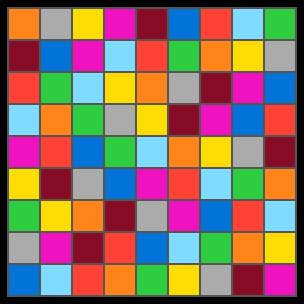} &
    \includegraphics[width=2.5cm,height=2.5cm]{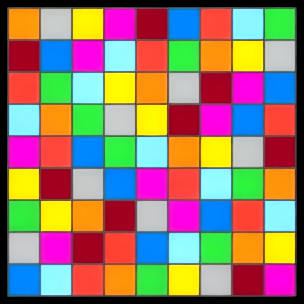} &
    \includegraphics[width=2.5cm,height=2.5cm]{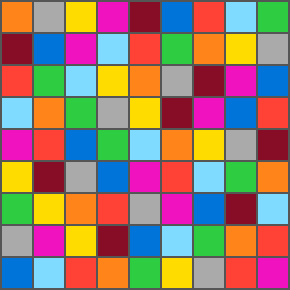} \\[0.1em]
    
    \includegraphics[width=2.5cm,height=2.5cm]{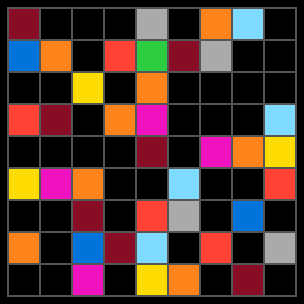} &
    \includegraphics[width=2.5cm,height=2.5cm]{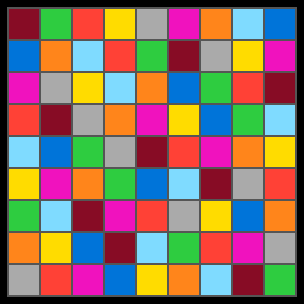} &
    \includegraphics[width=2.5cm,height=2.5cm]{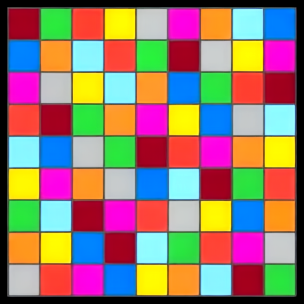} &
    \includegraphics[width=2.5cm,height=2.5cm]{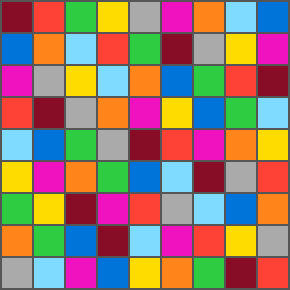} \\[0.1em]
    
    \includegraphics[width=2.5cm,height=2.5cm]{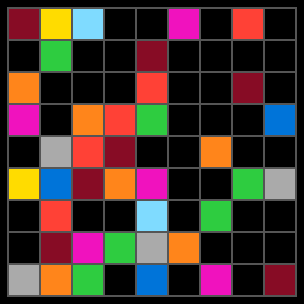} &
    \includegraphics[width=2.5cm,height=2.5cm]{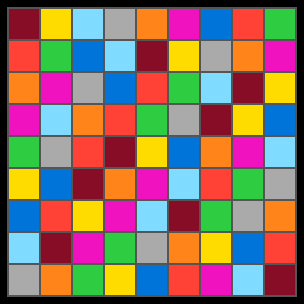} &
    \includegraphics[width=2.5cm,height=2.5cm]{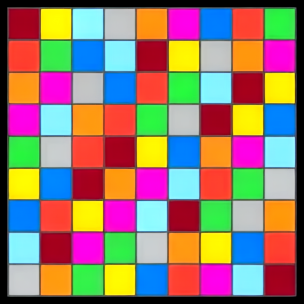} &
    \includegraphics[width=2.5cm,height=2.5cm]{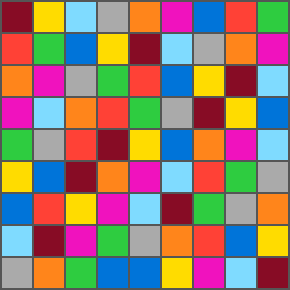} \\
\end{tabular}

\caption{Additional qualitative examples for the \textit{Sudoku} task, showing inputs, ground truth outputs, and model predictions after finetuning with \(n=1000\) samples.}
\end{figure}

\begin{figure}[htbp]
\centering
\begin{tabular}{cccc}
    \textbf{Input} & \textbf{Output} & \vdmbestnt{\textbf{CogVideoX1.5-5B}} & \llmbestnt{\shortstack{\textbf{Qwen3-4B}\\\textbf{Instruct-2507}}} \\[0.3em]
    
    \includegraphics[width=2.5cm,height=2.5cm]{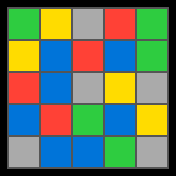} &
    \includegraphics[width=2.5cm,height=2.5cm]{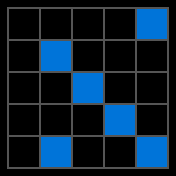} &
    \includegraphics[width=2.5cm,height=2.5cm]{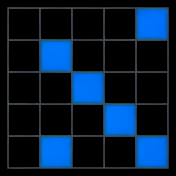} &
    \includegraphics[width=2.5cm,height=2.5cm]{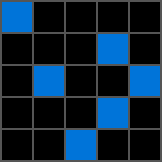} \\[0.1em]
    
    \includegraphics[width=2.5cm,height=2.5cm]{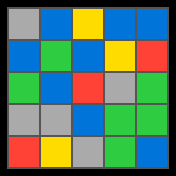} &
    \includegraphics[width=2.5cm,height=2.5cm]{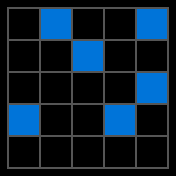} &
    \includegraphics[width=2.5cm,height=2.5cm]{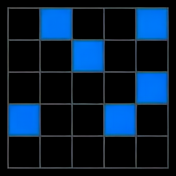} &
    \includegraphics[width=2.5cm,height=2.5cm]{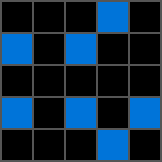} \\[0.1em]
    
    \includegraphics[width=2.5cm,height=2.5cm]{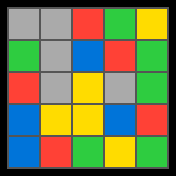} &
    \includegraphics[width=2.5cm,height=2.5cm]{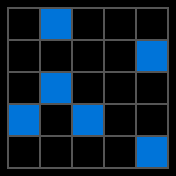} &
    \includegraphics[width=2.5cm,height=2.5cm]{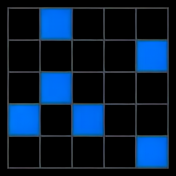} &
    \includegraphics[width=2.5cm,height=2.5cm]{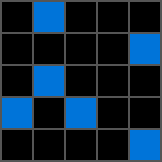} \\[0.1em]
    
    \includegraphics[width=2.5cm,height=2.5cm]{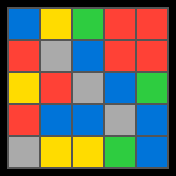} &
    \includegraphics[width=2.5cm,height=2.5cm]{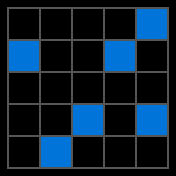} &
    \includegraphics[width=2.5cm,height=2.5cm]{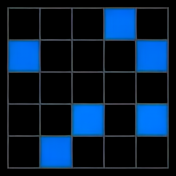} &
    \includegraphics[width=2.5cm,height=2.5cm]{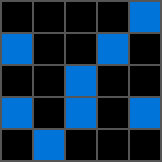} \\[0.1em]
    
    \includegraphics[width=2.5cm,height=2.5cm]{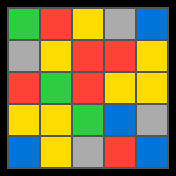} &
    \includegraphics[width=2.5cm,height=2.5cm]{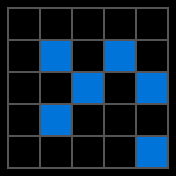} &
    \includegraphics[width=2.5cm,height=2.5cm]{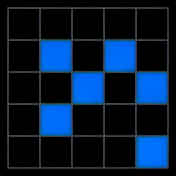} &
    \includegraphics[width=2.5cm,height=2.5cm]{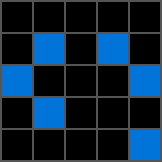} \\[0.1em]
    
    \includegraphics[width=2.5cm,height=2.5cm]{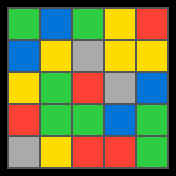} &
    \includegraphics[width=2.5cm,height=2.5cm]{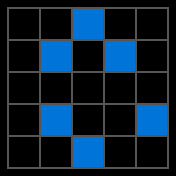} &
    \includegraphics[width=2.5cm,height=2.5cm]{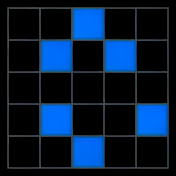} &
    \includegraphics[width=2.5cm,height=2.5cm]{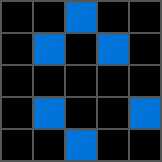} \\[0.1em]
    
    \includegraphics[width=2.5cm,height=2.5cm]{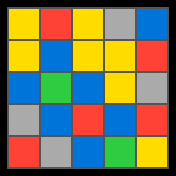} &
    \includegraphics[width=2.5cm,height=2.5cm]{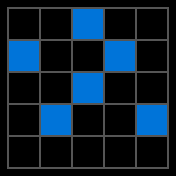} &
    \includegraphics[width=2.5cm,height=2.5cm]{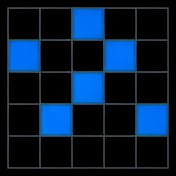} &
    \includegraphics[width=2.5cm,height=2.5cm]{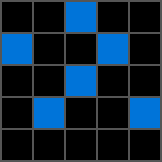} \\[0.1em]
    
    \includegraphics[width=2.5cm,height=2.5cm]{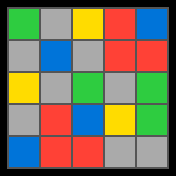} &
    \includegraphics[width=2.5cm,height=2.5cm]{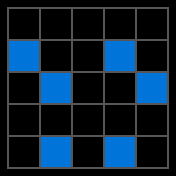} &
    \includegraphics[width=2.5cm,height=2.5cm]{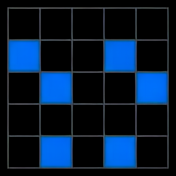} &
    \includegraphics[width=2.5cm,height=2.5cm]{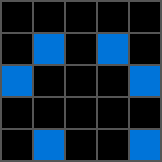} \\
\end{tabular}

\caption{Additional qualitative examples for the \textit{Hitori } task, showing inputs, ground truth outputs, and model predictions after finetuning with \(n=100\) samples.}
\end{figure}

\begin{figure}[htbp]
\centering
\begin{tabular}{cccc}
    \textbf{Input} & \textbf{Output} & \vdmbestnt{\textbf{CogVideoX1.5-5B}} & \llmbestnt{\shortstack{\textbf{Qwen3-4B}\\\textbf{Instruct-2507}}} \\[0.3em]
    
    \includegraphics[width=2.5cm,height=2.5cm]{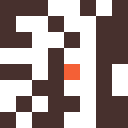} &
    \includegraphics[width=2.5cm,height=2.5cm]{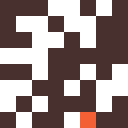} &
    \includegraphics[width=2.5cm,height=2.5cm]{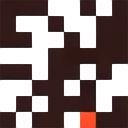} &
    \includegraphics[width=2.5cm,height=2.5cm]{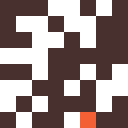} \\[0.1em]
    
    \includegraphics[width=2.5cm,height=2.5cm]{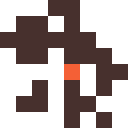} &
    \includegraphics[width=2.5cm,height=2.5cm]{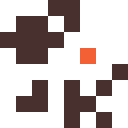} &
    \includegraphics[width=2.5cm,height=2.5cm]{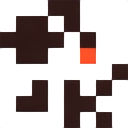} &
    \includegraphics[width=2.5cm,height=2.5cm]{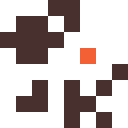} \\[0.1em]
    
    \includegraphics[width=2.5cm,height=2.5cm]{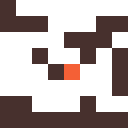} &
    \includegraphics[width=2.5cm,height=2.5cm]{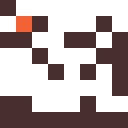} &
    \includegraphics[width=2.5cm,height=2.5cm]{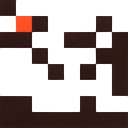} &
    \includegraphics[width=2.5cm,height=2.5cm]{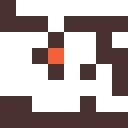} \\[0.1em]
    
    \includegraphics[width=2.5cm,height=2.5cm]{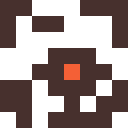} &
    \includegraphics[width=2.5cm,height=2.5cm]{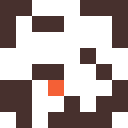} &
    \includegraphics[width=2.5cm,height=2.5cm]{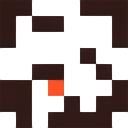} &
    \includegraphics[width=2.5cm,height=2.5cm]{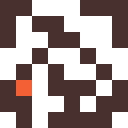} \\[0.1em]
    
    \includegraphics[width=2.5cm,height=2.5cm]{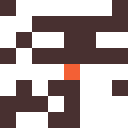} &
    \includegraphics[width=2.5cm,height=2.5cm]{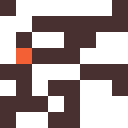} &
    \includegraphics[width=2.5cm,height=2.5cm]{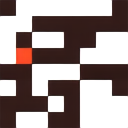} &
    \includegraphics[width=2.5cm,height=2.5cm]{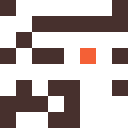} \\[0.1em]
    
    \includegraphics[width=2.5cm,height=2.5cm]{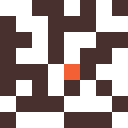} &
    \includegraphics[width=2.5cm,height=2.5cm]{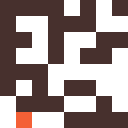} &
    \includegraphics[width=2.5cm,height=2.5cm]{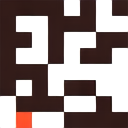} &
    \includegraphics[width=2.5cm,height=2.5cm]{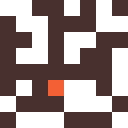} \\[0.1em]
    
    \includegraphics[width=2.5cm,height=2.5cm]{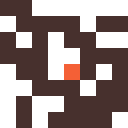} &
    \includegraphics[width=2.5cm,height=2.5cm]{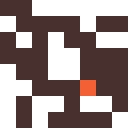} &
    \includegraphics[width=2.5cm,height=2.5cm]{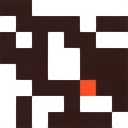} &
    \includegraphics[width=2.5cm,height=2.5cm]{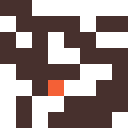} \\[0.1em]
    
    \includegraphics[width=2.5cm,height=2.5cm]{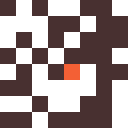} &
    \includegraphics[width=2.5cm,height=2.5cm]{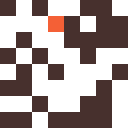} &
    \includegraphics[width=2.5cm,height=2.5cm]{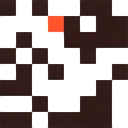} &
    \includegraphics[width=2.5cm,height=2.5cm]{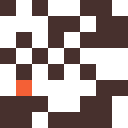} \\
\end{tabular}

\caption{Additional qualitative examples for the \textit{Langton Ant} (horizon 10) task, showing inputs, ground truth outputs, and model predictions after finetuning with \(n=1000\) samples.}
\end{figure}

\section{Additional Results}
\section{ARC Family}

Here, we include the comparison table for ConceptARC, by including finetuned LLMs (Qwen3-4B-Instruct, Qwen3-8B, LLama3.1-8B-Instruct) and GPT-4 [IC]\footnote{Added for reference with commercial models, this case is directly IC and not our finetune  approach.} \cite{moskvichev2023conceptarc}, as well as VDMs (CogVideoX1.5-5B, Wan2.1-14B, LTX-2B/13B). These additional results provide broader context and help reinforce the trends observed in the main text. See Table \ref{tab:appdx:concept-accuracy}. 

The relatively lower performance of LTX compared to other VDMs may stem from its aggressive VAE compression, which can discard structural information important for ConceptARC. This reflects a design tradeoff of the LTX models, aimed at enabling much faster video generation \cite{hacohen2024ltx}.

\begin{table}[htbp]
    \centering
    \caption{Concept-wise overall accuracy across models. Best values are highlighted for \vdmbestnt{VDMs} or \llmbestnt{LLMs}.}
    \label{tab:appdx:concept-accuracy}
    \resizebox{\linewidth}{!}{%
    \begin{tabular}{lccccccccc}
    \toprule
    \textbf{Concept} & \textbf{LTX-13B} & \textbf{LTX-2B} & \textbf{Wan2.1-14B} & \textbf{CogVideoX1.5-5B} & \makecell{\textbf{Qwen3-4B}\\\textbf{Instruct-2507}} & \textbf{Qwen3-8B} & \textbf{Llama3.1-8B} & \textbf{GPT-4 [IC]} \\
        \midrule
        AboveBelow & 0.30 & 0.17 & 0.37 & \vdmbest{0.40} & \llmbest{0.40} & \llmbest{0.40} & 0.17 & 0.23 \\
        TopBottom2D & 0.23 & 0.17 & \vdmbest{0.63} & 0.37 & 0.50 & 0.50 & 0.37 & 0.23 \\
        TopBottom3D & 0.27 & 0.17 & \vdmbest{0.47} & 0.33 & 0.13 & 0.20 & 0.17 & 0.20 \\
        HorizontalVertical & 0.13 & 0.20 & \vdmbest{0.53} & 0.47 & 0.43 & 0.47 & 0.33 & 0.27 \\
        Center & 0.33 & 0.30 & \vdmbest{0.57} & 0.37 & 0.20 & 0.20 & 0.13 & 0.33 \\
        FilledNotFilled & 0.30 & 0.27 & \vdmbest{0.50} & 0.37 & 0.27 & 0.23 & 0.20 & 0.17 \\
        CompleteShape & 0.20 & 0.10 & \vdmbest{0.40} & 0.37 & 0.23 & 0.30 & 0.13 & 0.23 \\
        InsideOutside & 0.27 & 0.27 & \vdmbest{0.37} & 0.33 & 0.13 & 0.20 & 0.13 & 0.10 \\
        ExtractObjects & 0.07 & 0.07 & \vdmbest{0.23} & 0.07 & 0.10 & 0.10 & 0.03 & 0.03 \\
        Count & 0.40 & 0.43 & \vdmbest{0.83} & 0.57 & 0.13 & 0.13 & 0.17 & 0.13 \\
        SameDifferent & 0.23 & 0.23 & 0.33 & \vdmbest{0.37} & 0.27 & 0.23 & 0.27 & 0.17 \\
        Order & 0.03 & 0.03 & 0.00 & 0.07 & \llmbest{0.27} & \llmbest{0.27} & 0.10 & \llmbest{0.27} \\
        MoveToBoundary & 0.17 & 0.00 & 0.13 & 0.17 & \llmbest{0.23} & 0.10 & 0.17 & 0.20 \\
        ExtendToBoundary & 0.20 & 0.23 & \vdmbest{0.50} & 0.40 & 0.13 & 0.17 & 0.10 & 0.07 \\
        Copy & 0.20 & 0.03 & 0.17 & 0.13 & 0.17 & 0.10 & 0.10 & \llmbest{0.23} \\
        CleanUp & 0.43 & 0.40 & \vdmbest{0.60} & 0.53 & 0.27 & 0.30 & 0.27 & 0.20 \\
        \midrule
        \textbf{Average Accuracy} & 0.24 & 0.19 & \vdmbest{0.41} & 0.33 & 0.24 & 0.24 & 0.18 & 0.19 \\
        \bottomrule
    \end{tabular}}
\end{table}

\subsection{Pitfalls of Vision Language Models}

Vision–Language Models (VLMs) promise to bridge the gap between visual perception and language by training on vast datasets of paired images and text. In principle, this multimodal pretraining should enable these models to solve visually grounded  tasks more effectively than language-only models. To test whether this promise holds in practice, we evaluate a representative VLM, Gemma-4B~\cite{gemma3}, on a structured visual task: \textit{Sudoku}.

We fine-tune the same model with \(n = 1000\) samples under three configurations: \textbf{text-only}, \textbf{image-only}, and \textbf{combined image–text}; keeping all other settings fixed. The results in Table~\ref{tab:appdx-vlms-sudoku} reveal a striking limitation: adding image input offers no measurable improvement, and the \textbf{image-only} variant performs worse than a trivial baseline. This suggests that the model is unable to extract meaningful information from visual inputs, even when explicitly trained to do so.

\begin{table}[htbp]
\centering
\caption{Relative Accuracy and Accuracy on \textit{Sudoku}.}
\begin{tabular}{lcc}
\toprule
\textbf{Model} & \textbf{Relative Accuracy} & \textbf{Accuracy} \\
\midrule
\textbf{Text-only} & 0.79 & 0.06 \\
\textbf{Combined image--text} & 0.78 & 0.06 \\
\textbf{Image-only} & 0.12 & 0.00 \\
\bottomrule
\end{tabular}
\label{tab:appdx-vlms-sudoku}
\end{table}

To investigate why, we train the \textbf{image-only} model on a simplified task: reconstructing the textual grid representation of its own image input rather than predicting a Sudoku solution. With small training sets (\(n=3,5,10\)), the model fails to interpret the images and instead memorizes training samples, reproducing them verbatim regardless of input (Table~\ref{tab:appdx-output-matching-sudoku}). The model learns little about the underlying structure of the visual input.

\begin{table}[htbp]
\centering
\caption{Distribution of outputs on the test set exactly matching training samples for different training set sizes.}
\begin{tabular}{lccc}
\toprule
\textbf{Training Set Size} & \textbf{Sample} & \textbf{Proportion} & \textbf{Total Proportion} \\
\midrule
\multirow{3}{*}{3} 
& Sample 1 & 0.385 & \multirow{3}{*}{\textbf{1.00}} \\
& Sample 2 & 0.010 & \\
& Sample 3 & 0.605 & \\
\midrule
\multirow{4}{*}{5} 
& Sample 1 & 0.490 & \multirow{4}{*}{\textbf{0.99}} \\
& Sample 2 & 0.030 & \\
& Sample 3 & 0.335 & \\
& Sample 4 & 0.135 & \\
\midrule
\multirow{8}{*}{10} 
& Sample 1 & 0.100 & \multirow{8}{*}{\textbf{0.96}} \\
& Sample 2 & 0.010 & \\
& Sample 3 & 0.030 & \\
& Sample 4 & 0.005 & \\
& Sample 5 & 0.015 & \\
& Sample 6 & 0.170 & \\
& Sample 7 & 0.615 & \\
& Sample 8 & 0.010 & \\
\bottomrule
\end{tabular}
\label{tab:appdx-output-matching-sudoku}
\end{table}

This experiment exposes a deeper issue: despite their multimodal pretraining, current VLMs struggle to extract structured information from images \cite{jing2025a, sim2025can}. They appear to rely primarily on semantics and basic pattern recognition rather than true visual understanding. Furthermore, VLMs inherit many of the limitations of LLMs, such as reliance on text-based outputs, without gaining meaningful visual understanding ability.

Because VLMs provide no measurable advantage over language-only models for these structured visual tasks, we focus on LLMs as the primary baseline. LLMs already demonstrate strong capabilities in structured prediction and symbolic manipulation, making them a fair and informative comparison point for VDMs. This framing keeps the evaluation focused on model families that offer complementary strengths.

\section{Results - Full Tables}
We provide the complete set of experimental results, which constitute the underlying data for the figures reported in the main paper.

\begin{table*}[htbp]
    \centering
    \caption{Comparison of CogVideoX1.5-5B and Qwen3-4B-Instruct-2507 accuracy on structured games. Missing values are shown as \texttt{-}.}
    \resizebox{\textwidth}{!}{
        \begin{tabular}{r|ccccc|ccccc}
            \toprule
            \textbf{n} &
            \multicolumn{5}{c|}{\textbf{CogVideoX1.5-5B}} &
            \multicolumn{5}{c}{\textbf{Qwen3-4B-Instruct-2507}} \\
            \midrule
            & \textbf{Chess-Mate-in-1} & \textbf{Connect 4} & \textbf{Hitori 5x5} & \textbf{Sudoku Mini} & \textbf{Sudoku} &
              \textbf{Chess-Mate-in-1} & \textbf{Connect 4} & \textbf{Hitori 5x5} & \textbf{Sudoku Mini} & \textbf{Sudoku} \\
            \midrule
            3   & 0.00 & 0.44 & 0.01 & 0.22 & 0.00 & 0.00 & 0.03 & 0.00 & 0.18 & --    \\
            5   & 0.00 & 0.62 & 0.02 & 0.36 & 0.00 & 0.02 & 0.05 & 0.00 & 0.22 & --    \\
            10  & 0.00 & 0.74 & 0.62 & 0.65 & 0.00 & 0.04 & 0.08 & 0.02 & 0.48 & --    \\
            30  & 0.02 & 0.78 & 0.72 & 0.78 & 0.20 & 0.13 & 0.38 & 0.02 & 0.64 & 0.00 \\
            50  & 0.04 & 0.80 & 0.84 & 0.90 & 0.34 & 0.15 & 0.38 & 0.10 & 0.68 & 0.00 \\
            100 & 0.08 & 0.85 & 0.92 & 0.91 & 0.60 & 0.24 & 0.69 & 0.28 & 0.78 & 0.01 \\
            300 & 0.14 & 0.84 & 0.94 & 0.90 & 0.55 & 0.38 & 0.71 & 0.57 & 0.80 & 0.01 \\
            500 & 0.20 & 0.89 & 0.94 & 0.94 & 0.60 & 0.44 & 0.69 & 0.64 & 0.86 & 0.06 \\
            1000& 0.22 & 0.90 & 0.96 & 0.91 & 0.79 & 0.56 & 0.76 & 0.86 & 0.90 & 0.14 \\
            3000& --    & 0.92 & 0.98 & 0.95 & 0.86 & --    & 0.78 & 0.94 & 0.92 & 0.32 \\
            5000& --    & 0.90 & 0.99 & 0.96 & 0.86 & --    & 0.82 & 0.96 & 0.96 & 0.55 \\
            \bottomrule
        \end{tabular}
    }
    \label{tab:cogvideo-vs-qwen-games}
\end{table*}

\begin{table*}[htbp]
    \centering
    \caption{Comparison of CogVideoX1.5-5B and Qwen3-4B-Instruct-2507 accuracy on Life-Like Cellular Automata variants. Missing values are shown as \texttt{-}.}
    \resizebox{\textwidth}{!}{%
        \begin{tabular}{r|ccccc|ccccc}
            \toprule
            \textbf{n} &
            \multicolumn{5}{c|}{\textbf{CogVideoX1.5-5B}} &
            \multicolumn{5}{c}{\textbf{Qwen3-4B-Instruct-2507}} \\
            \midrule
            & \textbf{Life\_B3S2} & \textbf{DayAndNight} & \textbf{Maze} & \textbf{Seeds} & \textbf{Game of Life} &
              \textbf{Life\_B3S2} & \textbf{DayAndNight} & \textbf{Maze} & \textbf{Seeds} & \textbf{Game Of Life} \\
            10  & 0.00  & 0.00  & 0.00  & 0.00  & 0.00  & --    & --    & --    & --    & --    \\
            30  & 1.00  & 0.81  & 0.87  & 1.00  & 0.96  & --    & 0.63  & 0.81  & 0.75  & 0.63  \\
            50  & 1.00  & 0.95  & 0.91  & 1.00  & 0.97  & --    & 0.64  & 0.80  & 0.78  & 0.64  \\
            100 & 1.00  & 1.00  & 0.96  & 1.00  & 1.00  & 0.61  & 0.70  & 0.87  & 0.63  & 0.73  \\
            300 & --    & --    & --    & --    & --    & 1.00  & 1.00  & 1.00  & 1.00  & 1.00  \\
            500 & --    & --    & --    & --    & --    & --    & 1.00  & 1.00  & 1.00  & 1.00  \\
            \bottomrule
        \end{tabular}
    }
    \label{tab:cogvideo-vs-qwen-gol}
\end{table*}

\begin{table*}[htbp]
    \centering
    \caption{Comparison of CogVideoX1.5-5B and Qwen3-4B-Instruct-2507 accuracy on Langton's Ant with respect to number of steps into the future. Missing values are shown as \texttt{-}.}
    %\resizebox{\textwidth}{!}{%
        \begin{tabular}{r|cccc|cccc}
            \toprule
            \textbf{n} &
            \multicolumn{4}{c|}{\textbf{CogVideoX1.5-5B}} &
            \multicolumn{4}{c}{\textbf{Qwen3-4B-Instruct-2507}} \\
            \midrule
            & \textbf{Step 2} & \textbf{Step 3} & \textbf{Step 5} & \textbf{Step 10} &
              \textbf{Step 2} & \textbf{Step 3} & \textbf{Step 5} & \textbf{Step 10} \\
            \midrule
            3   & 0.18 & 0.03 & 0.03 & --    & 0.32 & 0.03 & --    & --    \\
            5   & 0.23 & 0.07 & 0.04 & 0.00 & 0.21 & 0.04 & --    & --    \\
            10  & 0.67 & 0.29 & 0.06 & 0.01 & 0.51 & 0.19 & --    & --    \\
            30  & 1.00 & 0.76 & 0.25 & 0.01 & 0.79 & 0.46 & 0.06 & 0.00 \\
            50  & 1.00 & 0.99 & 0.41 & 0.01 & 0.950 & 0.58 & 0.14 & 0.010 \\
            100 & 1.00 & 1.000 & 0.88 & 0.08 & 0.99 & 0.910 & 0.39 & 0.01 \\
            300 & --    & --    & 1.00 & 0.42 & 1.00 & 1.00 & 0.98 & 0.12 \\
            500 & --    & --    & 1.00 & 0.83 & 1.00 & 1.00 & 1.00 & 0.21 \\
            1000& --    & --    & 1.00 & 0.98 & 1.00 & 1.00 & 1.00 & 0.47 \\
            3000& --    & --    & --    & 0.99 & --    & --    & --    & 0.71 \\
            5000& --    & --    & --    & --    & --    & --    & --    & 0.93 \\
            \bottomrule
        \end{tabular}
    %}
    \label{tab:cogvideo-vs-qwen-langton}
\end{table*}

\begin{table*}[htbp]
    \centering
    \caption{Comparison of CogVideoX1.5 and Qwen3-4B-Instruct-2507 accuracy on \textit{Maze} and \textit{Shortest Path} tasks. Missing values are shown as \texttt{-}.}
    \resizebox{\textwidth}{!}{%
        \begin{tabular}{r|ccc|ccc}
            \toprule
            \textbf{n} &
            \multicolumn{3}{c|}{\textbf{CogVideoX1.5}} &
            \multicolumn{3}{c}{\textbf{Qwen3-4B-Instruct-2507}} \\
            \midrule
            & \textbf{Base Maze} & \textbf{Maze Generalization} & \textbf{Shortest Path} &
              \textbf{Base Maze} & \textbf{Maze Generalization} & \textbf{Shortest Path} \\
            \midrule
            3   & 0.015 & --    & 0.010 & --    & --    & --    \\
            5   & 0.010 & --    & 0.025 & --    & --    & --    \\
            10  & 0.070 & 0.050 & 0.040 & --    & --    & --    \\
            30  & 0.550 & 0.175 & 0.330 & 0.000 & --    & 0.010 \\
            50  & 0.760 & 0.355 & 0.420 & 0.005 & 0.000 & 0.010 \\
            100 & 0.940 & 0.590 & 0.700 & 0.005 & 0.000 & 0.050 \\
            300 & 1.000 & 0.755 & 0.860 & 0.115 & 0.020 & 0.155 \\
            500 & 1.000 & 0.885 & 0.910 & 0.195 & 0.060 & 0.320 \\
            1000& --    & 0.865 & 0.945 & 0.500 & 0.335 & 0.500 \\
            3000& --    & 0.815 & 0.960 & 0.710 & 0.375 & 0.640 \\
            5000& --    & 0.940 & 0.975 & 0.925 & 0.525 & 0.770 \\
            \bottomrule
        \end{tabular}
    }
    \label{tab:cogvideo-vs-qwen-maze-nav}
\end{table*}

\begin{table*}[htbp]
    \centering
    \caption{Comparison of CogVideoX1.5-5B and Qwen3-4B-Instruct-2507 accuracy on cellular automata rules grouped by Wolfram classes. Missing values are shown as \texttt{-}.}
    %\resizebox{\textwidth}{!}{%
    \begin{tabular}{r|cccc|cccc}
        \toprule
        \textbf{n} &
        \multicolumn{4}{c|}{\textbf{CogVideoX1.5-5B}} &
        \multicolumn{4}{c}{\textbf{Qwen3-4B-Instruct-2507}} \\
        \midrule
        \multicolumn{9}{c}{\textbf{Class 1}} \\
        \midrule
        & \textbf{R8} & \textbf{R32} & \textbf{R128} & \textbf{R160} &
          \textbf{R8} & \textbf{R32} & \textbf{R128} & \textbf{R160} \\
        3   & 0.75 & 0.49 & 0.29 & 0.13 & 0.06 & 0.02 & 0.04 & 0.04 \\
        5   & 0.71 & 0.51 & 0.28 & 0.20 & 0.10 & 0.06 & 0.06 & 0.04 \\
        10  & 0.74 & 0.67 & 0.32 & 0.48 & 0.19 & 0.21 & 0.08 & 0.12 \\
        30  & 0.77 & 0.82 & 0.85 & 0.87 & 0.72 & 0.67 & 0.65 & 0.81 \\
        50  & 0.72 & 0.98 & 0.99 & 0.93 & 0.81 & 0.96 & 0.77 & 0.84 \\
        100 & 1.00   & --    & --    & --    & 0.97 & 0.93 & 0.90 & 0.99 \\
        300 & --    & --    & --    & --    & 0.98 & --    & --    & --    \\
        \midrule
        \multicolumn{9}{c}{\textbf{Class 2}} \\
        \midrule
        & \textbf{R4} & \textbf{R108} & \textbf{R170} & \textbf{R250} &
          \textbf{R4} & \textbf{R108} & \textbf{R170} & \textbf{R250} \\
        3   & 0.71 & 0.155 & 0.07 & 0.17 & --    & --    & --    & --    \\
        5   & 0.76 & 0.310 & 0.27 & 0.19 & --    & --    & --    & --    \\
        10  & 0.74 & 0.415 & 0.87 & 0.27 & --    & --    & 0.85    & --    \\
        30  & 0.85 & 0.640 & 1.00 & 0.59 & 0.72  & 0.47  & 0.99  & 0.52  \\
        50  & 0.93 & 0.785 & 1.00 & 0.90 & 0.82  & 0.82  & 0.98  & 0.86  \\
        100 & --    & --    & --    & --    & 0.90  & 0.90  & 1.00  & 1.00  \\
        300 & --    & --    & --    & --    & 1.00  & 1.00  & 1.00  & 0.99  \\
        \midrule
        \multicolumn{9}{c}{\textbf{Class 3}} \\
        \midrule
        & \textbf{R30} & \textbf{R45} & \textbf{R90} & \textbf{R150} &
          \textbf{R30} & \textbf{R45} & \textbf{R90} & \textbf{R150} \\
        3   & 0.00 & 0.00 & 0.00 & 0.00 & --    & --    & --    & --    \\
        5   & 0.00 & 0.00 & 0.00 & 0.00 & --    & --    & --    & --    \\
        10  & 0.00 & 0.00 & 0.00 & 0.00 & --    & --    & --    & --    \\
        30  & 0.07 & 0.07 & 0.10 & 0.00 & 0.18  & 0.03  & 0.03  & 0.01  \\
        50  & 0.55 & 0.53 & 0.25 & 0.01 & 0.83  & 0.71  & 0.08  & 0.97  \\
        100 & 0.97 & 1.00 & 0.99 & 0.65 & 0.97  & 0.98  & 0.27  & 0.99  \\
        300 & --    & --    & --  & 0.86    & 1.00  & 1.00  & 0.90  & 1.00  \\
        500 & --    & --    & --  & 0.98    & --  & --  & --  & --  \\
        \midrule
        \multicolumn{9}{c}{\textbf{Class 4}} \\
        \midrule
        & \textbf{R110} & \textbf{R54} & \textbf{R62} & \textbf{R106} &
          \textbf{R110} & \textbf{R54} & \textbf{R62} & \textbf{R106} \\
        3   & 0.00 & 0.00 & 0.02 & 0.00 & --    & --    & --    & --    \\
        5   & 0.00 & 0.00 & 0.02 & 0.00 & --    & --    & --    & --    \\
        10  & 0.00 & 0.01 & 0.03 & 0.00 & --    & --    & --    & --    \\
        30  & 0.42 & 0.54 & 0.31 & 0.09 & 0.87  & 0.31  & 0.13  & 0.18  \\
        50  & 0.90 & 0.99 & 0.53 & 0.57 & 0.95  & 0.78  & 0.79  & 0.63  \\
        100 & 1.00 & 1.00 & 0.97 & 0.97 & 1.00  & 0.94  & 0.93  & 1.00  \\
        300 & 1.00 & 1.00 & --    & 1.00 & 1.00  & 1.00  & 1.00  & 1.00  \\
        \bottomrule
    \end{tabular}%
    %}
    \label{tab:cogvideo-vs-qwen-by-class}
\end{table*}

\section{Exploring Generalization of I2I-Tuned VDMs}

While the main text emphasizes grid-structured visual prediction tasks, our framework extends naturally to a broad range of image-to-image problems. In this section, we briefly explore its applicability to classical computer vision tasks. Few-shot adaptation functions both as an efficient tuning strategy and as a probe of model competence: if the model succeeds with \textbf{very few paired examples}, it indicates that the underlying ability was already internalized during pretraining.

We fine-tune CogVideoX1.5-5B, across tasks using between one and thirty paired examples, maintaining the same architecture, optimization schedule, and hyperparameters as in the main experiments. No auxiliary losses or task-specific modifications are introduced, isolating the contribution of pretrained knowledge.

We explore this setup on several established datasets spanning diverse visual domains, including \textbf{NYUv2} \cite{Silberman:ECCV12}, \textbf{ADE20K} \cite{zhou2017scene, zhou2019semantic}, \textbf{ML-Hypersim} \cite{roberts:2021}, \textbf{COCO 2017} \cite{lin2014microsoft}, and \textbf{DreamBooth} \cite{ruiz2023dreambooth}. These benchmarks cover a wide range of classical computer vision problems, from structured scene understanding to generative image transformation.

\begin{figure*}[h]
    \centering
    \begin{tabular}{c c c c}
    % Larger Input spanning two rows
    \multirow{2}{*}{
        \begin{subfigure}{0.26\textwidth} % Increased from 0.22 to 0.26
            \centering
            \vspace{-1.8cm}
            \includegraphics[width=\linewidth]{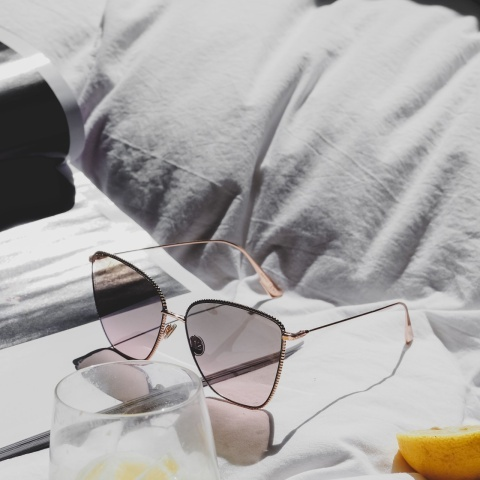}
            \caption{\textbf{Input}}
        \end{subfigure}
    } &
    
    % Top row: 1-shot
    \begin{subfigure}{0.21\textwidth}
        \centering
        \includegraphics[width=\linewidth]{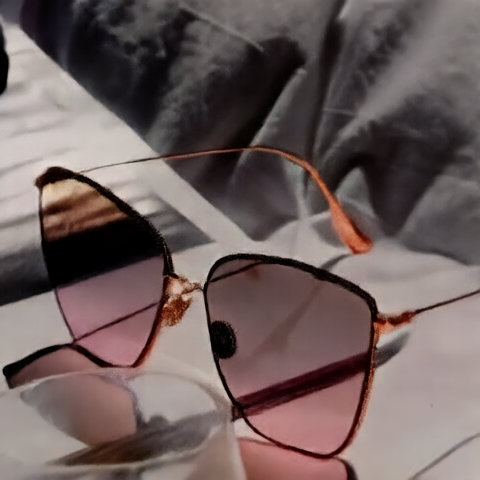}
        \caption{\textbf{Zoom}\\1-shot}
    \end{subfigure} &
    \begin{subfigure}{0.21\textwidth}
        \centering
        \includegraphics[width=\linewidth]{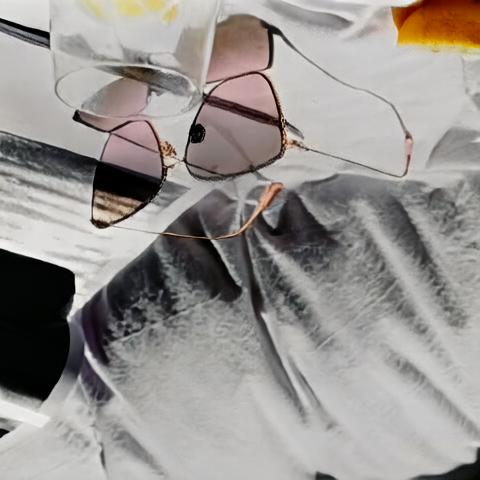}
        \caption{\textbf{Vertical Flip}\\1-shot}
    \end{subfigure} &
    \begin{subfigure}{0.21\textwidth}
        \centering
        \includegraphics[width=\linewidth]{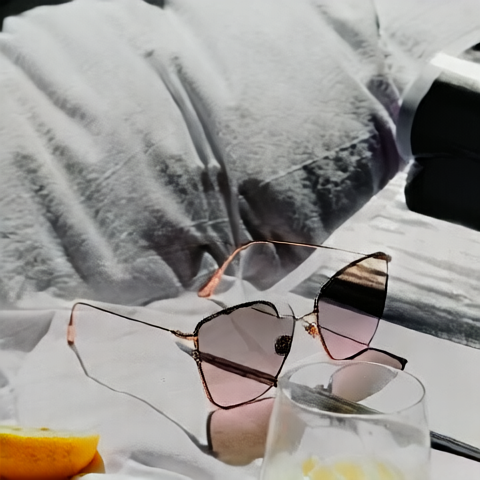}
        \caption{\textbf{Horizontal Flip}\\1-shot}
    \end{subfigure} \\
    
    % Bottom row: 3-shot
    & 
    \begin{subfigure}{0.21\textwidth}
        \centering
        \includegraphics[width=\linewidth]{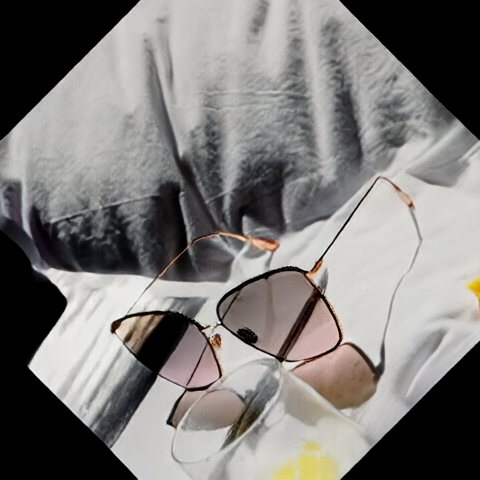}
        \caption{\textbf{Rotation45}\\3-shot}
    \end{subfigure} &
    \begin{subfigure}{0.21\textwidth}
        \centering
        \includegraphics[width=\linewidth]{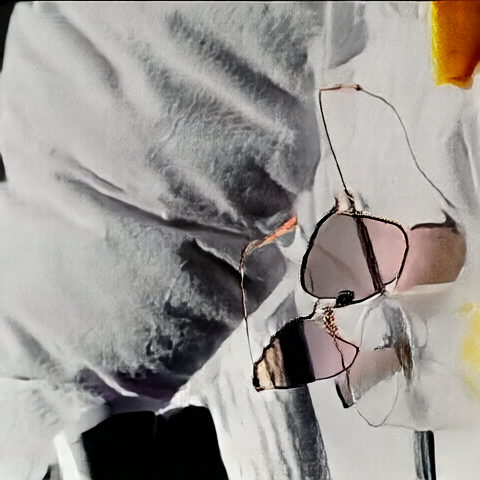}
        \caption{\textbf{Rotation90}\\3-shot}
    \end{subfigure} &
    \begin{subfigure}{0.21\textwidth}
        \centering
        \includegraphics[width=\linewidth]{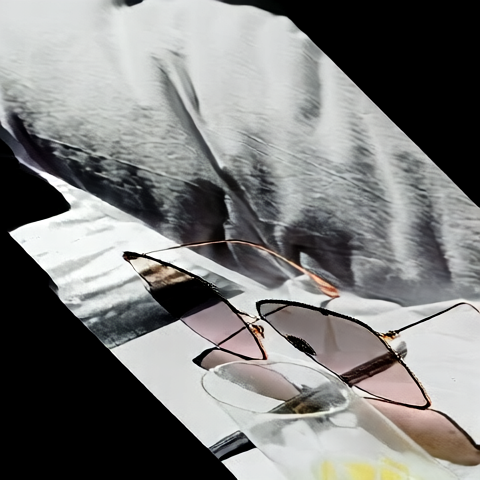}
        \caption{\textbf{Shearing}\\3-shot}
    \end{subfigure} \\
    
    \end{tabular}
    \caption{Geometric transformations learned in few-shot setting. Input is shown on the left, with 1-shot results on the top row and 3-shot results on the bottom row.}
    \label{fig:appdx:cv:geometric-transforms}
    \end{figure*}

\begin{figure}[ht]
\centering
\begin{tabular}{ccccc}
\textbf{Input} & \textbf{Starry Night} & \textbf{Pixel Art} & \textbf{Cubism} & \textbf{Ukiyo-e} \\[0.5em]
\includegraphics[width=2.4cm]{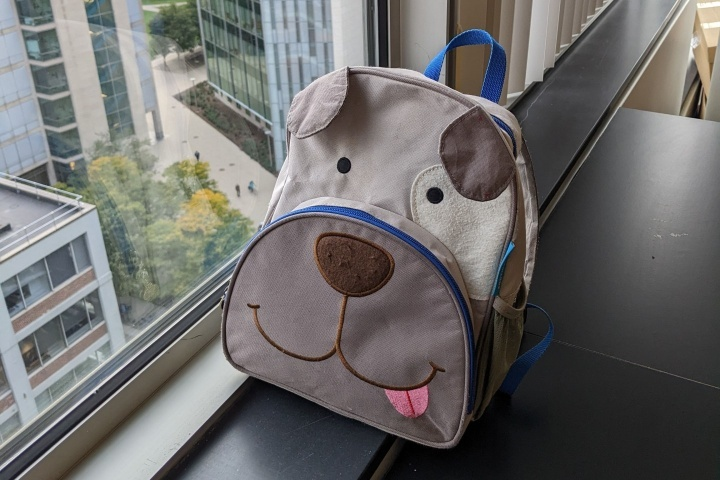} &
\includegraphics[width=2.4cm]{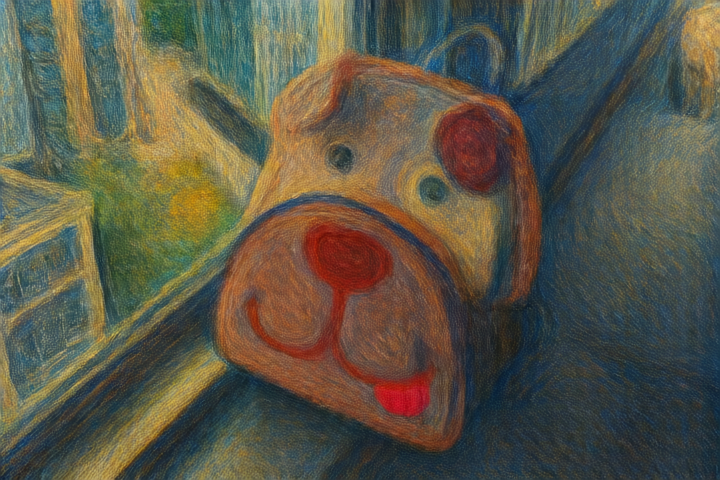} &
\includegraphics[width=2.4cm]{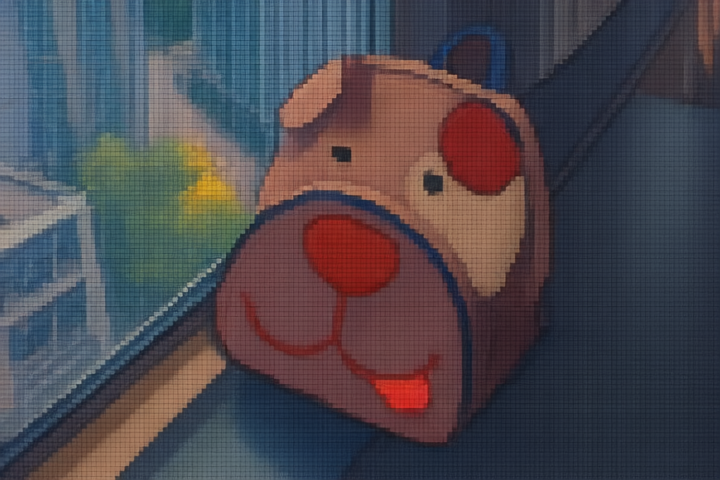} &
\includegraphics[width=2.4cm]{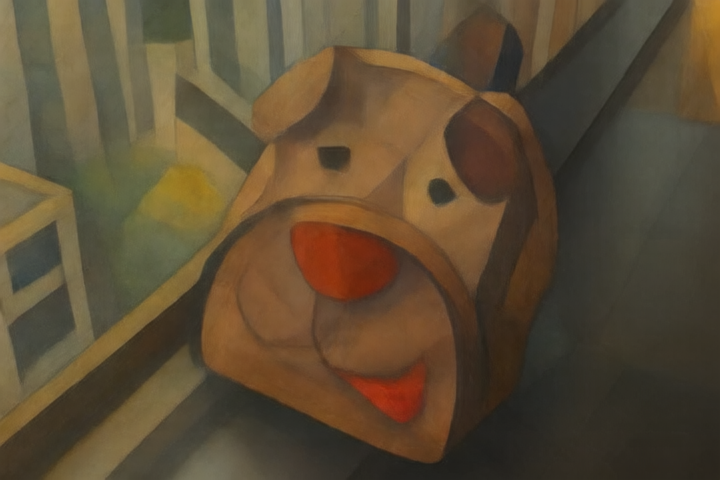} &
\includegraphics[width=2.4cm]{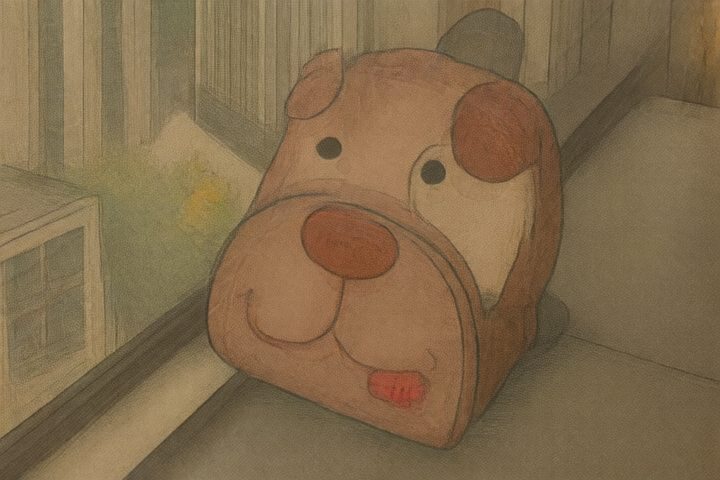} \\

\includegraphics[width=2.4cm]{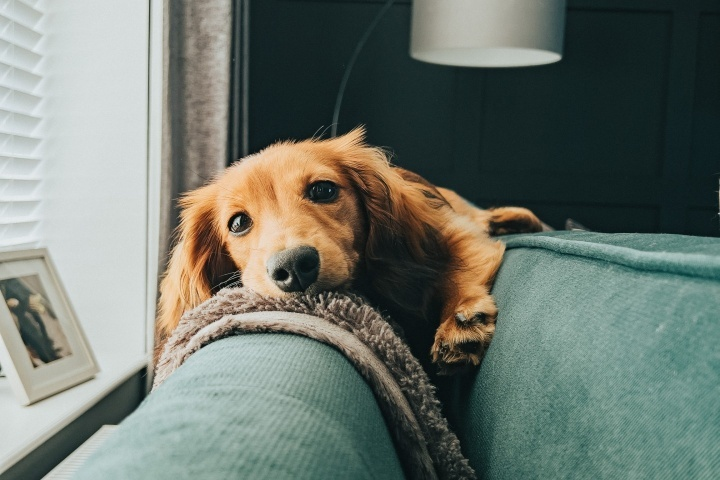} &
\includegraphics[width=2.4cm]{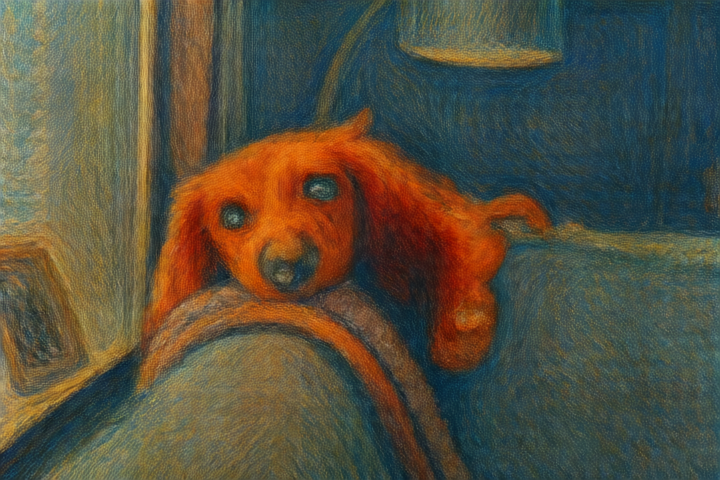} &
\includegraphics[width=2.4cm]{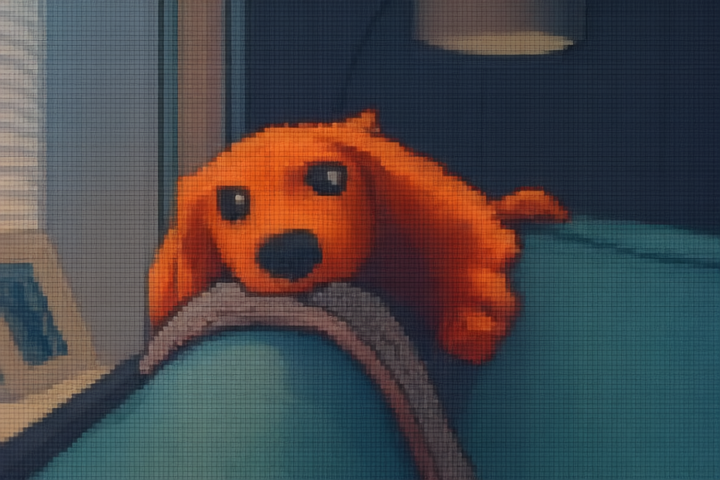} &
\includegraphics[width=2.4cm]{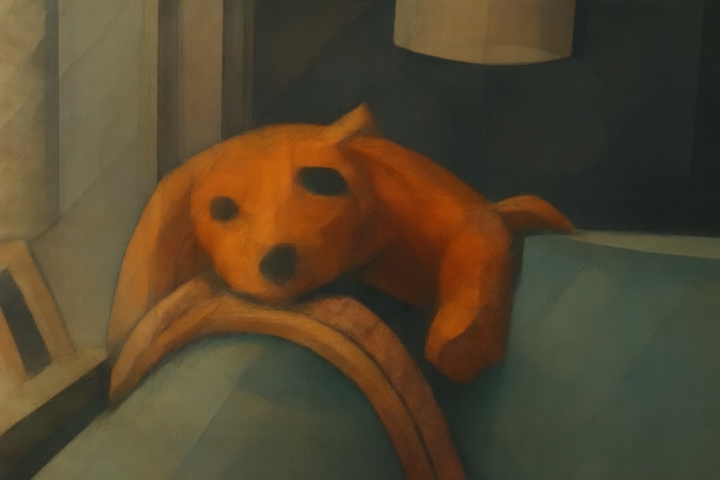} &
\includegraphics[width=2.4cm]{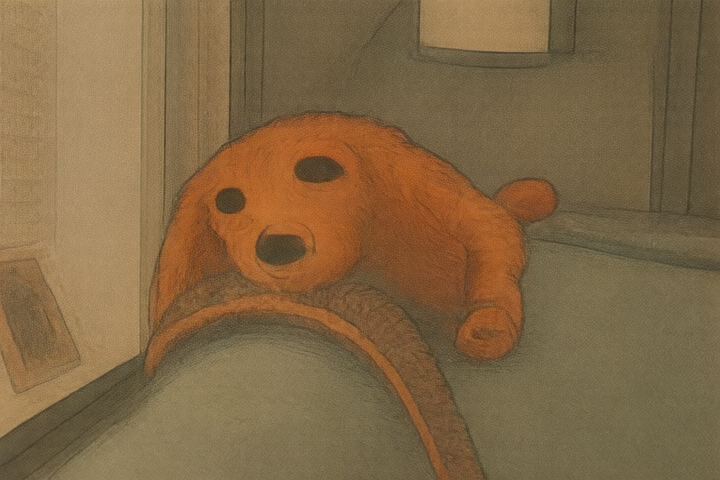} \\

\includegraphics[width=2.4cm]{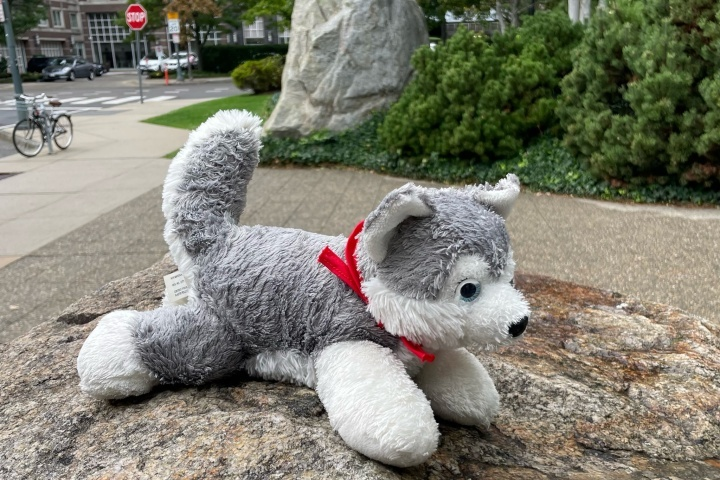} &
\includegraphics[width=2.4cm]{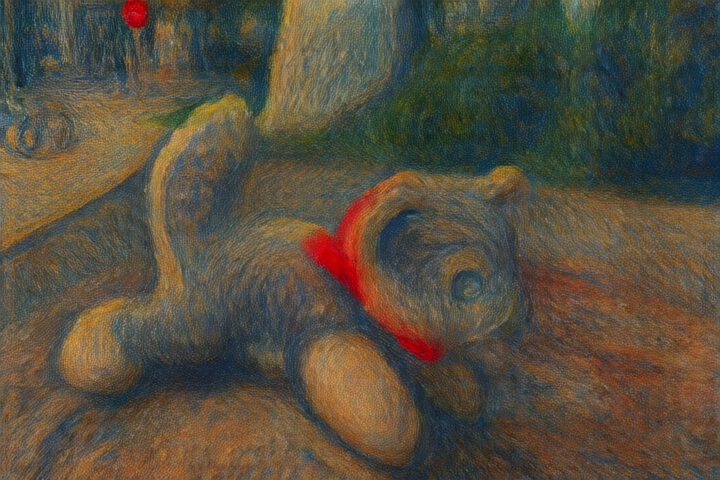} &
\includegraphics[width=2.4cm]{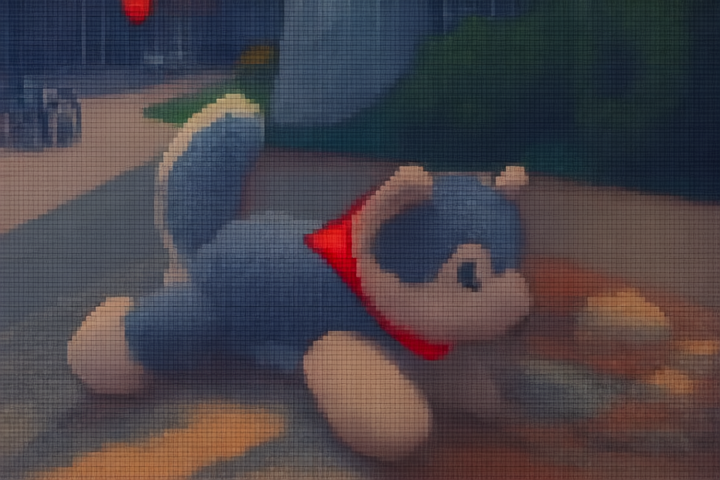} &
\includegraphics[width=2.4cm]{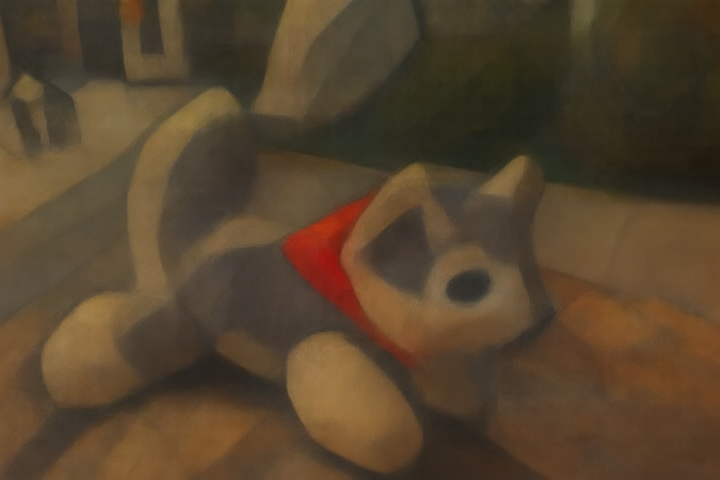} &
\includegraphics[width=2.4cm]{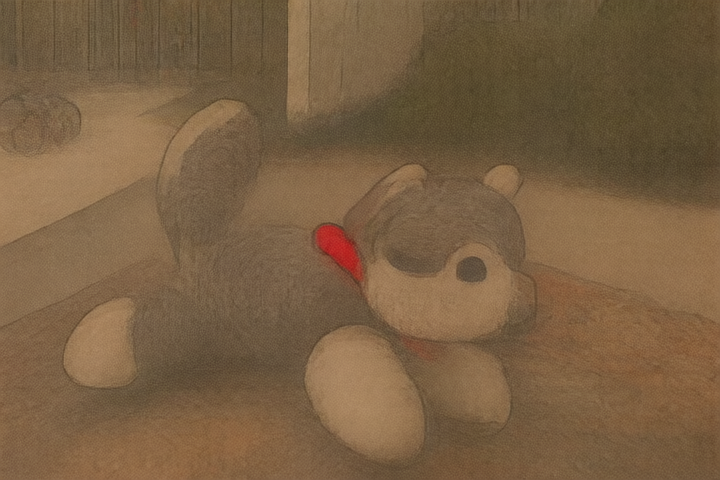} \\

\includegraphics[width=2.4cm]{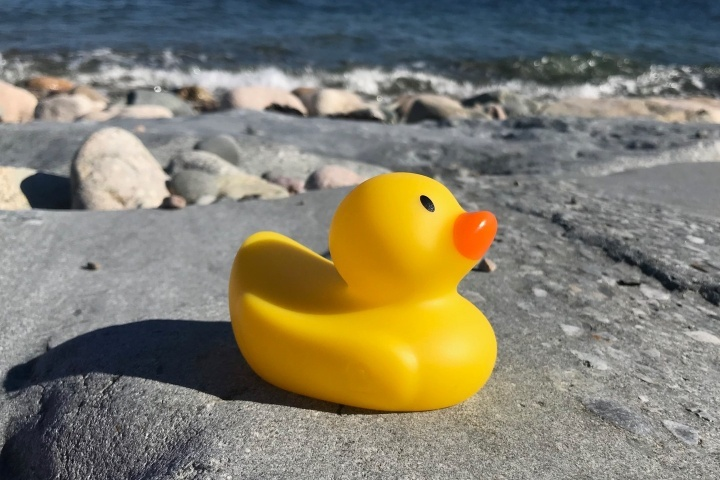} &
\includegraphics[width=2.4cm]{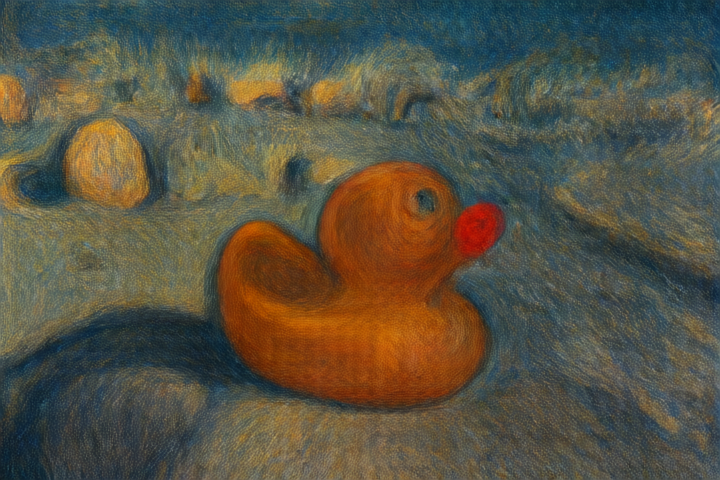} &
\includegraphics[width=2.4cm]{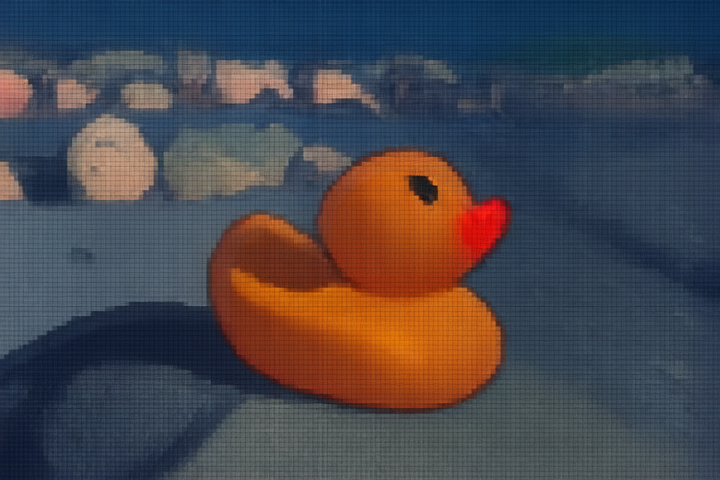} &
\includegraphics[width=2.4cm]{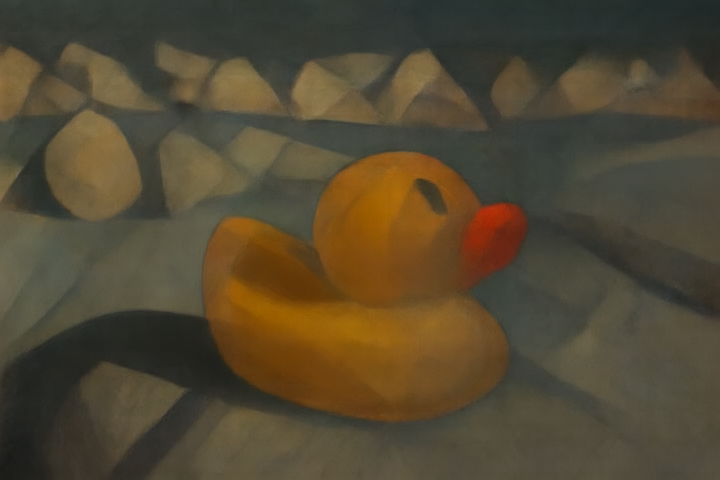} &
\includegraphics[width=2.4cm]{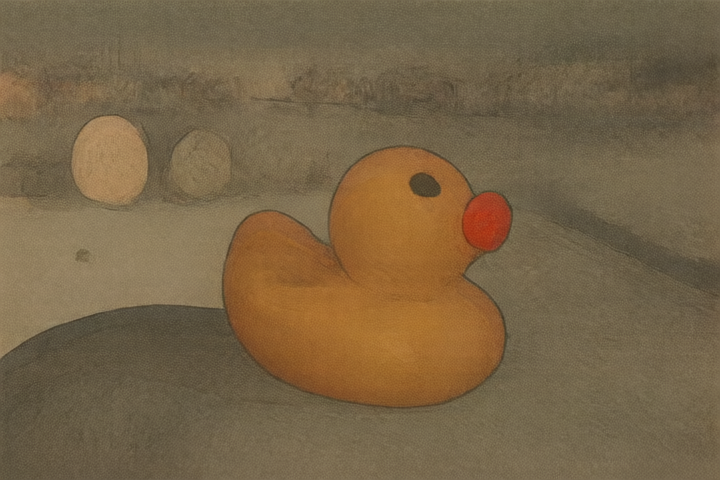} \\

\includegraphics[width=2.4cm]{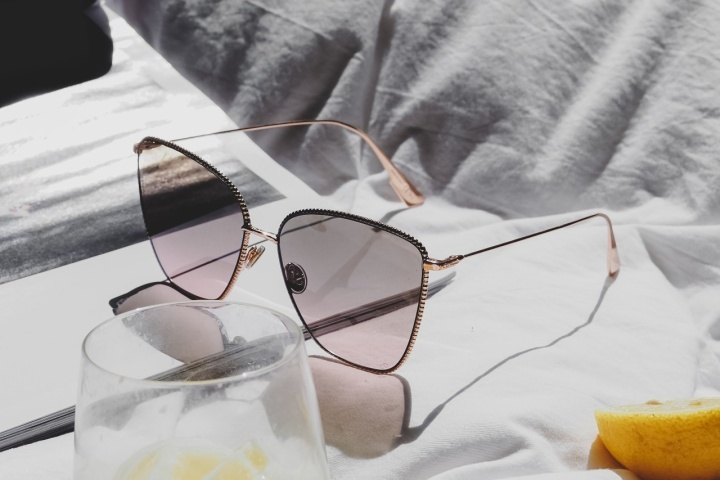} &
\includegraphics[width=2.4cm]{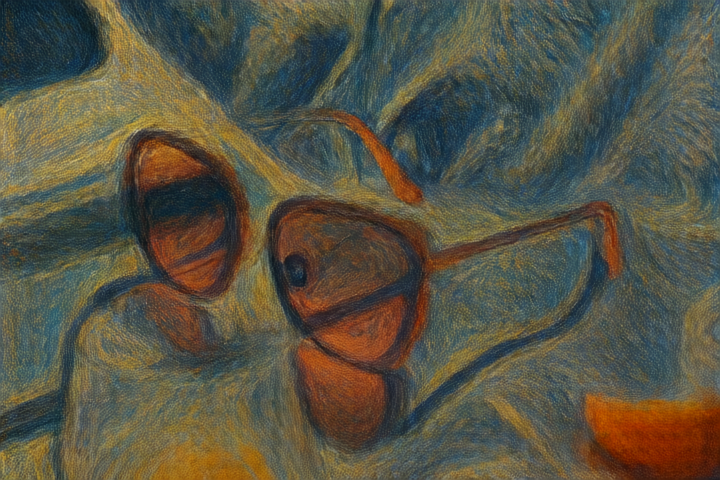} &
\includegraphics[width=2.4cm]{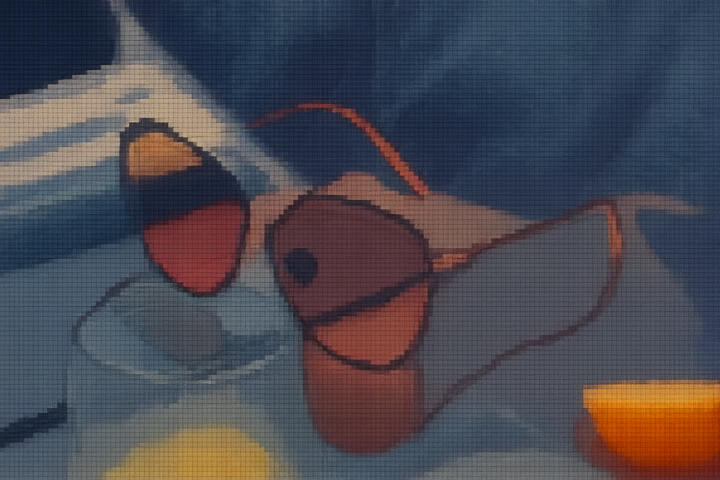} &
\includegraphics[width=2.4cm]{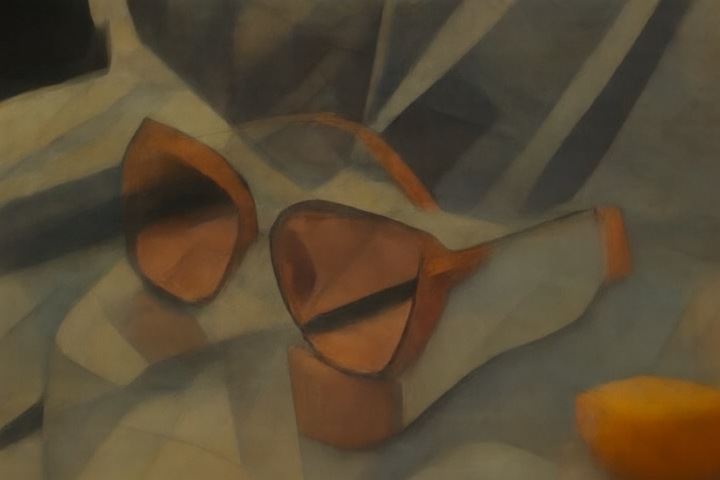} &
\includegraphics[width=2.4cm]{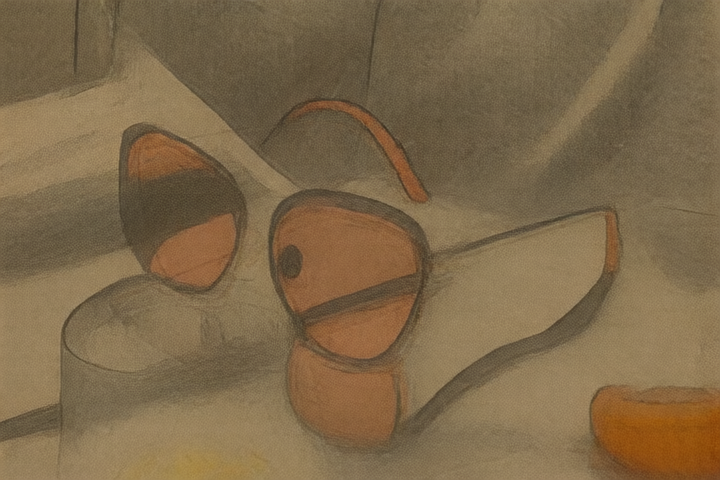} \\

\includegraphics[width=2.4cm]{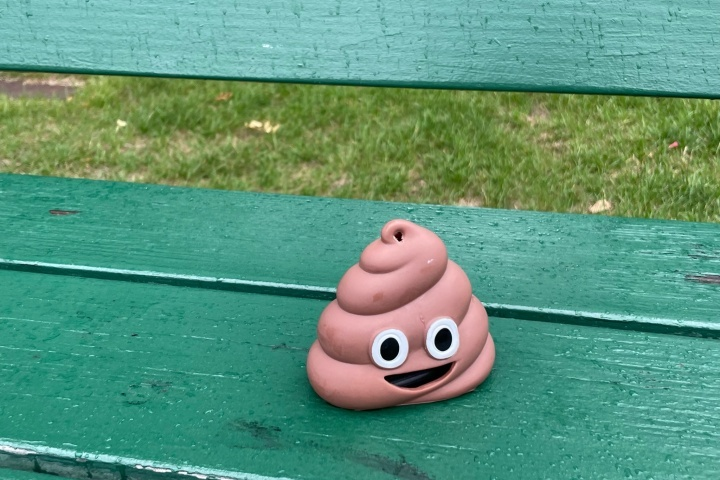} &
\includegraphics[width=2.4cm]{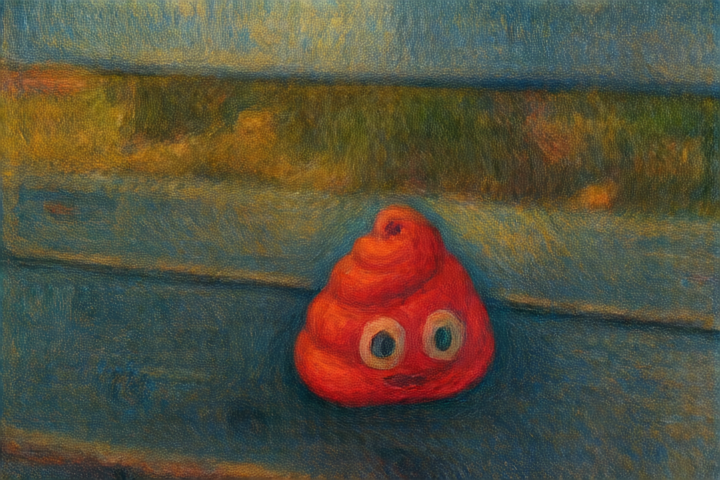} &
\includegraphics[width=2.4cm]{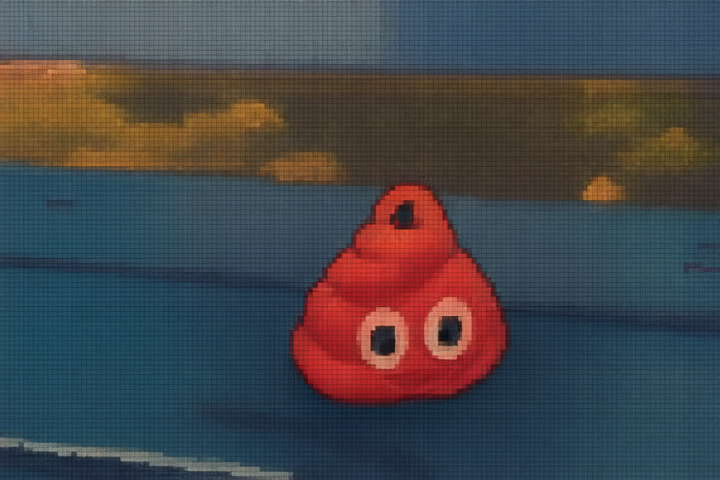} &
\includegraphics[width=2.4cm]{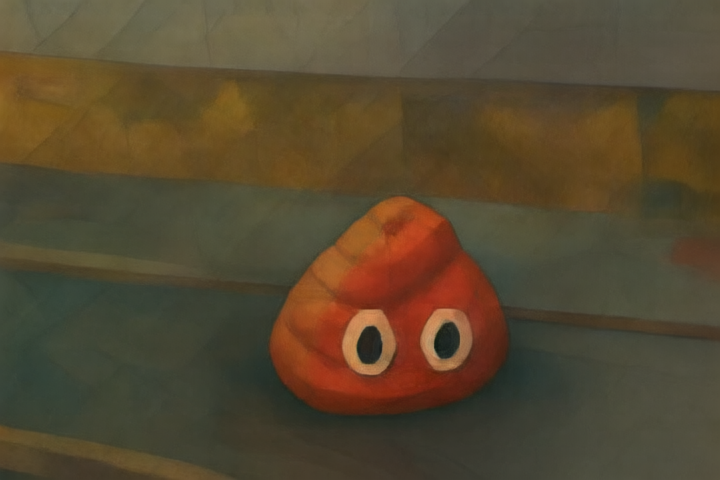} &
\includegraphics[width=2.4cm]{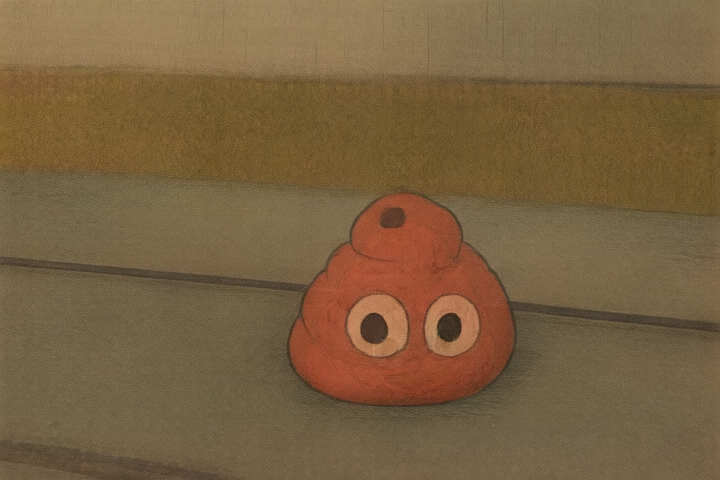} \\
\end{tabular}
\caption{1-shot style transfer results. The model adapts the input images to distinct artistic styles (\textit{Starry Night}, \textit{Pixel Art}, \textit{Cubism}, and \textit{Ukiyo-e}) using only a single reference example.}
\label{fig:appdx:cv:style-transfer}
\end{figure}

Figure \ref{fig:appdx:cv:geometric-transforms} illustrates that the model can capture geometric transformations under extreme few-shot conditions. We further show one-shot style transfer in Figure \ref{fig:appdx:cv:style-transfer}.

\begin{figure}[ht]
\centering
\begin{tabular}{cc@{\hskip 0.5cm}cc}
\textbf{Input} & \makecell{\textbf{Binary Segmentation}\\\textbf{Prediction}} & \textbf{Input} & \textbf{Pose Prediction} \\
\includegraphics[width=3cm]{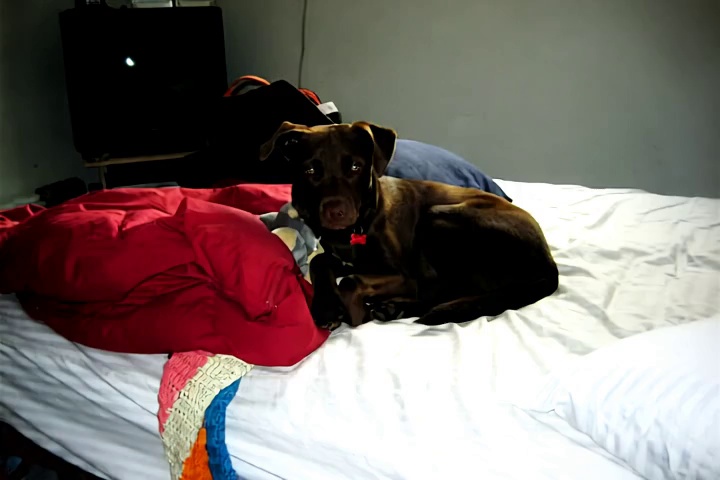} &
\includegraphics[width=3cm]{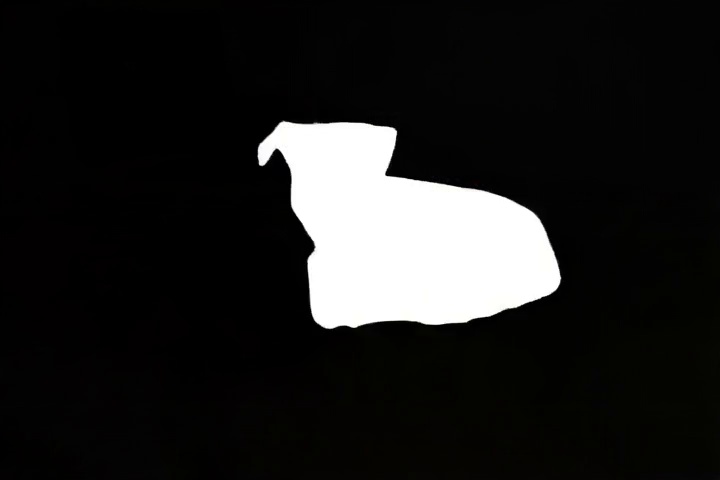} &
\includegraphics[width=3cm]{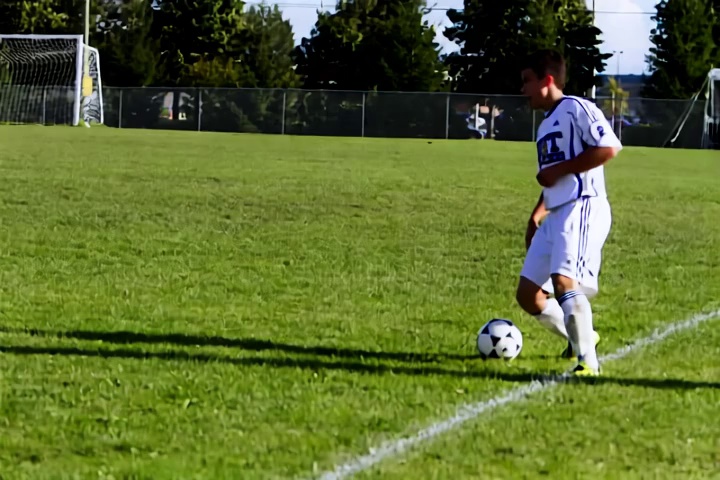} &
\includegraphics[width=3cm]{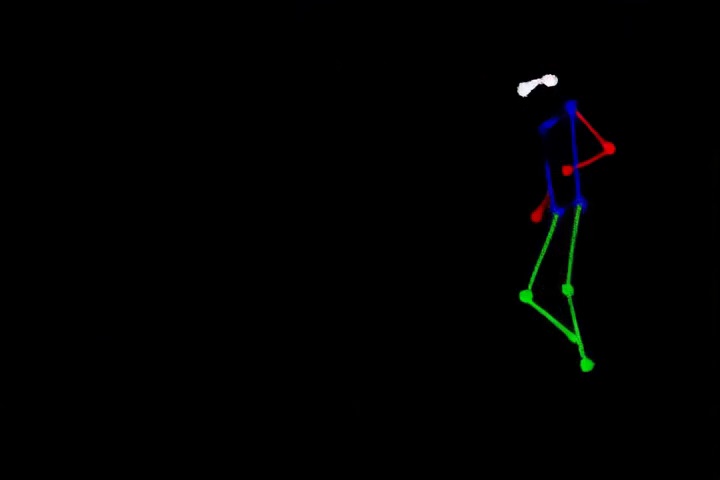}\\
\includegraphics[width=3cm]{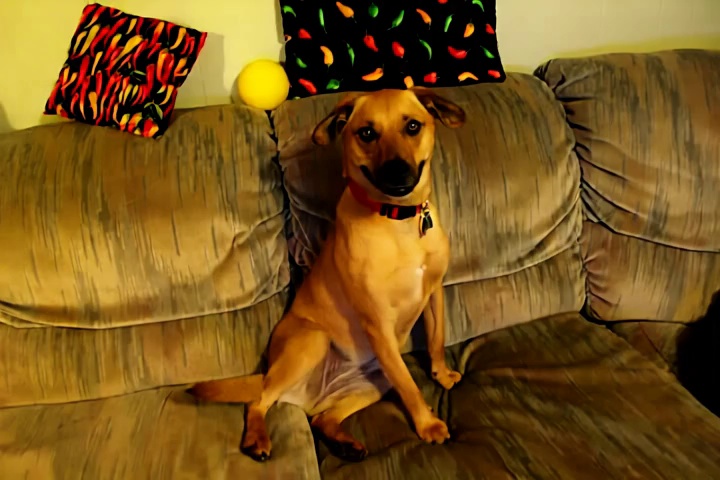} &
\includegraphics[width=3cm]{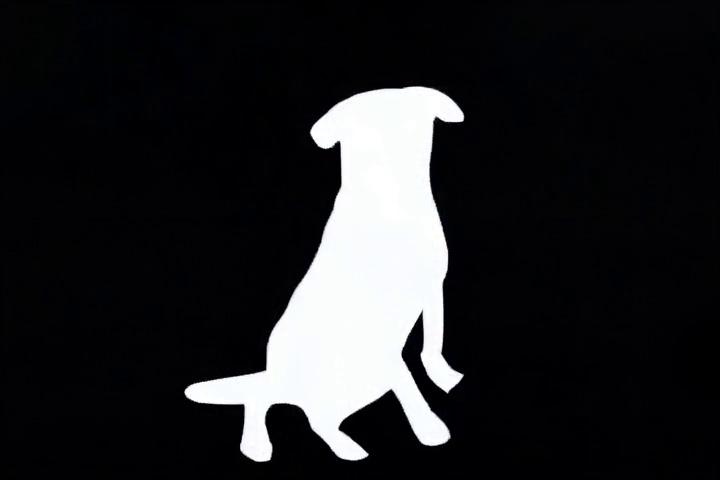} &
\includegraphics[width=3cm]{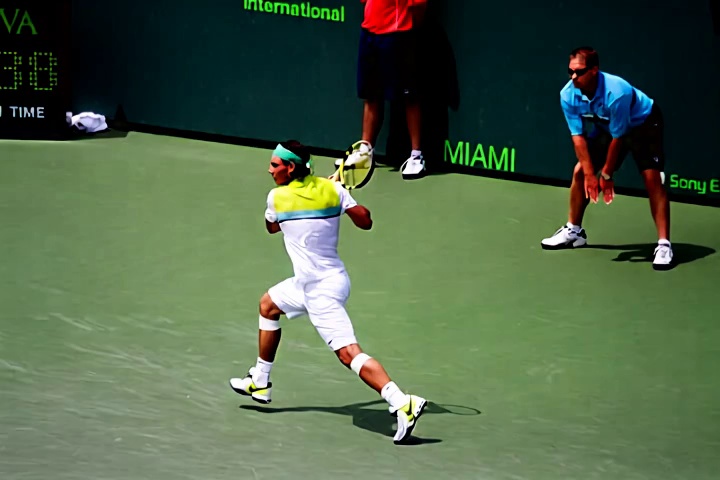} &
\includegraphics[width=3cm]{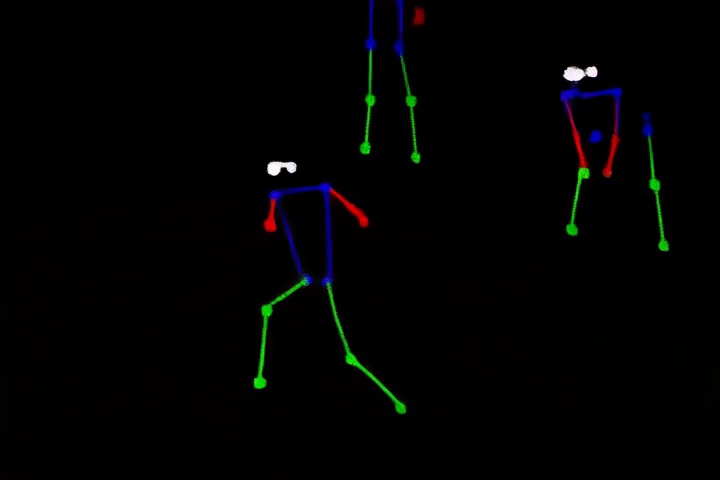}\\
\includegraphics[width=3cm]{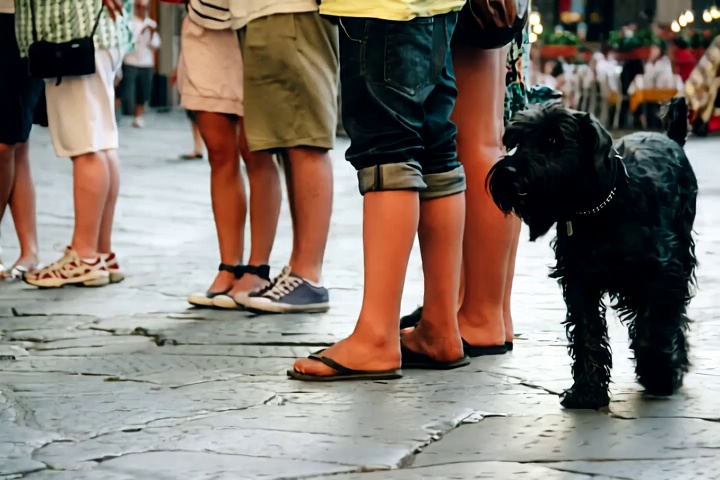} &
\includegraphics[width=3cm]{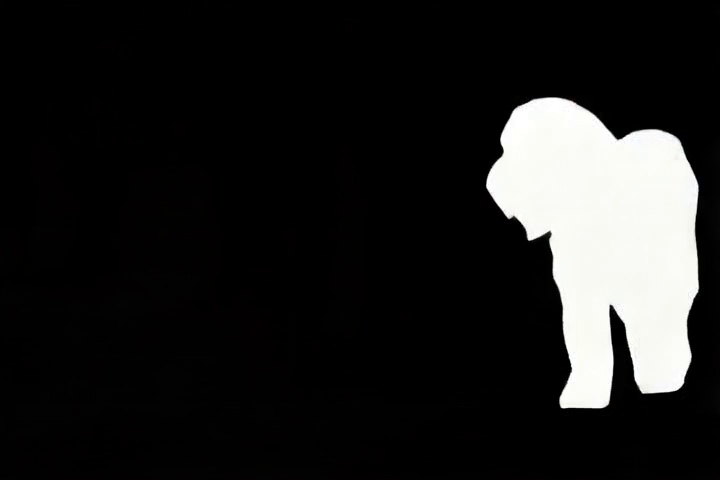} &
\includegraphics[width=3cm]{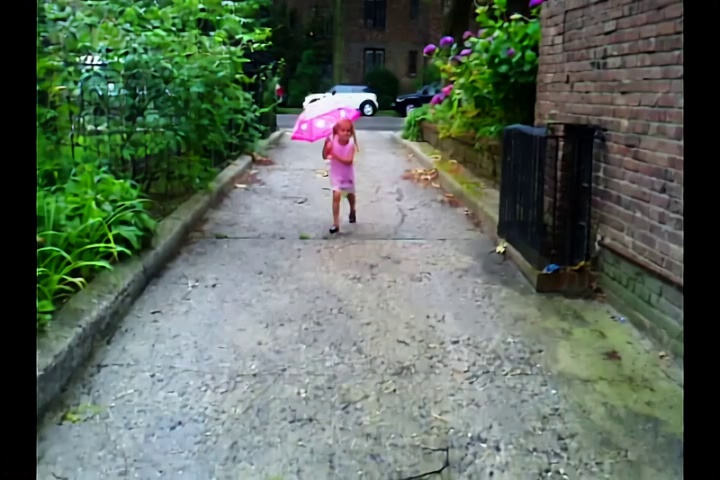} &
\includegraphics[width=3cm]{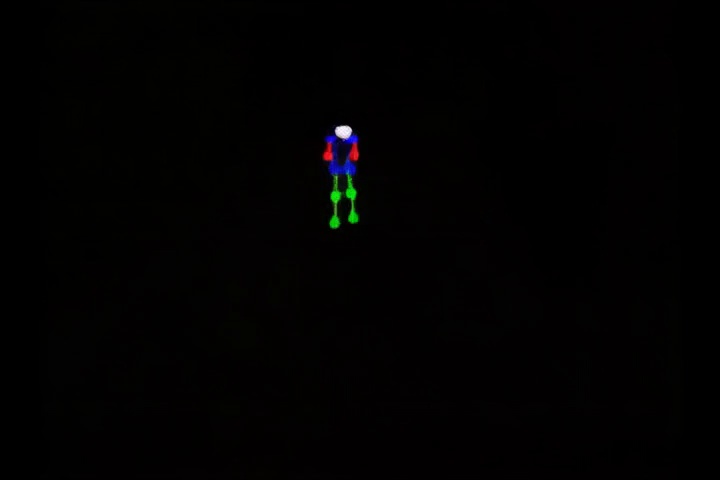}\\
\includegraphics[width=3cm]{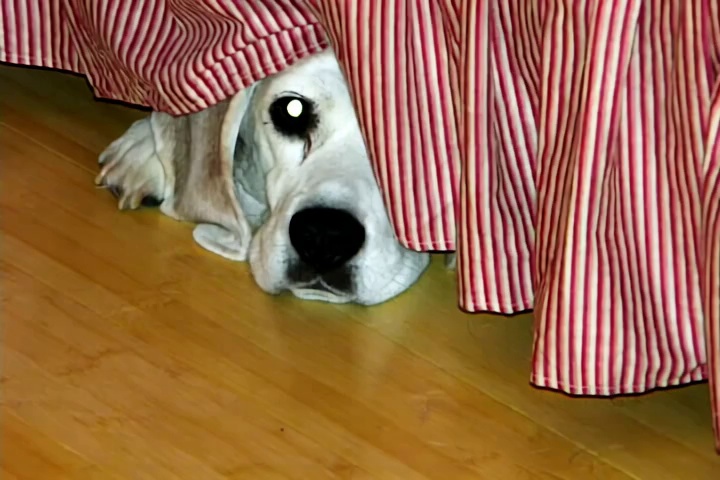} &
\includegraphics[width=3cm]{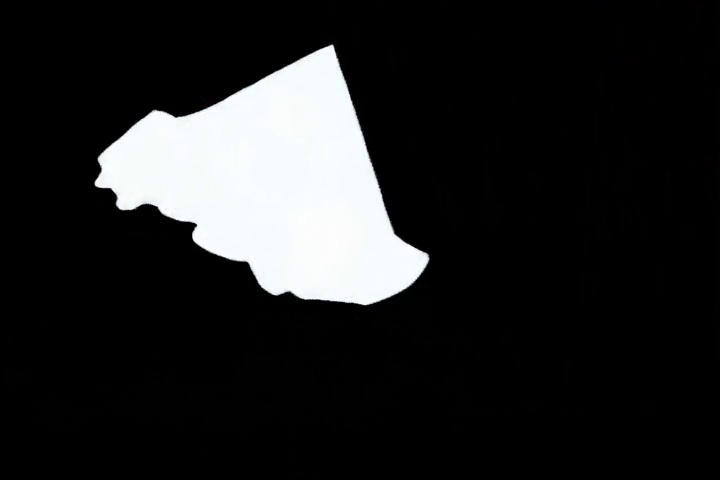} &
\includegraphics[width=3cm]{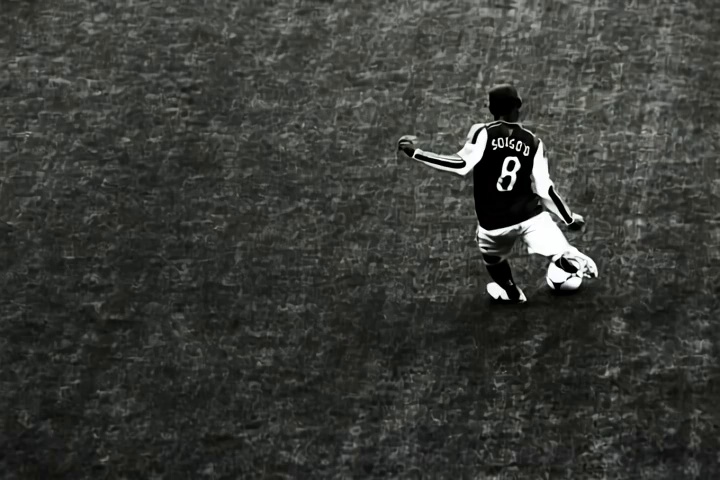} &
\includegraphics[width=3cm]{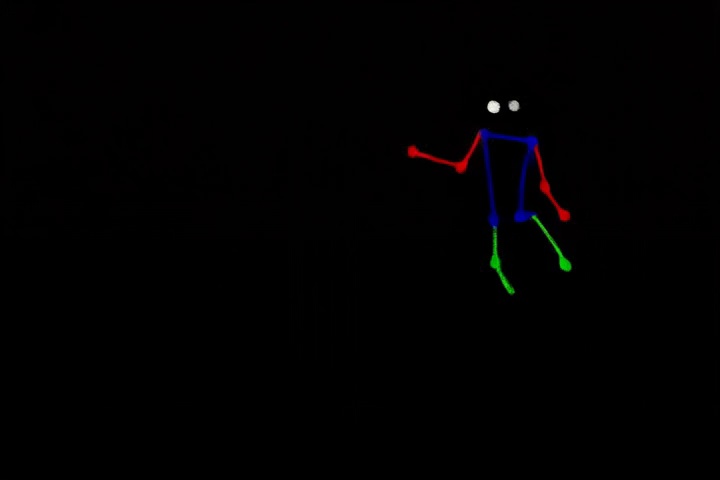}\\
\includegraphics[width=3cm]{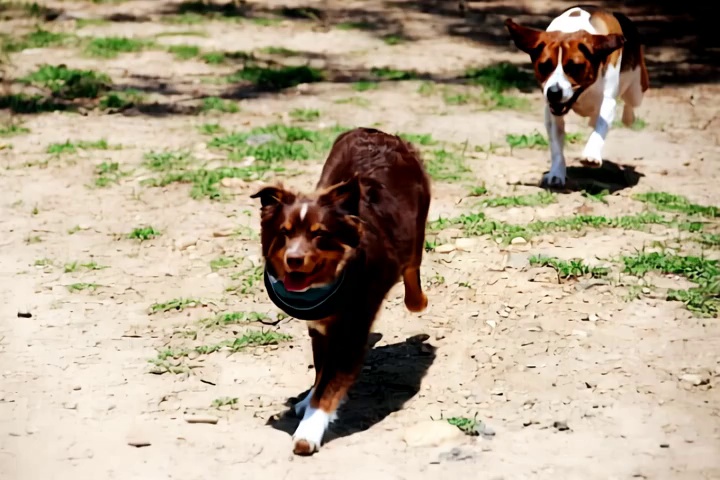} &
\includegraphics[width=3cm]{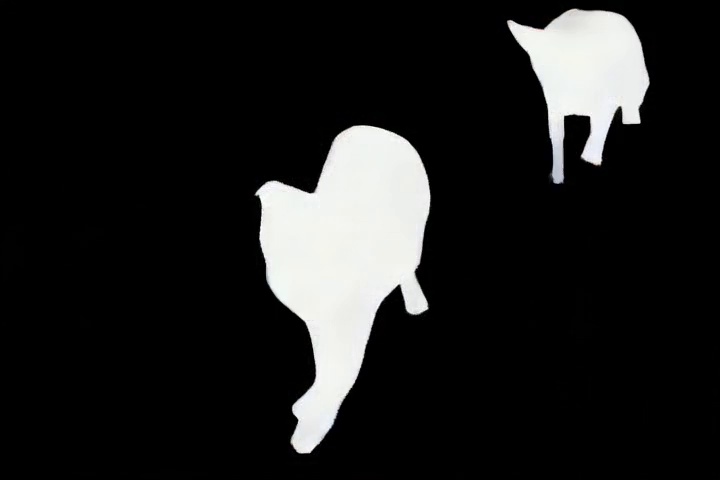} &
\includegraphics[width=3cm]{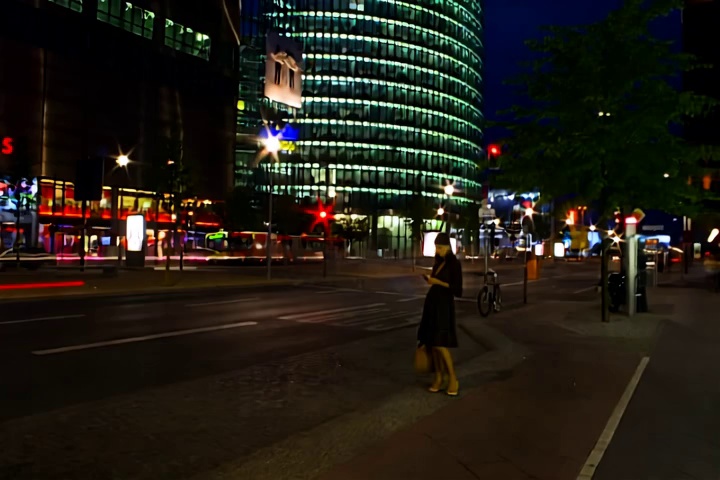} &
\includegraphics[width=3cm]{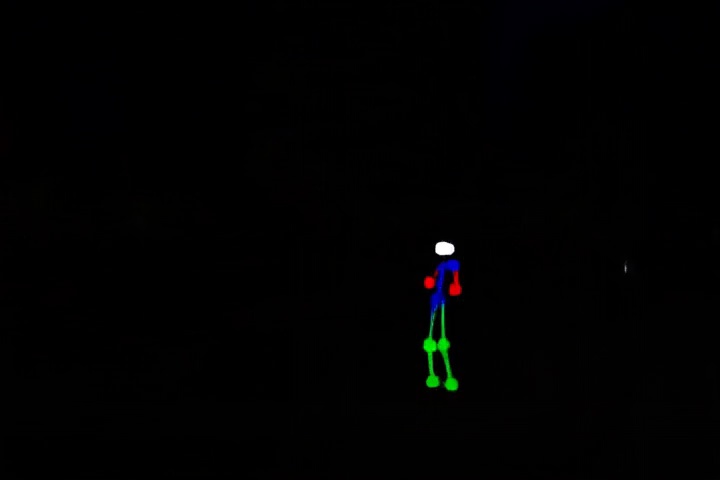}\\
\includegraphics[width=3cm]{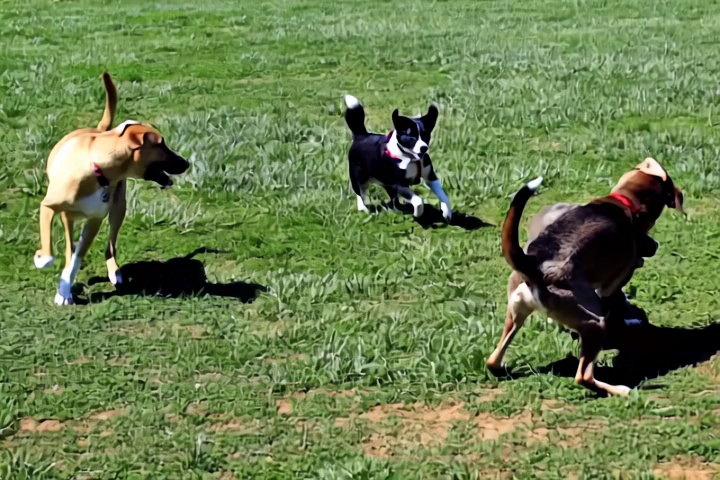} &
\includegraphics[width=3cm]{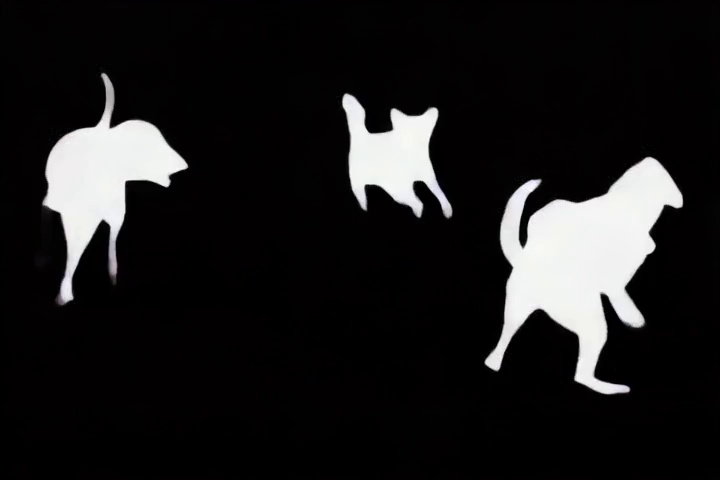} &
\includegraphics[width=3cm]{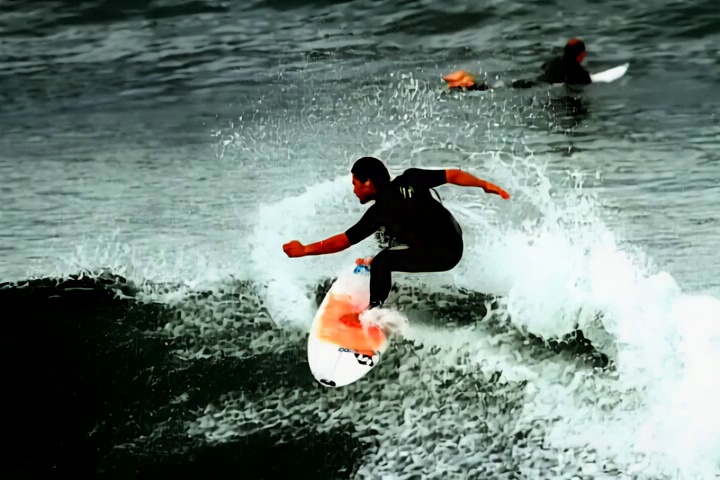} &
\includegraphics[width=3cm]{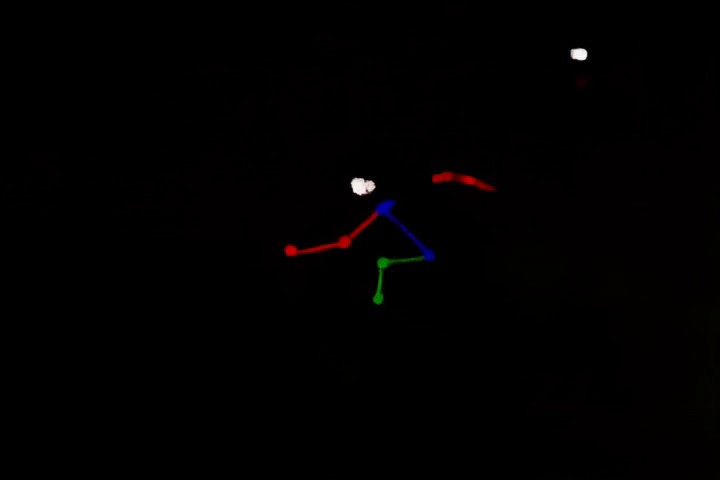}\\
\includegraphics[width=3cm]{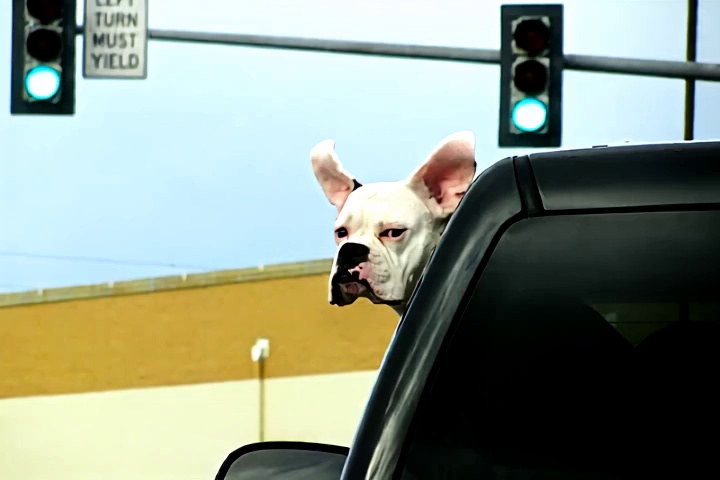} &
\includegraphics[width=3cm]{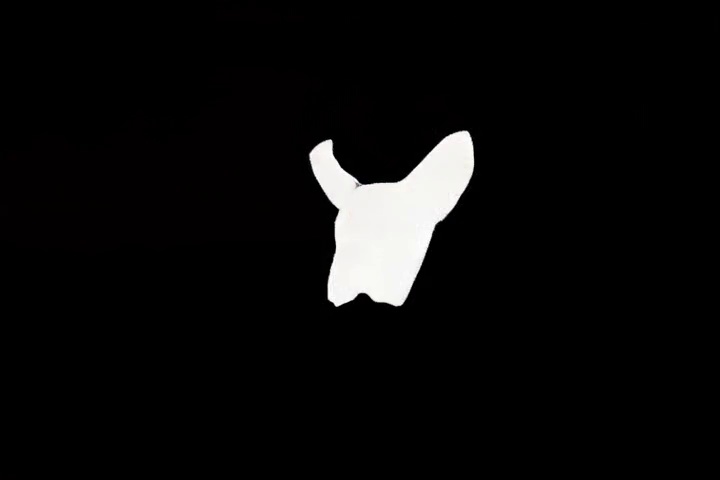} &
\includegraphics[width=3cm]{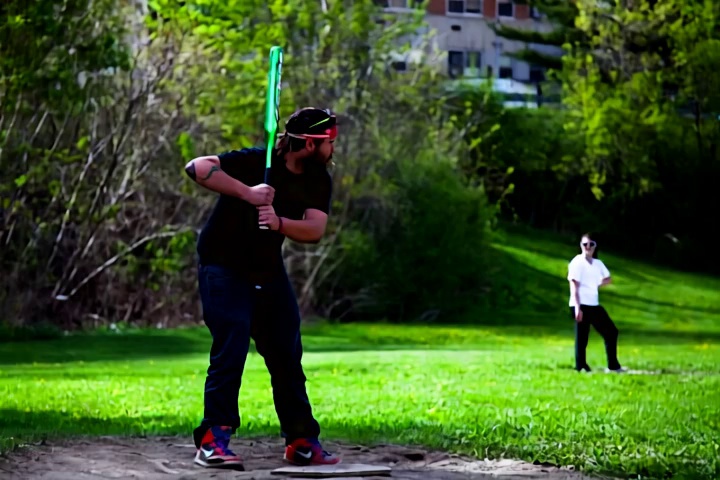} &
\includegraphics[width=3cm]{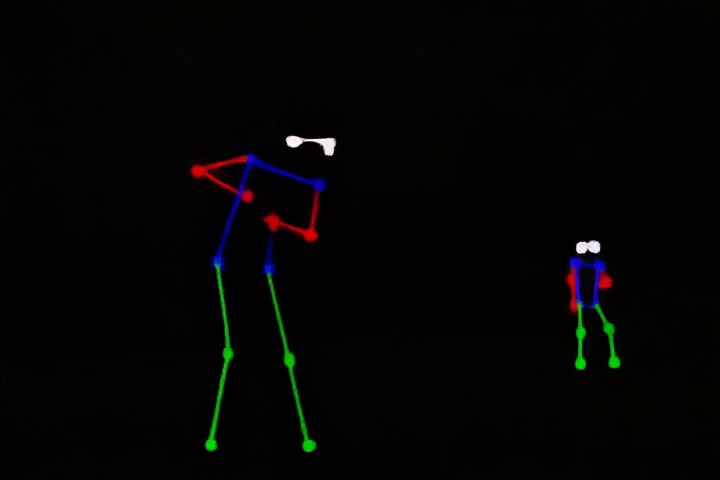}\\
\includegraphics[width=3cm]{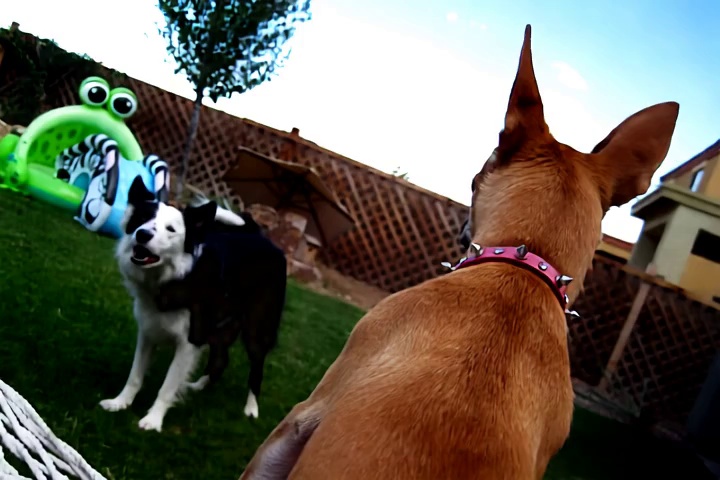} &
\includegraphics[width=3cm]{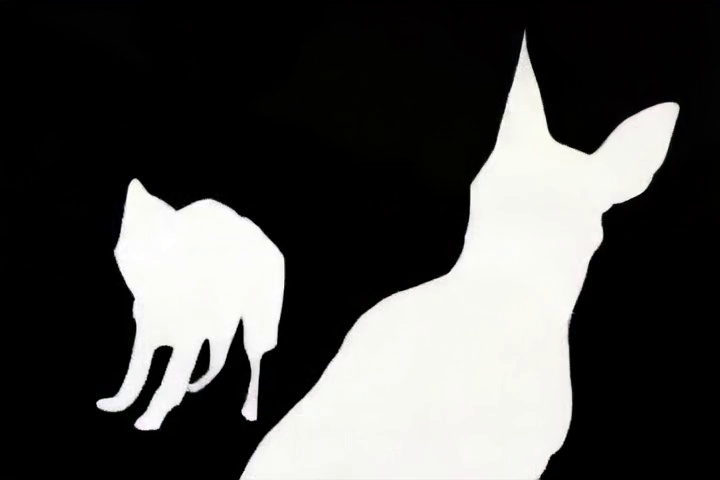} &
\includegraphics[width=3cm]{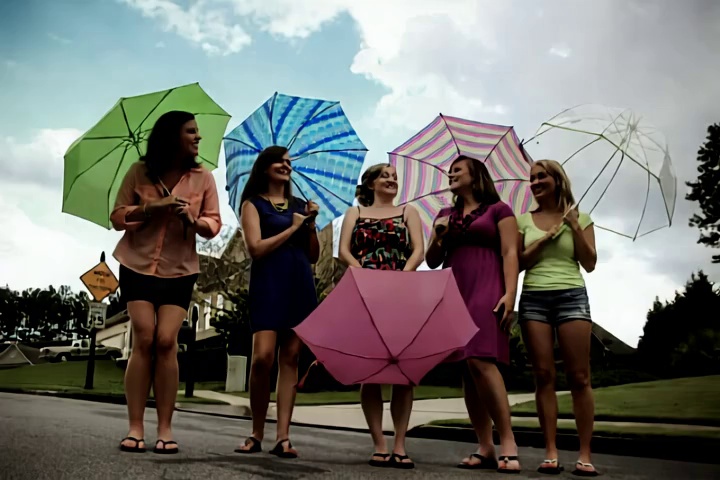} &
\includegraphics[width=3cm]{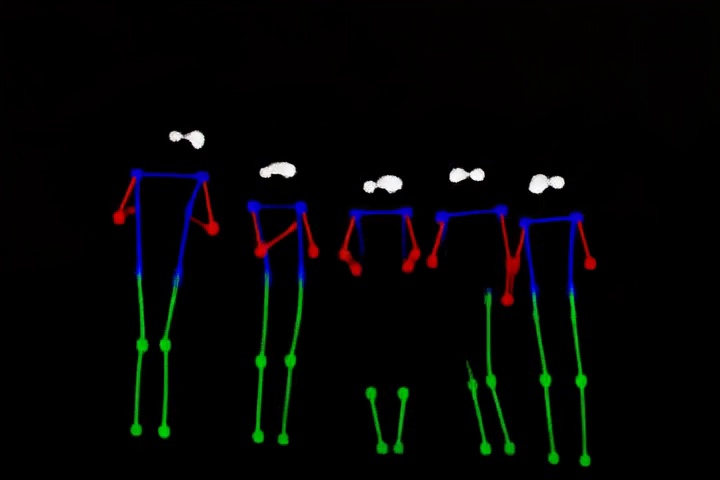}\\
\includegraphics[width=3cm]{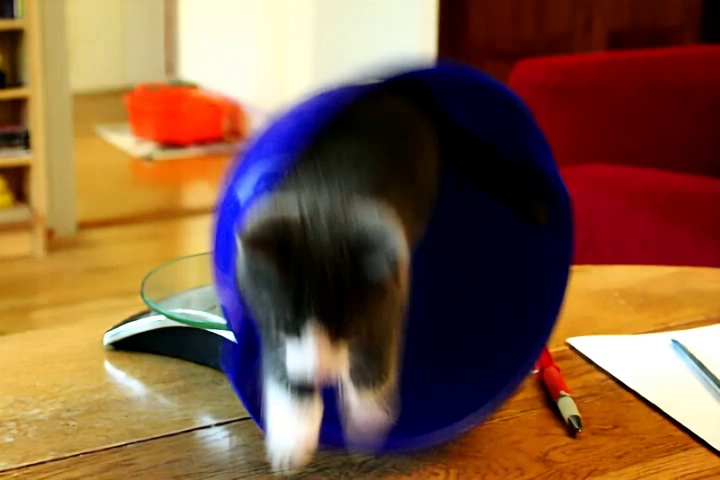} &
\includegraphics[width=3cm]{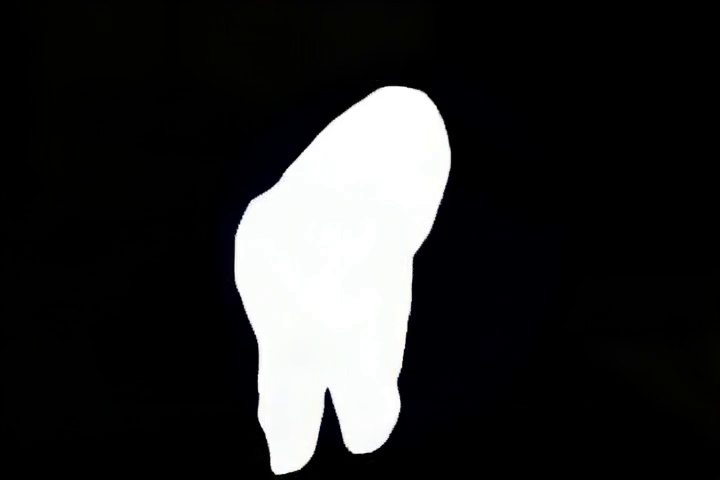} &
\includegraphics[width=3cm]{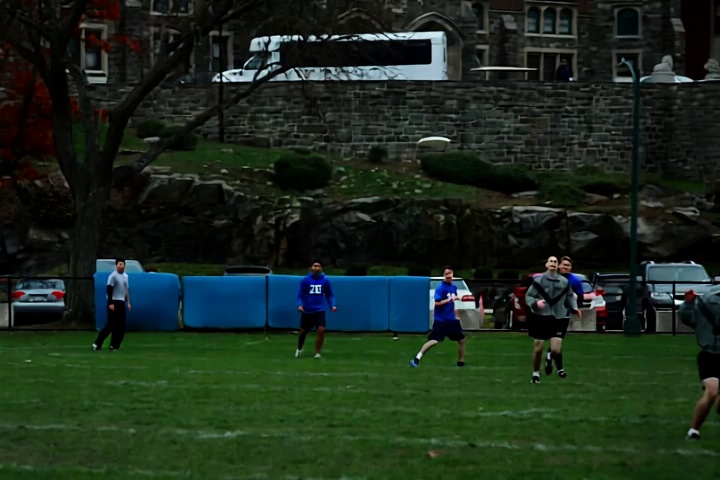} &
\includegraphics[width=3cm]{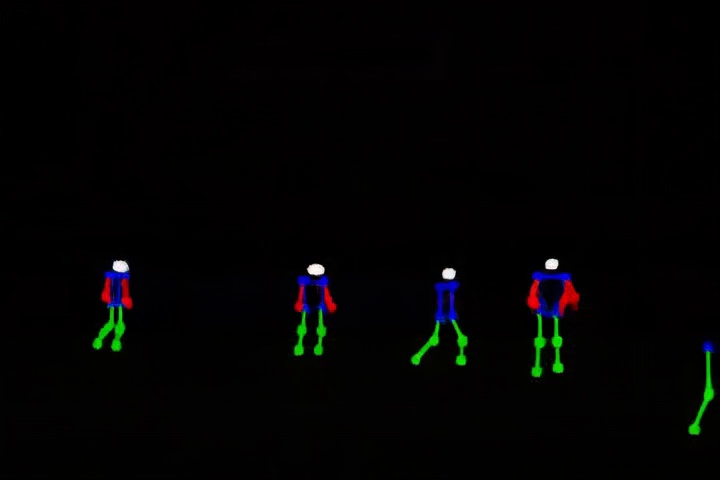}\\
\end{tabular}
\caption{Predictions after finetuning with \(n=30\) samples for \textit{Binary Segmentation} and \textit{Pose}.}
\label{fig:appdx:cv:seg-and-pose}
\end{figure}

We also qualitative show this framework can be used to solve some classical computer vision tasks. In Figure \ref{fig:appdx:cv:seg-and-pose} we show examples after training with only \(n=30\) samples for \textit{Binary Segmentation} for dogs and \textit{Pose} estimation for humans.

\begin{figure}[ht]
\centering
\begin{tabular}{lccccc}
& \textbf{Input} & \(\mathbf{n=1}\) & \(\mathbf{n=3}\) & \(\mathbf{n=5}\) & \(\mathbf{n=10}\) \\
\raisebox{0.4cm}{\rotatebox{90}{\textbf{Inpainting}}} &
\includegraphics[width=2.4cm]{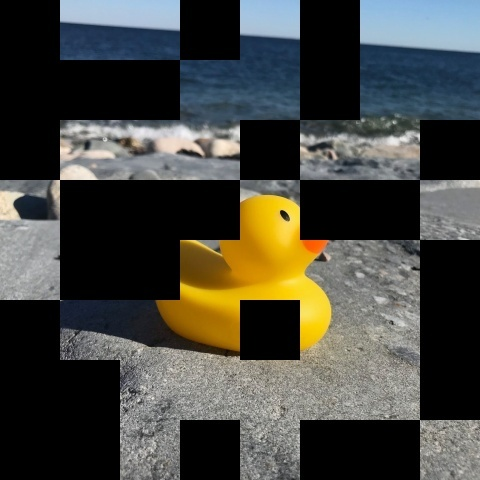} &
\includegraphics[width=2.4cm]{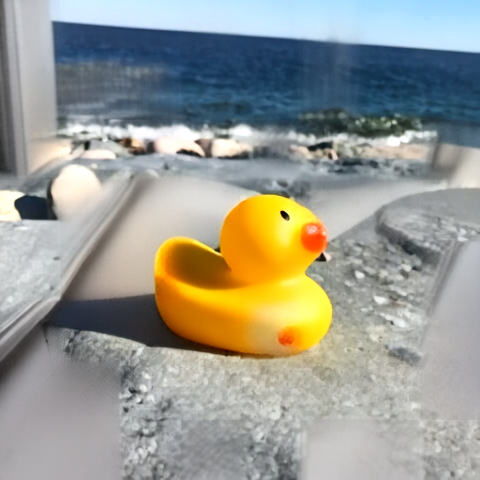} &
\includegraphics[width=2.4cm]{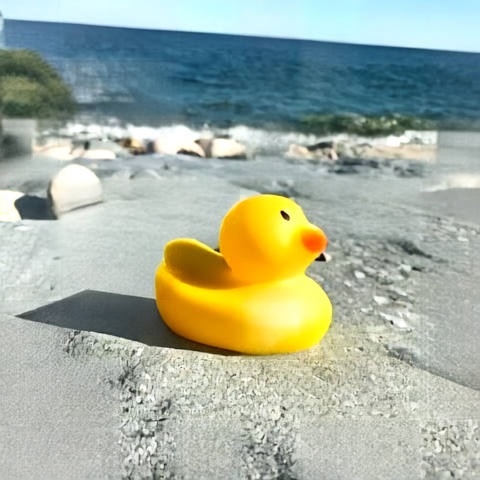} &
\includegraphics[width=2.4cm]{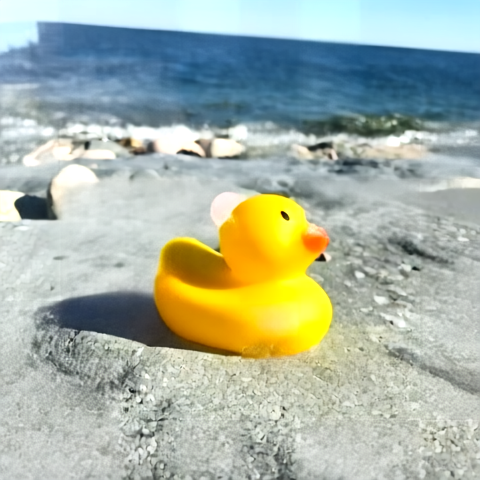} &
\includegraphics[width=2.4cm]{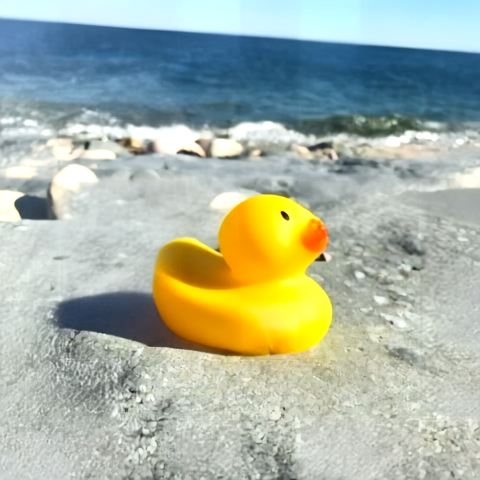} \\[0.5em]

\raisebox{0.25cm}{\rotatebox{90}{\textbf{Colorization}}} &
\includegraphics[width=2.4cm]{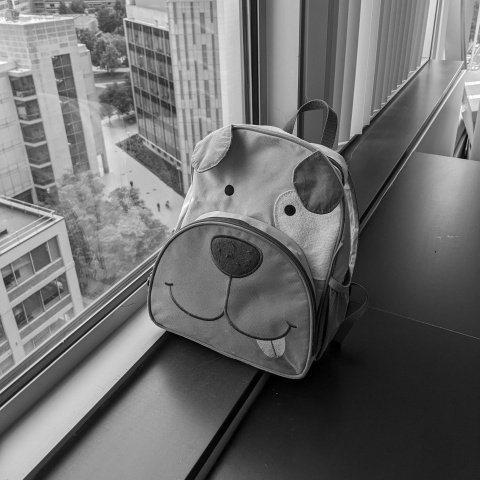} &
\includegraphics[width=2.4cm]{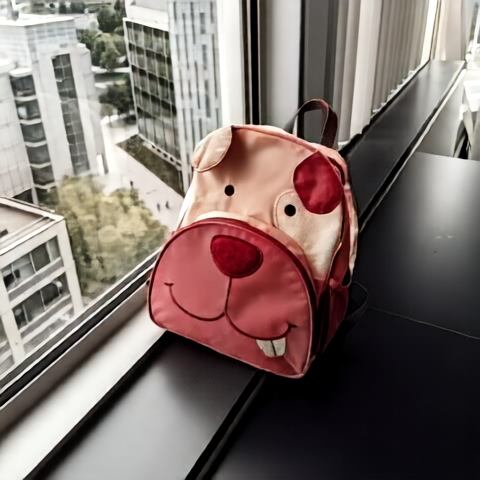} &
\includegraphics[width=2.4cm]{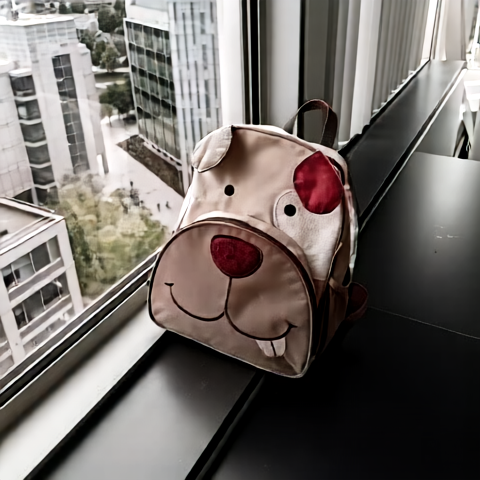} &
\includegraphics[width=2.4cm]{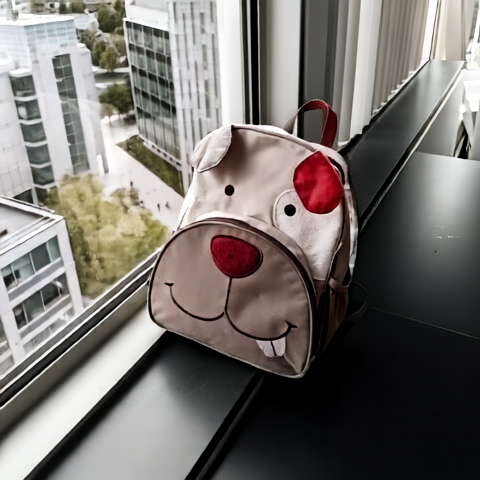} &
\includegraphics[width=2.4cm]{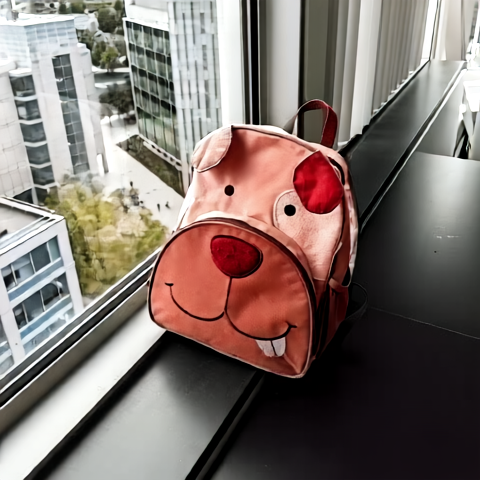} \\[0.5em]

\raisebox{0.6cm}{\rotatebox{90}{\textbf{Jigsaw}}} &
\includegraphics[width=2.4cm]{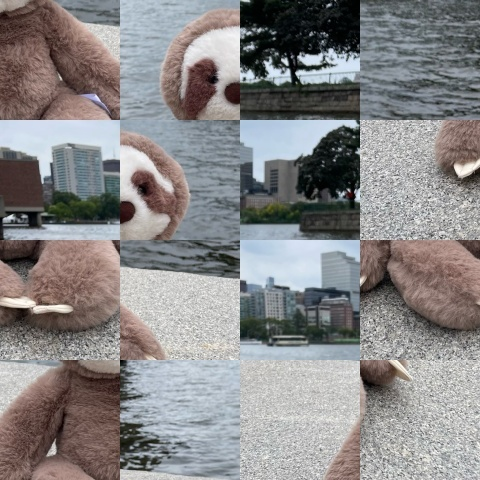} &
\includegraphics[width=2.4cm]{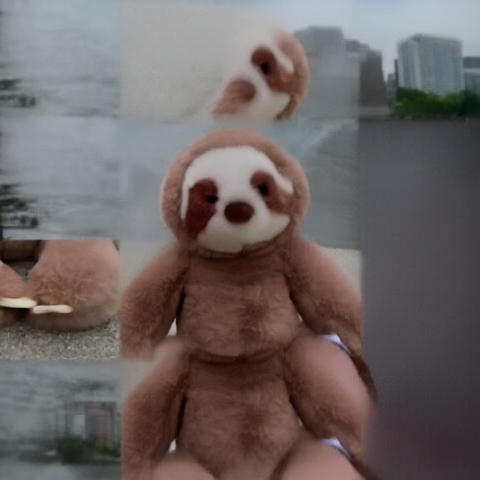} &
\includegraphics[width=2.4cm]{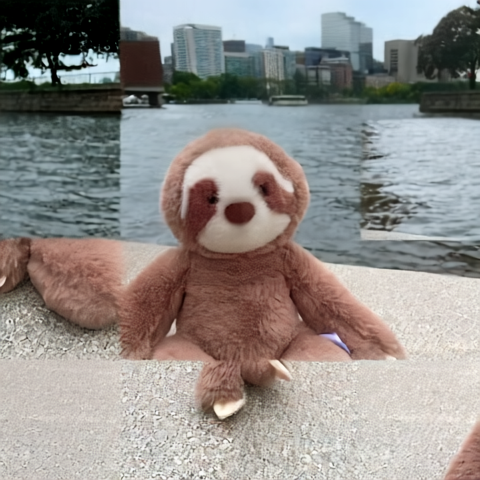} &
\includegraphics[width=2.4cm]{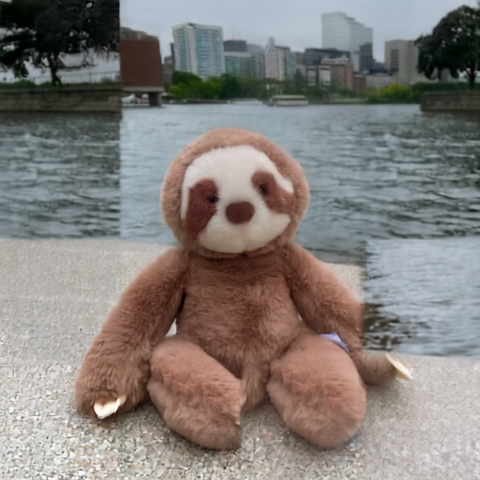} &
\includegraphics[width=2.4cm]{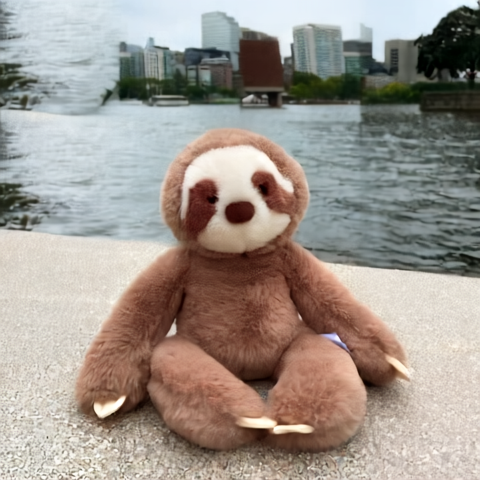} \\
\end{tabular}
\caption{Qualitative results for different tasks (\textit{Inpainting}, \textit{Colorization}, \textit{Jigsaw}) with different numbers of training examples.}
\label{fig:appdx:cv:misc-with-n}
\end{figure}

\begin{figure}[ht]
\centering
\begin{tabular}{cc}
\textbf{Input} & \textbf{Depth Prediction} \\
\includegraphics[width=3cm]{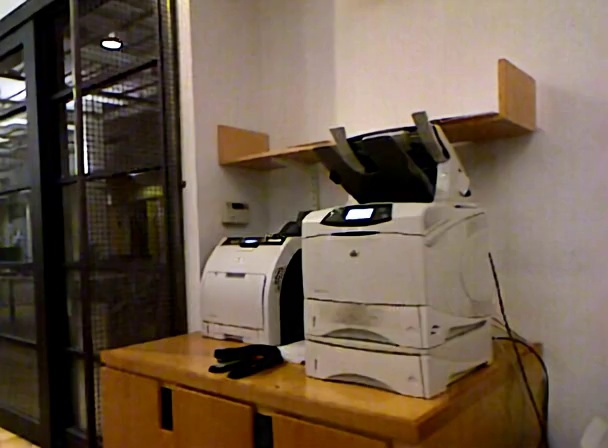} &
\includegraphics[width=3cm]{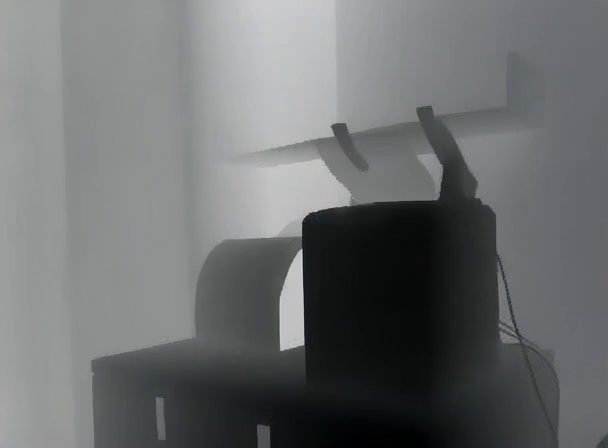} \\
\includegraphics[width=3cm]{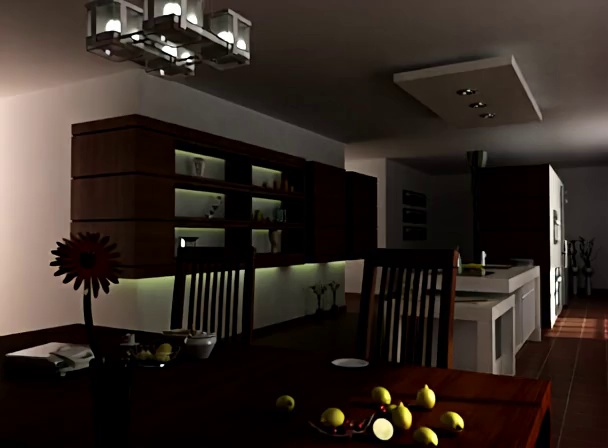} &
\includegraphics[width=3cm]{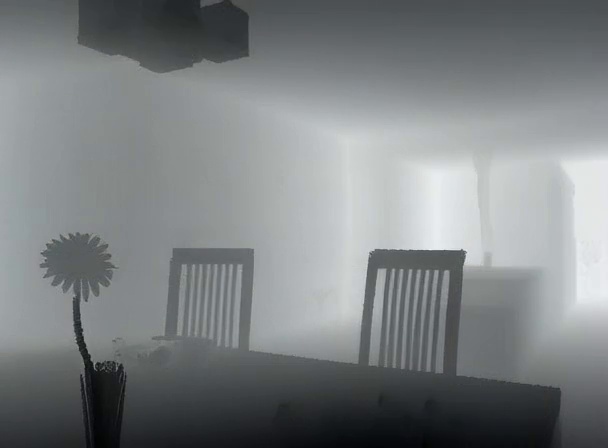} \\
\includegraphics[width=3cm]{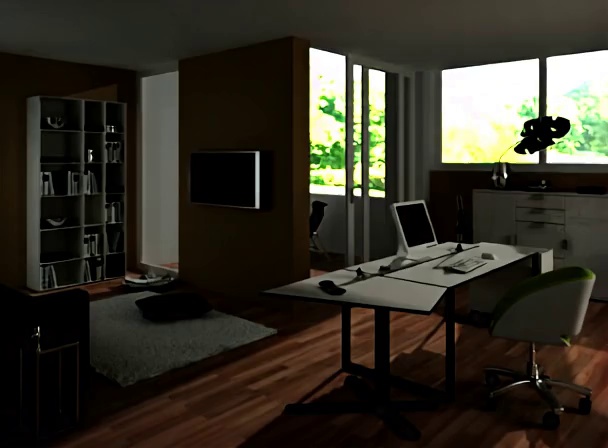} &
\includegraphics[width=3cm]{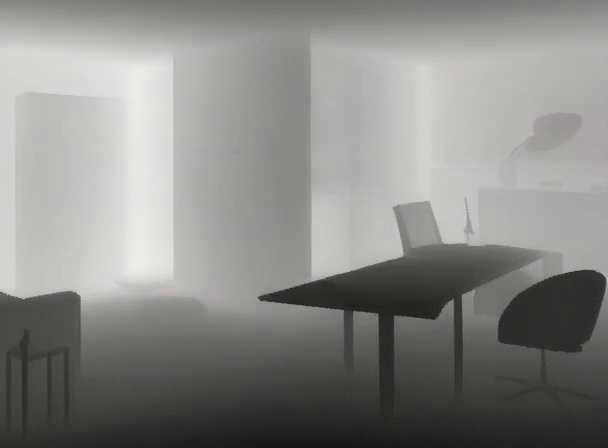} \\
\includegraphics[width=3cm]{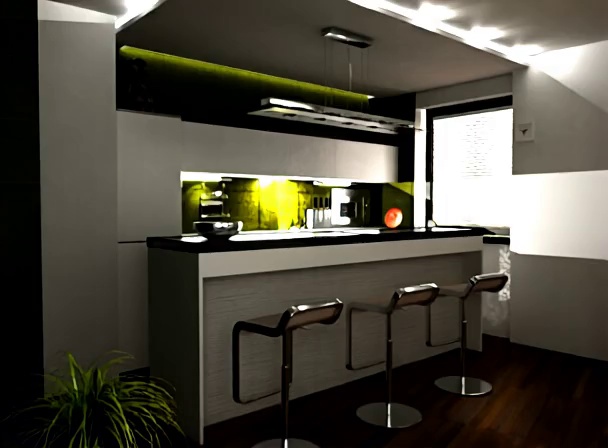} &
\includegraphics[width=3cm]{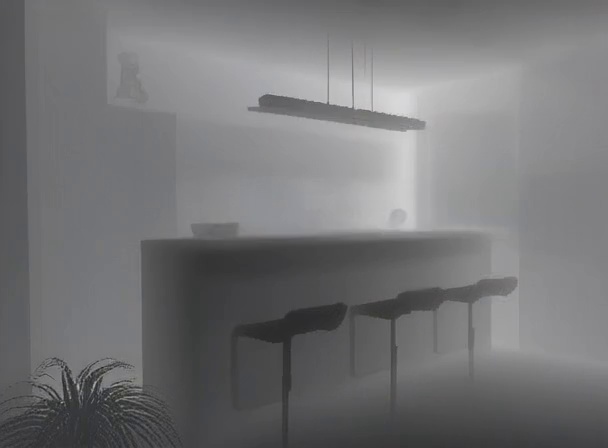} \\
\includegraphics[width=3cm]{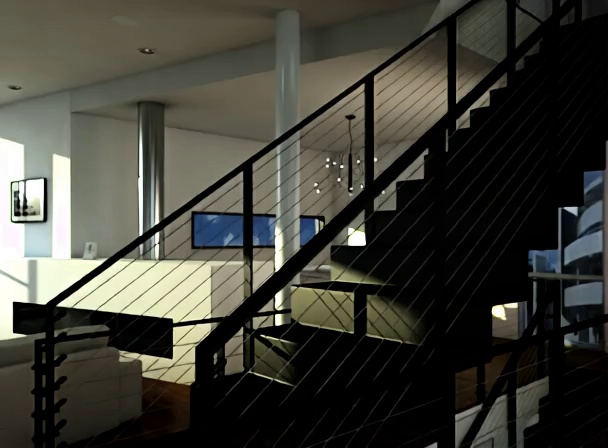} &
\includegraphics[width=3cm]{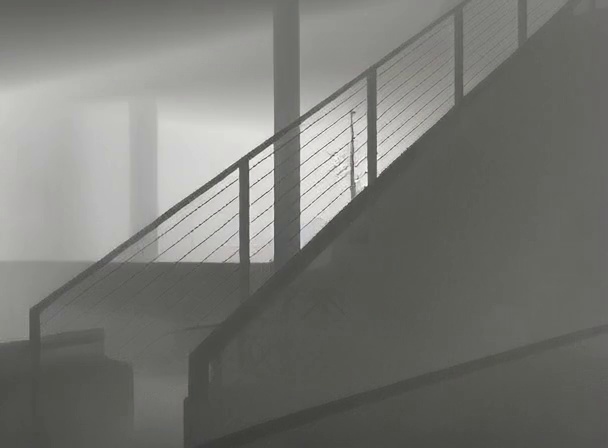} \\
\end{tabular}
\caption{Predictions after finetuning with \(n=30\) samples for \textit{Depth}.}
\label{fig:appdx:cv:depth}
\end{figure}

\begin{figure}[ht]
\centering
\begin{tabular}{cc}
\textbf{Input Image} & \textbf{Segmentation Prediction} \\
\includegraphics[width=3cm]{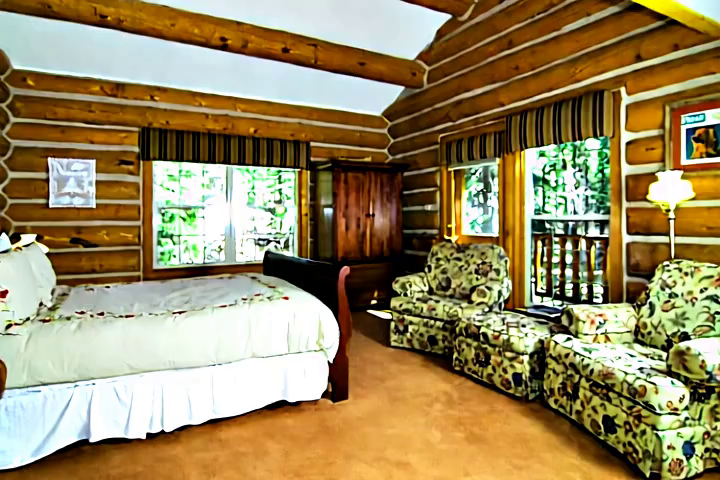} &
\includegraphics[width=3cm]{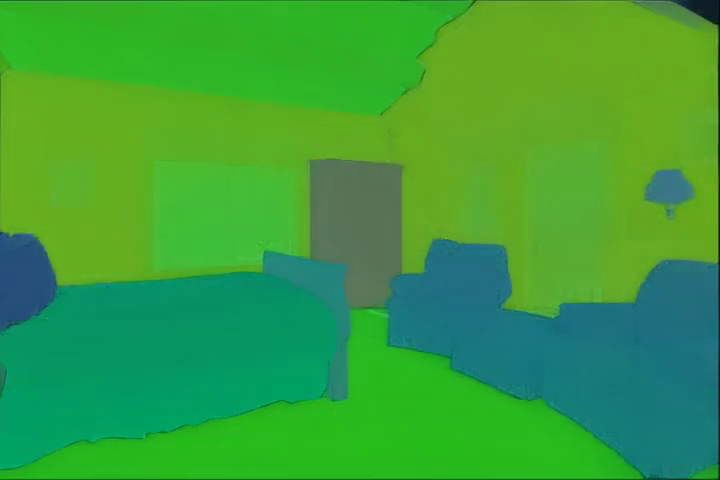} \\
\includegraphics[width=3cm]{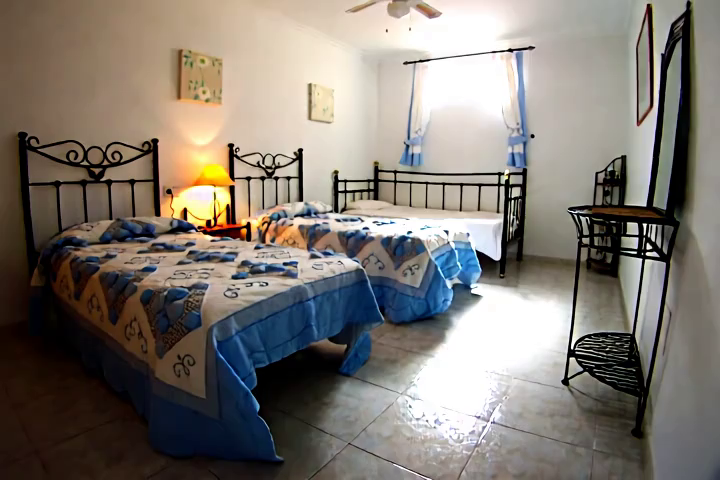} &
\includegraphics[width=3cm]{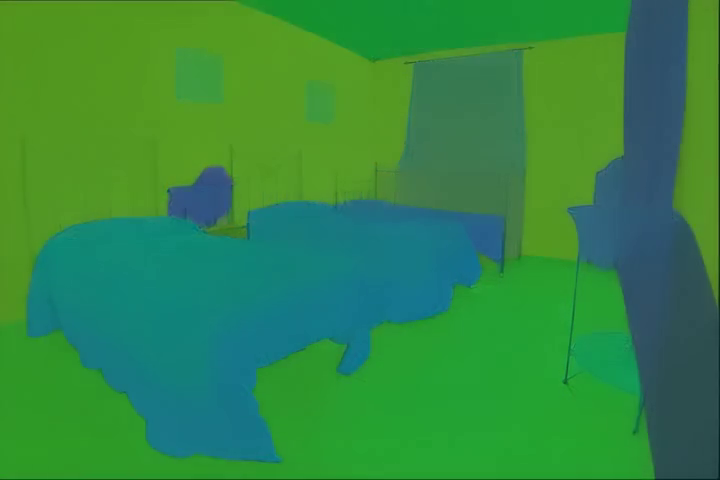} \\
\end{tabular}
\caption{Examples from the \textit{Image $\rightarrow$ Segmentation} in 1-shot setting for \textit{Chamber}.}
\label{fig:appdx:cv:img2seg}
\end{figure}

\begin{figure}[ht]
\centering
\begin{tabular}{c@{\hskip 0.5cm}cc}
& \textbf{Input Segmentation} & \textbf{Image Prediction} \\
% Chamber section - adjust the [0pt] value to move text up/down
\multirow{3}{*}[-30pt]{\rotatebox{90}{\textbf{Chamber}}} &
\includegraphics[width=3cm]{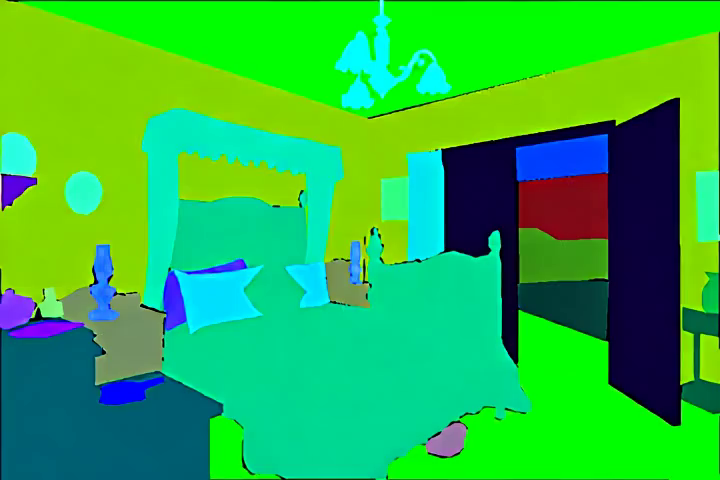} &
\includegraphics[width=3cm]{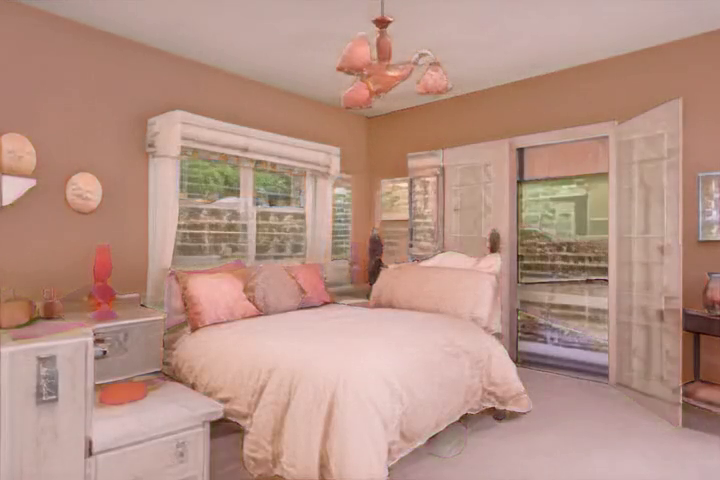} \\
& \includegraphics[width=3cm]{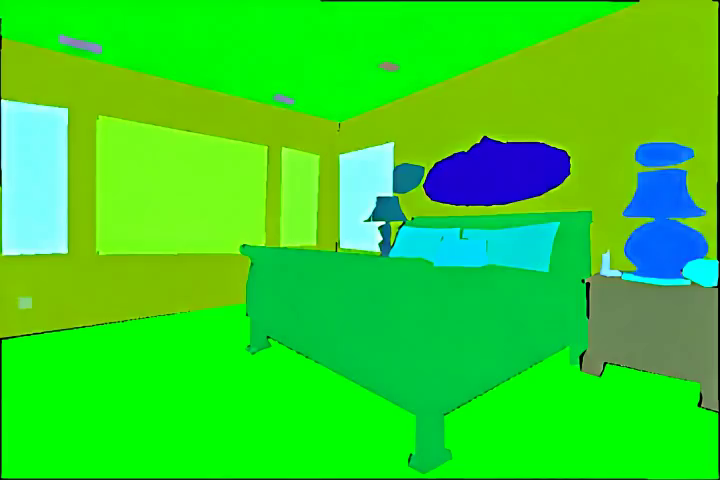} &
\includegraphics[width=3cm]{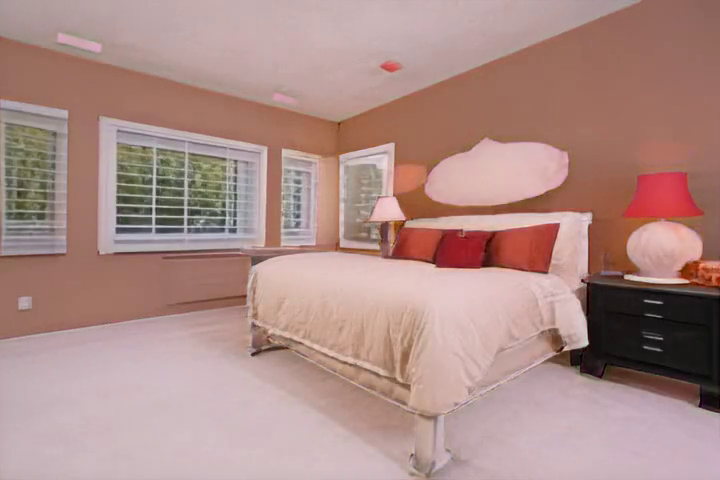} \\
& \includegraphics[width=3cm]{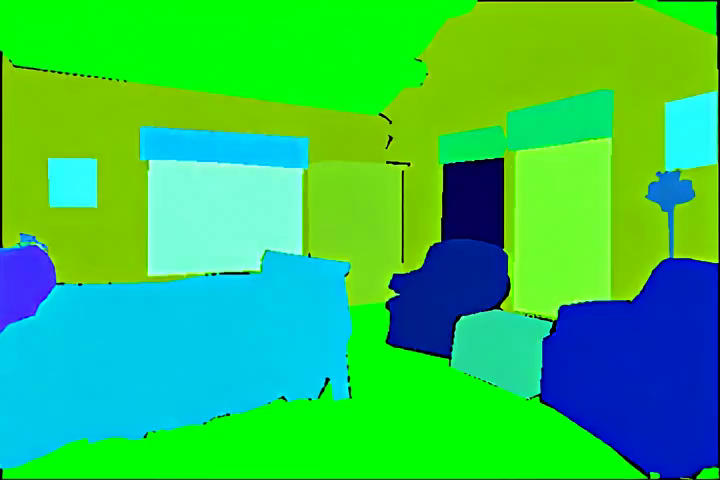} &
\includegraphics[width=3cm]{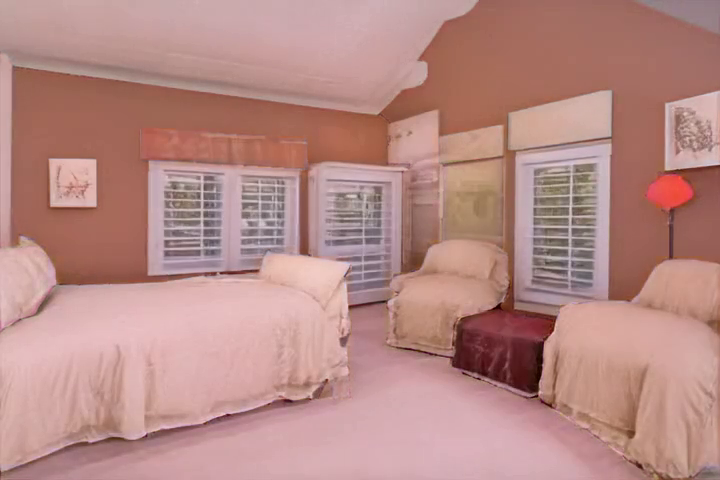} \\
% Add some space between sections
\\[0.3cm]
% Coast section - adjust the [0pt] value to move text up/down
\multirow{1}{*}[30pt]{\rotatebox{90}{\textbf{Coast}}} &
\includegraphics[width=3cm]{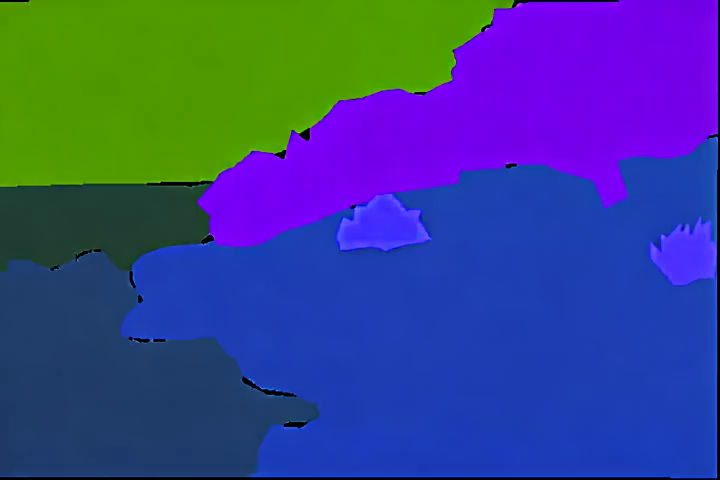} &
\includegraphics[width=3cm]{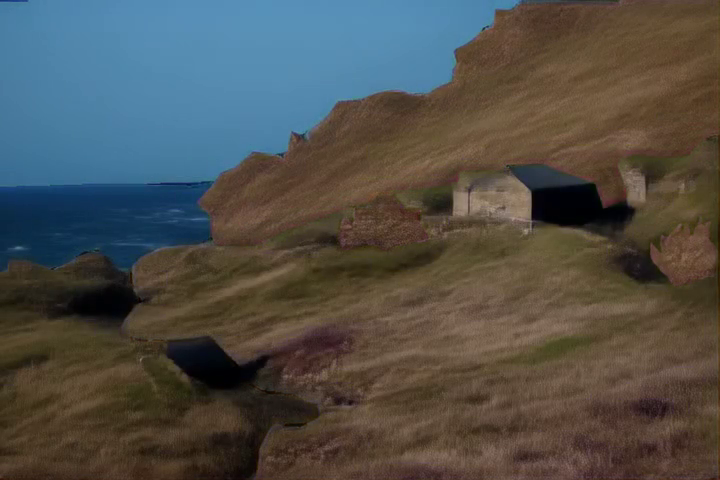} \\
% Add some space between sections
\\[0.3cm]
% Badlands section - adjust the [0pt] value to move text up/down
\multirow{1}{*}[40pt]{\rotatebox{90}{\textbf{Badlands}}} &
\includegraphics[width=3cm]{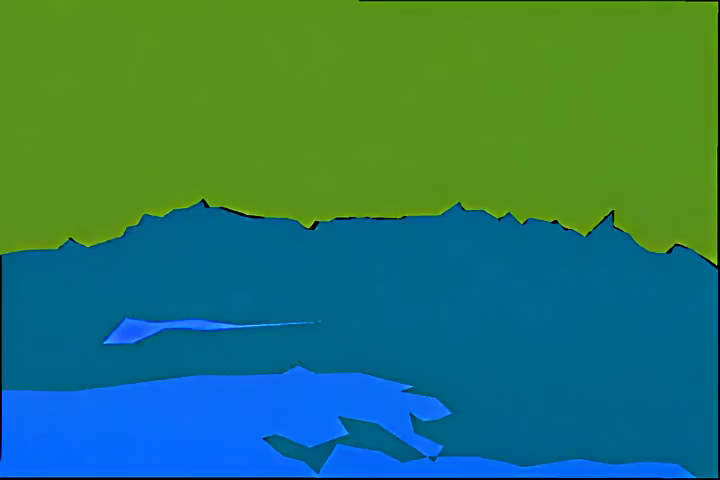} &
\includegraphics[width=3cm]{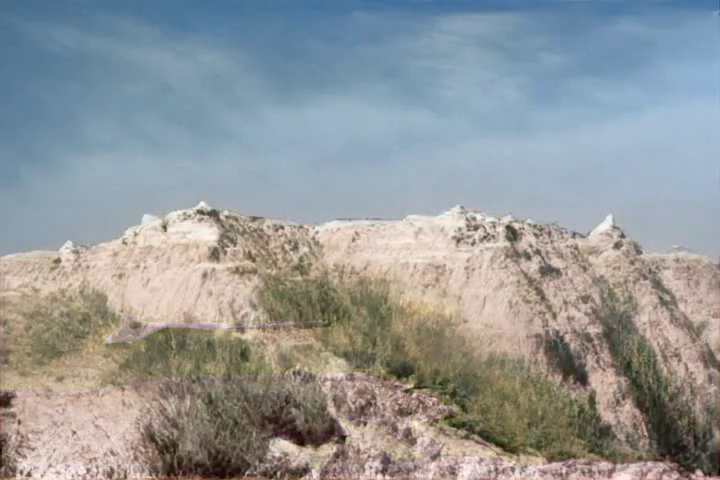} \\
\end{tabular}
\caption{Examples from the \textit{Segmentation $\rightarrow$ Image} task in the 1-shot setting. Each environment corresponds to a separate 1-shot training: for \textit{Chamber} we train on one chamber and test on others, while for \textit{Coast} and \textit{Badlands} the same protocol applies within their category.}
\label{fig:appdx:cv:seg2img}
\end{figure}

\end{document}